\documentclass[10pt,twocolumn,letterpaper]{article}

\usepackage[pagenumbers]{cvpr} %

\usepackage[linesnumbered,lined,ruled,commentsnumbered]{algorithm2e}
\usepackage{algpseudocode}
\usepackage[accsupp]{axessibility}  %
\usepackage{multirow}
\usepackage{array}     %
\usepackage{adjustbox}

\usepackage{xcolor}
\usepackage{colortbl}
\usepackage{soul}  %

\newcommand\Eq[1]{Equation~(\ref{eq:#1})}

\definecolor{colorR3}{RGB}{227, 26, 28}  %
\definecolor{colorR2}{RGB}{49, 163, 84}   %
\definecolor{colorR1}{RGB}{31, 120, 180}  %

\usepackage[frozencache,cachedir=.]{minted}
\definecolor{cvprblue}{rgb}{0.21,0.49,0.74}

\usepackage[pagebackref,breaklinks,colorlinks,allcolors=cvprblue]{hyperref}

\def\paperID{11278} %
\def\confName{CVPR}
\def\confYear{2025}

\title{LookingGlass: Generative Anamorphoses via Laplacian Pyramid Warping}

\author{Pascal Chang$^{1,2}$\\
\and
\kern-0.25em
Sergio Sancho$^{1,2}$\\
\and 
\kern-0.25em
Jingwei Tang$^{2}$\\
\and 
\kern-0.25em
Markus Gross$^{1,2}$\\
\and 
\kern-0.25em
Vinicius Azevedo$^{2}$\\
\and 
$^{1}$ETH Zürich
\and
$^{2}$DisneyResearch$|$Studios
}

\begin{document}

\twocolumn[{
\maketitle
\vspace{-27pt}
\renewcommand\twocolumn[1][]{#1}
\begin{center}
    \centering
     \includegraphics[width=0.98\linewidth]{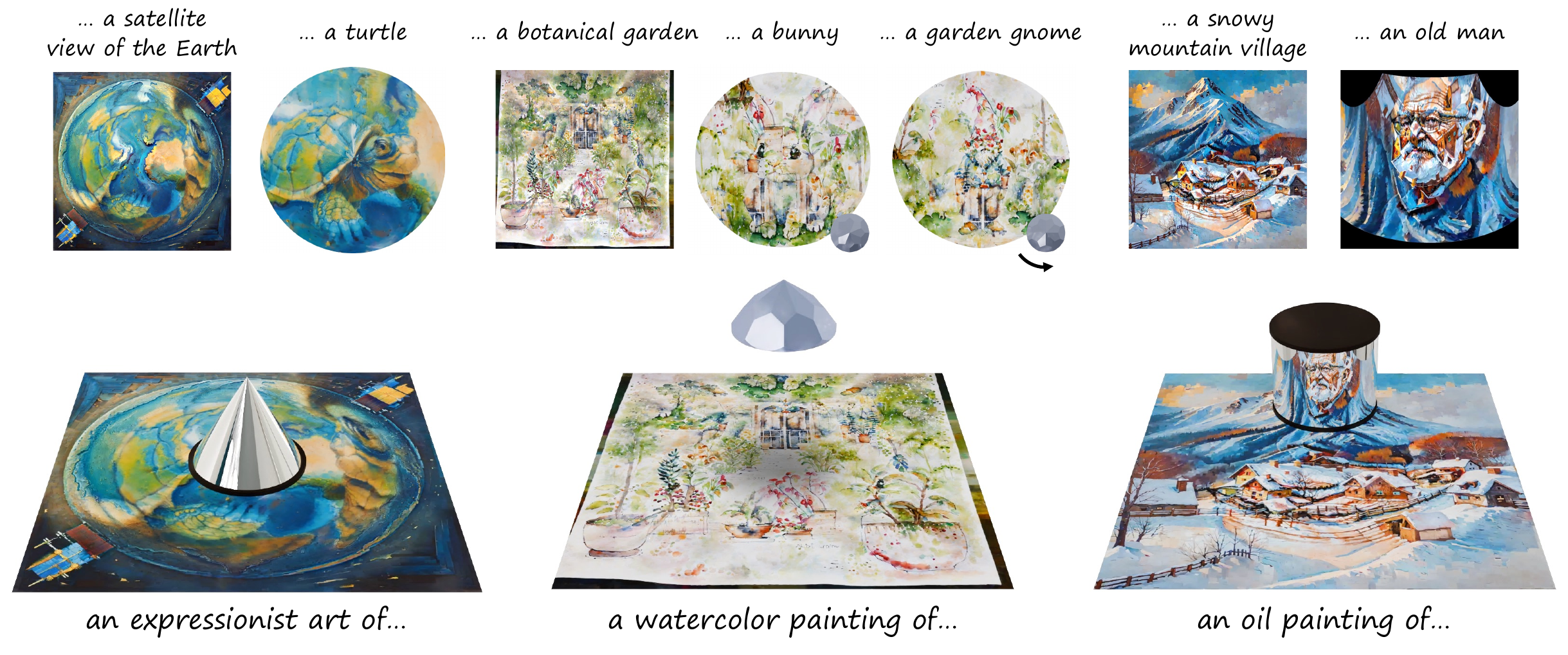}

\captionof{figure}{We propose a method to generate \emph{ambiguous anamorphoses}—images that reveal a hidden image when viewed through a mirror or lens. In the examples above, a conic mirror viewed from the top reveals a turtle hidden in an Earth image; a garden, seen through a lens, shows a bunny, and rotating the lens slightly reveals a gnome; a cylindrical mirror reflects a village painting into the face of an old man.}
   \label{fig:teaser}
\end{center}
}]

\begin{abstract}
    Anamorphosis refers to a category of images that are intentionally distorted, making them unrecognizable when viewed directly. Their true form only reveals itself when seen from a specific viewpoint, which can be through some catadioptric device like a mirror or a lens. While the construction of these mathematical devices can be traced back to as early as the 17th century \cite{niceron1638perspective}, they are only interpretable when viewed from a specific vantage point and tend to lose meaning when seen normally. In this paper, we revisit these famous optical illusions with a generative twist. With the help of latent rectified flow models, we propose a method to create anamorphic images that still retain a valid interpretation when viewed directly. To this end, we introduce \emph{Laplacian Pyramid Warping}, a frequency-aware image warping technique key to generating high-quality visuals. Our work extends Visual Anagrams \cite{geng2024visualanagrams} to latent space models and to a wider range of spatial transforms, enabling the creation of novel generative perceptual illusions.
\end{abstract}
    
\begin{figure*}
  \centering
  \centering
   \includegraphics[width=1.0\linewidth]{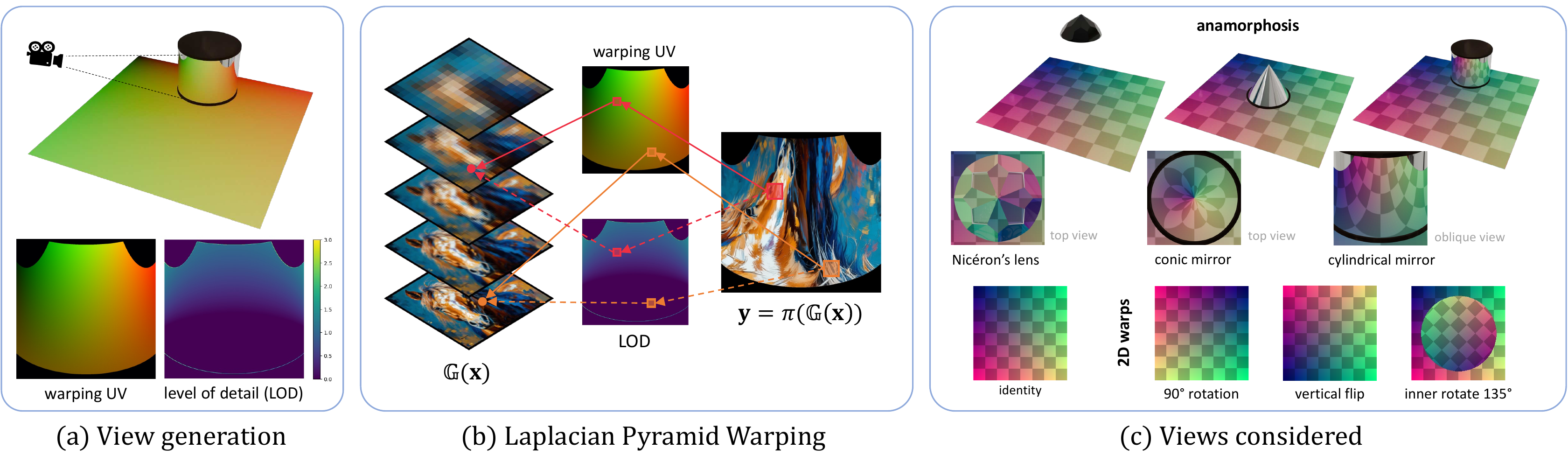}

   \caption{\textbf{Laplacian Pyramid Warping.} (a) The view mappings are generated using a ray tracer and the Level of Detail (LOD) map is computed. (b) For each pixel, our forward warping algorithm looks up in the warping UV mapping and LOD to fetch the corresponding value. (c) We consider different views, from 2D transformations like vertical flip and arbitrary angle rotation to complex 3D projections.}
   \label{fig:laplacian_warping}
\end{figure*}

\section{Introduction}
\label{sec:intro}

Anamorphosis, derived from the Greek \emph{ana} (``back" or ``again") and \emph{morphe} (``form"), refers to a category of images that are deliberately distorted, rendering them unrecognizable when viewed directly. These optical illusions reveal their true form only when observed from a precise vantage point or through reflective or refractive surfaces, such as mirrors or lenses—objects collectively known as \emph{anamorphoscopes} \cite{kuchel1979anamorphoscope}. These mathematical curiosities became more popular since the 17th century, when the pioneering treatise by the French mathematician J.-F. Nicéron, \textit{La Perspective Curieuse}, laid the foundation for their rigorous construction \cite{niceron1638perspective}. However, these images are typically interpretable only from specific angles, losing their meaning when viewed normally. 

In this paper, we propose a method for creating anamorphic images using latent text-to-image models. We focus on setups in which the image has a valid interpretation when viewed as-is without distortions. Our work is similar to the recent framework of Visual Anagrams proposed by Geng \etal \cite{geng2024visualanagrams}, which generates ambiguous images by synchronizing diffusion paths across multiple views. However, their method is constrained to pixel-space diffusion models and limited to orthogonal transformations of image pixels. We address these two limitations in this paper. First, we enable the use of latent diffusion and flow models in an artifact-free manner, improving the generation quality. We believe this will render the generation of these illusions more accessible. Second, we introduce \textit{Laplacian Pyramid Warping}, a robust image-warping technique that handles complex image transformations while preserving high-frequency details. This enables the generation of intricate anamorphoses involving complex reflective and refractive surfaces, with minimal sacrifice to image quality. Compared to previous work, our method demonstrates a significant boost in both the quality and expressiveness of the generated results.

\section{Related Work}
\label{sec:related_work}

\paragraph{Computational optical illusions.} Anamorphosis can be dated back to around the 16th century \cite{vaulezard1630perspective,niceron1638perspective,jurgis1977anamorphic}, when artists either hand drew the illusions on paper or used grids to create them systematically. Since then, the generation of optical illusions has seen significant progress, especially with the advent of computational methods in recent years. Earlier work focuses on creating illusion with 2D images, such as revealing an image by stacking transparent sheets of images \cite{nakajima2004picture}, achieving appearance change of images at different viewing distances \cite{oliva2006hybrid}, creating static images that appear to move \cite{chi2008self}, and designing a refractive lens for revealing a hidden image from dots \cite{papas12magic}.
Beyond image manipulation, several works explored 3D illusions. 
Hsiao \etal \cite{hsiao2018multiview} introduced multi-view wire art, where a single 3D wireframe produces different projected images from various perspectives.
Perroni-Scharf and Rusinkiewicz \cite{perronischarf2023printable} extended this idea to 3D-printed view-dependent surfaces. Apart from illusion based on 3D geometries, Chu \etal \cite{chu2010camouflage} explored camouflaging objects by retexturing them, while Chandra \etal \cite{chandra2022designing} developed models that shift in perception based on lighting changes. In contrast, we  focus on 2D illusions that require 3D objects to reveal the hidden views.

\paragraph{Illusions with diffusion models.}

Recent work has revealed the potential of diffusion models in creating optical illusions. Burgert \etal \cite{burgert2024diffusionillusions} employ score distillation sampling (SDS) to generate images that align with multiple prompts from different viewpoints. Although their optimization-based method can theoretically produce anamorphoses, it suffers from lower image quality and long inference times. Visual Anagrams \cite{geng2024visualanagrams} introduces a formal framework for illusion generation in a single diffusion pass. However, their approach is limited to orthogonal transformations, making it unsuitable for generating the complex deformations needed for anamorphoses. Subsequent studies have also explored various types of illusions, such as visually meaningful spectrograms \cite{chen2024images} and generative hybrid images \cite{geng2024factorized}. Our proposed method is most similar to Visual Anagrams. Key differences, however, are that we extend to latent space models and a broader range of transformations. A concurrent work, Illusion3D \cite{feng2024illusion3d3dmultiviewillusion}, builds on \cite{burgert2024diffusionillusions} to generate 3D anamorphic illusions, but appears constrained in quality and artistic flexibility. We outline the key differences with our method in the supplementary material.

Beyond academic research, the artistic community has also explored diffusion models for optical illusions. Notably, an anonymous artist known as MrUgleh \cite{ugleh2023spiral} repurposed a model fine-tuned for generating QR codes \cite{monster2023controlnetqr, zhang2023adding} to create images that subtly mimic the global structure of a specified template image. Our focus is on generating ambiguous images and anamorphoses based on text prompts, which does not require an image template.

\paragraph{Synchronized diffusion.}

In Visual Anagrams \cite{geng2024visualanagrams}, diffusion paths from different viewpoints are synchronized by averaging the predicted noise at each timestep. Numerous studies have explored merging diffusion paths, often in the context of controlled image generation. MultiDiffusion \cite{bar2023multidiffusion} proposes a least-squares formulation for merging views, which simplifies to averaging in the special case of equal-size crops—a setup they apply to panorama generation. DiffCollage \cite{zhange2023diffcollage} synthesizes large-scale content by merging outputs from diffusion models trained on segments of the larger composition. SyncTweedies \cite{Kim2024SyncTweedies} thoroughly examines synchronization techniques, finding that averaging the predicted clean images yields the best quality. Closer to our approach, Generative Powers of Ten \cite{wang2023generativepowers} creates infinite zoom videos by merging concentric views at different resolutions using Laplacian pyramids. But their method is tailored to the specific use case of zooming. One of our contributions, \textit{Laplacian Pyramid Warping}, generalizes this approach to arbitrary views.

\paragraph{Image pyramids in vision and graphics.}

Image pyramids, particularly Gaussian and Laplacian pyramids, are widely used in computer vision for their multi-scale representation capabilities \cite{burt1983laplacian, Clark1976, Williams1983}. By decomposing images hierarchically, pyramids enable efficient compression, progressive image reconstruction, and seamless blending—essential in applications like panorama stitching and HDR imaging \cite{burt1983multiresolution}. Beyond blending, Gaussian pyramids are central to scale-invariant object detection and recognition, where they assist in feature detection for algorithms like SIFT \cite{lowe2004distinctive}. They are also valuable in texture analysis and synthesis \cite{heeger1995pyramidtexture}, and optical flow estimation \cite{ranjan2017optical}, where multi-scale representations enhance accuracy and reduce artifacts.

In computer graphics, pyramids relate closely to techniques like texture MIP-mapping \cite{ewins1998mip} and antialiasing, which address the challenges of rendering textures at varying distances and viewing angles. MIP-maps, essentially a Gaussian pyramid form, allow graphics engines to select the appropriate level of detail (LOD) based on screen space, minimizing artifacts like flickering and enhancing both quality and efficiency. We repurpose these texture MIP-mapping techniques in our proposed method for frequency-aware image warping.

\begin{figure}[t]
    \centering
    \includegraphics[width=0.168\textwidth]{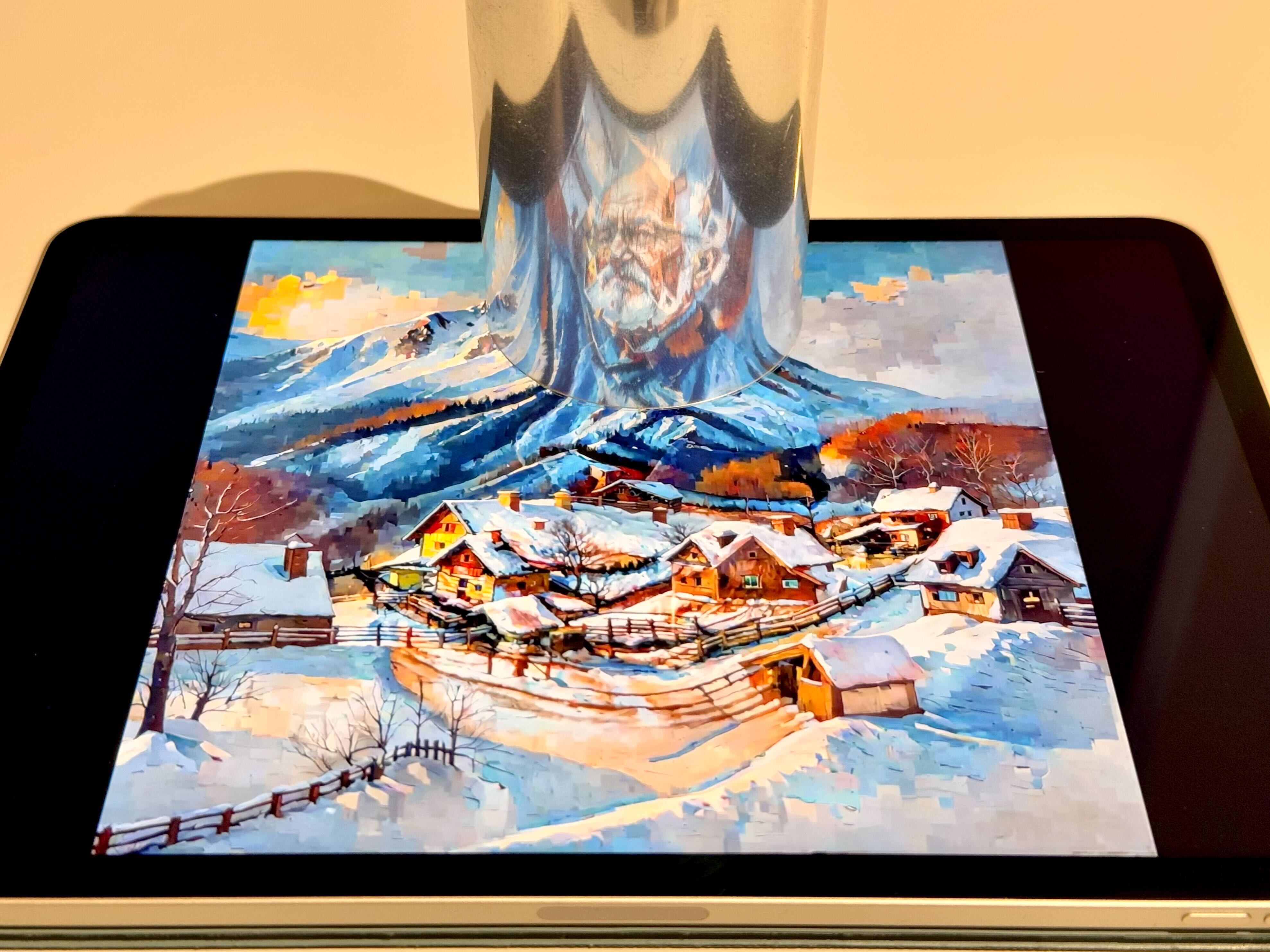}\hspace{6pt}
\includegraphics[width=0.126\textwidth]{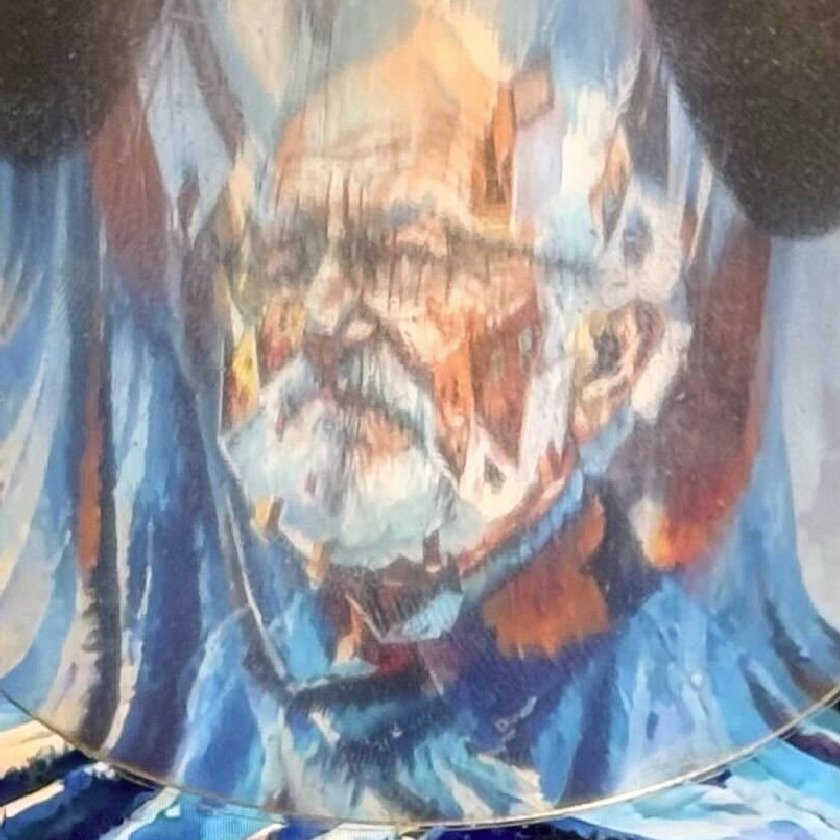}\hspace{6pt}
\includegraphics[width=0.126\textwidth]{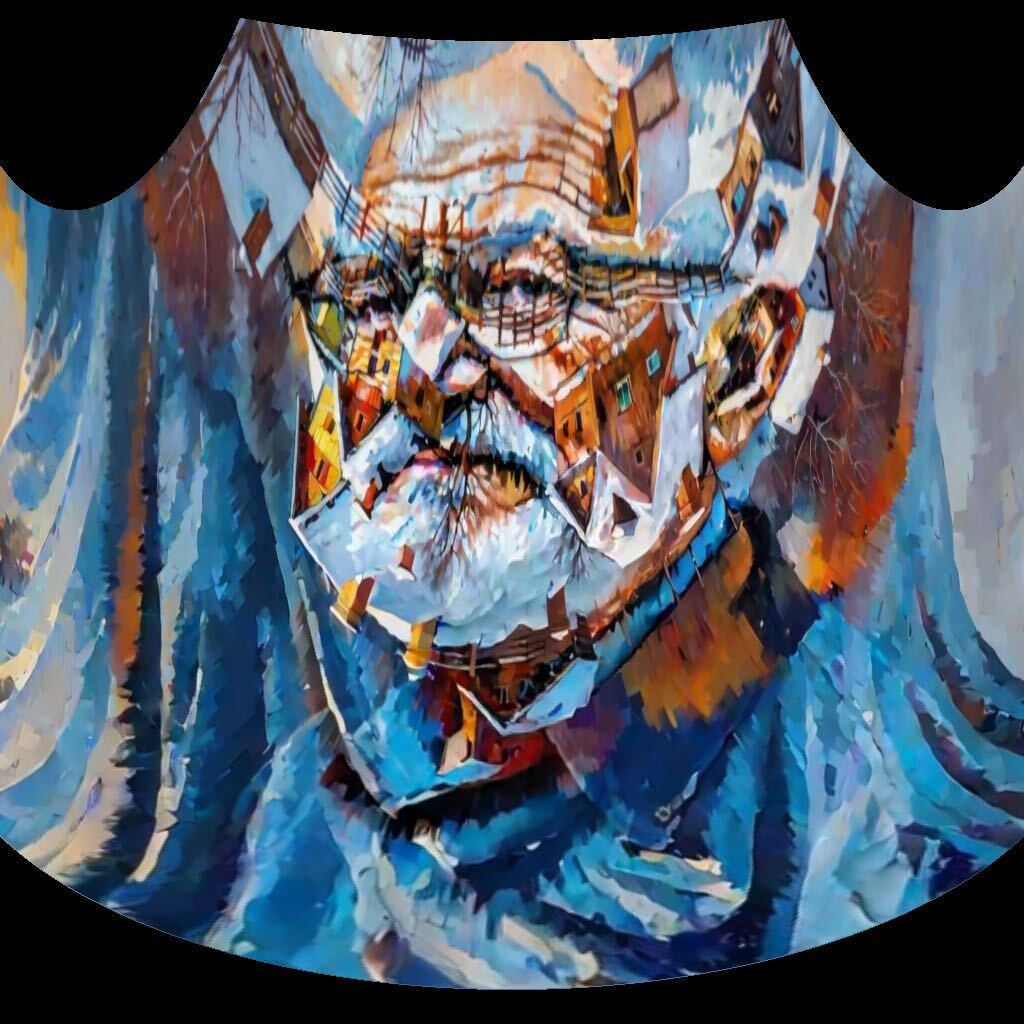}
    \caption{A real life demo of the cylindrical mirror illusion.}
    \label{fig:reallife}
\end{figure}

\begin{figure*}
  \centering
  \centering
   \includegraphics[width=1.0\linewidth]{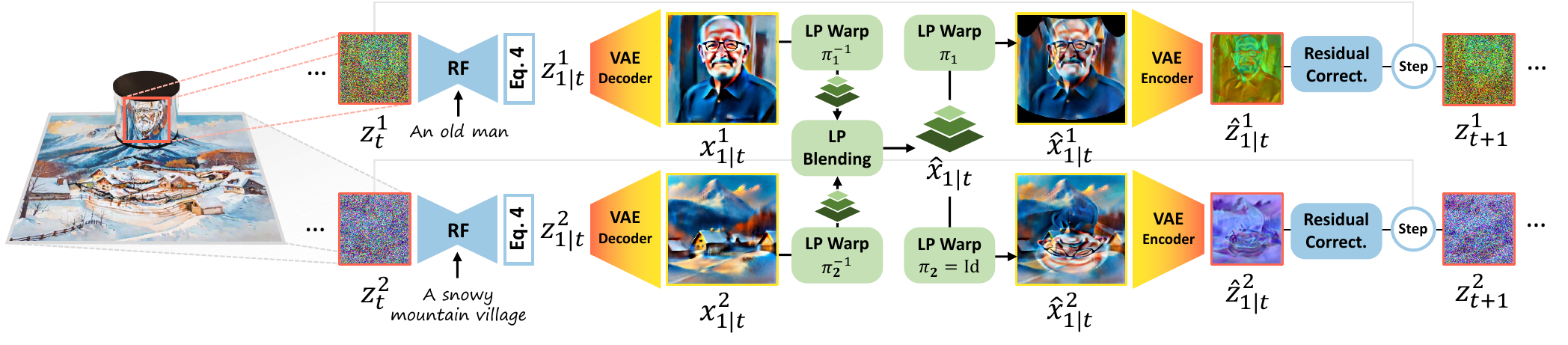}

   \caption{\textbf{Our Proposed Pipeline.} At each denoising step, the estimated final image is computed from the network velocity estimate and decoded into image space. Image warping and view aggregation is performed in image space using Laplacian pyramids, before encoding back into latent space for the diffusion step.}
   \label{fig:pipeline}
\end{figure*}

\section{Preliminaries}
\label{sec:preliminaries}

\subsection{Text-conditioned Rectified Flows}

In Rectified Flows (RFs), a noise sample $\mathbf{z}_0 \sim \mathcal N (\mathbf{0}, \mathbf{I})$ is mapped to an image $\mathbf{z}_1 \sim p_1$ through the ODE:
\begin{equation}\label{eq:flow_ode}
    d \mathbf{z}_t = \boldsymbol{u}_t(\mathbf{z}_t, y)dt,
\end{equation}
where $t \in [0, 1]$, $y$ is an optional text prompt conditioning, and the velocity field is typically parameterized with a neural network, \ie $\boldsymbol{u}_t(\mathbf{z}_t, y) = \boldsymbol{u}_{\theta}(\mathbf{z}_t; t, y)$. At inference, the ODE is discretized, and solved with classical integration schemes such as forward Euler:
\begin{equation}\label{eq:flow_euler}
    \mathbf{z}_{t+\Delta t} = \mathbf{z}_t + \boldsymbol{u}_{\theta}(\mathbf{z}_t; t, y)\Delta t,
\end{equation}

\paragraph{Classifier-free guidance (CFG).} As in diffusion models, classifier-free guidance \cite{ho2022classifier} can be used to improve sample quality in RFs. The final velocity interpolates between a text-conditioned and an unconditional prediction:
\begin{equation}
\label{eq:cfg}
    \hat{\boldsymbol{u}}_t = (1+\omega) \boldsymbol{u}_\theta(\mathbf{z}_t ; t, y) - \omega \boldsymbol{u}_\theta(\mathbf{z}_t ; t, \varnothing),
\end{equation}
where $\omega$ is the classifier-free \textit{guidance scale}. Higher guidance scales typically improve sample quality at the expense of diversity, but also tend to produce over-saturated images.

\paragraph{Predicted clean image.} At any intermediate timestep $t$, an estimate of the clean image, denoted $\mathbf{z}_{1|t}$, can be obtained by a single Euler step to $t=1$ using the current velocity estimate:
\begin{equation}
\label{eq:final_estimate}
    \mathbf{z}_{1|t} = \mathbf{z}_t + \boldsymbol{u}_{\theta}(\mathbf{z}_t; t, y)(1-t).
\end{equation}
\Eq{final_estimate} can be seen as the flow matching equivalent of the Tweedie's formula \cite{robbins1992empirical} in diffusion models.
\subsection{Gaussian \& Laplacian Pyramid}

A Gaussian pyramid is a multi-scale representation of an image obtained by iteratively applying a Gaussian blur kernel $\kappa$ and downsampling $\mathbf{D}(\cdot)$. Given an image $\mathbf{x}$, the image \( \mathbb G_l \) at level \(l\) is computed from the previous level as
\begin{equation*}
\mathbb G_l(\mathbf{x}) = \mathbf{D}(\kappa(\mathbb G_{l-1}(\mathbf{x}))),
\end{equation*}
where \( \mathbb G_0(\mathbf{x}) =\mathbf{x} \) is the original image.

A Laplacian pyramid stores the high-frequency details between each level of a Gaussian pyramid. Each Laplacian level \( \mathbb L_l \) is defined as the difference between a Gaussian level and the upsampled version of the next level 
\begin{equation*}
\mathbb L_l(\mathbf{x}) = \mathbb G_l(\mathbf{x}) - \mathbf{U}(\mathbb G_{l+1}(\mathbf{x})),
\end{equation*}
with \(\mathbb L_{L-1}(\mathbf{x}) = \mathbb G_{L-1}(\mathbf{x}) \) for a pyramid of depth $L$ and $\mathbf{U}(\cdot)$ being the upsample operator. To reconstruct the image, we recursively add each Laplacian level back to the upsampled version of the next level.

\subsection{Visual Anagrams}
The work of Geng \etal \cite{geng2024visualanagrams} proposes creating multi-view images by using a text-to-image generative model to simultaneously denoise multiple views of an image. The original paper utilizes diffusion models. We summarize the method here through the terminologies of RFs for simplicity. 

A canonical space $\mathcal{C}$ is defined for an image. A set of prompts $y_i$ are associated with different view functions $\pi_i$, which transform the image from the canonical space to the target space $\mathcal{T}$ where rectified flow models are applied. At each timestep of the inference, $\mathbf{z}^i_t$ of each view is transformed into the canonical space, and averaged together with the other views. After averaging, the noisy images in the canonical space are transformed back to the target space as $\hat{\mathbf{z}}^{i}_{t}$, replacing the original $\mathbf{z}^i_t$ as

\begin{equation}\label{eq:visual_anagram}
    \hat{\mathbf{z}}^{i}_{t}=\pi_i\left(\frac{1}{N} \sum_j \pi_j^{-1}\left(\mathbf{z}^{j}_{t}\right)\right).
\end{equation}

Transforming noisy samples, as in Geng \etal \cite{geng2024visualanagrams}, limits possible mappings between the canonical and target spaces to be orthogonal transformations such as flipping, rotation and permutation of pixels. This happens since arbitrarily warping a noise sample is generally more difficult than warping images, as bilinear and bicubic interpolations can destroy the Gaussian noise properties \cite{chang2024how}. Moreover, SyncTweedies \cite{Kim2024SyncTweedies} showed that averaging the predicted clean image $\mathbf{z}^i_{1|t}$, instead of noisy samples, produces higher quality results. Thus, to allow more general transformations and improve generation quality, we modify \Eq{visual_anagram} to average predicted clean image using Tweedie's formula as

\begin{equation}\label{eq:sync_tweedies}
    \hat{\mathbf{z}}^{i}_{1|t}=\pi_i\left(\frac{1}{N} \sum_j \pi_j^{-1}\left(\mathbf{z}^{j}_{1|t}\right)\right).
\end{equation}

\section{Generative Anamorphosis}
\label{sec:method}

Our goal is to generate high-quality anamorphoses from text prompts. Anamorphoses involve view functions beyond simple transformations such as rotation, flipping and pixel permutations where no analytical transformations can be defined. We opt for a more general representation: $\pi$ is now a 2-channel image of UV coordinates indicating where to fetch values in the canonical view for each pixel $\pi (x, y) = (u,v)$. We implement the transformation with a simple raytracer by placing a UV coordinate texture on the main image plane, and rendering the result when viewing through mirrors or lenses (see Figure \ref{fig:laplacian_warping}).

Simply adopting SyncTweedies \cite{Kim2024SyncTweedies}, however, is not enough when considering latent diffusion models. Thus, in \Cref{subsec: glva}, we present a generalization of previous approaches to latent diffusion models. Naively averaging arbitrary transformations with highly distorted regions, typical when looking through curved mirrors or lenses, results in visual artifacts. We introduce a novel Laplacian Image Warping method that utilizes a multi-level texture structure inspired by classic works in computer graphics \cite{Clark1976, Williams1983} in \Cref{subsec:lpw} to alleviate this problem. Additional design choices are then discussed in \Cref{subsec:designchoice}. \Cref{fig:pipeline} and \Cref{algo:main} show an overview of our method.

\subsection{Latent Visual Anagrams}\label{subsec: glva}

\begin{figure}
  \centering
  \centering
   \includegraphics[width=1.0\linewidth]{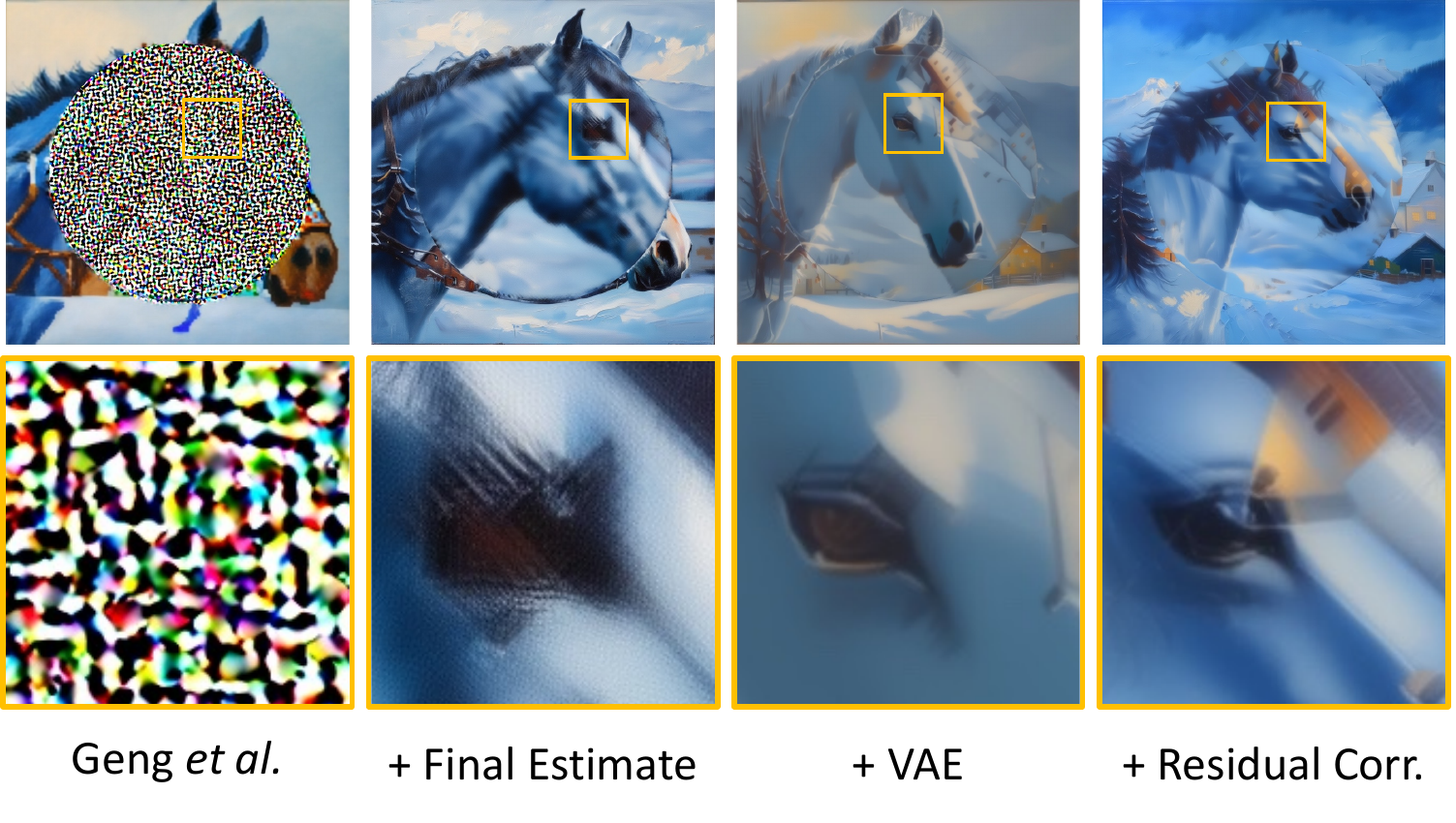}

    \vspace{-3mm}

   \caption{\textbf{Latent Visual Anagrams.} In this 135$^\circ$ rotation example, we demonstrate that contributions from \cref{subsec: glva} improve the generation of visual anagrams with latent models. While the final estimate from SyncTweedies \cite{Kim2024SyncTweedies} partially addresses noise issues, artifacts from the VAE persist. Our VAE encoding/decoding process and residual correction further enhance image quality.}
   \label{fig:ablation_comp}
\end{figure}

Tancik \cite{tancik2023illusion} generates multi-view illusions with Stable Diffusion 1.5 \cite{rombach2022high} using a similar pipeline to Geng \etal \cite{geng2024visualanagrams}. Images are transformed to canonical space through views in the latent space, and decoded to the clean image after the denoising process is finished. The results contain visual artifacts due to the fact that VAEs for latent diffusion and flow models are generally not trained to be equivariant. When the latent images are deformed, the corresponding decoded image does not necessarily share the same transformation. While these artifacts are less pronounced in recent models with larger latent spaces (\eg Stable Diffusion 3 \cite{esser2024scaling}), they still persist and create undesirable strokes (see Figure \ref{fig:ablation_comp}).

\vspace{-1.0em}

\paragraph{VAE encoding-decoding.} To tackle this issue, we propose to keep all image transformations in image space. This is done by decoding the estimated clean image latent to pixel space, applying the transformation, and re-encoding it to latent space. Denoting $\mathcal E$, $\mathcal D$ the encoder and decoder of the VAE respectively, our estimated clean latent at timestep $t$, in view $i$, becomes
\begin{equation}\label{eq:latent_aggreg}
    \hat{\mathbf{z}}^{i}_{1|t}=\mathcal E \circ \pi_i\left(\frac{1}{N} \sum_j \pi_j^{-1} \circ \mathcal D \left(\mathbf{z}^{j}_{1|t}\right)\right).
\end{equation}

\vspace{-1.0em}

\paragraph{Residual correction.} While \cref{eq:latent_aggreg} proved to be effective in removing the artifacts, we observed that it can be sensitive to the VAE reconstruction quality. In particular, when the input latent is far from the training distribution of the VAE (\eg predicted clean image from early steps of denoising), the reconstruction typically fails to match the value range of the input. This creates washed-out colors in the final image. We correct the reconstruction failure through a first-order term $\Delta\hat{\mathbf{z}}^{i}_{1|t}$. This is done by first computing the \textit{residual} between the latent and the decoded-encoded latent. This residual is then transformed to a target view as
\begin{equation}\label{eq:delta_aggreg}
    \Delta\hat{\mathbf{z}}^{i}_{1|t}=\pi_i\left(\frac{1}{N} \sum_j \pi_j^{-1} \left(\mathbf{z}^{j}_{1|t} - \mathcal E \circ \mathcal D \left(\mathbf{z}^{j}_{1|t}\right)\right)\right).
\end{equation}

This correction term is then added to \cref{eq:latent_aggreg} as the final estimated clean latent. The correction has two nice properties. First, if the VAE reconstruction is perfect, then $\mathcal E ( \mathcal D (\mathbf{z})) = \mathbf{z}$, and $\Delta\mathbf{z} = 0$. Therefore, no correction is done. Second, when there is only one single identity view ($N=1$), we recover $\hat{\mathbf{z}}^{j}_{1|t} = \mathbf{z}^{j}_{1|t}$, the original predicted clean latent, as expected. Figure \ref{fig:ablation_comp} compares different steps presented in this section, and shows the effectiveness of the residual correction in maintaining the vibrant colors.

\subsection{Laplacian Pyramid Warping (LPW)}\label{subsec:lpw}

\begin{figure}
  \centering
  \centering
   \includegraphics[width=1.0\linewidth]{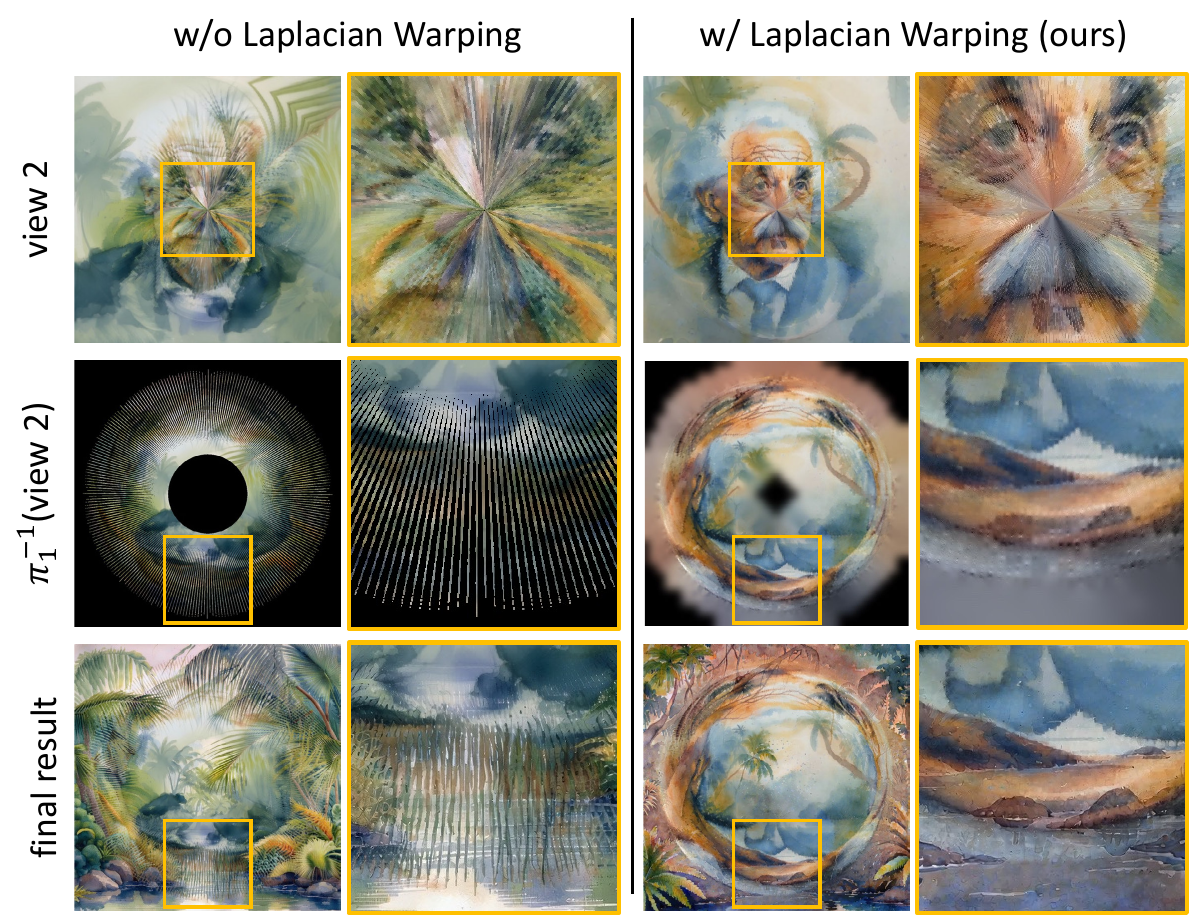}

   \caption{\textbf{Inverse Laplacian Warping.} In this conic mirror example, we use a challenging pair of prompts to demonstrate our inverse warping. The main view shows \emph{``a jungle,"} while the mirror view reveals \emph{``a portrait of Einstein."} Without inverse Laplacian warping, gaps from the inverse transformation cause striped artifacts and distortions. Our Inverse Laplacian Warping (\cref{subsec:lpw}) correctly assigns values at the right frequencies, eliminating artifacts and making the mirror view more recognizable.}
   \label{fig:laplacian_pb}
\end{figure}

While the above modifications enable using latent diffusion and flow models for creating ambiguous images with arbitrary 2D transformations, applying it to the generation of anamorphoses still presents a few challenges. First, a view in anamorphosis rarely covers the entirety of the main image. As such, boundaries can create visible artifacts and seams that are undesirable. Second, having views that have varying degrees of stretching can lead to degraded high frequency details (see \Cref{fig:laplacian_pb}).

We identify the key problem being using the averaging operation as aggregation of views. When images of different views are transformed to the canonical view, they can have different frequency components. One view may map to a small pixel subset in the canonical view or undergo large stretching. Averaging pixels of these views with another view not stretched so much ignores the frequency mismatching problem. To solve this problem, we propose to use Laplacian pyramid \cite{burt1987laplacian} for blending the views. 

After decoding the predicted clean latent, \emph{Inverse Laplacian Warping} $\pi^{-1}$ is used to map the image to a Laplacian pyramid in the canonical view. The canonical views are then aggregated through Laplacian Pyramid Blending. Afterwards, the canonical views are transformed back to images through \emph{Forward Laplacian Warping} $\pi$.

\vspace{-0.08em}

\paragraph{Forward Laplacian Warping.} Given an image $\mathbf{x}$ and a view projection $\pi$, we propose a Level-of-Detail-Aware (LOD-aware) method to compute $\mathbf{y} = \pi(\mathbf{x})$. First, we build a Gaussian pyramid from our image $\mathbb G(\mathbf{x}) = \{ \mathbb G_0(\mathbf{x}), ..., \mathbb G_{L-1}(\mathbf{x}) \}$. Then, we compute a LOD level map using $\pi$ and image-space derivatives. For a given pixel $(x,y)$, the LOD level is given by:
\begin{equation}\label{eq:lod}
\adjustbox{max width=0.85\linewidth}{
$l = \log_2 \left( \max \left( \sqrt{\left( \frac{\partial u}{\partial x} \right)^2 + \left( \frac{\partial u}{\partial y} \right)^2}, \sqrt{\left( \frac{\partial v}{\partial x} \right)^2 + \left( \frac{\partial v}{\partial y} \right)^2} \right) \right).$
}
\end{equation}
The transformed image is obtained by sampling the pyramid $\mathbb G(\mathbf{x})$ with the LOD map using nearest or trilinear interpolation. This approach is commonly used in computer graphics when rendering textures to avoid aliasing. Our contribution lies in connecting the method to our problem and repurposing the idea for image warping. For clarity, we simplify the equation as $\mathbf{y} = \pi(\mathbb G (\mathbf x))$.

\vspace{-0.08em}

\paragraph{Inverse Laplacian Warping.} To transform an image $\mathbf{y}$, into a Laplacian pyramid in the canonical view $\mathbb G(\mathbf{x}) = \pi^{-1}(\mathbf{y})$, we define the inverse Laplacian warping operation. As inverting an arbitrarily complex image transformation is infeasible, we make use of the gradient of Forward Laplacian Warping. Consider a dummy zero image $\mathbf{x}^0 = \boldsymbol{0}$, our inverted image pyramid is given by:
\begin{equation}\label{eq:inversion}
    \mathbb G(\mathbf{x}) = -\nabla_{\mathbf{x}^0}\left[\frac 1 2 \| \pi(\mathbf{x}^0) - \mathbf{y}\|^2\right].
\end{equation}
This effectively transports the pixels in $\mathbf{y}$ to the corresponding location and level in the pyramid. We then propagate the changes of lower levels to higher ones, and extract a Laplacian pyramid out of the resulting pyramid. More details can be found in the supplementary material.

\paragraph{Laplacian Pyramid Blending} is used for aggregating different canonical views  to obtain a synchronized image for the given denoising step. Each level is averaged and the final image is reconstructed from the resulting Laplacian pyramid. In the case of anamorphic illusions, most views will only cover the identity space partially, so the blending is weighted by a set of masks at each level. Special care need to be taken at the boundary of the masks, as well as during averaging,  which we further discuss in the supplementary material. \\

\begin{algorithm}[t]

\SetKwInput{KwInput}{Input}                %
\SetKwInput{KwOutput}{Output}              %
\SetKwInput{KwParameter}{Param}       %
\SetAlgoLined
\KwInput{
    $ \forall i \in 0, ..., N-1$: view transformations $\pi_i$ and text prompts $y_i$, with $N$ being the number of views. Pretrained RF model $\boldsymbol{u}_\theta$. \\
    
}
\KwOutput{Final images in each view ${\mathbf{x}}^0_{1},..., {\mathbf{x}}^{N-1}_{1}$}
\BlankLine

$\mathbf{z}^0_0, ..., \mathbf{z}^{N-1}_0 \sim \mathcal N (0, I)$ \\
\For{$t \leftarrow 0: T$} {
    \For{$i \leftarrow 0 : N-1$} { 
        ${\mathbf{z}}^i_{1|t} \leftarrow   {\mathbf{z}}^i_t + \boldsymbol{u}_{\theta}({\mathbf{z}}^i_t; t, y)(1-t)$ \Comment{Eq. (\ref{eq:final_estimate})} \\
        ${\mathbf{x}}^i_{1|t} \leftarrow$ \textrm{\textsc{Vae\_Decode}} $( {\mathbf{z}}^i_{1|t} )$ \\
        $\Delta{\mathbf{z}}^i_{1|t} \leftarrow {\mathbf{z}}^i_{1|t} -$ \textrm{\textsc{Vae\_Encode}} $(\mathbf{x}^i_{1|t})$ \\ 
    }
    $\hat{\mathbf{x}}_{1|t} \leftarrow$ \textrm{\textsc{Laplacian\_Blending}} $(\pi_0^{-1}({\mathbf{x}}^0_{1|t}),..., \pi_{N-1}^{-1}({\mathbf{x}}^{N-1}_{1|t}))$ \\ 
    $\Delta\hat{\mathbf{z}}_{1|t} \leftarrow$ \textrm{\textsc{Laplacian\_Blending}} $(\pi_0^{-1}({\Delta\mathbf{z}}^0_{1|t}),..., \pi_{N-1}^{-1}({\Delta\mathbf{z}}^{N-1}_{1|t}))$ \\ 
    \For{$i \leftarrow 0: N-1$} { 
        $\hat{\mathbf{x}}^i_{1|t} \leftarrow \pi_i (\hat{\mathbf{x}}_{1|t})$ \\
        $\hat{\mathbf{z}}^i_{1|t} \leftarrow $ \textrm{\textsc{Vae\_Encode}} $(\hat{\mathbf{x}}^i_{1|t})$ \Comment{Eq. (\ref{eq:latent_aggreg})} \\
        $\Delta\hat{\mathbf{z}}^i_{1|t} \leftarrow \pi_i\left(\Delta\hat{\mathbf{z}}_{1|t}\right)$ \Comment{Eq. (\ref{eq:delta_aggreg})} \\ 
        $\hat{\mathbf{z}}^i_{1|t} \leftarrow \hat{\mathbf{z}}^i_{1|t} + \Delta\hat{\mathbf{z}}^i_{1|t}$ \\
        $\mathbf{z}^i_{t+1} \leftarrow $ \textrm{\textsc{Denoising\_Step}} $\left(\hat{\mathbf{z}}^i_{1|t},  \mathbf{z}^i_{t} \right)$ \\
    }
}

$\{\mathbf{x}^{i}_{1}\}_i \leftarrow$ \textrm{\textsc{Vae\_Decode}} $\left( \{{\mathbf{z}}^i_{T}\}_i \right)$ \\

\Return {$\mathbf{x}^0_{1},..., \mathbf{x}^{N-1}_{1}$}

\caption{LookingGlass}
\label{algo:main}
\end{algorithm}

\subsection{Design Choices \& Further Improvements}\label{subsec:designchoice}

\paragraph{Model choice.} Our method is designed to work with latent rectified flow and diffusion models. However, we observed that different models behave differently to our synchronization scheme, which we discuss in the supplementary. For the majority of our experiments, we use Stable Diffusion 3.5 because of their higher visual quality.

\paragraph{Improved consistency with time travel.} Similar to DDNM \cite{wang2023ddnm}, RePaint \cite{lugmayr2022repaint} and Bansal \etal \cite{bansal2023universal}, we found repeating segments of the denoising process allows the model to blend different views better. To keep the inference efficient, similar to FreeDoM \cite{yu2023freedom}, we only apply time traveling at intermediate timesteps between 20\% and 80\% of the denoising process and use segments of size 1.

\begin{figure*}
  \centering
  \centering
   \includegraphics[width=\linewidth]{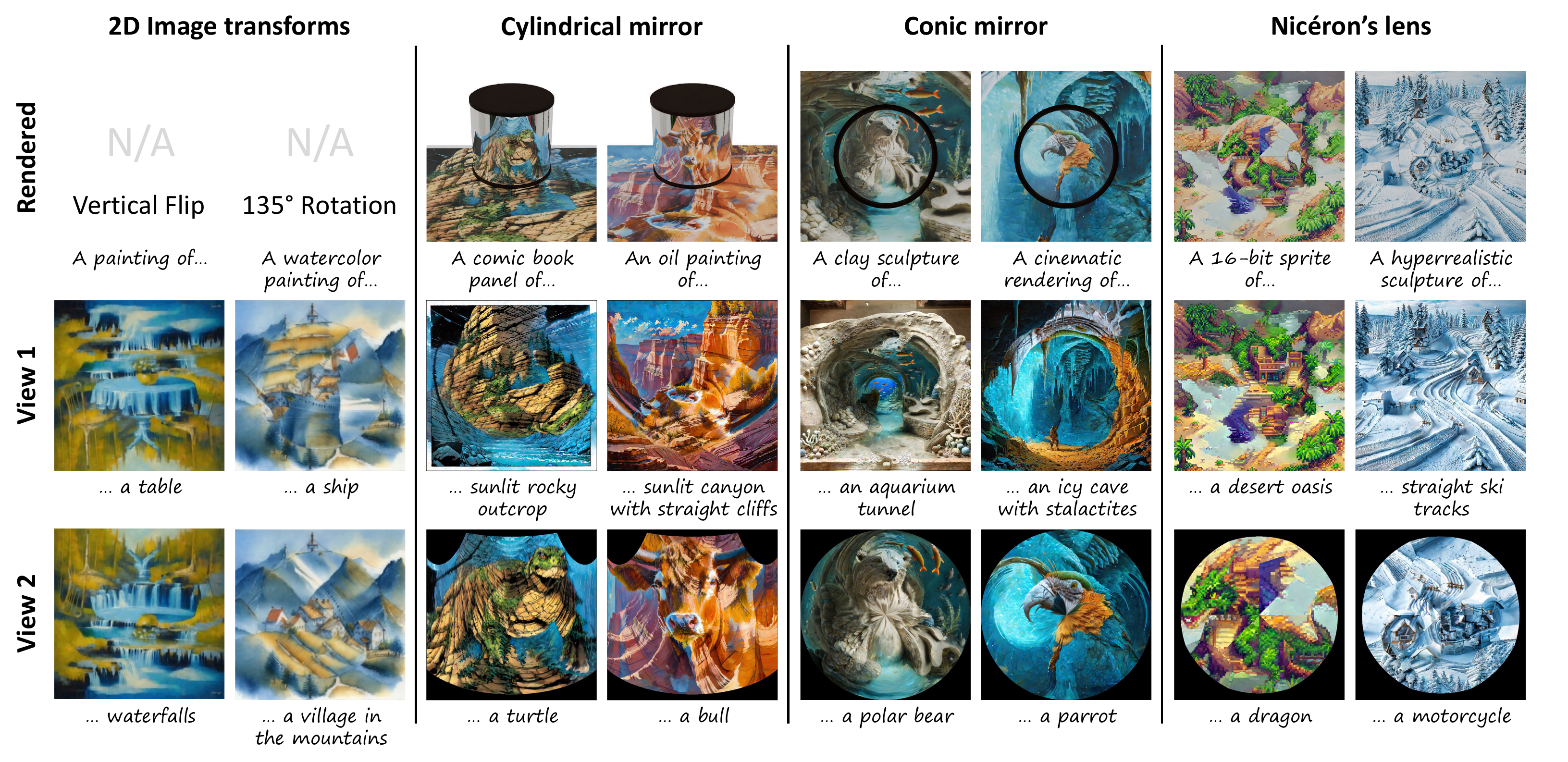}

   \caption{\textbf{Generative Anamorphoses.} Generated results with our approach for 2D transformations, and the three types of anamorphoses: a cylindrical mirror, a conic mirror, and Nicéron's lens. A rendering of the physical setup is shown in the top row when applicable.}
   \label{fig:results_ana}
\end{figure*}

\paragraph{Prioritizing a single view.} We observed that a key component to the success of previous works \cite{geng2024visualanagrams, tancik2023illusion, burgert2024diffusionillusions} lies in the fact that generated images generally lack detail. This makes it easier for human imagination to interpret image features differently based on different prompts. With latent flow models, our method generates highly detailed images at 1K resolution. This poses a new challenge, as high-frequency details are rarely compatible between the views and easily give away hidden views. As this is an inherent problem with high resolution images, we propose to prioritize one of the views, which is defined by the user. We set a portion of the last timesteps of the denoising process to be solely denoising for the chosen view. This encourages the model to create coherent details towards the end, while hiding the remaining views better.

\section{Experiments}
\label{sec:experiments}

\begin{figure}[t]
  \centering
  \centering
   \includegraphics[width=\linewidth]{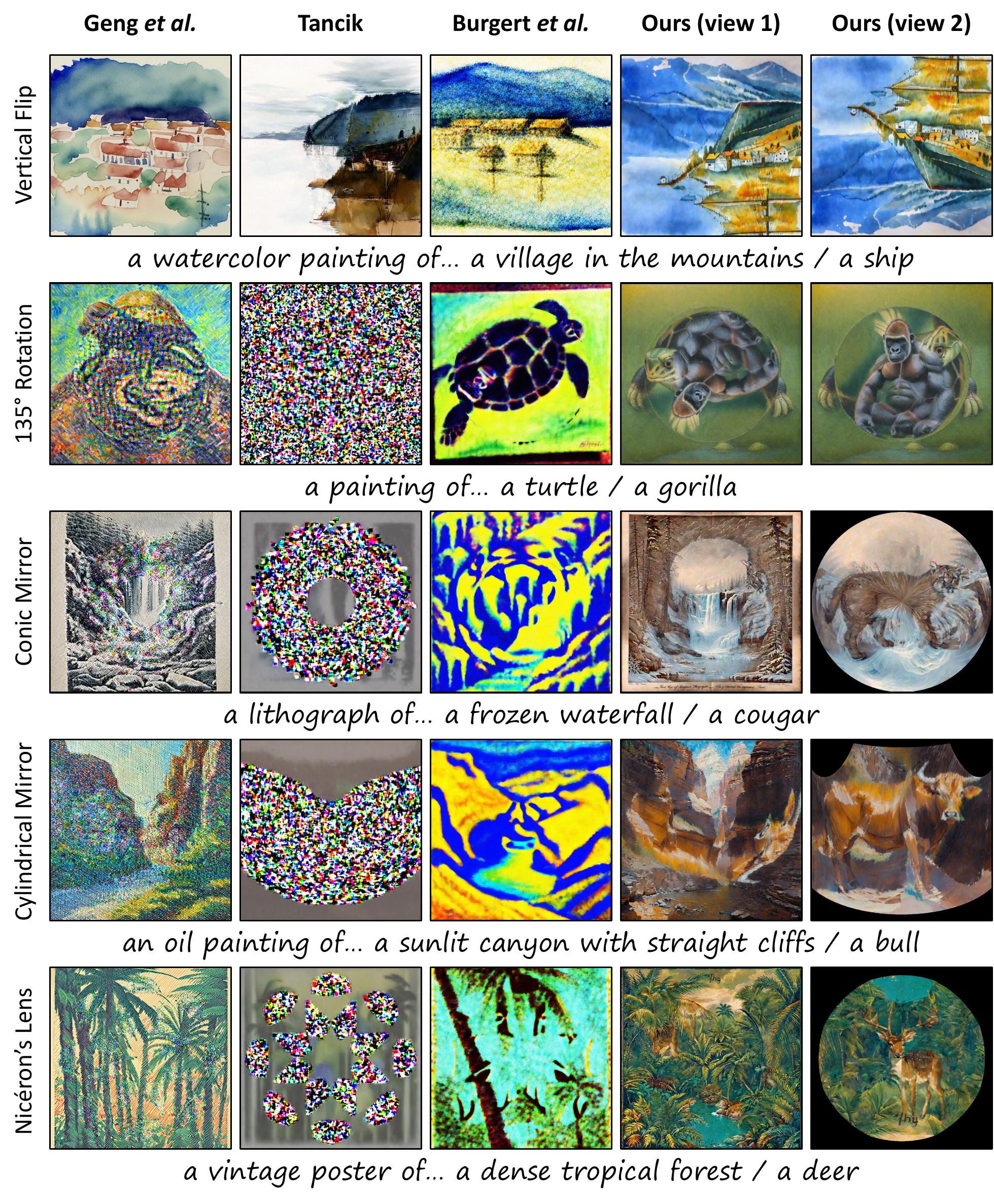}

   \caption{\textbf{Qualitative Comparison.} We compare our method against prior work for the considered views (\cref{subsec:view_considered}). Note that all transformations except the vertical flip are not supported by Geng \etal \cite{geng2024visualanagrams} and Tancik \cite{tancik2023illusion} due to their inherent limitations.}
   \label{fig:quali_comp}
\end{figure}

\subsection{Views Considered}\label{subsec:view_considered}

In this section we briefly describe the views considered for generating illusions, which are shown in Figure \ref{fig:laplacian_warping} (c).

\begin{itemize}
    \item {\textbf{2D transforms.} As in Visual Anagrams \cite{geng2024visualanagrams}, we generate vertical flip and 90$^\circ$ rotation illusions. Since our method can handle arbitrary view projections, we also create optical illusions involving 135$^\circ$ rotations.}
    \item {\textbf{Mirror cone.} A conic mirror placed at the center of an image reveals a hidden picture when viewed from the top.}
    \item {\textbf{Mirror cylinder.} A cylindrical mirror located over an image shows a hidden picture when viewed from an angle.}
    \item {\textbf{Nicéron's lens.} We replicate the setting described by Nicéron \cite{niceron1638perspective}. Given an image, and looking at it through a lens sculpted with polygonal faces, the irregular refraction will generate novel images.}
\end{itemize}

Note that in our examples, we always set one view as the identity view (\ie the canonical view), but this is not required. Transforms $\pi_i$ can be defined relative to a canonical space $\mathcal{C}$ without explicitly specifying the latter.

\subsection{Quantitative Results}

\begin{table}[b!]
\centering
\vspace{-0.5cm}
\resizebox{\columnwidth}{!}{%
\begin{tabular}{m{1.5cm}lcccccc}
\toprule
 & Method & $\mathcal{A}$ $\uparrow$ & $\mathcal{A}_{0.9}$ $\uparrow$ & $\mathcal{C}$ $\uparrow$ & $\mathcal{C}_{0.9}$ $\uparrow$ & FID $\downarrow$ & KID $\downarrow$ \\
\midrule
\multirow{5}{=}{\centering Vertical\ Flip}
& Geng \etal \cite{geng2024visualanagrams} & \cellcolor{red!20}0.306 & \cellcolor{yellow!20}0.340 & \cellcolor{orange!20}0.695 & \cellcolor{yellow!20}0.786 & 149.24 & 0.057 \\
& Tancik SD 3.5 \cite{tancik2023illusion} & \cellcolor{orange!20}0.306 & \cellcolor{red!20}0.349 & \cellcolor{yellow!20}0.693 & \cellcolor{red!20}0.806 & \cellcolor{orange!20}132.52 & \cellcolor{orange!20}0.049 \\
& Burgert \etal \cite{burgert2024diffusionillusions} & 0.281 & 0.324 & 0.679 & 0.778 & 219.84 & 0.115 \\
& SyncTweedies \cite{Kim2024SyncTweedies} & \cellcolor{yellow!20}0.302 & \cellcolor{orange!20}0.341 & \cellcolor{red!20}0.707 & \cellcolor{orange!20}0.801 & \cellcolor{yellow!20}132.62 & \cellcolor{yellow!20}0.054 \\
& \textbf{LookingGlass (ours)} & \textbf{0.297} & \textbf{0.338} & \textbf{0.680} & \textbf{0.779} & \cellcolor{red!20}\textbf{124.67} & \cellcolor{red!20}\textbf{0.049} \\
\midrule
\multirow{5}{=}{\centering $135^{\circ}$\ Rotation}
& Geng \etal \cite{geng2024visualanagrams} & 0.262 & 0.308 & 0.563 & 0.652 & 293.00 & 0.254 \\
& Tancik SD 3.5 \cite{tancik2023illusion} & 0.194 & 0.216 & 0.498 & 0.509 & 439.35 & 0.545 \\
& Burgert \etal \cite{burgert2024diffusionillusions} & \cellcolor{yellow!20}0.280 & \cellcolor{yellow!20}0.326 & \cellcolor{orange!20}0.654 & \cellcolor{orange!20}0.760 & \cellcolor{yellow!20}223.21 & \cellcolor{yellow!20}0.120 \\
& SyncTweedies \cite{Kim2024SyncTweedies} & \cellcolor{orange!20}0.283 & \cellcolor{orange!20}0.335 & \cellcolor{yellow!20}0.647 & \cellcolor{yellow!20}0.753 & \cellcolor{orange!20}166.03 & \cellcolor{orange!20}0.083 \\
& \textbf{LookingGlass (ours)} & \cellcolor{red!20}\textbf{0.295} & \cellcolor{red!20}\textbf{0.338} & \cellcolor{red!20}\textbf{0.666} & \cellcolor{red!20}\textbf{0.767} & \cellcolor{red!20}\textbf{129.74} & \cellcolor{red!20}\textbf{0.055} \\
\midrule
\multirow{5}{=}{\centering Cylindrical\ Mirror}
& Geng \etal \cite{geng2024visualanagrams} & 0.171 & 0.198 & 0.506 & 0.546 & 285.23 & 0.216 \\
& Tancik SD 3.5 \cite{tancik2023illusion} & 0.171 & 0.198 & 0.505 & 0.547 & 284.97 & 0.215 \\
& Burgert \etal \cite{burgert2024diffusionillusions} & \cellcolor{orange!20}0.261 & \cellcolor{orange!20}0.304 & \cellcolor{red!20}0.706 & \cellcolor{orange!20}0.795 & \cellcolor{yellow!20}229.65 & \cellcolor{yellow!20}0.138 \\
& SyncTweedies \cite{Kim2024SyncTweedies} & \cellcolor{yellow!20}0.241 & \cellcolor{yellow!20}0.284 & \cellcolor{yellow!20}0.673 & \cellcolor{yellow!20}0.763 & \cellcolor{orange!20}138.69 & \cellcolor{orange!20}0.082 \\
& \textbf{LookingGlass (ours)} & \cellcolor{red!20}\textbf{0.272} & \cellcolor{red!20}\textbf{0.318} & \cellcolor{orange!20}\textbf{0.698} & \cellcolor{red!20}\textbf{0.810} & \cellcolor{red!20}\textbf{130.27} & \cellcolor{red!20}\textbf{0.070} \\
\bottomrule
\end{tabular}

}

\caption{\textbf{Quantitative Comparison.} Sample quality is assessed with FID/KID against a reference dataset of 3.2k images generated from the same set of prompts. Image-prompt alignment is assessed using CLIP alignment score $\mathcal A$, and concealment score $\mathcal C$ introduced in \cite{geng2024visualanagrams}. While all methods achieve comparable results for the vertical flip, LookingGlass surpasses previous approaches on more complex transformations, including anamorphoses. Please see the supplementary material for more quantitative evaluations.}
\label{tab:main_comparison}
\end{table}

We compare our method quantitatively with Visual Anagrams \cite{geng2024visualanagrams}, Diffusion Illusions \cite{burgert2024diffusionillusions}, SyncTweedies \cite{Kim2024SyncTweedies}, and Tancik \cite{tancik2023illusion}. Following Geng \etal, CLIP is used \cite{radford2021learning} to measure the \textbf{alignment score} $\mathcal A$ and the \textbf{concealment score} $\mathcal C$. We generate 50 pairs of prompts and create images for these prompts using standard denoising (no optical illusion), as well as with our method. The results are reported in Table \ref{tab:main_comparison} for three tasks: simple vertical flip, a more complex $135^{\circ}$ rotation, and the cylindrical mirror anamorphosis. Our results are generated with Stable Diffusion 3.5 Medium on a Nvidia GeForce RTX 4090 GPU. Using 30 inference steps, with time-traveling between 20\% and 80\% of the diffusion process repeated twice, we generate an image pair in approximately 80 seconds.

\subsection{Ablations}

We perform an ablation of the proposed approach. \Cref{fig:ablation_comp} shows that warping the clean image estimate significantly improves image quality compared to warping the predicted noise. Additionally, the proposed VAE encoding/decoding and latent residual correction enhance detail preservation and reduce reconstruction errors. \Cref{tab:main_ablations} suggests that time traveling improves visual quality, as reflected by the FID metric, but may lead to reduced prompt alignment. Further qualitative ablations on time traveling and the effects of prioritizing a single view can be found in the Appendix.

\begin{table}[b!]
\centering
\resizebox{\columnwidth}{!}{
\begin{tabular}{lcccccc}
\toprule
& $\mathcal A$ $\uparrow$ & $\mathcal A_{0.9}$ $\uparrow$ & $\mathcal C$ $\uparrow$ & $\mathcal C_{0.9}$ $\uparrow$ & FID $\downarrow$ & KID $\downarrow$ \\
\midrule
Geng et al. SD 3 \cite{geng2024visualanagrams} & 0.219 & 0.249 & 0.516 & 0.571 & 335.24 & 0.297 \\
+ Final Estimate (Eq. \ref{eq:final_estimate}) & 0.273 & 0.320 & 0.657 & 0.757 & 171.61 & 0.106 \\
+ VAE (Sec. \ref{subsec: glva}) & 0.293 & 0.333 & 0.717 & 0.814 & 160.17 & 0.083 \\
+ LPW (Sec. \ref{subsec:lpw}) & \textbf{0.300} & \textbf{0.336} & \textbf{0.723} & 0.814 & 153.27 & \textbf{0.074} \\
+ Time Travel (Sec. \ref{subsec:designchoice}) & 0.295 & 0.331 & 0.716 & \textbf{0.816} & \textbf{150.01} & \textbf{0.074} \\
\bottomrule
\end{tabular}
}

\caption{\textbf{Ablation Study.} Starting from the baseline of Geng \etal \cite{geng2024visualanagrams}, we ablate the contributions introduced by our approach. We also show the 90th-percentile of the CLIP-based metrics, as we are interested in the best case performance.}
\label{tab:main_ablations}
\end{table}

\subsection{Qualitative Results}

Figures \ref{fig:quali_comp} and \ref{fig:results_ana} show example images generated using our method in combination with Stable Diffusion 3.5. \Cref{fig:reallife} features a real-life demonstration of the cylindrical example, confirming that the generated results function as intended in practice. Additional results for all three types of anamorphoses are provided in the Appendix. These results are selected from a set of curated prompts that worked best, as not all prompt combinations are expected to work well.

\subsection{User Study}\label{subsec:user_study}

\begin{figure}[h]
  \centering
  \includegraphics[width=\linewidth]{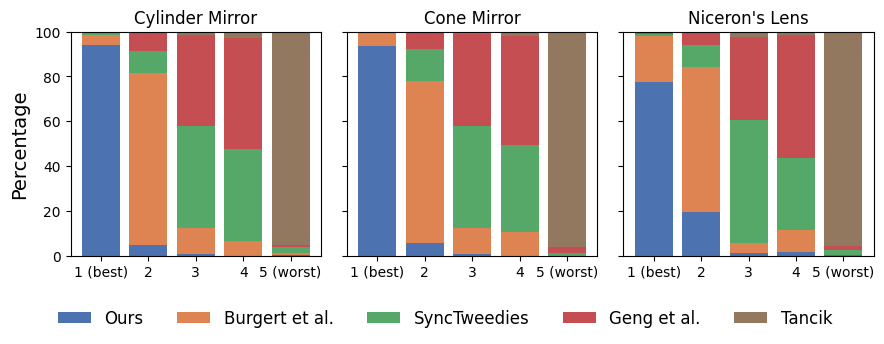}
   \caption{\textbf{User Study.} A survey of 27 participants shows that our method (blue) is consistently preferred to prior works.}
   \label{fig:userstudy}
   \vspace{-1em}
\end{figure}

We conducted a user study with 27 participants. A total of 10 prompt pairs, the same for all three types of illusions, were selected to generate results using different methods, with the same random seed (no hand-picked samples). Participants ranked the samples from 1 (best) to 5 (worst) based on prompt fidelity, style adherence and overall visual quality. The results are reported in \Cref{fig:userstudy}.

\section{Conclusion and Discussion}
\label{sec:conclusion}

We presented LookingGlass, a method for generating high-quality, ambiguous anamorphic illusions with latent RF models. Our contributions are twofold. First, we extended Geng \etal \cite{geng2024visualanagrams} to latent space models without introducing artifacts, enabling the first feed-forward generation of high-quality illusions with common models. This makes illusion generation more accessible and allows for possible integration with modern generative techniques like ControlNet and DreamBooth. Second, we introduced \textit{Laplacian Pyramid Warping} (LPW), a warping method that preserves fine details while handling extreme distortions. While essential for illusion generation, LPW is also compatible with pixel diffusion models and has potential applications in generative mesh texturing and panorama synthesis.

Despite these advances, selecting effective prompts remains a challenge, as not all prompts lead to high-quality illusions. Additionally, while our method is significantly more efficient than optimization-based approaches like Burgert \etal \cite{burgert2024diffusionillusions}, it is computationally more expensive than Geng \etal \cite{geng2024visualanagrams} due to the VAE intermediate steps. We leave this to future work.

{
    \small
    \bibliographystyle{ieeenat_fullname}
    \bibliography{references}

\begin{thebibliography}{45}
\providecommand{\natexlab}[1]{#1}
\providecommand{\url}[1]{\texttt{#1}}
\expandafter\ifx\csname urlstyle\endcsname\relax
  \providecommand{\doi}[1]{doi: #1}\else
  \providecommand{\doi}{doi: \begingroup \urlstyle{rm}\Url}\fi

\bibitem[at~StabilityAI(2023)]{DeepFloydIF}
DeepFloyd~Lab at StabilityAI.
\newblock {DeepFloyd IF}: a novel state-of-the-art open-source text-to-image
  model with a high degree of photorealism and language understanding.
\newblock \url{https://www.deepfloyd.ai/deepfloyd-if}, 2023.
\newblock Retrieved on 2023-11-08.

\bibitem[Bansal et~al.(2023)Bansal, Chu, Schwarzschild, Sengupta, Goldblum,
  Geiping, and Goldstein]{bansal2023universal}
Arpit Bansal, Hong-Min Chu, Avi Schwarzschild, Soumyadip Sengupta, Micah
  Goldblum, Jonas Geiping, and Tom Goldstein.
\newblock Universal guidance for diffusion models.
\newblock In \emph{Proceedings of the IEEE/CVF Conference on Computer Vision
  and Pattern Recognition}, pages 843--852, 2023.

\bibitem[Bar-Tal et~al.(2023)Bar-Tal, Yariv, Lipman, and
  Dekel]{bar2023multidiffusion}
Omer Bar-Tal, Lior Yariv, Yaron Lipman, and Tali Dekel.
\newblock Multidiffusion: Fusing diffusion paths for controlled image
  generation.
\newblock \emph{arXiv preprint arXiv:2302.08113}, 2023.

\bibitem[Burgert et~al.(2024)Burgert, Li, Leite, Ranasinghe, and
  Ryoo]{burgert2024diffusionillusions}
Ryan Burgert, Xiang Li, Abe Leite, Kanchana Ranasinghe, and Michael Ryoo.
\newblock Diffusion illusions: Hiding images in plain sight.
\newblock In \emph{ACM SIGGRAPH 2024 Conference Papers}, New York, NY, USA,
  2024. Association for Computing Machinery.

\bibitem[Burt and Adelson(1983{\natexlab{a}})]{burt1983laplacian}
P. Burt and E. Adelson.
\newblock The laplacian pyramid as a compact image code.
\newblock \emph{IEEE Transactions on Communications}, 31\penalty0 (4):\penalty0
  532--540, 1983{\natexlab{a}}.

\bibitem[Burt and Adelson(1983{\natexlab{b}})]{burt1983multiresolution}
Peter~J. Burt and Edward~H. Adelson.
\newblock A multiresolution spline with application to image mosaics.
\newblock \emph{ACM Trans. Graph.}, 2\penalty0 (4):\penalty0 217–236,
  1983{\natexlab{b}}.

\bibitem[Burt and Adelson(1987)]{burt1987laplacian}
Peter~J Burt and Edward~H Adelson.
\newblock The laplacian pyramid as a compact image code.
\newblock In \emph{Readings in computer vision}, pages 671--679. Elsevier,
  1987.

\bibitem[Chandra et~al.(2022)Chandra, Li, Tenenbaum, and
  Ragan-Kelley]{chandra2022designing}
Kartik Chandra, Tzu-Mao Li, Joshua Tenenbaum, and Jonathan Ragan-Kelley.
\newblock Designing perceptual puzzles by differentiating probabilistic
  programs.
\newblock In \emph{Special Interest Group on Computer Graphics and Interactive
  Techniques Conference Proceedings (SIGGRAPH '22 Conference Proceedings)},
  2022.

\bibitem[Chang et~al.(2024)Chang, Tang, Gross, and Azevedo]{chang2024how}
Pascal Chang, Jingwei Tang, Markus Gross, and Vinicius~C. Azevedo.
\newblock How i warped your noise: a temporally-correlated noise prior for
  diffusion models.
\newblock In \emph{The Twelfth International Conference on Learning
  Representations}, 2024.

\bibitem[Chen et~al.(2024)Chen, Geng, and Owens]{chen2024images}
Ziyang Chen, Daniel Geng, and Andrew Owens.
\newblock Images that sound: Composing images and sounds on a single canvas.
\newblock \emph{Neural Information Processing Systems (NeurIPS)}, 2024.

\bibitem[Chi et~al.(2008)Chi, Lee, Qu, and Wong]{chi2008self}
Ming-Te Chi, Tong-Yee Lee, Yingge Qu, and Tien-Tsin Wong.
\newblock Self-animating images: Illusory motion using repeated asymmetric
  patterns.
\newblock \emph{ACM Transactions on Graphics (SIGGRAPH 2008 issue)},
  27\penalty0 (3):\penalty0 62:1--62:8, 2008.

\bibitem[Chu et~al.(2010)Chu, Hsu, Mitra, Cohen-Or, Wong, and
  Lee]{chu2010camouflage}
Hung-Kuo Chu, Wei-Hsin Hsu, Niloy~J. Mitra, Daniel Cohen-Or, Tien-Tsin Wong,
  and Tong-Yee Lee.
\newblock Camouflage images.
\newblock \emph{ACM Trans. Graph.}, 29\penalty0 (4), 2010.

\bibitem[Clark(1976)]{Clark1976}
James~H. Clark.
\newblock Hierarchical geometric models for visible surface algorithms.
\newblock \emph{Commun. ACM}, 19\penalty0 (10):\penalty0 547–554, 1976.

\bibitem[Esser et~al.(2024)Esser, Kulal, Blattmann, Entezari, M{\"u}ller,
  Saini, Levi, Lorenz, Sauer, Boesel, et~al.]{esser2024scaling}
Patrick Esser, Sumith Kulal, Andreas Blattmann, Rahim Entezari, Jonas
  M{\"u}ller, Harry Saini, Yam Levi, Dominik Lorenz, Axel Sauer, Frederic
  Boesel, et~al.
\newblock Scaling rectified flow transformers for high-resolution image
  synthesis.
\newblock In \emph{Forty-first International Conference on Machine Learning},
  2024.

\bibitem[Ewins et~al.(1998)Ewins, Waller, White, and Lister]{ewins1998mip}
Jon~P Ewins, Marcus~D Waller, Martin White, and Paul~F Lister.
\newblock Mip-map level selection for texture mapping.
\newblock \emph{IEEE Transactions on Visualization and Computer Graphics},
  4\penalty0 (4):\penalty0 317--329, 1998.

\bibitem[Feng et~al.(2024)Feng, Sanjay, Lutz, AlBahar, Ge, and
  Huang]{feng2024illusion3d3dmultiviewillusion}
Yue Feng, Vaibhav Sanjay, Spencer Lutz, Badour AlBahar, Songwei Ge, and Jia-Bin
  Huang.
\newblock Illusion3d: 3d multiview illusion with 2d diffusion priors.
\newblock 2024.

\bibitem[Geng et~al.(2024{\natexlab{a}})Geng, Park, and
  Owens]{geng2024factorized}
Daniel Geng, Inbum Park, and Andrew Owens.
\newblock Factorized diffusion: Perceptual illusions by noise decomposition.
\newblock In \emph{European Conference on Computer Vision (ECCV)},
  2024{\natexlab{a}}.

\bibitem[Geng et~al.(2024{\natexlab{b}})Geng, Park, and
  Owens]{geng2024visualanagrams}
Daniel Geng, Inbum Park, and Andrew Owens.
\newblock Visual anagrams: Generating multi-view optical illusions with
  diffusion models.
\newblock In \emph{Conference on Computer Vision and Pattern Recognition
  (CVPR)}, 2024{\natexlab{b}}.

\bibitem[Heeger and Bergen(1995)]{heeger1995pyramidtexture}
David~J. Heeger and James~R. Bergen.
\newblock Pyramid-based texture analysis/synthesis.
\newblock In \emph{Proceedings of the 22nd Annual Conference on Computer
  Graphics and Interactive Techniques}, page 229–238, New York, NY, USA,
  1995. Association for Computing Machinery.

\bibitem[Ho and Salimans(2022)]{ho2022classifier}
Jonathan Ho and Tim Salimans.
\newblock Classifier-free diffusion guidance.
\newblock \emph{arXiv preprint arXiv:2207.12598}, 2022.

\bibitem[Hsiao et~al.(2018)Hsiao, Huang, and Chu]{hsiao2018multiview}
Kai-Wen Hsiao, Jia-Bin Huang, and Hung-Kuo Chu.
\newblock Multi-view wire art.
\newblock \emph{ACM Trans. Graph.}, 37\penalty0 (6), 2018.

\bibitem[Jurgis and Strachan(1977)]{jurgis1977anamorphic}
Baltrusaitis Jurgis and WJ Strachan.
\newblock \emph{Anamorphic Art}.
\newblock Abrams, New York, 1977.

\bibitem[Kim et~al.(2024)Kim, Koo, Yeo, and Sung]{Kim2024SyncTweedies}
Jaihoon Kim, Juil Koo, Kyeongmin Yeo, and Minhyuk Sung.
\newblock Synctweedies: A general generative framework based on synchronized
  diffusions.
\newblock \emph{arXiv:2403.14370}, 2024.

\bibitem[Kuchel(1979)]{kuchel1979anamorphoscope}
Philip Kuchel.
\newblock Anamorphoscopes: A visual aid for circle inversion.
\newblock \emph{The Mathematical Gazette}, 63:\penalty0 82, 1979.

\bibitem[Laabs(2023)]{monster2023controlnetqr}
Monster Laabs.
\newblock Controlnet qr code monster v2 for sd-1.5, 2023.

\bibitem[Lowe(2004)]{lowe2004distinctive}
David~G Lowe.
\newblock Distinctive image features from scale-invariant keypoints.
\newblock \emph{International journal of computer vision}, 60:\penalty0
  91--110, 2004.

\bibitem[Lugmayr et~al.(2022)Lugmayr, Danelljan, Romero, Yu, Timofte, and
  Van~Gool]{lugmayr2022repaint}
Andreas Lugmayr, Martin Danelljan, Andres Romero, Fisher Yu, Radu Timofte, and
  Luc Van~Gool.
\newblock Repaint: Inpainting using denoising diffusion probabilistic models.
\newblock In \emph{Proceedings of the IEEE/CVF conference on computer vision
  and pattern recognition}, pages 11461--11471, 2022.

\bibitem[Nakajima and Yamaguchi(2004)]{nakajima2004picture}
Mizuho Nakajima and Yasushi Yamaguchi.
\newblock Picture illusion by overlap.
\newblock In \emph{ACM SIGGRAPH 2004 Posters}, page~56, New York, NY, USA,
  2004. Association for Computing Machinery.

\bibitem[Niceron(1638)]{niceron1638perspective}
Jean~Fran{\c{c}}ois Niceron.
\newblock \emph{La perspective curieuse}.
\newblock P. Billaine, Paris, 1638.

\bibitem[Oliva et~al.(2006)Oliva, Torralba, and Schyns]{oliva2006hybrid}
Aude Oliva, Antonio Torralba, and Philippe~G. Schyns.
\newblock Hybrid images.
\newblock In \emph{ACM SIGGRAPH 2006 Papers}, page 527–532, New York, NY,
  USA, 2006. Association for Computing Machinery.

\bibitem[Papas et~al.(2012)Papas, Houit, Nowrouzezahrai, Gross, and
  Jarosz]{papas12magic}
Marios Papas, Thomas Houit, Derek Nowrouzezahrai, Markus Gross, and Wojciech
  Jarosz.
\newblock The magic lens: Refractive steganography.
\newblock \emph{ACM Transactions on Graphics (Proceedings of SIGGRAPH Asia)},
  31\penalty0 (6), 2012.

\bibitem[Perroni-Scharf and Rusinkiewicz(2023)]{perronischarf2023printable}
Maxine Perroni-Scharf and Szymon Rusinkiewicz.
\newblock Constructing printable surfaces with view-dependent appearance.
\newblock In \emph{ACM SIGGRAPH 2023 Conference Proceedings}, New York, NY,
  USA, 2023. Association for Computing Machinery.

\bibitem[Radford et~al.(2021)Radford, Kim, Hallacy, Ramesh, Goh, Agarwal,
  Sastry, Askell, Mishkin, Clark, et~al.]{radford2021learning}
Alec Radford, Jong~Wook Kim, Chris Hallacy, Aditya Ramesh, Gabriel Goh,
  Sandhini Agarwal, Girish Sastry, Amanda Askell, Pamela Mishkin, Jack Clark,
  et~al.
\newblock Learning transferable visual models from natural language
  supervision.
\newblock In \emph{International conference on machine learning}, pages
  8748--8763. PMLR, 2021.

\bibitem[Ranjan and Black(2017)]{ranjan2017optical}
Anurag Ranjan and Michael~J Black.
\newblock Optical flow estimation using a spatial pyramid network.
\newblock In \emph{Proceedings of the IEEE conference on computer vision and
  pattern recognition}, pages 4161--4170, 2017.

\bibitem[Robbins(1992)]{robbins1992empirical}
Herbert~E Robbins.
\newblock An empirical bayes approach to statistics.
\newblock In \emph{Breakthroughs in Statistics: Foundations and basic theory},
  pages 388--394. Springer, 1992.

\bibitem[Rombach et~al.(2022)Rombach, Blattmann, Lorenz, Esser, and
  Ommer]{rombach2022high}
Robin Rombach, Andreas Blattmann, Dominik Lorenz, Patrick Esser, and Bj{\"o}rn
  Ommer.
\newblock High-resolution image synthesis with latent diffusion models.
\newblock In \emph{Proceedings of the IEEE/CVF conference on computer vision
  and pattern recognition}, pages 10684--10695, 2022.

\bibitem[Tancik(2023)]{tancik2023illusion}
Matthew Tancik.
\newblock Illusion diffusion: optical illusions using stable diffusion.
\newblock \url{https://github.com/tancik/Illusion-Diffusion}, 2023.

\bibitem[Ugleh(2023)]{ugleh2023spiral}
Ugleh.
\newblock
  \url{https://www.reddit.com/r/StableDiffusion/comments/16ew9fz/spiral_town_different_approach_to_qr_monster/},
  2023.

\bibitem[Vaulezard(1630)]{vaulezard1630perspective}
Jean-Louis Vaulezard.
\newblock \emph{Perspective cilindrique et conique, concave et convexe ou
  traité des apparences vueus par le moyen des miroirs}.
\newblock J. Jacquin, Paris, 1630.

\bibitem[Wang et~al.(2023{\natexlab{a}})Wang, Kontkanen, Curless, Seitz,
  Kemelmacher, Mildenhall, Srinivasan, Verbin, and
  Holynski]{wang2023generativepowers}
Xiaojuan Wang, Janne Kontkanen, Brian Curless, Steve Seitz, Ira Kemelmacher,
  Ben Mildenhall, Pratul Srinivasan, Dor Verbin, and Aleksander Holynski.
\newblock Generative powers of ten.
\newblock \emph{arXiv preprint arXiv:2312.02149}, 2023{\natexlab{a}}.

\bibitem[Wang et~al.(2023{\natexlab{b}})Wang, Yu, and Zhang]{wang2023ddnm}
Yinhuai Wang, Jiwen Yu, and Jian Zhang.
\newblock Zero-shot image restoration using denoising diffusion null-space
  model.
\newblock \emph{The Eleventh International Conference on Learning
  Representations}, 2023{\natexlab{b}}.

\bibitem[Williams(1983)]{Williams1983}
Lance Williams.
\newblock Pyramidal parametrics.
\newblock \emph{SIGGRAPH Comput. Graph.}, 17\penalty0 (3):\penalty0 1–11,
  1983.

\bibitem[Yu et~al.(2023)Yu, Wang, Zhao, Ghanem, and Zhang]{yu2023freedom}
Jiwen Yu, Yinhuai Wang, Chen Zhao, Bernard Ghanem, and Jian Zhang.
\newblock Freedom: Training-free energy-guided conditional diffusion model.
\newblock In \emph{Proceedings of the IEEE/CVF International Conference on
  Computer Vision}, pages 23174--23184, 2023.

\bibitem[Zhang et~al.(2023{\natexlab{a}})Zhang, Rao, and
  Agrawala]{zhang2023adding}
Lvmin Zhang, Anyi Rao, and Maneesh Agrawala.
\newblock Adding conditional control to text-to-image diffusion models,
  2023{\natexlab{a}}.

\bibitem[Zhang et~al.(2023{\natexlab{b}})Zhang, Song, Huang, Chen, and
  yu~Liu]{zhange2023diffcollage}
Qinsheng Zhang, Jiaming Song, Xun Huang, Yongxin Chen, and Ming yu Liu.
\newblock Diffcollage: Parallel generation of large content with diffusion
  models.
\newblock In \emph{CVPR}, 2023{\natexlab{b}}.

\end{thebibliography}
}

\clearpage
\setcounter{section}{0}
\renewcommand*{\thesection}{\Alph{section}}
\maketitlesupplementary

\section{Laplacian Pyramid Warping}

In this section, we provide additional details about \cref{subsec:lpw}. In \Cref{subsec:suppl_lpw_forward} and \ref{subsec:suppl_lpw_backward}, we illustrate the forward and inverse warping with the example of the conic mirror, and cover some implementation subtleties for the inverse operation. \Cref{subsec:suppl_lpw_blending} explains how we properly blend pyramid levels in the case of partial views. Lastly, we provide pseudo-code for Laplacian Pyramid Warping in \Cref{subsec:suppl_pseudocode}.

\begin{figure}[h]
  \centering
  \centering
   \includegraphics[width=1.0\linewidth]{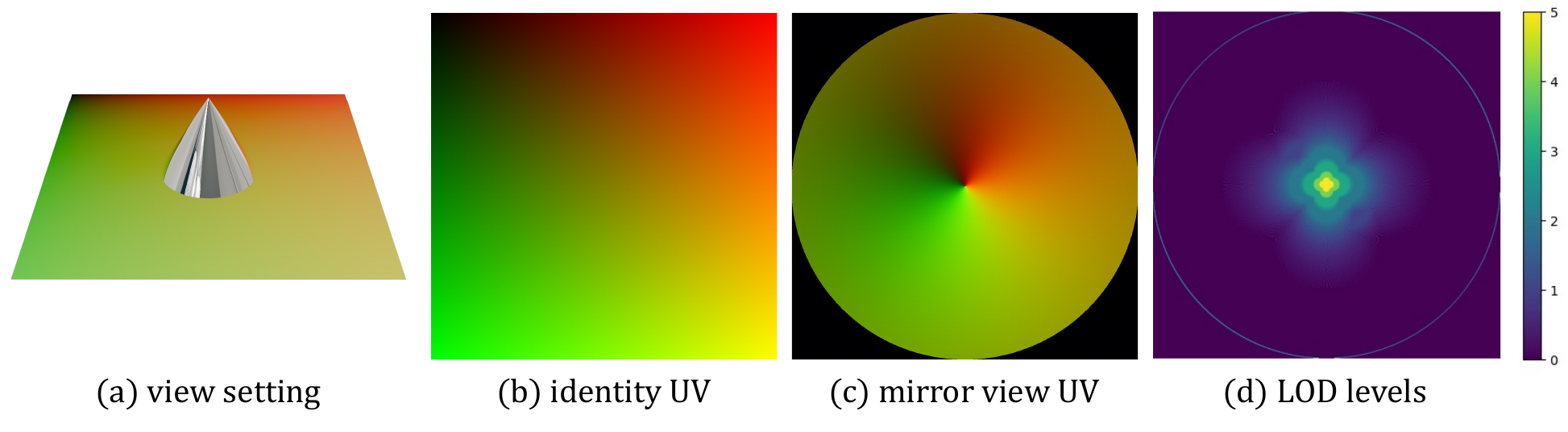}

   \caption{\textbf{Conic mirror view.} Rendering the scene (a) with a top down camera yields the UV map (c) for the conic mirror case. Using \cref{eq:lod}, the associated level-of-detail map (d) is computed.}
   \label{fig:suppl_cone_view}
\end{figure}

\subsection{Forward Warping}\label{subsec:suppl_lpw_forward}

\begin{figure}[h]
  \centering
  \centering
   \includegraphics[width=1.0\linewidth]{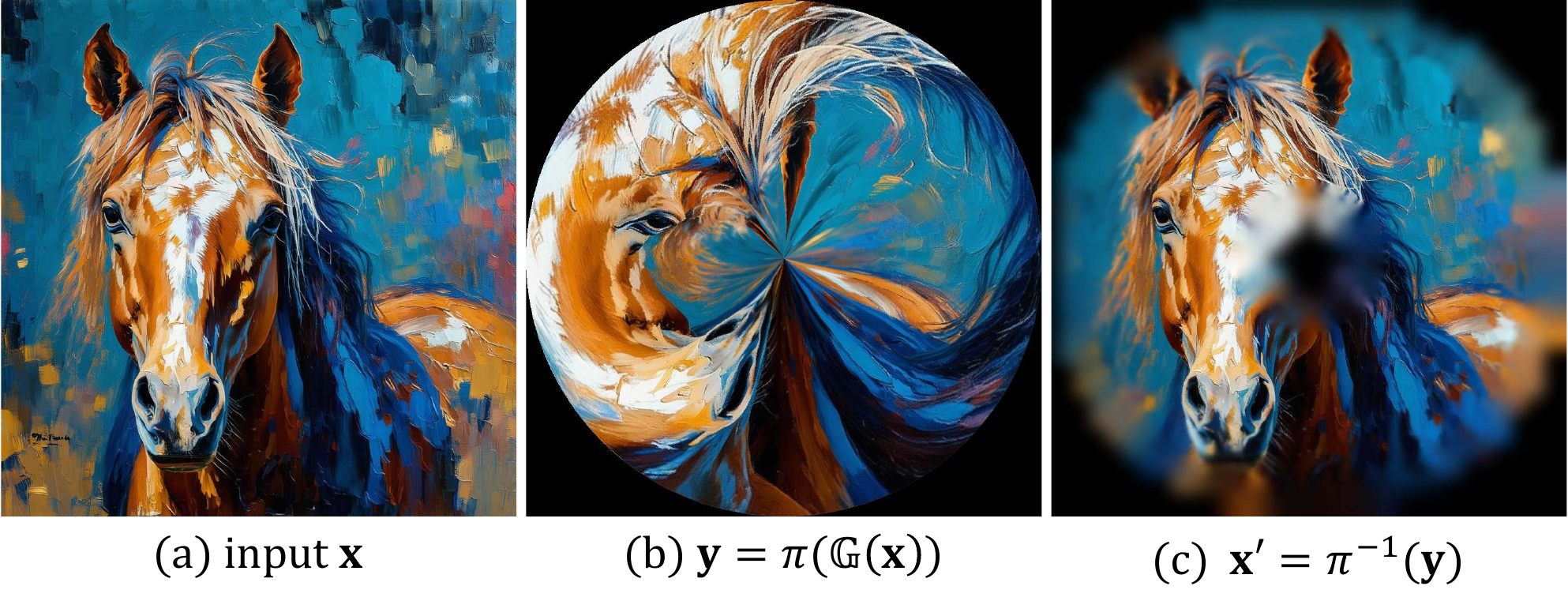}
    \vspace{-0.5cm}
   \caption{\textbf{Warp example.} Given an image (a), we warp it to the conic mirror view (b), and warp back to the original view (c). Our multi-scale approach warps values to different frequency bands based on the local compression of the view function $\pi$.}
   \label{fig:suppl_warp_result}
\end{figure}

We consider the conic mirror example and its corresponding warping function (\cref{fig:suppl_cone_view}). The forward warping operation was explained in \cref{subsec:lpw}. \Cref{fig:suppl_warp_result}.b) illustrates the result of applying forward warping to an image. In \Cref{fig:suppl_lpw_interm_steps}.b), we visualize the pyramid levels in the identity view, highlighting in white the pixels used in the forward warping. For instance, pixels closer to the cone's center are mapped to an outer ring in the identity view. Because the view function is locally more compressed for these pixels, they are sampled from a higher level of the pyramid (\ie, lower resolution).

\subsection{Inverse Warping}\label{subsec:suppl_lpw_backward}

\begin{figure}[h]
  \centering
  \centering
   \includegraphics[width=1.0\linewidth]{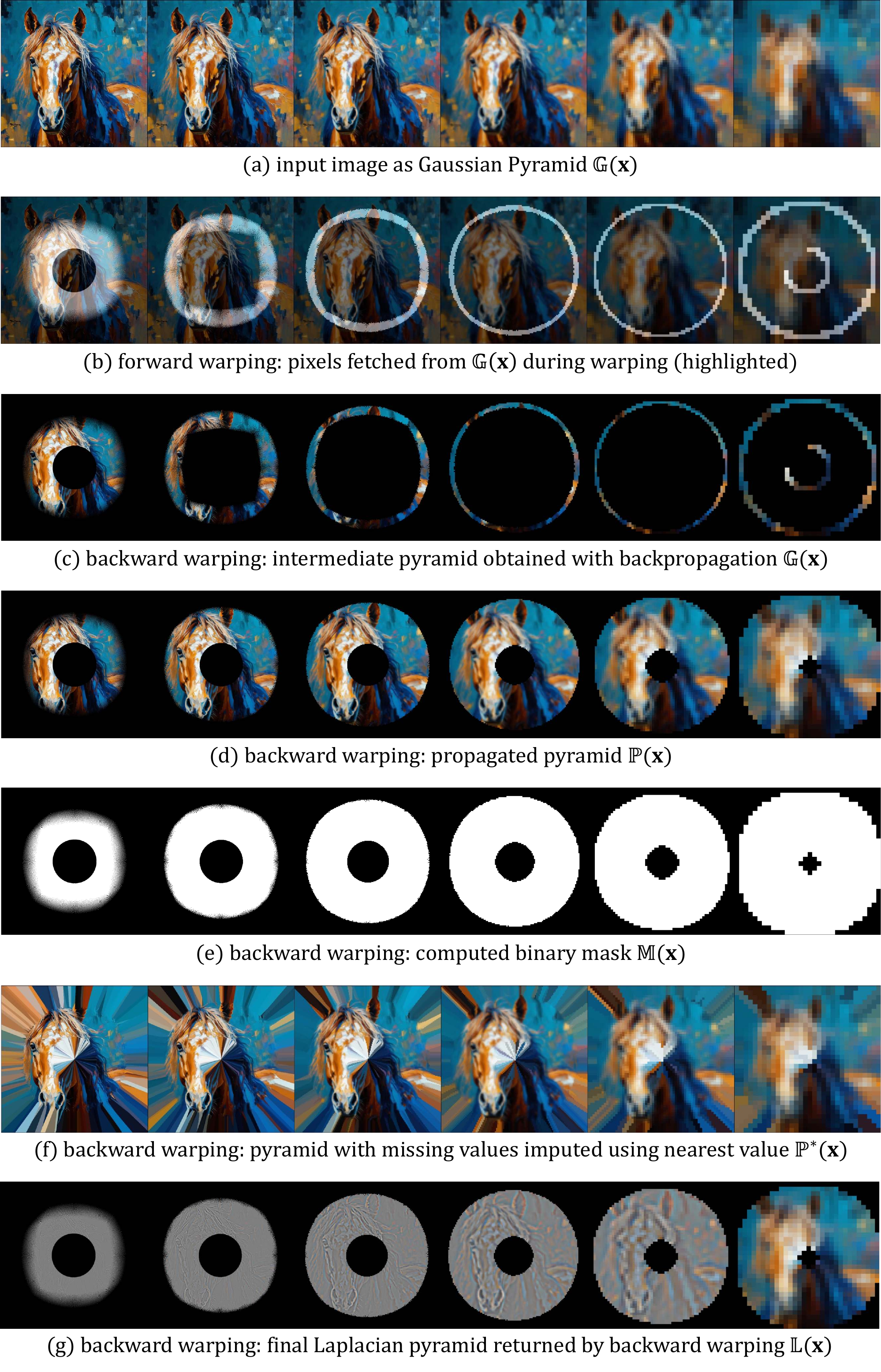}

   \caption{\textbf{Intermediate pyramids.} We visualize some intermediate pyramids that appear in forward (a, b) and backward warping (c-g). Refer to the text and \cref{fig:suppl_lpw_backward} for more context.}
   \label{fig:suppl_lpw_interm_steps}
\end{figure}

\Cref{fig:suppl_warp_result}.c) shows the result of warping the transformed image back to the identity view: pixels closer to the center of the cone in the warped view naturally map back to a lower frequency band in the identity view, occupying larger regions at the boundary in the reconstructed image. Rigorously speaking, the output of the inverse warping is a Laplacian pyramid (see \cref{fig:suppl_lpw_interm_steps}.g), not an image.

\begin{figure*}[t]
  \centering
  \centering
   \includegraphics[width=1.0\linewidth]{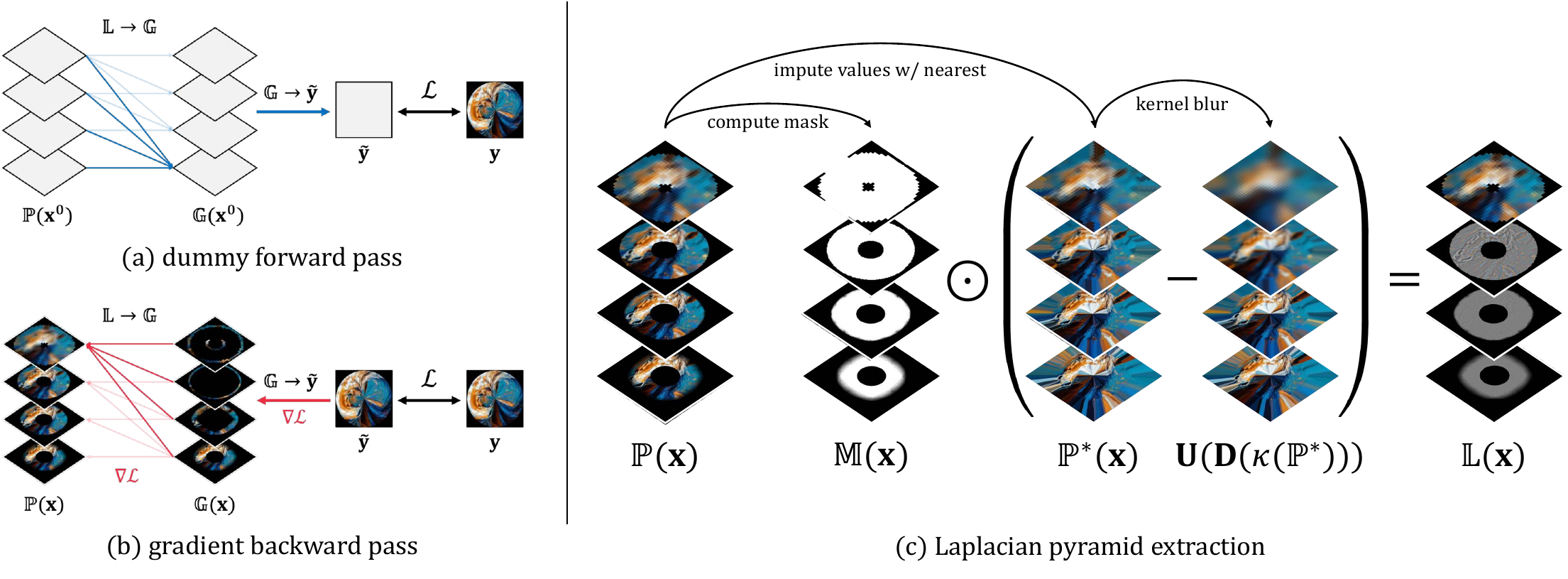}

   \caption{\textbf{Detailed look at inverse warping.} We provide an illustration to \Cref{eq:inversion}, explaining how to obtain the final Laplacian pyramid using back-propagation, as mentioned in the main paper.}
   \label{fig:suppl_lpw_backward}
\end{figure*}

\paragraph{Implementation details.} We provide a more detailed illustration of the inverse operation in \Cref{fig:suppl_lpw_backward}. The subfigures show:
\begin{enumerate}
    \item[a)] Starting from a dummy pyramid $\mathbb P (\mathbf{x}^0)$, we perform a Laplacian-to-Gaussian pyramid conversion to get $\mathbb G (\mathbf{x}^0)$ (\ie \textit{reparameterization}). Forward warping is then applied to obtain a warped dummy image $\tilde{\mathbf{y}}$.
    \item[b)] An $L_2$ loss is computed between $\tilde{\mathbf{y}}$ and $\mathbf{y}$, the image we wish to warp back. The gradients populate each level of the dummy pyramid, yielding $\mathbb P (\mathbf{x})$. The reparameterization ensures that gradients from each level flows to all levels above it.
    \item[c)] Lastly, we extract the final Laplacian pyramid $\mathbb L (\mathbf{x})$ from $\mathbb P (\mathbf{x})$ following
    \begin{equation}\label{eq:suppl_backward}
        \mathbb L (\mathbf{x}) = \mathbb M (\mathbf{x}) \odot \left( \mathbb P^* (\mathbf{x}) - \mathbf{U}(\mathbf{D}(\kappa(\mathbb P^* (\mathbf{x}))))\right),
    \end{equation}
    where $\mathbb M$ is a pyramid of binary masks indicating pixels that have gradients, $\mathbb P^*$ is the result of imputing missing values in $\mathbb P$ with nearest color (more details below).
\end{enumerate}

\begin{figure}[t]
  \centering
  \centering
   \includegraphics[width=1.0\linewidth]{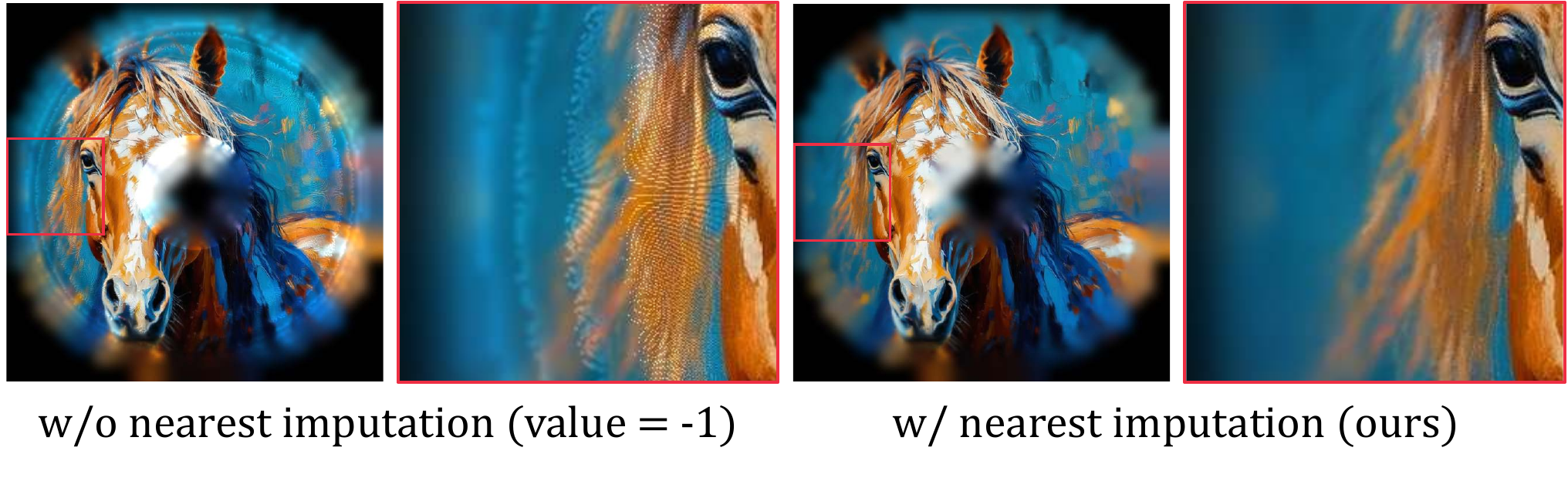}
\vspace{-0.5cm}
   \caption{\textbf{Imputation.} We use nearest neighbor imputation for pixels in $\mathbb P(\mathbf
   x)$ that has no colors (\ie did not receive gradients). We consider images with values in [-1, 1].}
   \label{fig:suppl_impute}
\end{figure}

\paragraph{Imputation.} It is important to impute the pixels that received no gradients with meaningful colors. Otherwise, if left black, the pyramid computation will pick up these discontinuities at the mask boundaries as high-frequency details, leading to incorrect values in those regions. We opt for a simple nearest neighbor imputation that fills pixels that have no gradients with the nearest pixel value. \Cref{fig:suppl_impute} illustrates the difference between no imputation (default value 0) and our nearest imputation.

\begin{figure}[h]
  \centering
  \centering
   \includegraphics[width=1.0\linewidth]{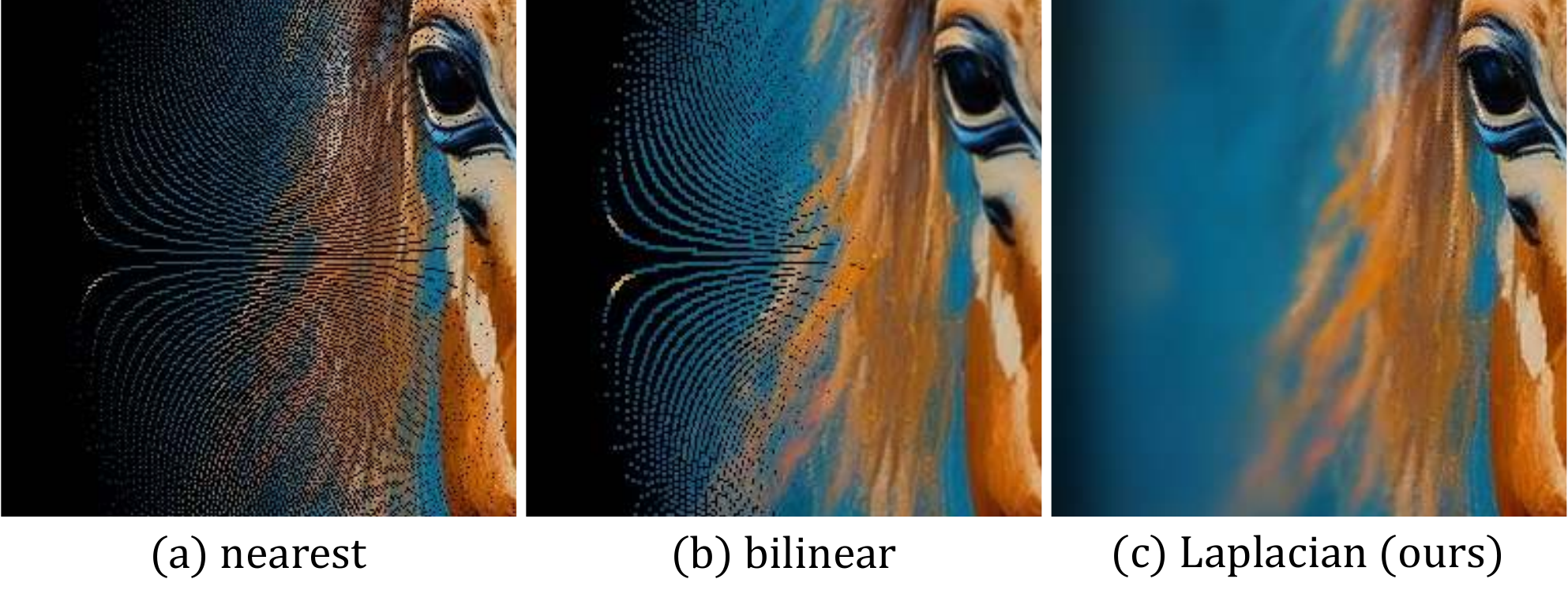}
\vspace{-0.5cm}
   \caption{\textbf{Backward warping baseline comparison.} In regions where the view function is compressing, standard interpolation like nearest (a) or bilinear interpolation (b) creates holes, while our Laplacian Pyramid Warping ensures smooth results (c).}
   \label{fig:suppl_warp_comp}
\end{figure}

\paragraph{Comparison with baselines.} In \Cref{fig:suppl_warp_comp}, we compare our Laplacian-based inverse warping with standard warping using nearest or bilinear interpolation. Because the view function has extreme compression around the cone center, this translates to missing values on the outer ring when warping back with nearest or bilinear, as some pixels in the identity view are not sampled during forward warping due to the compression. We showed in \cref{fig:laplacian_pb} that this can lead to artifacts in the generated image.

It is worth noting that another way to avoid missing values could be to do a \textit{forward} warping with the \textit{inverse} UV map. However, the inverse map is usually not trivial to compute. Additional problems arise when the mapping is not bijective or has discontinuities, which can be quite tricky to solve in a robust way. In contrast, our method automatically handles these cases, and only requires the forward UV map, which is easily obtainable through simple rendering of the desired view.

\subsection{Pyramid Blending}\label{subsec:suppl_lpw_blending}

As explained in \cref{subsec:lpw}, special care needs to be taken when blending the pyramids.

\paragraph{Blending with partial views.} In standard Laplacian Pyramid Blending, the blended pyramid is obtained by averaging each level of the input pyramids. Given two pyramids $\mathbb L^0, \mathbb L^1$, the level $k$ of the blended pyramid $\mathbb L$ is given by:
\begin{equation}
    \mathbb L_k = \frac{1}{2} \left(\mathbb L_k^0 + \mathbb L_k^1\right).
\end{equation}

In our case, Laplacian pyramids come from inverse warping of views, and might look like \cref{fig:suppl_warp_result}.c), with missing values. In this case, the average should only be computed over defined pixels. Assuming each level $\mathbb L_k$ is associated with a binary mask $\mathbb M_k$, the blending is:
\begin{equation}
    \mathbb L_k = \frac{1}{\mathbb M_k^0 + \mathbb M_k^1}\left(\mathbb M_k^0 \odot \mathbb L_k^0 + \mathbb M_k^1 \odot \mathbb L_k^1\right).
\end{equation}

In practice, we map missing values to $\texttt{torch.nan}$, and use $\texttt{torch.nanmean()}$ to perform averaging.

\paragraph{Detail-preserving averaging.} Another issue with averaging in general is that it reduces variance and washes out details. While this is already improved with the Laplacian pyramid, averaging still leads to the loss of sharp details. Given two pixel values $x, y \in [-1, 1]$, we define a normal averaging $\texttt{avg}$ and a value-weighted averaging $\texttt{vavg}$:
\begin{equation}\label{eq:suppl_vavg}
    \texttt{avg}(x,y) = \frac{1}{2}(x+y), \ \texttt{vavg}(x,y) = \frac{|x|x + |y|y}{|x|+|y|}
\end{equation}

The value-weighted average will give more weights to extreme pixel values, hence preserving better the value range. In the results shown, we linearly interpolate between the two types of averaging through a parameter $\alpha \in [0,1]$:
\begin{equation}\label{eq: alpha_interp}
    z = \texttt{avg}(x, y) + \alpha (\texttt{vavg}(x, y) - \texttt{avg}(x, y)).
\end{equation}

\begin{figure}[h]
  \centering
  \centering
   \includegraphics[width=1.0\linewidth]{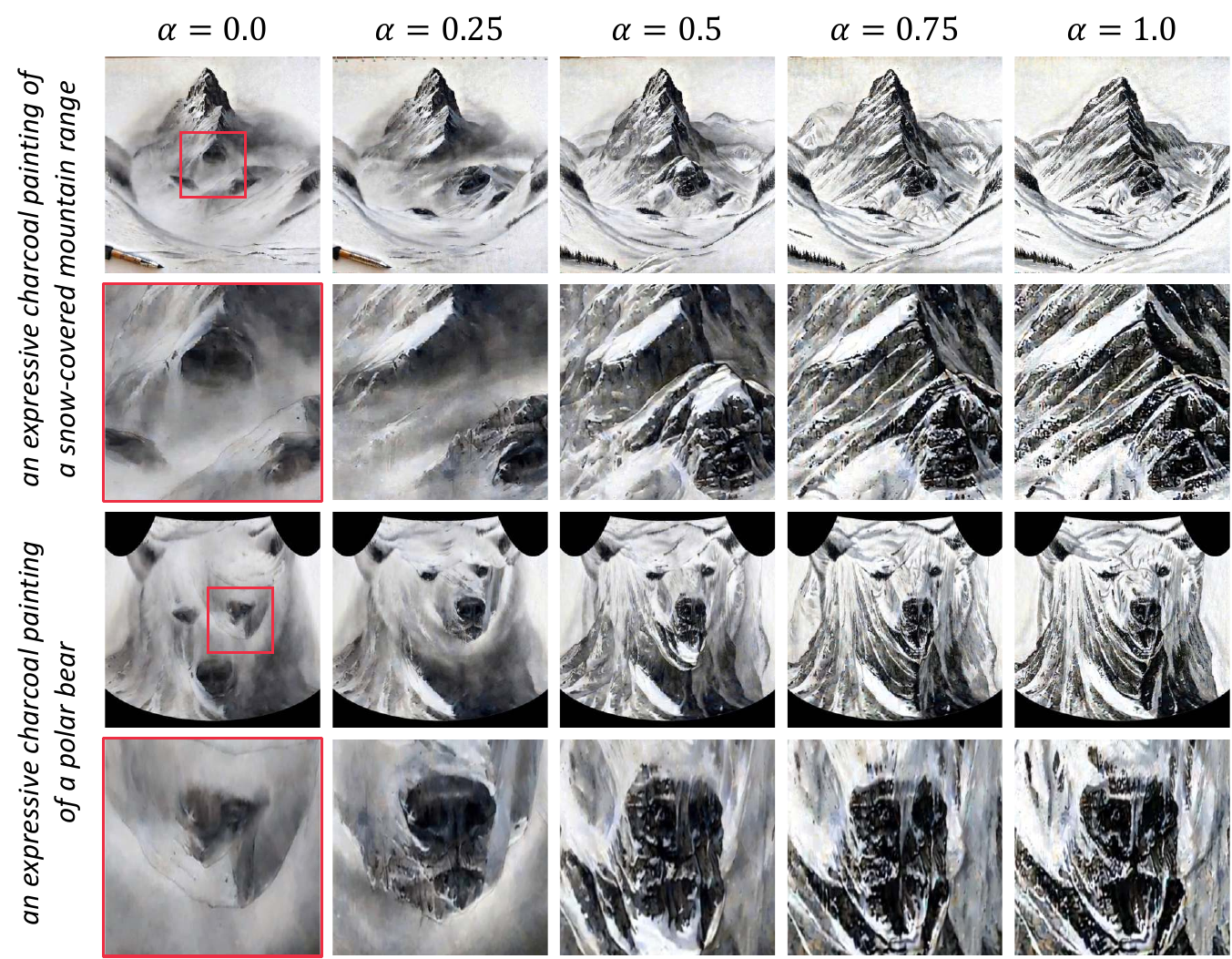}
   \includegraphics[width=1.0\linewidth]{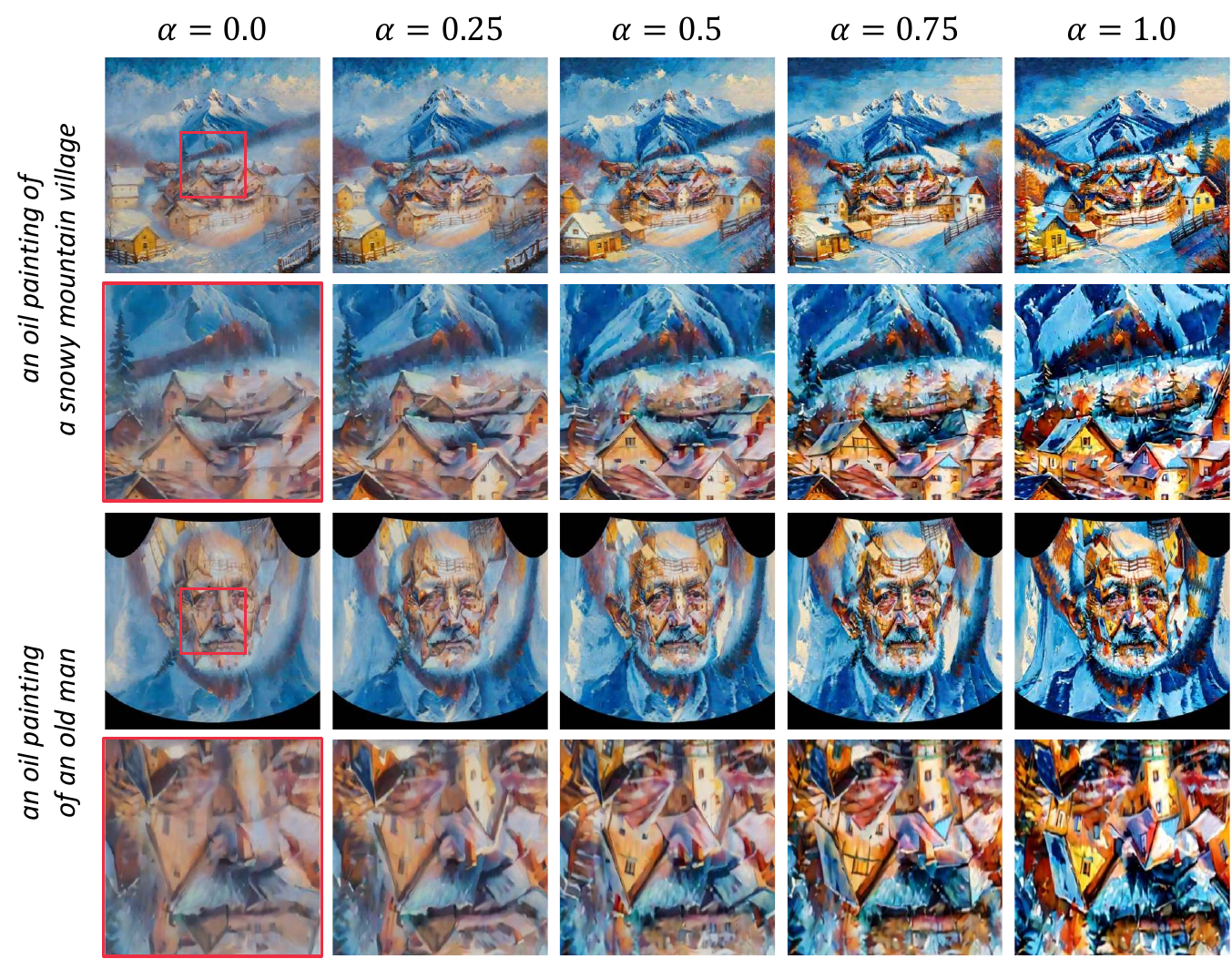}

   \caption{\textbf{Effects of parameter $\alpha$.} The parameter affects how views are merged together at each level of the pyramid. $\alpha=0$ corresponds to standard average, while $\alpha=1$ gives more weights to pixels with extreme values, preserving the details in all the views.}
   \label{fig:suppl_lp_alpha}
\end{figure}

\Cref{fig:suppl_lp_alpha} shows the effect of $\alpha$ for two examples with the cylinder mirror. When standard averaging is used ($\alpha = 0$), the image looks blurrier. However, when value-weighted average is used ($\alpha=1$), all details from both views are preserved, creating very saturated, over-sharpened results. In most of our results, we opt for $\alpha \in [0.25, 0.5]$.

\subsection{Pseudo-code}\label{subsec:suppl_pseudocode}

We provide a pseudo-code for our novel Laplacian Pyramid warping. Some notes:

\begin{itemize}
    \item l.8: \texttt{\_compute\_lod\_level} returns the LOD level map as a $(H, W)$ tensor using \Cref{eq:lod}; \\
    \item l.29: \texttt{pyrStack} takes a pyramid, upsamples all levels to highest resolution, and stacks them along a dimension; \\
    \item l.73: \texttt{impute\_with\_nearest} fills missing values with nearest ones in the image given the mask.
\end{itemize}

\begin{figure}[H]
    \begin{minted}[fontsize=\scriptsize, frame=lines, linenos]{python}

def _get_grid(warp, maxLOD):
    """
    Takes in numpy array warp of shape (1, 3, 1024, 1024)
    """
    # compute lod and normalize to (-1, 1)
    mapping = 1024 * warp[:, :2]
    lod = _compute_lod_level(mapping, maxLOD=maxLOD)
    lod = 2 * lod / maxLOD - 1.0  # (h, w)
    lod = lod.unsqueeze(0).unsqueeze(-1)

    # normalize uv to (-1, 1)
    grid = torch.tensor(warp[:, :2]).permute(0, 2, 3, 1)
    mask = torch.all(grid == 0.0, dim=-1, keepdim=True)
    mask = mask.expand(-1, -1, -1, 2)
    grid[mask] = torch.nan  # replace undefined with NaN
    grid = 2 * grid - 1

    # combine into a 3D coordinate grid (lod, u, v)
    grid = torch.cat([lod, grid], -1).unsqueeze(1)
    return grid  # (1, 1, dim, dim, 3)
    

def view(lp, warp, leveln):
    # get 3d coordinate grid
    grid = _get_grid(warp, maxLOD=leveln-1)

    # sample values
    layers = pyrStack(lp, dim=-1)
    new_im = F.grid_sample(
        layers, 
        grid, 
        mode='nearest',
        padding_mode="zeros",
        align_corners=True,
    ).squeeze(2)

    # replace by NaN where image is 0
    new_im[new_im == 0.0] = torch.nan
    new_im = torch.nanmean(new_im, 0).half()
    return new_im


def inverse_view(im, warp, leveln):
    c, h, w = im.shape
    grid = _get_grid(warp, maxLOD=leveln - 1)
    with torch.enable_grad():
        # create an empty pyramid
        opt_var = torch.zeros(1, c + 1, h, w)
        opt_var = LaplacianPyramid(opt_var, leveln)
        for lvl in opt_var:
            lvl.requires_grad_()

        # convert to Gaussian pyramid
        opt_gp = Laplacian2Gaussian(opt_var)
        target = torch.cat([im, torch.ones_like(im[:1])])
        layers = pyrStack(opt_gp, -1).float()
        warped = F.grid_sample(
            layers,
            grid,
            mode='nearest',
            padding_mode="zeros",
            align_corners=True,
        ).squeeze(2)
        loss = 0.5 * ((warped - target) ** 2).sum()
        loss.backward()
        result = [-l.grad.detach() for l in opt_var]
        result = [(r[:, :c] / r[:, -1:]) for r in result]

    # extract laplacian pyramid
    for k in range(leveln - 1):
        mask = torch.isnan(result[k])
        imputed = impute_with_nearest(result[k], ~mask)
        result[k] = imputed - pyrUp(pyrDown(imputed))
        result[k][mask] = torch.nan
        
    return result
    
\end{minted}
    \label{fig:pseudocode}
\end{figure}

\section{Additional Ablations}

We provide more qualitative ablations to show the effects of Laplacian warping, view prioritization, and time travel.

\subsection{Prioritizing a Single View}\label{subsec:suppl_abl_ratiox}

\begin{figure}[h]
  \centering
  \centering
   \includegraphics[width=1.0\linewidth]{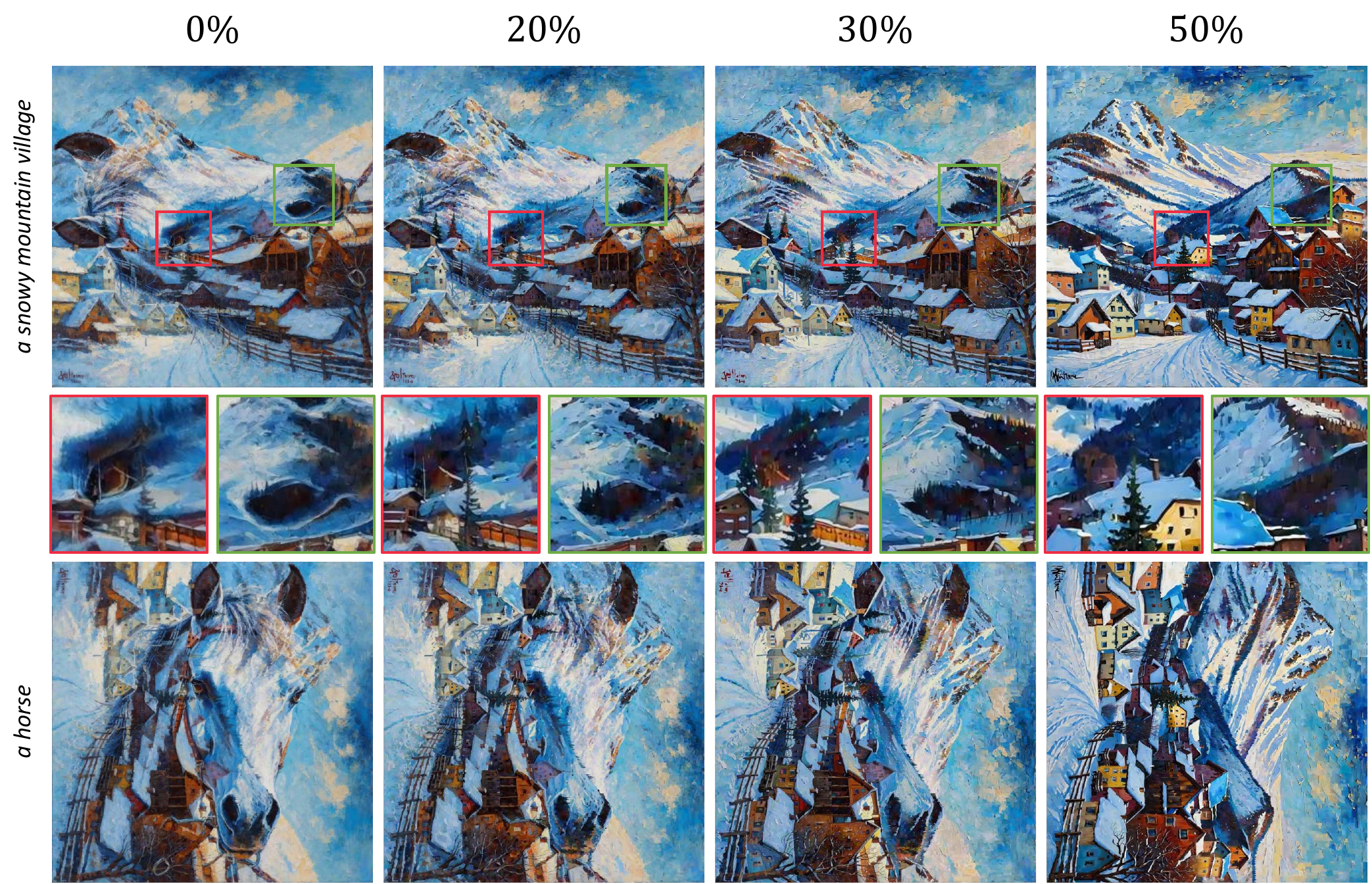}

   \caption{\textbf{Prioritizing a single view.} We show the effect of prioritizing the first view (\textit{``a snowy mountain village"}) over the second (\textit{``a horse"}) in the example of $90^\circ$ rotation.}
   \label{fig:suppl_lastx}
\end{figure}

In \Cref{subsec:designchoice}, we proposed to let the last $x\%$ of the denoising process only denoise a single one of the views. \Cref{fig:suppl_lastx} shows the effect of varying values of $x$ for an example with $90^\circ$ rotation. As the ratio increases, details from the first view (\textit{``a snowy mountain village"}) dominate over the second view (\textit{``a horse"}). This is a useful parameter to control the trade-off between views.

\begin{figure}[h]
  \centering
  \centering
   \includegraphics[width=1.0\linewidth]{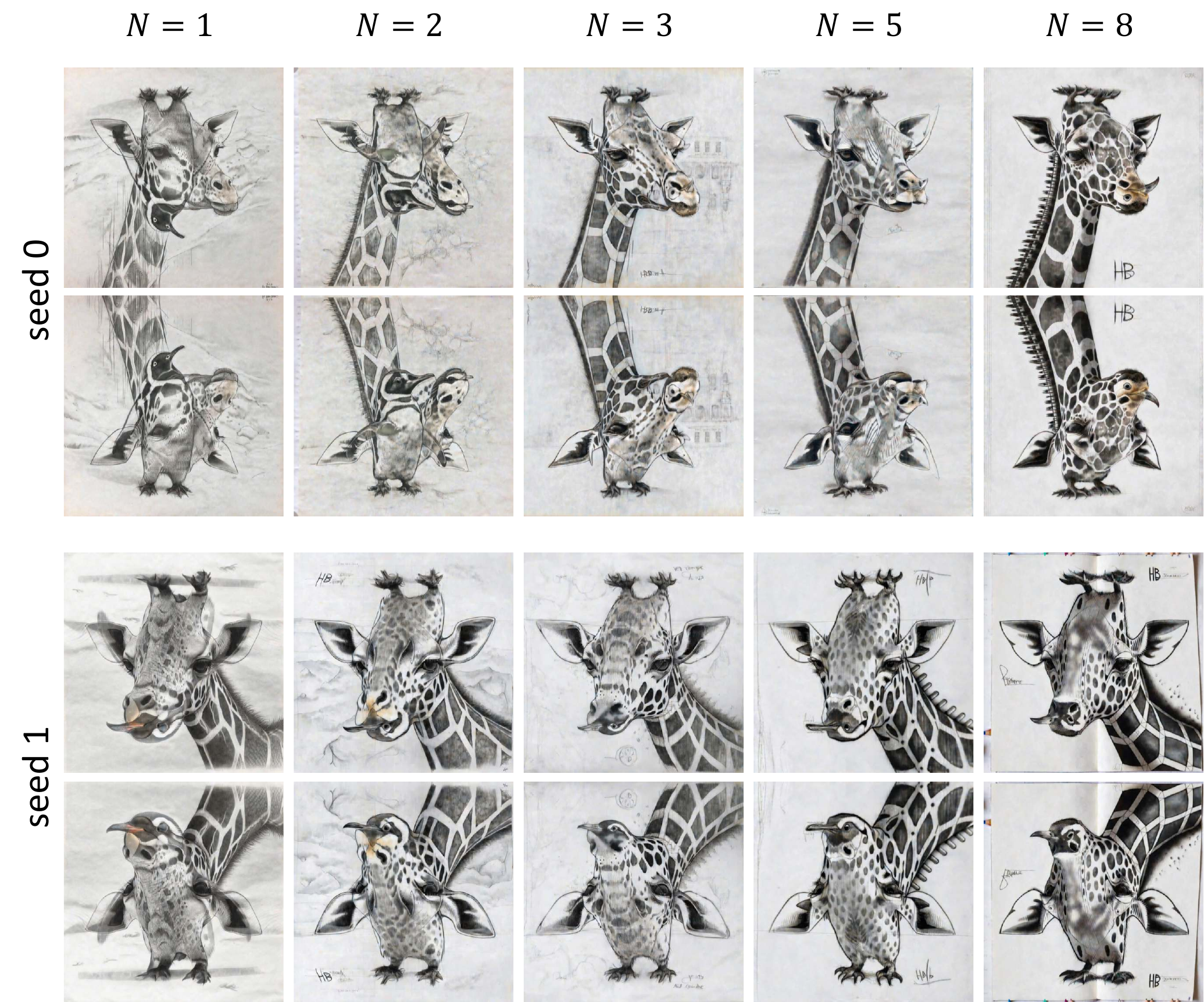}

   \caption{\textbf{Time travel.} We show the effect of time traveling for the flip example (``a giraffe head" / ``a penguin") with two different sampled seeds. Time traveling is only applied for timesteps between $20\%$ and $80\%$, by repeating each step $N$ times.}
   \label{fig:suppl_ttravel}
\end{figure}

\subsection{Time travel}\label{subsec:suppl_abl_ttravel}

We show in \Cref{fig:suppl_ttravel} that time traveling is an effective strategy for improving image consistency and blending between views. Without time traveling ($N=1$), the \textit{penguin} and the \textit{giraffe head} seem like two independent entities overlayed on top of each other in the image. As the repeating number $N$ increases, the two views align better, resulting in a sharper image with more coherent details. 

Obviously, runtime scales linearly with the repeating number. In Burgert et al. \cite{burgert2024diffusionillusions}, image quality and blending quality are entangled, and improving with the number of optimization steps. Here, time traveling is only responsible for blending quality, but is independent of the generated image quality. We believe this provides a more intuitive control parameter than the number of optimization steps.

Naturally, better blending still yields better overall quality, which explains the lower FID in \Cref{tab:main_ablations}. However, it seems to come at the cost of slightly worse prompt alignment. We hypothetize that blending better means making more compromise between views, which in turn makes it harder to match the prompts as well.

\begin{figure}[h]
  \centering
  \centering
   \includegraphics[width=1.0\linewidth]{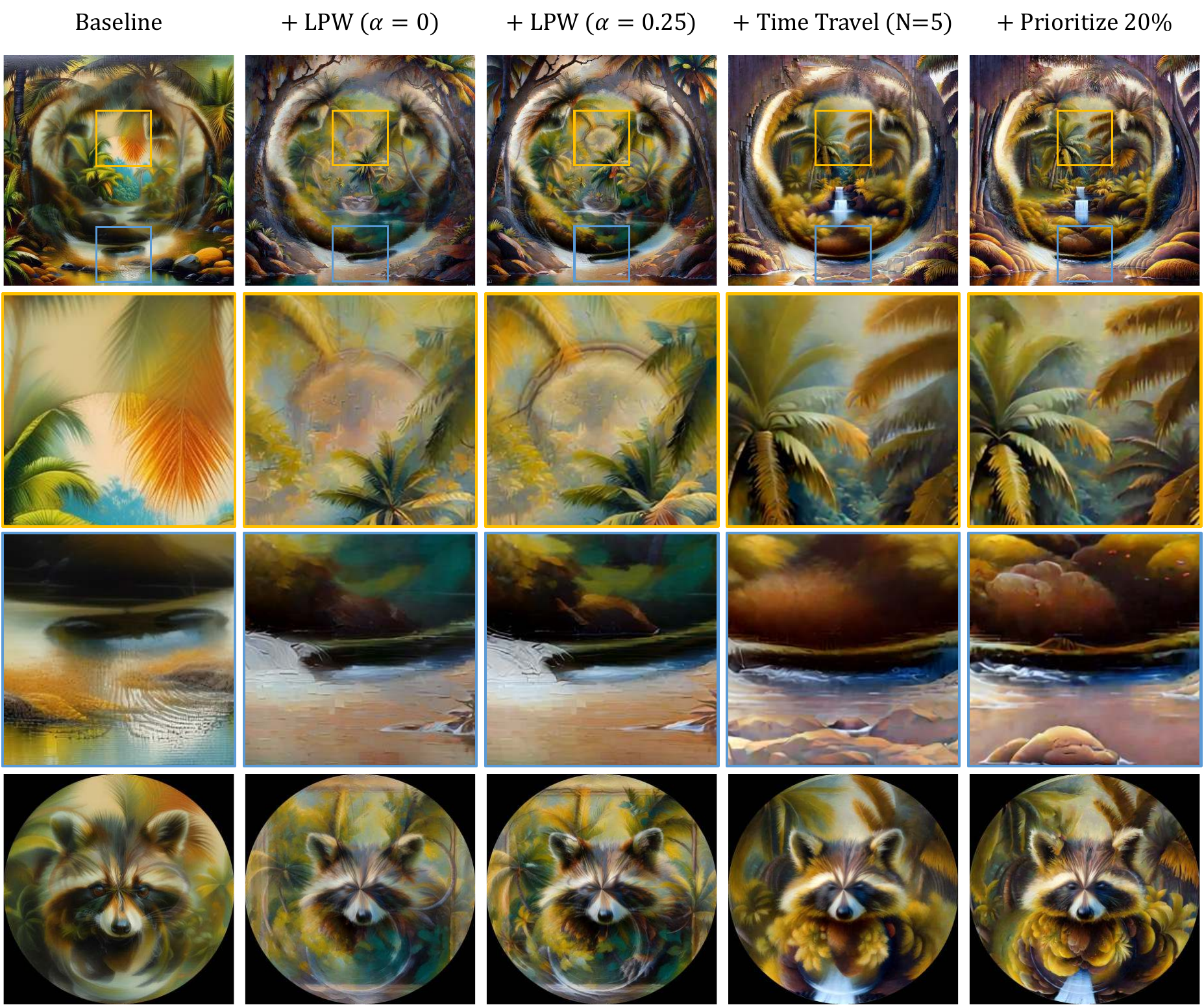}

   \caption{\textbf{Baseline ablation.} We consider the cone mirror example with two views (\textit{``a tropical jungle forest"-``a raccoon"}. We show successively the effect of Laplacian Pyramid Warping (LPW), value-weighted average ($\alpha$), time traveling, and single-view prioritization.}
   \label{fig:suppl_baseline}
\end{figure}

\subsection{Comparison with Baseline}\label{subsec:suppl_abl_comp}

Lastly, we show in \Cref{fig:suppl_baseline} the effects of these different features. Laplacian Pyramid Warping addresses the artifacts at the edge of the ring, visible in the baseline. Value-weighted average recovers sharp details lost in standard averaging. Time traveling ensures a smoother blending, while single-view prioritization adds back the final details in the identity view without destroying the second view. 

\section{Quantitative Evaluation Details}

We provide additional details regarding the quantitative evaluation in this section.

\paragraph{Dataset generation.} Similar to Geng \etal \cite{geng2024visualanagrams}, we use a custom list of 50 prompt pairs in the form of \texttt{[<style>, <prompt1>, <prompt2>]} for our quantitative evaluation. These are a mix between prompt pairs from Visual Anagrams' paper \cite{geng2024visualanagrams}, and from querying ChatGPT. For each method, we generate 10 samples per prompt pair. This results in 500 pairs of images, \ie 1k images per method. For FID/KID computation, we generated a reference dataset comprised of 3.2k images from SD3 and 3.2k images from SD3.5 following the same prompts (only single view).

\paragraph{Comparing with prior work.} For Visual Anagrams \cite{geng2024visualanagrams}, we used the original codebase from Geng \etal: \url{https://github.com/dangeng/visual_anagrams}. Results are generated with DeepFloyd IF \cite{DeepFloydIF} in two stages, and subsequently up-sampled to $1024\times 1024$ using Stable Diffusion x4 Upscaler, as provided by the code. 

For Burgert et al. \cite{burgert2024diffusionillusions}, we used the available codebase at (\url{https://github.com/RyannDaGreat/Diffusion-Illusions}) and added a function to rotate the inner circle of an image by $135^\circ$. Each image is generated with 1000 optimization steps using Stable Diffusion v1.4, the default model from their codebase. 

Lastly, we set the hyperparameters of our pipeline to replicate the original implementation of Tancik \cite{tancik2023illusion} and SyncTweedies \cite{Kim2024SyncTweedies}. The results are generated with Stable Diffusion 3.5 Medium, as in our method. 

\paragraph{Runtime.} We generate our results with Stable Diffusion 3.5 Medium on a Nvidia GeForce RTX 4090 GPU. With 30 steps of inference and time traveling between $20\%$ and $80\%$ repeating $2$ times, we can generate an image pair in $\sim 80$s.

\begin{table}[h]
    \centering
    \resizebox{\columnwidth}{!}{
    \begin{tabular}{ccccc}
&&&& \\ 
\toprule
  Geng \etal \cite{geng2024visualanagrams} & Tancik \etal \cite{tancik2023illusion} & SyncTweedies \cite{Kim2024SyncTweedies} & Burgert \etal \cite{burgert2024diffusionillusions} &  Ours SD 3.5 \\
\midrule
 18.6s & 17.2s & 18.8s & 176.0s & 79.4s \\
 \midrule
\end{tabular}
}
    \caption{\textbf{Inference time comparison.}}
    \label{tab:runtime}
\end{table}

\section{Using Other Latent Models}

Most of our results were generated using Stable Diffusion 3.5. However, we also experimented with other models such as SD2.1 and SDXL. 

\begin{figure}[h]
  \centering
  \centering
   \includegraphics[width=1.0\linewidth]{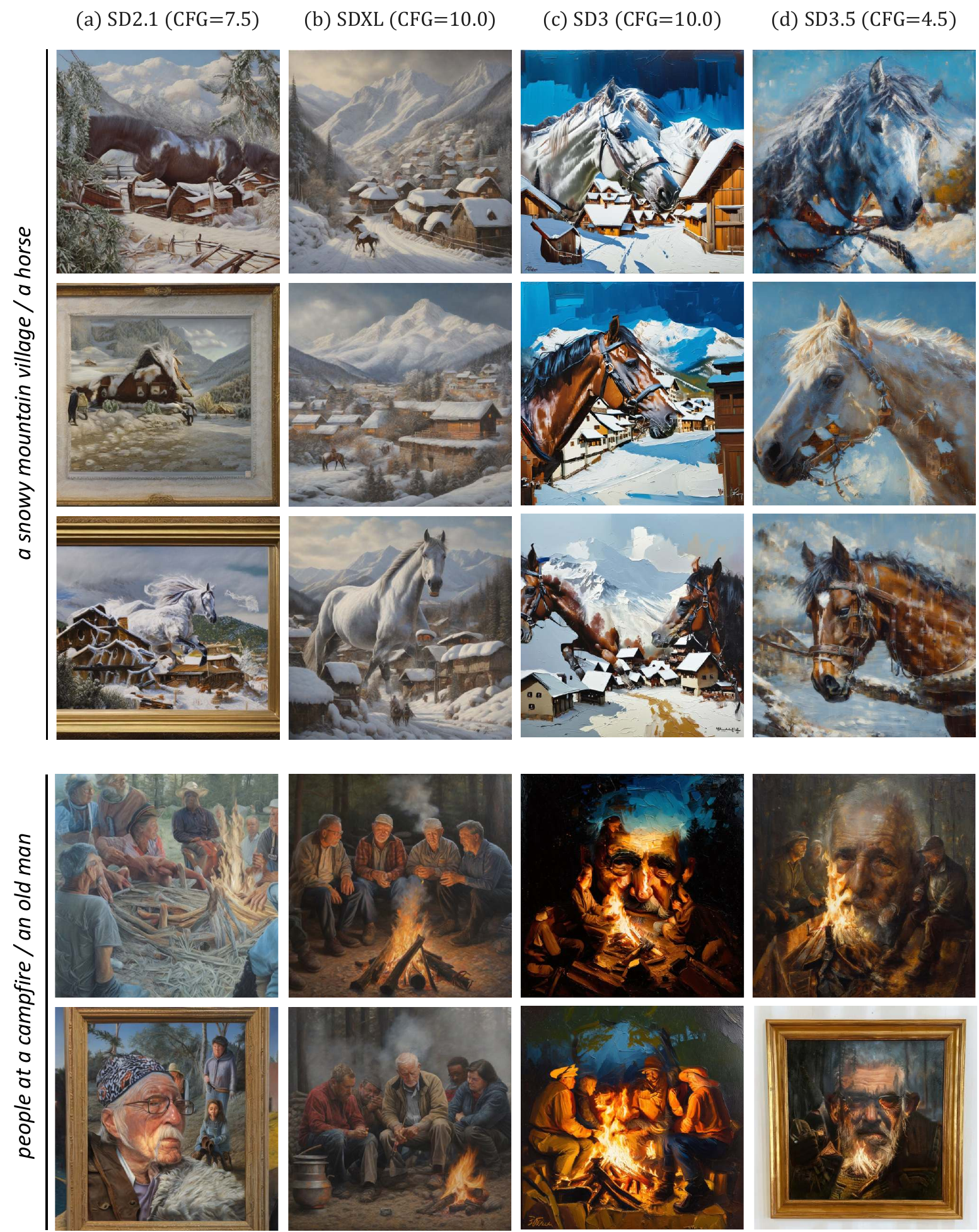}

   \caption{\textbf{Comparing different models for two-view setting.} We consider the case of two identity views, associated with two distinct prompts. While most models blend the two concepts \textit{spatially}, SDXL seems to blend them \textit{semantically}.}
   \label{fig:different_models}
\end{figure}

Here, we show a simple experiment with \textit{two identity views} associated with two distinct prompts. This way, we abstract out the VAE, as well as other problems that come from warping. The results are shown in \Cref{fig:different_models} for two pairs of prompts. Interestingly, even in such a simple setting, not all models behave the same: SD2.1 tend to generate images with poor quality, while SD3+ attempts to blend the two concepts by compositing them as much as possible. Curiously, SDXL seems to blend the concepts \textit{semantically}: ``a snowy mountain village" and ``a horse" give horses \textit{in} a village, while ``people at a campfire" and ``an old man" produce ``old men at a campfire". 

While intriguing, this is not ideal for ambiguous images, as the goal is more to blend \textit{spatially} the different views. Further investigation is needed to explain the discrepancy between SDXL and SD3+, possibly due to the switch from diffusion to flow matching or from U-Net to a transformer backbone. A deeper study is left for future work.

\section{User Study}

As mentioned in \Cref{subsec:user_study}, we conducted a user study over 27 participants to compare human preferences between our proposed method and existing prior work.

\begin{figure}[t!]
  \centering
  \centering
   \includegraphics[width=1.0\linewidth]{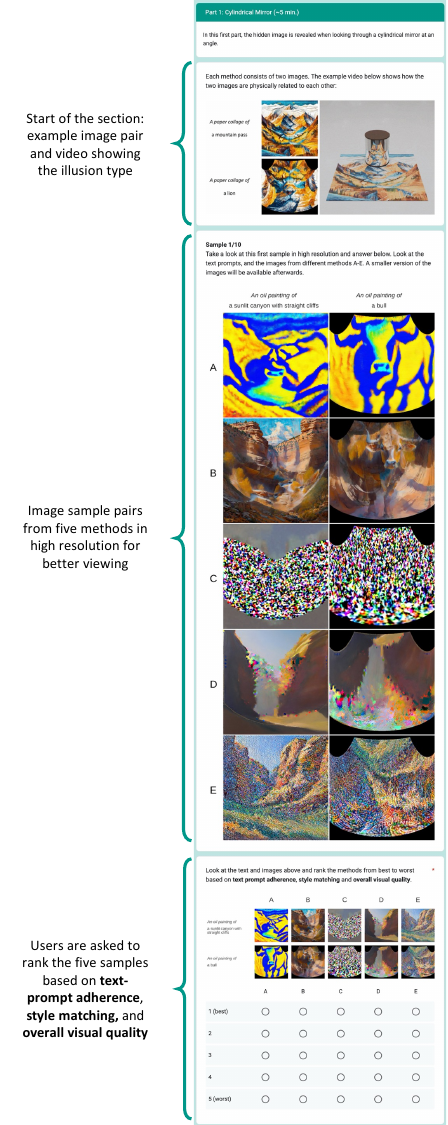}

   \caption{\textbf{Sample from the user study.}}
   \label{fig:user_study_sample}
\end{figure}

\paragraph{Study structure.} The study consists of three sections, each corresponding to a different type of anamorphosis: cylindrical mirror, conic mirror, and Niceron's lens. Each section begins with an example image and video demonstrating how the illusion works. Participants are then shown 10 different samples generated from 10 pairs of prompts. These prompts remain the same across all three sections to avoid bias. Additionally, the order in which the methods are presented is randomized for each sample. To ensure a fair comparison, we first display the results in high resolution before asking users to rank them (see \cref{fig:user_study_sample}).

\paragraph{Ranking criteria.} Participants are asked to provide a ranking of the five methods from 1 (best) to 5 (worst). At the beginning of the study, they are instructed to evaluate the images based on the following criteria:

\begin{itemize} 
\item \textbf{Match both text prompts:} When viewed directly (resp. through a mirror or lens), the image should correspond to the first (resp. second) prompt.
\item \textbf{Maintain the specified style:} The image should adhere to the given style prompt (\eg painting, photograph). 
\item \textbf{Be high-quality:} The final image should be sharp, detailed, and visually appealing. 
\end{itemize}

\section{Failed Experiments}

\paragraph{Relative negative prompting.} In Visual Anagrams \cite{geng2024visualanagrams}, the authors note that including other views in the negative prompt does not improve substantially the visual quality. Moreover, prompt conflicts can arise. For example, the prompts \textit{``an oil painting of a village"} and \textit{``an oil painting of a horse"} both share the style descriptor \textit{``an oil painting of"}, which should not be included in the negative prompt. Similarly, prompts like \textit{``a cat"} and \textit{``a dog"} share features such as fur. If these shared features appear in both the negative and positive prompts, it can lead to suboptimal results. 

We experimented with a variant of this, which we dub \textit{relative negative prompting}. The idea is to put in the negative prompt only the relative direction between the positive prompt (from the current view) and the negative prompt (from the other view). Thus, we subtract the positive prompt embedding from the negative one. Our experiments showed that this removes the need to manually select which part to keep for the negative prompt. In the example of the shared style, our difference vector cancels it out automatically, and the model is thus still able to generate the correct style. However, similar to Geng \etal \cite{geng2024visualanagrams}, we did not observe substantial improvement using this method. 

\section{Choosing Prompts \& Failure Cases}

The quality of the generation relies on choosing a good style and pair of prompts. Here are some tips we found for generating good anamorphoses:
\begin{enumerate}
    \item a place or location (\eg jungle, desert, library etc.) gives a lot of freedom to the composition and generally works well for the identity view;
    \item the second view is generally seen through some mirror or a lens, which is smaller than the main image. For this view, easily recognizable subjects like animals or faces are good prompts in most cases;
    \item artistic styles are more prone to produce good results than photorealistic styles;
    \item styles with no colors (\eg sketches, ink, marble sculpture) will generate better results when the two prompts have very different color palettes. \\
\end{enumerate}

\begin{figure}[h]
  \centering
  \centering
   \includegraphics[width=0.7\linewidth]{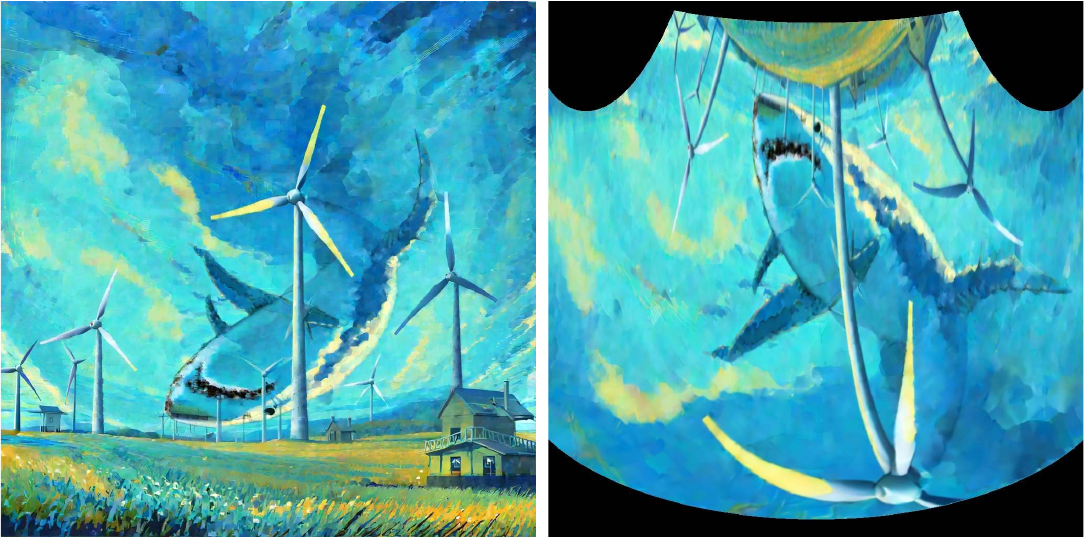}

   \caption{\textbf{Failure case.} Similar to Geng \etal, our method can cheat and put both views in the image without properly mixing them. In this example, the shark from the cylinder mirror view can be seen in the sky behind the wind turbines.}
   \label{fig:suppl_failure}
\end{figure}

Our method is still prone to fail in certain cases. For example, the model can still cheat and put all the views in the image without properly blending them (see \Cref{fig:suppl_failure}).

\section{Concurrent Work}

After our initial submission, a preprint titled ``\textit{Illusion3D: 3D Multiview Illusion with 2D Diffusion Priors}" \cite{feng2024illusion3d3dmultiviewillusion} appeared on arXiv, addressing a similar problem. While their code is not available at the time of writing, we identified a few key differences between our method and theirs.

First, Illusion3D builds on optimization-based approaches like Burgert et al. \cite{burgert2024diffusionillusions}, whereas our method improves upon feedforward techniques \cite{geng2024visualanagrams, tancik2023illusion}, generating images in a single inference pass. Their approach replaces Score Distillation Sampling (SDS) \cite{burgert2024diffusionillusions} with Variational Score Distillation (VSD), which requires training a LoRA module during optimization. Due to the inherent limitations of score distillation methods, we believe our approach produces higher-quality images while supporting a broader range of styles, from artistic to photorealistic, as demonstrated in \Cref{sec:app_results}.

A key advantage of Illusion3D, however, is its ability to generate full 3D structures, which our method does not support. That said, our approach can still generate 2D textures for mapping onto 3D surfaces, similar to their sphere or cube illusions. Lastly, we expect our method to have significantly lower generation time and memory requirements compared to Illusion3D.

\begin{table*}[t!]
\centering
\vspace{-0.5cm}
\begin{tabular}{m{1.5cm}lcccccccc}
\toprule
 & Method & $\mathcal{S}$ $\uparrow$ & $\mathcal{S}_{0.9}$ $\uparrow$ & $\mathcal{A}$ $\uparrow$ & $\mathcal{A}_{0.9}$ $\uparrow$ & $\mathcal{C}$ $\uparrow$ & $\mathcal{C}_{0.9}$ $\uparrow$ & FID $\downarrow$ & KID $\downarrow$ \\
\midrule
\multirow{5}{=}{\centering Vertical\ Flip}
& Geng \etal \cite{geng2024visualanagrams} & \cellcolor{orange!20}0.325 & \cellcolor{orange!20}0.362 & \cellcolor{red!20}0.306 & \cellcolor{yellow!20}0.340 & \cellcolor{orange!20}0.695 & \cellcolor{yellow!20}0.786 & 149.24 & 0.057 \\
& Tancik SD 3.5 \cite{tancik2023illusion} & \cellcolor{red!20}0.328 & \cellcolor{red!20}0.367 & \cellcolor{orange!20}0.306 & \cellcolor{red!20}0.349 & \cellcolor{yellow!20}0.693 & \cellcolor{red!20}0.806 & \cellcolor{orange!20}132.52 & \cellcolor{orange!20}0.049 \\
& Burgert \etal \cite{burgert2024diffusionillusions} & 0.303 & 0.347 & 0.281 & 0.324 & 0.679 & 0.778 & 219.84 & 0.115 \\
& SyncTweedies \cite{Kim2024SyncTweedies} & \cellcolor{yellow!20}0.323 & \cellcolor{yellow!20}0.360 & \cellcolor{yellow!20}0.302 & \cellcolor{orange!20}0.341 & \cellcolor{red!20}0.707 & \cellcolor{orange!20}0.801 & \cellcolor{yellow!20}132.62 & \cellcolor{yellow!20}0.054 \\
& \textbf{LookingGlass (ours)} & \textbf{0.320} & \textbf{0.358} & \textbf{0.297} & \textbf{0.338} & \textbf{0.680} & \textbf{0.779} & \cellcolor{red!20}\textbf{124.67} & \cellcolor{red!20}\textbf{0.049} \\
\midrule
\multirow{5}{=}{\centering $135^{\circ}$\ Rotation}
& Geng \etal \cite{geng2024visualanagrams} & 0.284 & 0.340 & 0.262 & 0.308 & 0.563 & 0.652 & 293.00 & 0.254 \\
& Tancik SD 3.5 \cite{tancik2023illusion} & 0.203 & 0.225 & 0.194 & 0.216 & 0.498 & 0.509 & 439.35 & 0.545 \\
& Burgert \etal \cite{burgert2024diffusionillusions} & \cellcolor{yellow!20}0.301 & \cellcolor{yellow!20}0.347 & \cellcolor{yellow!20}0.280 & \cellcolor{yellow!20}0.326 & \cellcolor{orange!20}0.654 & \cellcolor{orange!20}0.760 & \cellcolor{yellow!20}223.21 & \cellcolor{yellow!20}0.120 \\
& SyncTweedies \cite{Kim2024SyncTweedies} & \cellcolor{orange!20}0.308 & \cellcolor{orange!20}0.354 & \cellcolor{orange!20}0.283 & \cellcolor{orange!20}0.335 & \cellcolor{yellow!20}0.647 & \cellcolor{yellow!20}0.753 & \cellcolor{orange!20}166.03 & \cellcolor{orange!20}0.083 \\
& \textbf{LookingGlass (ours)} & \cellcolor{red!20}\textbf{0.319} & \cellcolor{red!20}\textbf{0.357} & \cellcolor{red!20}\textbf{0.295} & \cellcolor{red!20}\textbf{0.338} & \cellcolor{red!20}\textbf{0.666} & \cellcolor{red!20}\textbf{0.767} & \cellcolor{red!20}\textbf{129.74} & \cellcolor{red!20}\textbf{0.055} \\
\midrule
\multirow{5}{=}{\centering Cylindrical\ Mirror}
& Geng \etal \cite{geng2024visualanagrams} & 0.190 & 0.228 & 0.171 & 0.198 & 0.506 & 0.546 & 285.23 & 0.216 \\
& Tancik SD 3.5 \cite{tancik2023illusion} & 0.189 & 0.225 & 0.171 & 0.198 & 0.505 & 0.547 & 284.97 & 0.215 \\
& Burgert \etal \cite{burgert2024diffusionillusions} & \cellcolor{orange!20}0.285 & \cellcolor{orange!20}0.334 & \cellcolor{orange!20}0.261 & \cellcolor{orange!20}0.304 & \cellcolor{red!20}0.706 & \cellcolor{orange!20}0.795 & \cellcolor{yellow!20}229.65 & \cellcolor{yellow!20}0.138 \\
& SyncTweedies \cite{Kim2024SyncTweedies} & \cellcolor{yellow!20}0.285 & \cellcolor{yellow!20}0.348 & \cellcolor{yellow!20}0.241 & \cellcolor{yellow!20}0.284 & \cellcolor{yellow!20}0.673 & \cellcolor{yellow!20}0.763 & \cellcolor{orange!20}138.69 & \cellcolor{orange!20}0.082 \\
& \textbf{LookingGlass (ours)} & \cellcolor{red!20}\textbf{0.307} & \cellcolor{red!20}\textbf{0.360} & \cellcolor{red!20}\textbf{0.272} & \cellcolor{red!20}\textbf{0.318} & \cellcolor{orange!20}\textbf{0.698} & \cellcolor{red!20}\textbf{0.810} & \cellcolor{red!20}\textbf{130.27} & \cellcolor{red!20}\textbf{0.070} \\
\bottomrule
\end{tabular}

\caption{\textbf{Additional quantitative comparison.} We additionally assess image-prompt alignment using CLIP similarity score $\mathcal S$ on all three transformations evaluated in the main paper. While all methods achieve comparable results for the vertical flip, LookingGlass surpasses previous approaches on more complex transformations, including anamorphoses.}
\label{tab:suppl_quant}
\end{table*}

\section{Additional  Results}\label{sec:app_results}

In the next pages, we show additional qualitative results for the three anamorphic views: cylinder mirror, conic mirror, and Nicéron's lens. Please refer to the supplementary videos to see these anamorphoses in action. \Cref{tab:suppl_quant} shows additional quantitative evaluations.

\begin{figure*}[h]
  \centering
  \centering
   \includegraphics[width=1.0\linewidth]{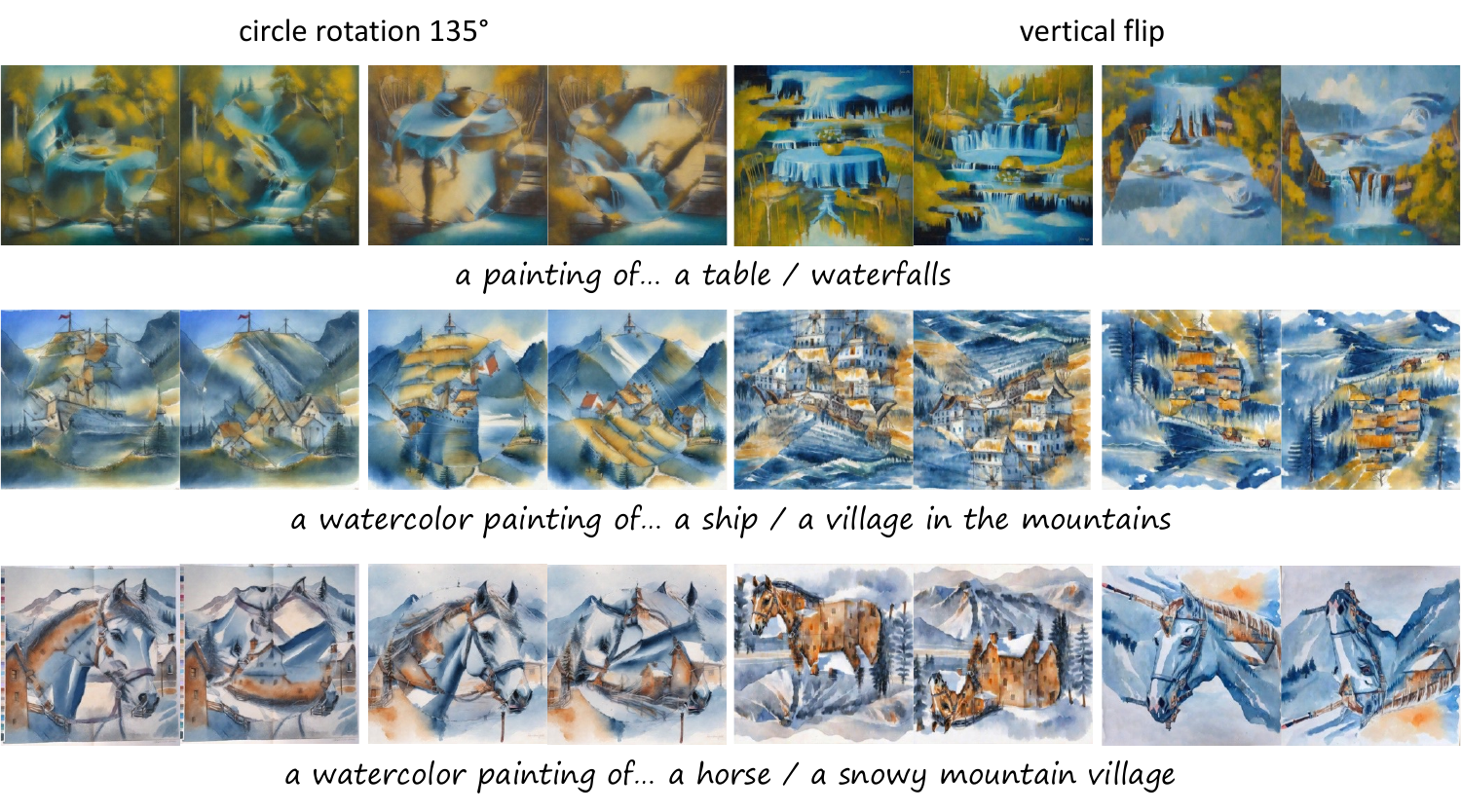}

   \caption{\textbf{2D transform results.} Here are some generated results for the two 2D transforms: vertical flip, and 135$^\circ$ rotation (not supported by Geng \etal \cite{geng2024visualanagrams}).}
   \label{fig:suppl_2d_results}
\end{figure*}

\begin{figure*}[h!]
    \centering
    \begin{subfigure}{0.490\textwidth}
        \centering
        \includegraphics[width=\textwidth]{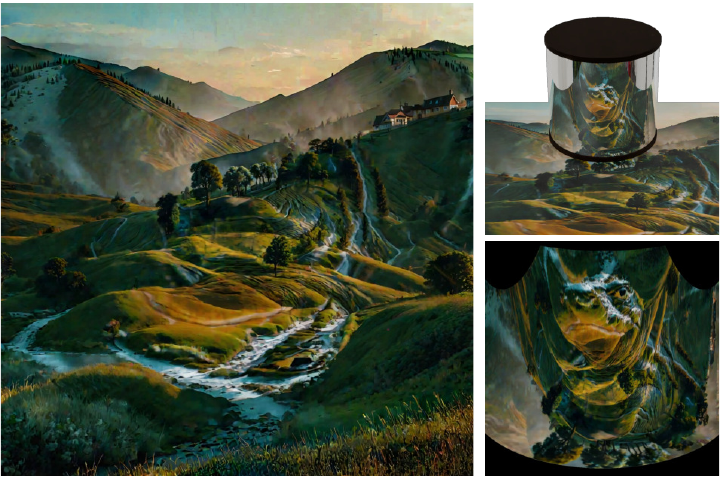}
        \caption*{\normalsize \centering \textit{a cinematic rendering of} \par\nobreak rolling hills in golden light / turtle}
    \end{subfigure}    \hfill
    \begin{subfigure}{0.490\textwidth}
        \centering
        \includegraphics[width=\textwidth]{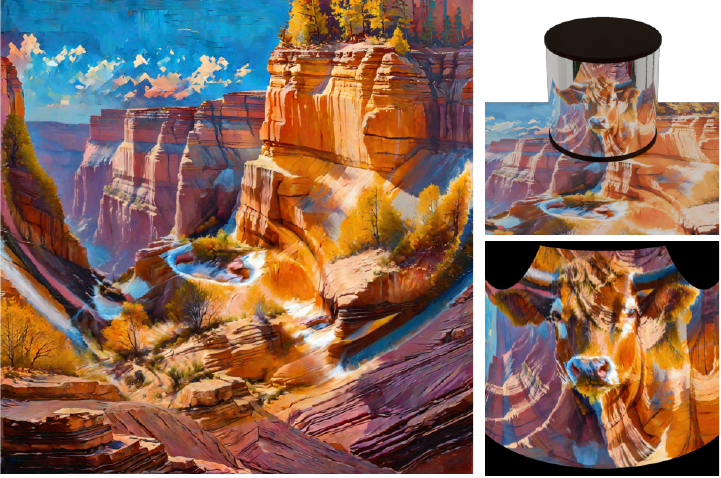}
        \caption*{\normalsize \centering \textit{an oil painting of} \par\nobreak sunlit canyon with straight cliffs / bull}
    \end{subfigure}
    
    \vspace{1.0cm} %
    \begin{subfigure}{0.490\textwidth}
        \centering
        \includegraphics[width=\textwidth]{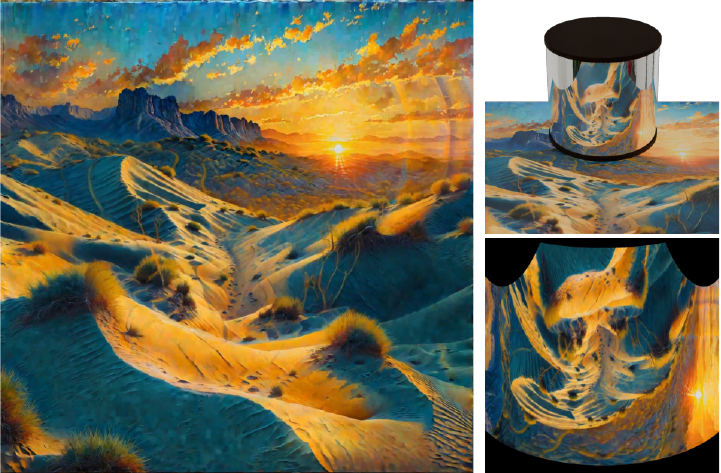}
        \caption*{\normalsize \centering \textit{an oil painting of} \par\nobreak desert dunes at sunset / jellyfish}
    \end{subfigure}    \hfill
    \begin{subfigure}{0.490\textwidth}
        \centering
        \includegraphics[width=\textwidth]{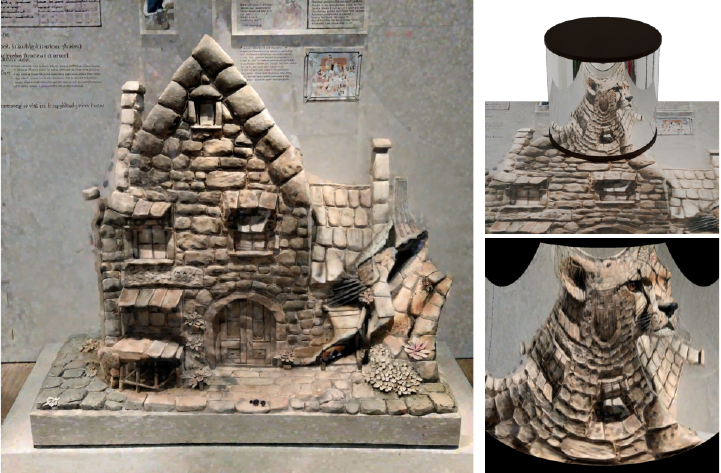}
        \caption*{\normalsize \centering \textit{a clay sculpture of} \par\nobreak cobblestone street / cheetah}
    \end{subfigure}
    
    \vspace{1.0cm} %
    \begin{subfigure}{0.490\textwidth}
        \centering
        \includegraphics[width=\textwidth]{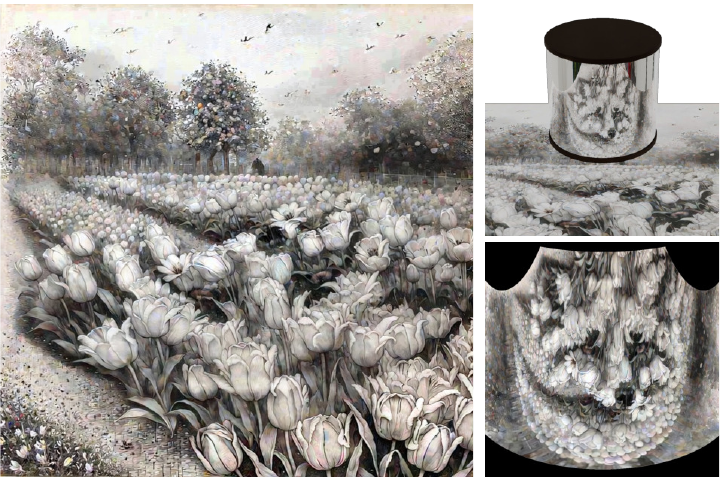}
        \caption*{\normalsize \centering \textit{a charcoal drawing of} \par\nobreak flower garden with rows of tulips / fox}
    \end{subfigure}    \hfill
    \begin{subfigure}{0.490\textwidth}
        \centering
        \includegraphics[width=\textwidth]{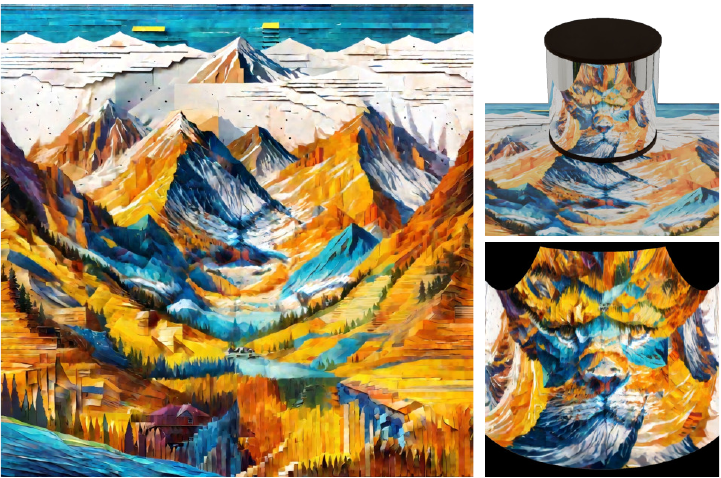}
        \caption*{\normalsize \centering \textit{a paper collage of} \par\nobreak mountain pass / lion}
    \end{subfigure}
    \caption{\textbf{Cylinder mirror anamorphosis.} In this figure and the two following ones, we show additional results for the cylinder mirror example. Each example contains the identity view, the mirror view as predicted by the flow model, and a rendering of the actual physical setting to validate our examples. Kindly refer to the supplementary videos to see these results in action.}
    \label{fig:suppl_cylinder_b}
\end{figure*}

\begin{figure*}[h!]
    \centering
    \begin{subfigure}{0.158\textwidth}
        \centering
        \includegraphics[width=\textwidth]{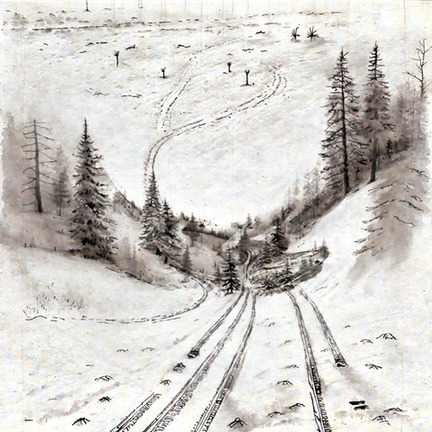}
        \vspace*{-6mm}
        \caption*{\centering \tiny \textit{an ink wash drawing of} \par\nobreak straight ski tracks}
    \end{subfigure}
    \begin{subfigure}{0.158\textwidth}
        \centering
        \includegraphics[width=\textwidth]{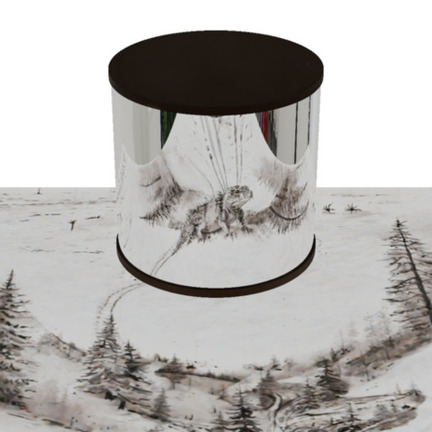}
        \vspace*{-6mm}
        \caption*{\centering \tiny \textit{\quad} \par\nobreak \quad}
    \end{subfigure}
    \begin{subfigure}{0.158\textwidth}
        \centering
        \includegraphics[width=\textwidth]{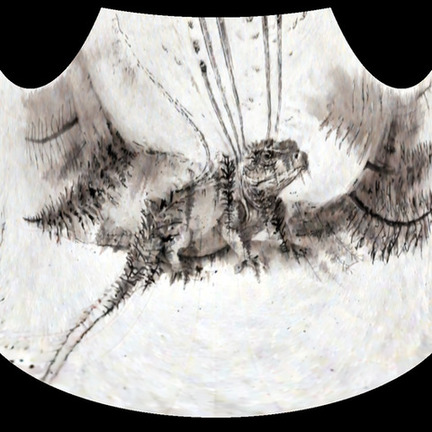}
        \vspace*{-6mm}
        \caption*{\centering \tiny \textit{an ink wash drawing of} \par\nobreak iguana}
    \end{subfigure}    \hfill
    \begin{subfigure}{0.158\textwidth}
        \centering
        \includegraphics[width=\textwidth]{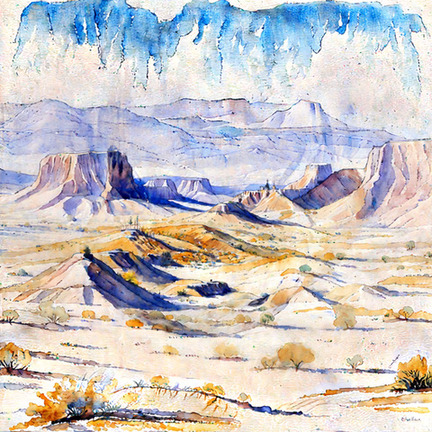}
        \vspace*{-6mm}
        \caption*{\centering \tiny \textit{a watercolor painting of} \par\nobreak desert plateau}
    \end{subfigure}
    \begin{subfigure}{0.158\textwidth}
        \centering
        \includegraphics[width=\textwidth]{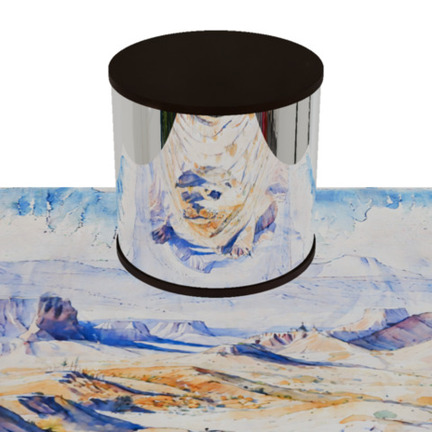}
        \vspace*{-6mm}
        \caption*{\centering \tiny \textit{\quad} \par\nobreak \quad}
    \end{subfigure}
    \begin{subfigure}{0.158\textwidth}
        \centering
        \includegraphics[width=\textwidth]{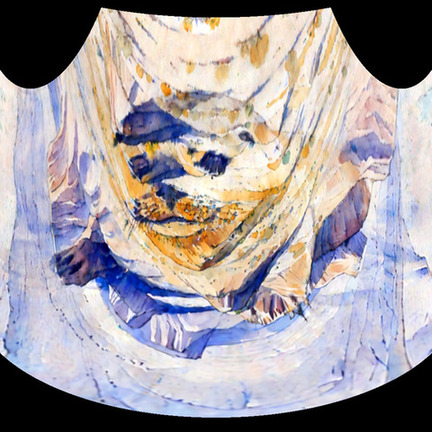}
        \vspace*{-6mm}
        \caption*{\centering \tiny \textit{a watercolor painting of} \par\nobreak seal}
    \end{subfigure}
    \begin{subfigure}{0.158\textwidth}
        \centering
        \includegraphics[width=\textwidth]{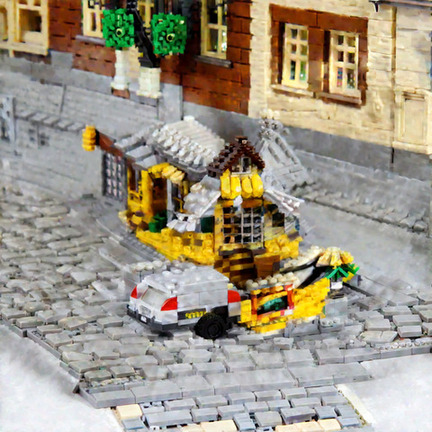}
        \vspace*{-6mm}
        \caption*{\centering \tiny \textit{a LEGO model of} \par\nobreak cobblestone street}
    \end{subfigure}
    \begin{subfigure}{0.158\textwidth}
        \centering
        \includegraphics[width=\textwidth]{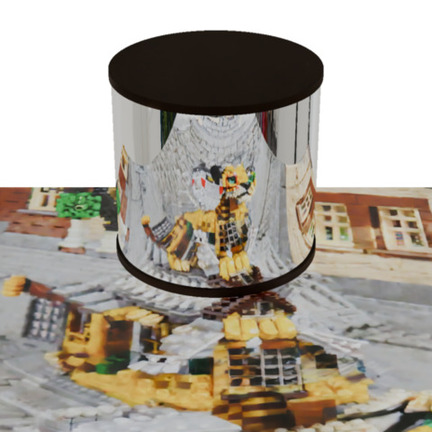}
        \vspace*{-6mm}
        \caption*{\centering \tiny \textit{\quad} \par\nobreak \quad}
    \end{subfigure}
    \begin{subfigure}{0.158\textwidth}
        \centering
        \includegraphics[width=\textwidth]{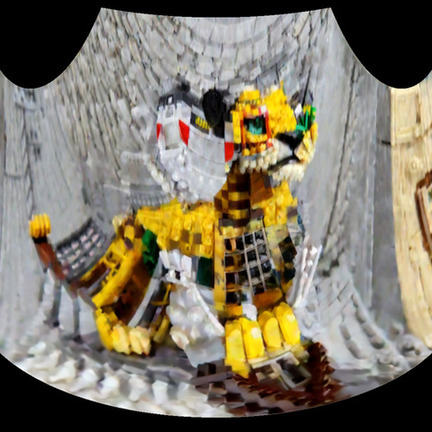}
        \vspace*{-6mm}
        \caption*{\centering \tiny \textit{a LEGO model of} \par\nobreak cheetah}
    \end{subfigure}    \hfill
    \begin{subfigure}{0.158\textwidth}
        \centering
        \includegraphics[width=\textwidth]{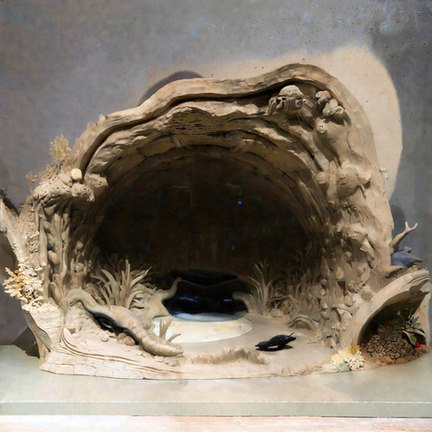}
        \vspace*{-6mm}
        \caption*{\centering \tiny \textit{a clay sculpture of} \par\nobreak aquarium tunnel}
    \end{subfigure}
    \begin{subfigure}{0.158\textwidth}
        \centering
        \includegraphics[width=\textwidth]{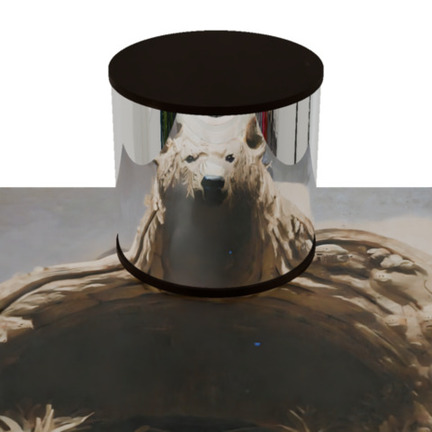}
        \vspace*{-6mm}
        \caption*{\centering \tiny \textit{\quad} \par\nobreak \quad}
    \end{subfigure}
    \begin{subfigure}{0.158\textwidth}
        \centering
        \includegraphics[width=\textwidth]{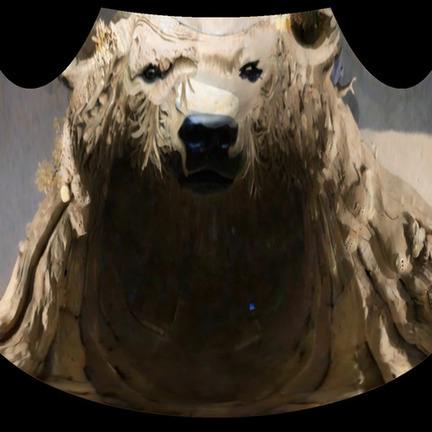}
        \vspace*{-6mm}
        \caption*{\centering \tiny \textit{a clay sculpture of} \par\nobreak polar bear}
    \end{subfigure}
    \begin{subfigure}{0.158\textwidth}
        \centering
        \includegraphics[width=\textwidth]{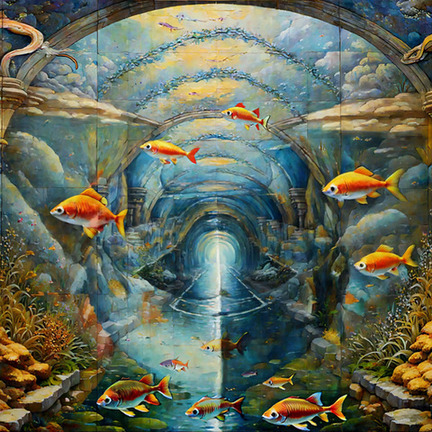}
        \vspace*{-6mm}
        \caption*{\centering \tiny \textit{a fresco painting of} \par\nobreak aquarium tunnel}
    \end{subfigure}
    \begin{subfigure}{0.158\textwidth}
        \centering
        \includegraphics[width=\textwidth]{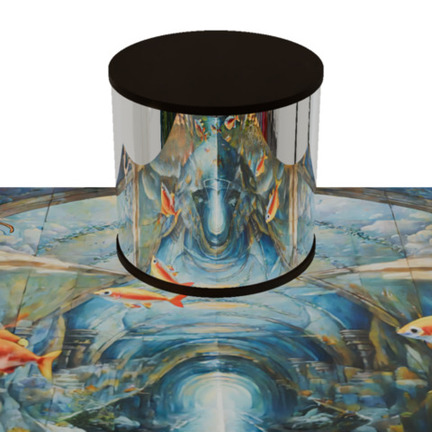}
        \vspace*{-6mm}
        \caption*{\centering \tiny \textit{\quad} \par\nobreak \quad}
    \end{subfigure}
    \begin{subfigure}{0.158\textwidth}
        \centering
        \includegraphics[width=\textwidth]{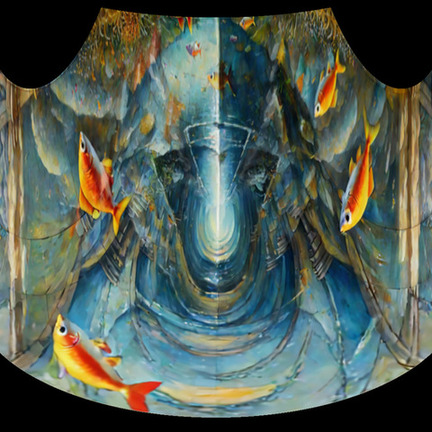}
        \vspace*{-6mm}
        \caption*{\centering \tiny \textit{a fresco painting of} \par\nobreak knight's helmet}
    \end{subfigure}    \hfill
    \begin{subfigure}{0.158\textwidth}
        \centering
        \includegraphics[width=\textwidth]{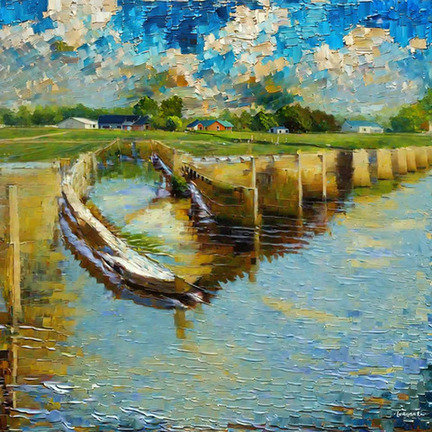}
        \vspace*{-6mm}
        \caption*{\centering \tiny \textit{an oil painting of} \par\nobreak straight levee dividing water}
    \end{subfigure}
    \begin{subfigure}{0.158\textwidth}
        \centering
        \includegraphics[width=\textwidth]{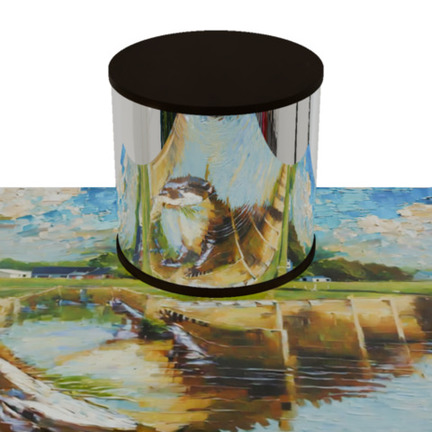}
        \vspace*{-6mm}
        \caption*{\centering \tiny \textit{\quad} \par\nobreak \quad}
    \end{subfigure}
    \begin{subfigure}{0.158\textwidth}
        \centering
        \includegraphics[width=\textwidth]{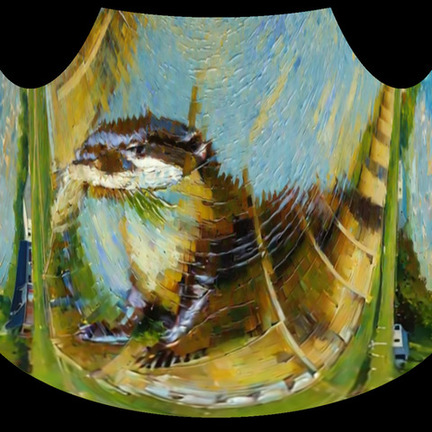}
        \vspace*{-6mm}
        \caption*{\centering \tiny \textit{an oil painting of} \par\nobreak otter}
    \end{subfigure}
    \begin{subfigure}{0.158\textwidth}
        \centering
        \includegraphics[width=\textwidth]{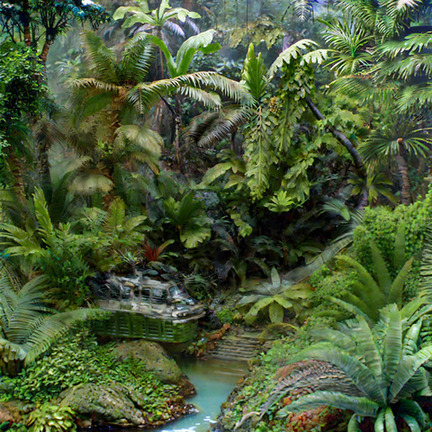}
        \vspace*{-6mm}
        \caption*{\centering \tiny \textit{a diorama of} \par\nobreak dense tropical rainforest}
    \end{subfigure}
    \begin{subfigure}{0.158\textwidth}
        \centering
        \includegraphics[width=\textwidth]{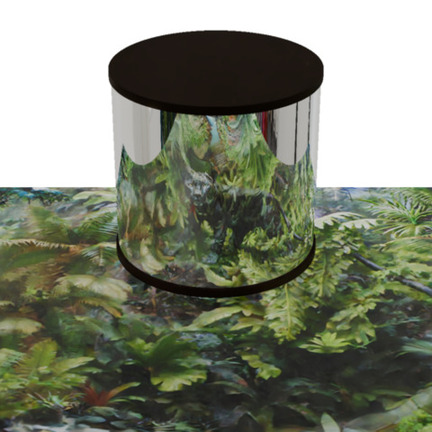}
        \vspace*{-6mm}
        \caption*{\centering \tiny \textit{\quad} \par\nobreak \quad}
    \end{subfigure}
    \begin{subfigure}{0.158\textwidth}
        \centering
        \includegraphics[width=\textwidth]{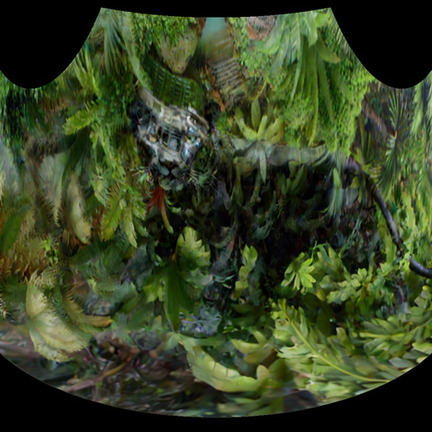}
        \vspace*{-6mm}
        \caption*{\centering \tiny \textit{a diorama of} \par\nobreak panther}
    \end{subfigure}    \hfill
    \begin{subfigure}{0.158\textwidth}
        \centering
        \includegraphics[width=\textwidth]{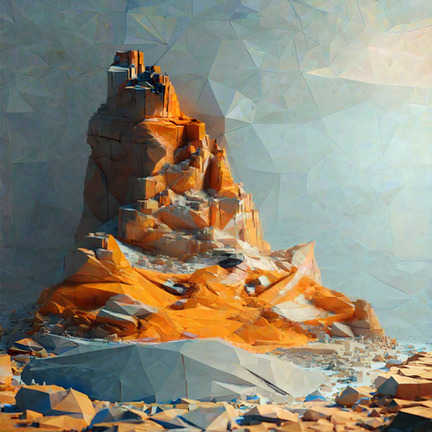}
        \vspace*{-6mm}
        \caption*{\centering \tiny \textit{a low-poly model of} \par\nobreak sunlit rocky outcrop}
    \end{subfigure}
    \begin{subfigure}{0.158\textwidth}
        \centering
        \includegraphics[width=\textwidth]{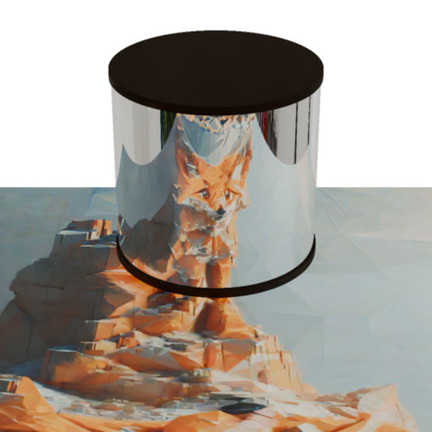}
        \vspace*{-6mm}
        \caption*{\centering \tiny \textit{\quad} \par\nobreak \quad}
    \end{subfigure}
    \begin{subfigure}{0.158\textwidth}
        \centering
        \includegraphics[width=\textwidth]{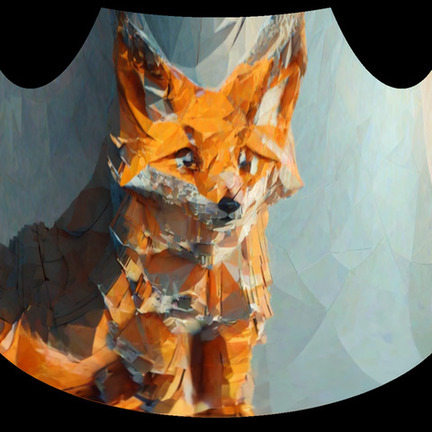}
        \vspace*{-6mm}
        \caption*{\centering \tiny \textit{a low-poly model of} \par\nobreak fox}
    \end{subfigure}
    \begin{subfigure}{0.158\textwidth}
        \centering
        \includegraphics[width=\textwidth]{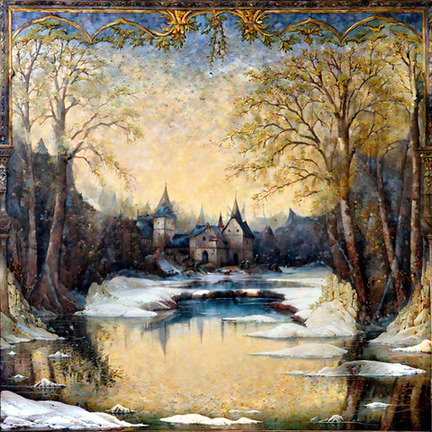}
        \vspace*{-6mm}
        \caption*{\centering \tiny \textit{a fresco painting of} \par\nobreak mirror-like frozen pond}
    \end{subfigure}
    \begin{subfigure}{0.158\textwidth}
        \centering
        \includegraphics[width=\textwidth]{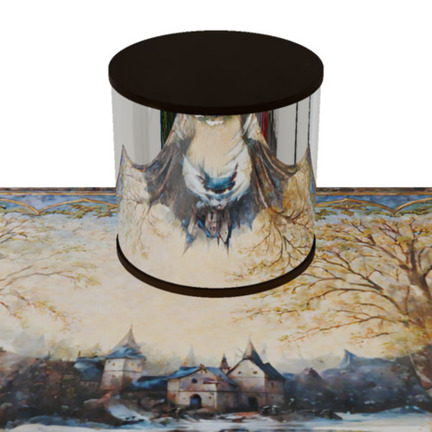}
        \vspace*{-6mm}
        \caption*{\centering \tiny \textit{\quad} \par\nobreak \quad}
    \end{subfigure}
    \begin{subfigure}{0.158\textwidth}
        \centering
        \includegraphics[width=\textwidth]{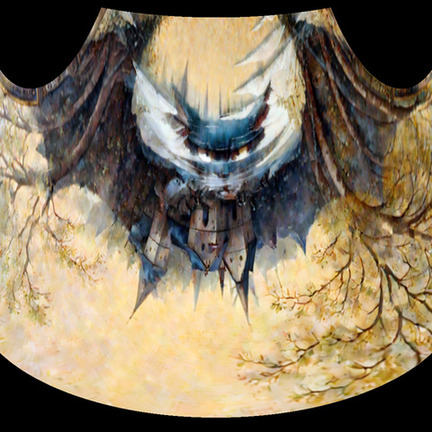}
        \vspace*{-6mm}
        \caption*{\centering \tiny \textit{a fresco painting of} \par\nobreak bat}
    \end{subfigure}    \hfill
    \begin{subfigure}{0.158\textwidth}
        \centering
        \includegraphics[width=\textwidth]{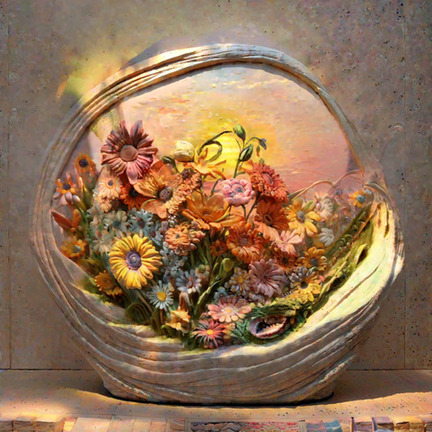}
        \vspace*{-6mm}
        \caption*{\centering \tiny \textit{a clay sculpture of} \par\nobreak flower meadow at sunrise}
    \end{subfigure}
    \begin{subfigure}{0.158\textwidth}
        \centering
        \includegraphics[width=\textwidth]{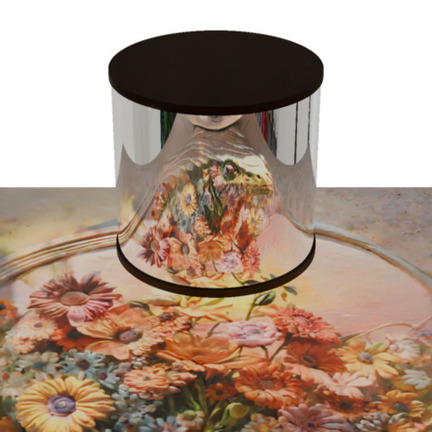}
        \vspace*{-6mm}
        \caption*{\centering \tiny \textit{\quad} \par\nobreak \quad}
    \end{subfigure}
    \begin{subfigure}{0.158\textwidth}
        \centering
        \includegraphics[width=\textwidth]{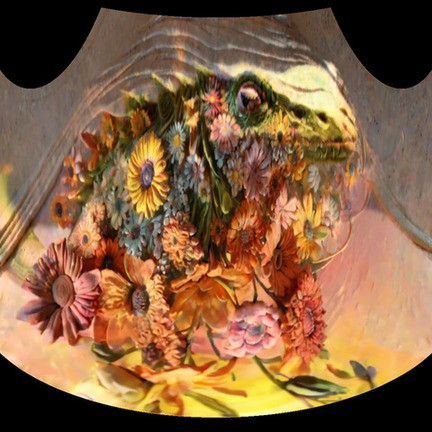}
        \vspace*{-6mm}
        \caption*{\centering \tiny \textit{a clay sculpture of} \par\nobreak iguana}
    \end{subfigure}
    \begin{subfigure}{0.158\textwidth}
        \centering
        \includegraphics[width=\textwidth]{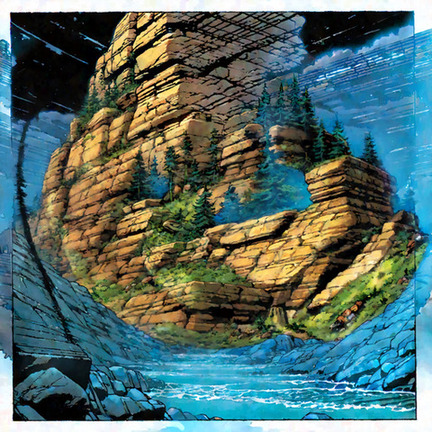}
        \vspace*{-6mm}
        \caption*{\centering \tiny \textit{a comic book panel of} \par\nobreak sunlit rocky outcrop}
    \end{subfigure}
    \begin{subfigure}{0.158\textwidth}
        \centering
        \includegraphics[width=\textwidth]{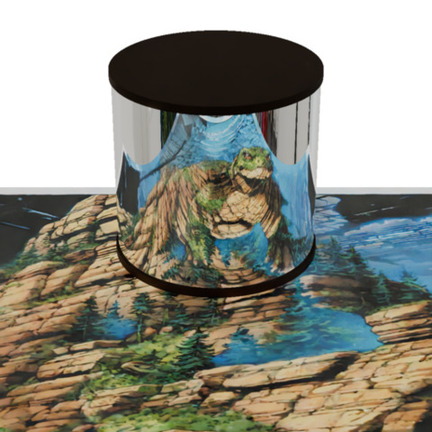}
        \vspace*{-6mm}
        \caption*{\centering \tiny \textit{\quad} \par\nobreak \quad}
    \end{subfigure}
    \begin{subfigure}{0.158\textwidth}
        \centering
        \includegraphics[width=\textwidth]{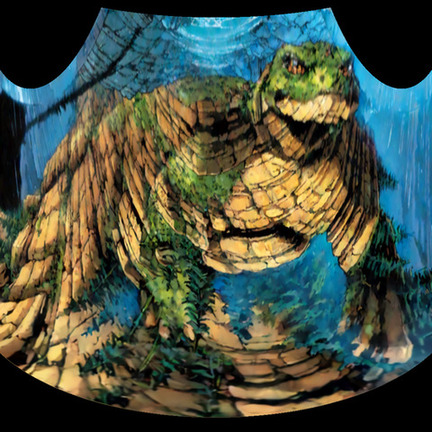}
        \vspace*{-6mm}
        \caption*{\centering \tiny \textit{a comic book panel of} \par\nobreak turtle}
    \end{subfigure}    \hfill
    \begin{subfigure}{0.158\textwidth}
        \centering
        \includegraphics[width=\textwidth]{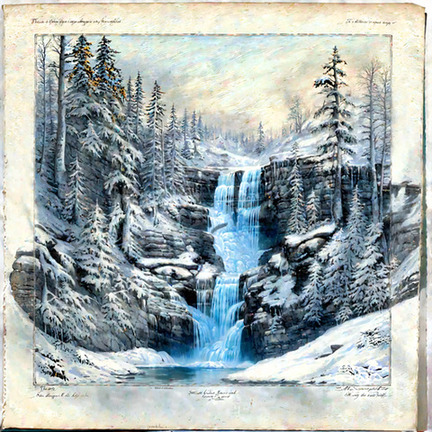}
        \vspace*{-6mm}
        \caption*{\centering \tiny \textit{a lithograph of} \par\nobreak frozen waterfall}
    \end{subfigure}
    \begin{subfigure}{0.158\textwidth}
        \centering
        \includegraphics[width=\textwidth]{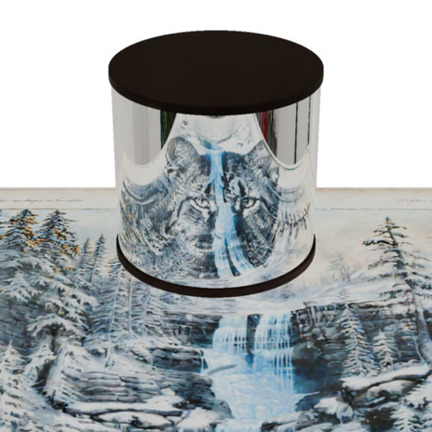}
        \vspace*{-6mm}
        \caption*{\centering \tiny \textit{\quad} \par\nobreak \quad}
    \end{subfigure}
    \begin{subfigure}{0.158\textwidth}
        \centering
        \includegraphics[width=\textwidth]{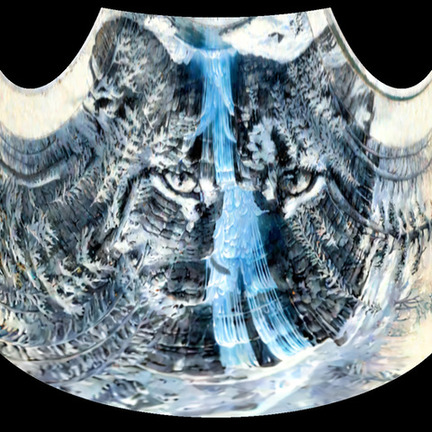}
        \vspace*{-6mm}
        \caption*{\centering \tiny \textit{a lithograph of} \par\nobreak cougar}
    \end{subfigure}
    \begin{subfigure}{0.158\textwidth}
        \centering
        \includegraphics[width=\textwidth]{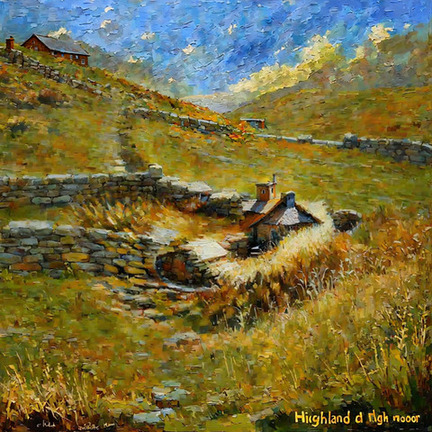}
        \vspace*{-6mm}
        \caption*{\centering \tiny \textit{an oil painting of} \par\nobreak highland moor with stone walls}
    \end{subfigure}
    \begin{subfigure}{0.158\textwidth}
        \centering
        \includegraphics[width=\textwidth]{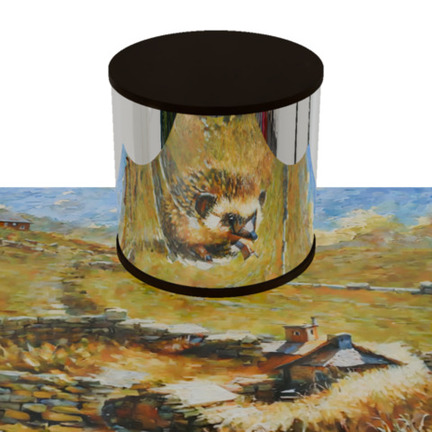}
        \vspace*{-6mm}
        \caption*{\centering \tiny \textit{\quad} \par\nobreak \quad}
    \end{subfigure}
    \begin{subfigure}{0.158\textwidth}
        \centering
        \includegraphics[width=\textwidth]{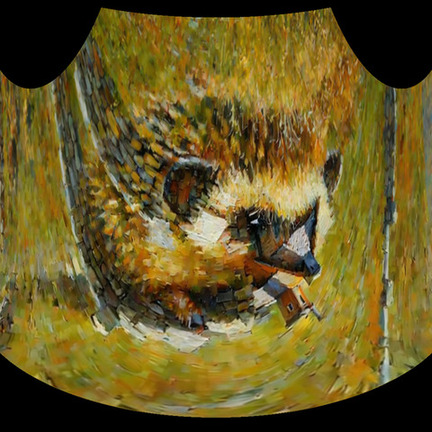}
        \vspace*{-6mm}
        \caption*{\centering \tiny \textit{an oil painting of} \par\nobreak hedgehog}
    \end{subfigure}    \hfill
    \begin{subfigure}{0.158\textwidth}
        \centering
        \includegraphics[width=\textwidth]{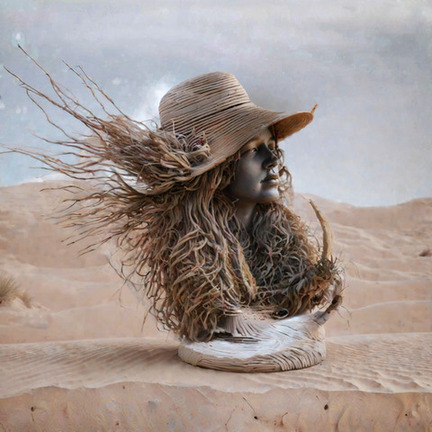}
        \vspace*{-6mm}
        \caption*{\centering \tiny \textit{a hyperrealistic sculpture of} \par\nobreak windy desert}
    \end{subfigure}
    \begin{subfigure}{0.158\textwidth}
        \centering
        \includegraphics[width=\textwidth]{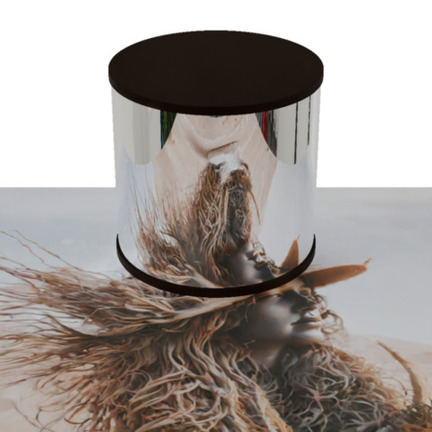}
        \vspace*{-6mm}
        \caption*{\centering \tiny \textit{\quad} \par\nobreak \quad}
    \end{subfigure}
    \begin{subfigure}{0.158\textwidth}
        \centering
        \includegraphics[width=\textwidth]{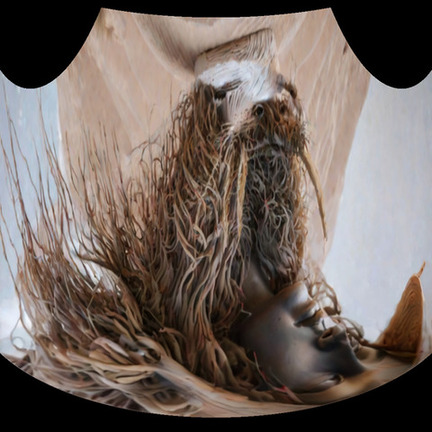}
        \vspace*{-6mm}
        \caption*{\centering \tiny \textit{a hyperrealistic sculpture of} \par\nobreak walrus}
    \end{subfigure}
\end{figure*}

\begin{figure*}[h!]
    \centering
    \begin{subfigure}{0.106\textwidth}
        \centering
        \includegraphics[width=\textwidth]{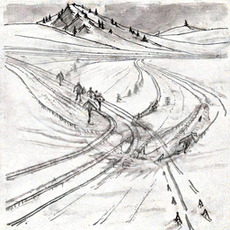}
        \vspace*{-6mm}
        \caption*{\centering \tiny \textit{an ink wash drawing of} \par\nobreak straight ski tracks}
    \end{subfigure}
    \begin{subfigure}{0.106\textwidth}
        \centering
        \includegraphics[width=\textwidth]{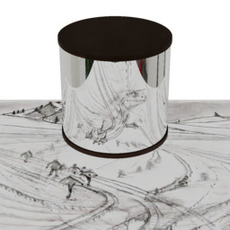}
        \vspace*{-6mm}
        \caption*{\centering \tiny \textit{\quad} \par\nobreak \quad}
    \end{subfigure}
    \begin{subfigure}{0.106\textwidth}
        \centering
        \includegraphics[width=\textwidth]{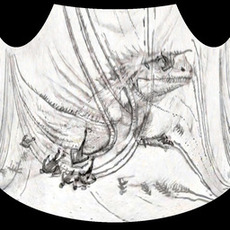}
        \vspace*{-6mm}
        \caption*{\centering \tiny \textit{an ink wash drawing of} \par\nobreak iguana}
    \end{subfigure}    \hfill
    \begin{subfigure}{0.106\textwidth}
        \centering
        \includegraphics[width=\textwidth]{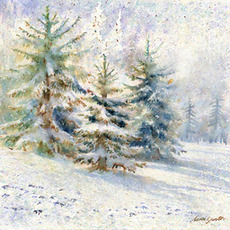}
        \vspace*{-6mm}
        \caption*{\centering \tiny \textit{a watercolor painting of} \par\nobreak snow-covered green trees}
    \end{subfigure}
    \begin{subfigure}{0.106\textwidth}
        \centering
        \includegraphics[width=\textwidth]{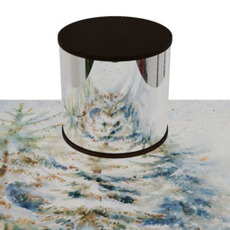}
        \vspace*{-6mm}
        \caption*{\centering \tiny \textit{\quad} \par\nobreak \quad}
    \end{subfigure}
    \begin{subfigure}{0.106\textwidth}
        \centering
        \includegraphics[width=\textwidth]{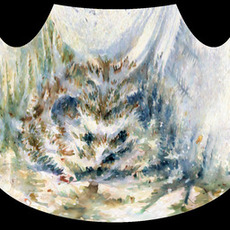}
        \vspace*{-6mm}
        \caption*{\centering \tiny \textit{a watercolor painting of} \par\nobreak hedgehog}
    \end{subfigure}    \hfill
    \begin{subfigure}{0.106\textwidth}
        \centering
        \includegraphics[width=\textwidth]{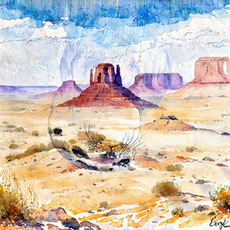}
        \vspace*{-6mm}
        \caption*{\centering \tiny \textit{a watercolor painting of} \par\nobreak desert plateau}
    \end{subfigure}
    \begin{subfigure}{0.106\textwidth}
        \centering
        \includegraphics[width=\textwidth]{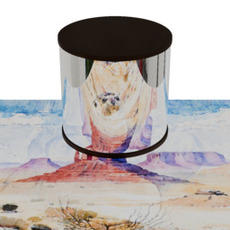}
        \vspace*{-6mm}
        \caption*{\centering \tiny \textit{\quad} \par\nobreak \quad}
    \end{subfigure}
    \begin{subfigure}{0.106\textwidth}
        \centering
        \includegraphics[width=\textwidth]{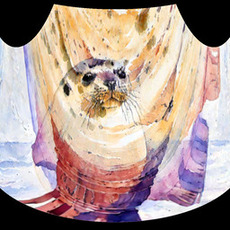}
        \vspace*{-6mm}
        \caption*{\centering \tiny \textit{a watercolor painting of} \par\nobreak seal}
    \end{subfigure}
    \begin{subfigure}{0.106\textwidth}
        \centering
        \includegraphics[width=\textwidth]{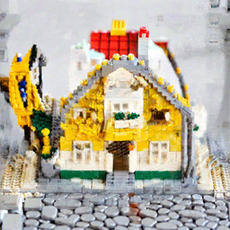}
        \vspace*{-6mm}
        \caption*{\centering \tiny \textit{a LEGO model of} \par\nobreak cobblestone street}
    \end{subfigure}
    \begin{subfigure}{0.106\textwidth}
        \centering
        \includegraphics[width=\textwidth]{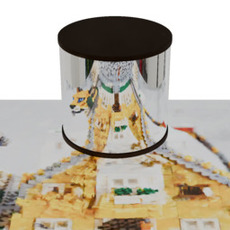}
        \vspace*{-6mm}
        \caption*{\centering \tiny \textit{\quad} \par\nobreak \quad}
    \end{subfigure}
    \begin{subfigure}{0.106\textwidth}
        \centering
        \includegraphics[width=\textwidth]{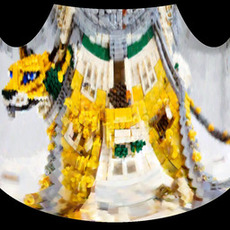}
        \vspace*{-6mm}
        \caption*{\centering \tiny \textit{a LEGO model of} \par\nobreak cheetah}
    \end{subfigure}    \hfill
    \begin{subfigure}{0.106\textwidth}
        \centering
        \includegraphics[width=\textwidth]{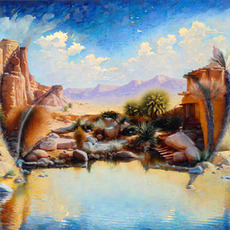}
        \vspace*{-6mm}
        \caption*{\centering \tiny \textit{a photorealistic painting} \par\nobreak \textit{of} desert oasis}
    \end{subfigure}
    \begin{subfigure}{0.106\textwidth}
        \centering
        \includegraphics[width=\textwidth]{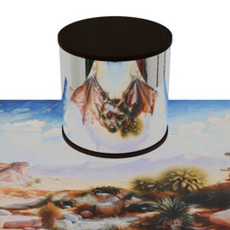}
        \vspace*{-6mm}
        \caption*{\centering \tiny \textit{\quad} \par\nobreak \quad}
    \end{subfigure}
    \begin{subfigure}{0.106\textwidth}
        \centering
        \includegraphics[width=\textwidth]{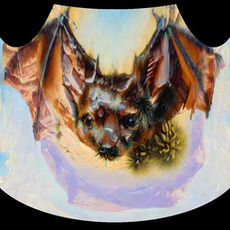}
        \vspace*{-6mm}
        \caption*{\centering \tiny \textit{a photorealistic painting} \par\nobreak \textit{of} bat}
    \end{subfigure}    \hfill
    \begin{subfigure}{0.106\textwidth}
        \centering
        \includegraphics[width=\textwidth]{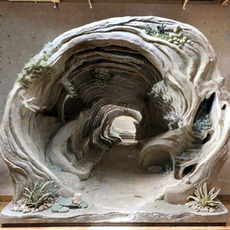}
        \vspace*{-6mm}
        \caption*{\centering \tiny \textit{a clay sculpture of} \par\nobreak aquarium tunnel}
    \end{subfigure}
    \begin{subfigure}{0.106\textwidth}
        \centering
        \includegraphics[width=\textwidth]{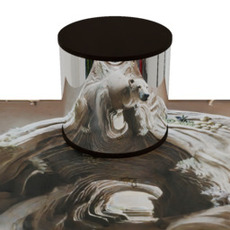}
        \vspace*{-6mm}
        \caption*{\centering \tiny \textit{\quad} \par\nobreak \quad}
    \end{subfigure}
    \begin{subfigure}{0.106\textwidth}
        \centering
        \includegraphics[width=\textwidth]{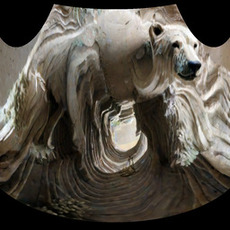}
        \vspace*{-6mm}
        \caption*{\centering \tiny \textit{a clay sculpture of} \par\nobreak polar bear}
    \end{subfigure}
    \begin{subfigure}{0.106\textwidth}
        \centering
        \includegraphics[width=\textwidth]{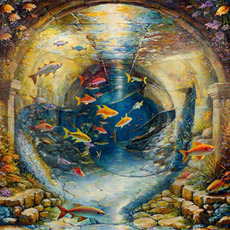}
        \vspace*{-6mm}
        \caption*{\centering \tiny \textit{a fresco painting of} \par\nobreak aquarium tunnel}
    \end{subfigure}
    \begin{subfigure}{0.106\textwidth}
        \centering
        \includegraphics[width=\textwidth]{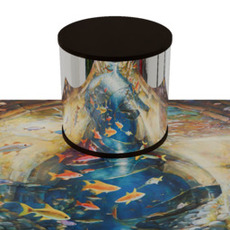}
        \vspace*{-6mm}
        \caption*{\centering \tiny \textit{\quad} \par\nobreak \quad}
    \end{subfigure}
    \begin{subfigure}{0.106\textwidth}
        \centering
        \includegraphics[width=\textwidth]{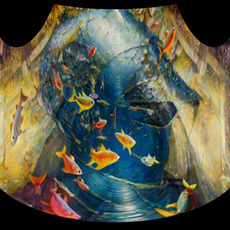}
        \vspace*{-6mm}
        \caption*{\centering \tiny \textit{a fresco painting of} \par\nobreak knight's helmet}
    \end{subfigure}    \hfill
    \begin{subfigure}{0.106\textwidth}
        \centering
        \includegraphics[width=\textwidth]{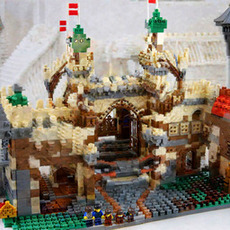}
        \vspace*{-6mm}
        \caption*{\centering \tiny \textit{a LEGO model of} \par\nobreak medieval castle gate}
    \end{subfigure}
    \begin{subfigure}{0.106\textwidth}
        \centering
        \includegraphics[width=\textwidth]{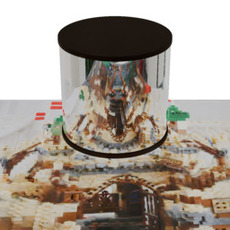}
        \vspace*{-6mm}
        \caption*{\centering \tiny \textit{\quad} \par\nobreak \quad}
    \end{subfigure}
    \begin{subfigure}{0.106\textwidth}
        \centering
        \includegraphics[width=\textwidth]{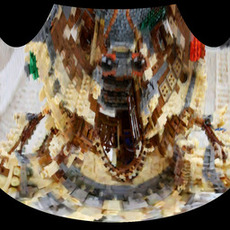}
        \vspace*{-6mm}
        \caption*{\centering \tiny \textit{a LEGO model of} \par\nobreak ant}
    \end{subfigure}    \hfill
    \begin{subfigure}{0.106\textwidth}
        \centering
        \includegraphics[width=\textwidth]{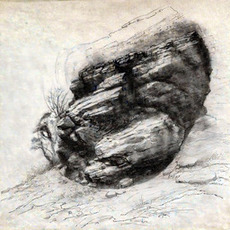}
        \vspace*{-6mm}
        \caption*{\centering \tiny \textit{a charcoal drawing of} \par\nobreak sunlit rocky outcrop}
    \end{subfigure}
    \begin{subfigure}{0.106\textwidth}
        \centering
        \includegraphics[width=\textwidth]{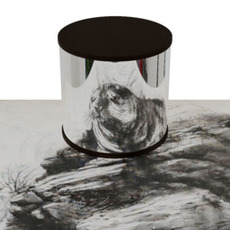}
        \vspace*{-6mm}
        \caption*{\centering \tiny \textit{\quad} \par\nobreak \quad}
    \end{subfigure}
    \begin{subfigure}{0.106\textwidth}
        \centering
        \includegraphics[width=\textwidth]{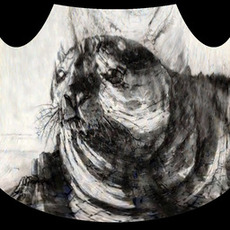}
        \vspace*{-6mm}
        \caption*{\centering \tiny \textit{a charcoal drawing of} \par\nobreak seal}
    \end{subfigure}
    \begin{subfigure}{0.106\textwidth}
        \centering
        \includegraphics[width=\textwidth]{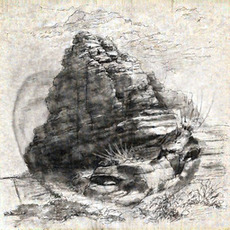}
        \vspace*{-6mm}
        \caption*{\centering \tiny \textit{a charcoal drawing of} \par\nobreak sunlit rocky outcrop}
    \end{subfigure}
    \begin{subfigure}{0.106\textwidth}
        \centering
        \includegraphics[width=\textwidth]{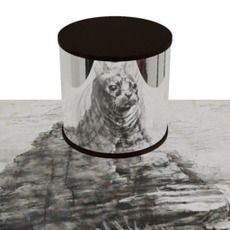}
        \vspace*{-6mm}
        \caption*{\centering \tiny \textit{\quad} \par\nobreak \quad}
    \end{subfigure}
    \begin{subfigure}{0.106\textwidth}
        \centering
        \includegraphics[width=\textwidth]{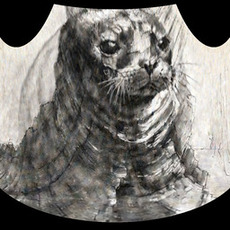}
        \vspace*{-6mm}
        \caption*{\centering \tiny \textit{a charcoal drawing of} \par\nobreak seal}
    \end{subfigure}    \hfill
    \begin{subfigure}{0.106\textwidth}
        \centering
        \includegraphics[width=\textwidth]{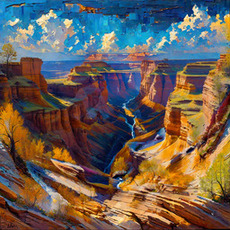}
        \vspace*{-6mm}
        \caption*{\centering \tiny \textit{an oil painting of} sunlit canyon with straight cliffs}
    \end{subfigure}
    \begin{subfigure}{0.106\textwidth}
        \centering
        \includegraphics[width=\textwidth]{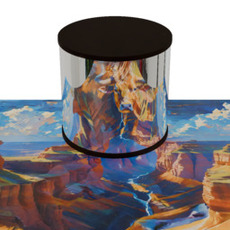}
        \vspace*{-6mm}
        \caption*{\centering \tiny \textit{\quad} \par\nobreak \quad}
    \end{subfigure}
    \begin{subfigure}{0.106\textwidth}
        \centering
        \includegraphics[width=\textwidth]{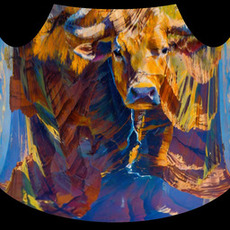}
        \vspace*{-6mm}
        \caption*{\centering \tiny \textit{an oil painting of} \par\nobreak bull}
    \end{subfigure}    \hfill
    \begin{subfigure}{0.106\textwidth}
        \centering
        \includegraphics[width=\textwidth]{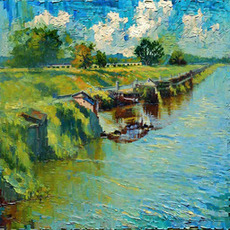}
        \vspace*{-6mm}
        \caption*{\centering \tiny \textit{an oil painting of} straight levee dividing water}
    \end{subfigure}
    \begin{subfigure}{0.106\textwidth}
        \centering
        \includegraphics[width=\textwidth]{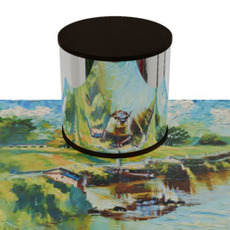}
        \vspace*{-6mm}
        \caption*{\centering \tiny \textit{\quad} \par\nobreak \quad}
    \end{subfigure}
    \begin{subfigure}{0.106\textwidth}
        \centering
        \includegraphics[width=\textwidth]{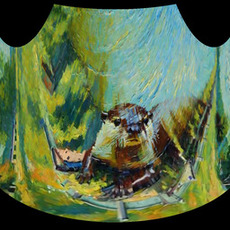}
        \vspace*{-6mm}
        \caption*{\centering \tiny \textit{an oil painting of} \par\nobreak otter}
    \end{subfigure}
    \begin{subfigure}{0.106\textwidth}
        \centering
        \includegraphics[width=\textwidth]{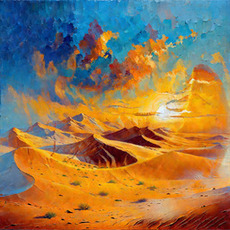}
        \vspace*{-6mm}
        \caption*{\centering \tiny \textit{an oil painting of} \par\nobreak desert dunes at sunset}
    \end{subfigure}
    \begin{subfigure}{0.106\textwidth}
        \centering
        \includegraphics[width=\textwidth]{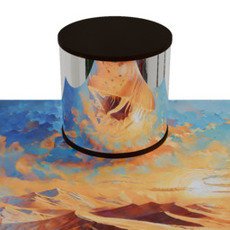}
        \vspace*{-6mm}
        \caption*{\centering \tiny \textit{\quad} \par\nobreak \quad}
    \end{subfigure}
    \begin{subfigure}{0.106\textwidth}
        \centering
        \includegraphics[width=\textwidth]{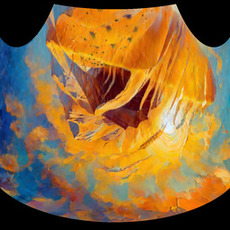}
        \vspace*{-6mm}
        \caption*{\centering \tiny \textit{an oil painting of} \par\nobreak jellyfish}
    \end{subfigure}    \hfill
    \begin{subfigure}{0.106\textwidth}
        \centering
        \includegraphics[width=\textwidth]{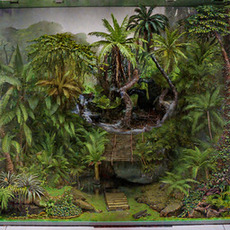}
        \vspace*{-6mm}
        \caption*{\centering \tiny \textit{a diorama of} \par\nobreak dense tropical rainforest}
    \end{subfigure}
    \begin{subfigure}{0.106\textwidth}
        \centering
        \includegraphics[width=\textwidth]{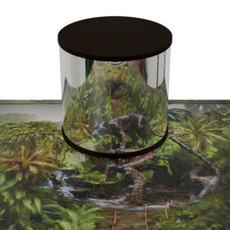}
        \vspace*{-6mm}
        \caption*{\centering \tiny \textit{\quad} \par\nobreak \quad}
    \end{subfigure}
    \begin{subfigure}{0.106\textwidth}
        \centering
        \includegraphics[width=\textwidth]{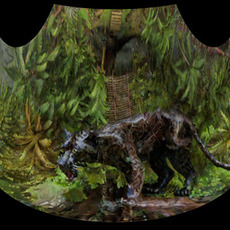}
        \vspace*{-6mm}
        \caption*{\centering \tiny \textit{a diorama of} \par\nobreak panther}
    \end{subfigure}    \hfill
    \begin{subfigure}{0.106\textwidth}
        \centering
        \includegraphics[width=\textwidth]{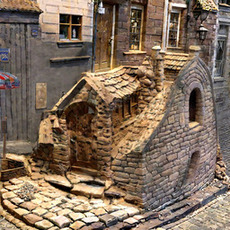}
        \vspace*{-6mm}
        \caption*{\centering \tiny \textit{a clay sculpture of} \par\nobreak cobblestone street}
    \end{subfigure}
    \begin{subfigure}{0.106\textwidth}
        \centering
        \includegraphics[width=\textwidth]{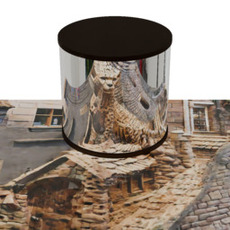}
        \vspace*{-6mm}
        \caption*{\centering \tiny \textit{\quad} \par\nobreak \quad}
    \end{subfigure}
    \begin{subfigure}{0.106\textwidth}
        \centering
        \includegraphics[width=\textwidth]{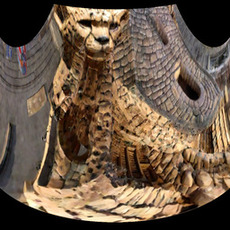}
        \vspace*{-6mm}
        \caption*{\centering \tiny \textit{a clay sculpture of} \par\nobreak cheetah}
    \end{subfigure}
    \begin{subfigure}{0.106\textwidth}
        \centering
        \includegraphics[width=\textwidth]{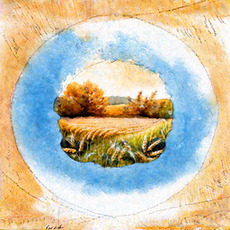}
        \vspace*{-6mm}
        \caption*{\centering \tiny \textit{an ink wash drawing of} \par\nobreak horizon of a wheat field}
    \end{subfigure}
    \begin{subfigure}{0.106\textwidth}
        \centering
        \includegraphics[width=\textwidth]{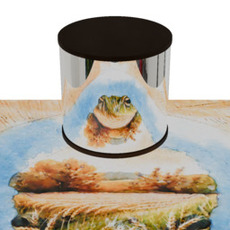}
        \vspace*{-6mm}
        \caption*{\centering \tiny \textit{\quad} \par\nobreak \quad}
    \end{subfigure}
    \begin{subfigure}{0.106\textwidth}
        \centering
        \includegraphics[width=\textwidth]{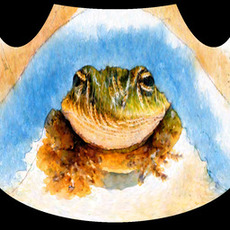}
        \vspace*{-6mm}
        \caption*{\centering \tiny \textit{an ink wash drawing of} \par\nobreak frog}
    \end{subfigure}    \hfill
    \begin{subfigure}{0.106\textwidth}
        \centering
        \includegraphics[width=\textwidth]{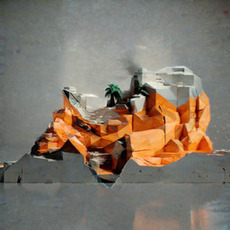}
        \vspace*{-6mm}
        \caption*{\centering \tiny \textit{a low-poly model of} \par\nobreak sunlit rocky outcrop}
    \end{subfigure}
    \begin{subfigure}{0.106\textwidth}
        \centering
        \includegraphics[width=\textwidth]{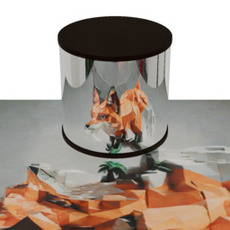}
        \vspace*{-6mm}
        \caption*{\centering \tiny \textit{\quad} \par\nobreak \quad}
    \end{subfigure}
    \begin{subfigure}{0.106\textwidth}
        \centering
        \includegraphics[width=\textwidth]{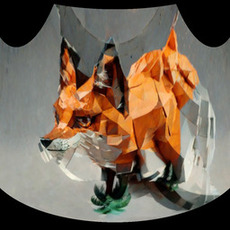}
        \vspace*{-6mm}
        \caption*{\centering \tiny \textit{a low-poly model of} \par\nobreak fox}
    \end{subfigure}    \hfill
    \begin{subfigure}{0.106\textwidth}
        \centering
        \includegraphics[width=\textwidth]{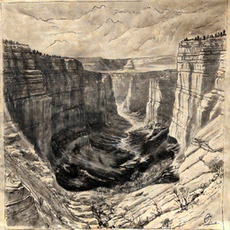}
        \vspace*{-6mm}
        \caption*{\centering \tiny \textit{a charcoal drawing of} sunlit canyon with cliffs}
    \end{subfigure}
    \begin{subfigure}{0.106\textwidth}
        \centering
        \includegraphics[width=\textwidth]{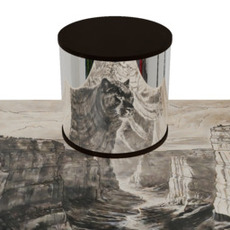}
        \vspace*{-6mm}
        \caption*{\centering \tiny \textit{\quad} \par\nobreak \quad}
    \end{subfigure}
    \begin{subfigure}{0.106\textwidth}
        \centering
        \includegraphics[width=\textwidth]{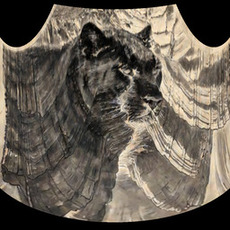}
        \vspace*{-6mm}
        \caption*{\centering \tiny \textit{a charcoal drawing of} \par\nobreak panther}
    \end{subfigure}
    \begin{subfigure}{0.106\textwidth}
        \centering
        \includegraphics[width=\textwidth]{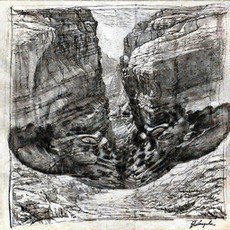}
        \vspace*{-6mm}
        \caption*{\centering \tiny \textit{a charcoal drawing of} sunlit canyon with cliffs}
    \end{subfigure}
    \begin{subfigure}{0.106\textwidth}
        \centering
        \includegraphics[width=\textwidth]{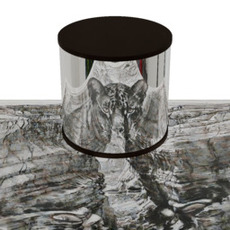}
        \vspace*{-6mm}
        \caption*{\centering \tiny \textit{\quad} \par\nobreak \quad}
    \end{subfigure}
    \begin{subfigure}{0.106\textwidth}
        \centering
        \includegraphics[width=\textwidth]{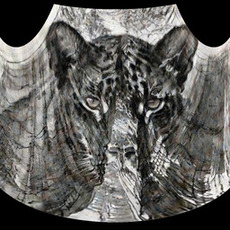}
        \vspace*{-6mm}
        \caption*{\centering \tiny \textit{a charcoal drawing of} \par\nobreak panther}
    \end{subfigure}    \hfill
    \begin{subfigure}{0.106\textwidth}
        \centering
        \includegraphics[width=\textwidth]{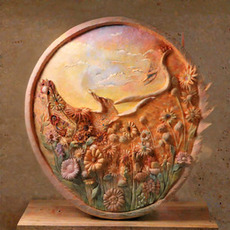}
        \vspace*{-6mm}
        \caption*{\centering \tiny \textit{a clay sculpture of} \par\nobreak flower meadow at sunrise}
    \end{subfigure}
    \begin{subfigure}{0.106\textwidth}
        \centering
        \includegraphics[width=\textwidth]{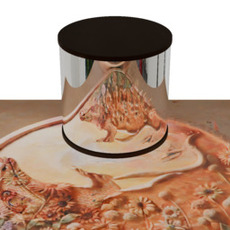}
        \vspace*{-6mm}
        \caption*{\centering \tiny \textit{\quad} \par\nobreak \quad}
    \end{subfigure}
    \begin{subfigure}{0.106\textwidth}
        \centering
        \includegraphics[width=\textwidth]{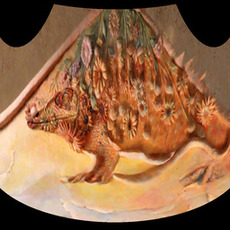}
        \vspace*{-6mm}
        \caption*{\centering \tiny \textit{a clay sculpture of} \par\nobreak iguana}
    \end{subfigure}    \hfill
    \begin{subfigure}{0.106\textwidth}
        \centering
        \includegraphics[width=\textwidth]{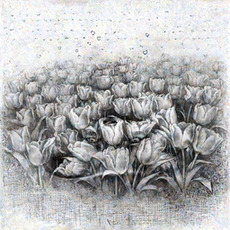}
        \vspace*{-6mm}
        \caption*{\centering \tiny \textit{a charcoal drawing of} \par\nobreak flower garden with tulips}
    \end{subfigure}
    \begin{subfigure}{0.106\textwidth}
        \centering
        \includegraphics[width=\textwidth]{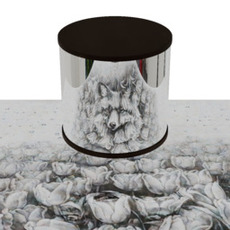}
        \vspace*{-6mm}
        \caption*{\centering \tiny \textit{\quad} \par\nobreak \quad}
    \end{subfigure}
    \begin{subfigure}{0.106\textwidth}
        \centering
        \includegraphics[width=\textwidth]{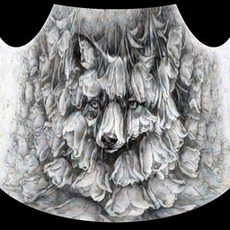}
        \vspace*{-6mm}
        \caption*{\centering \tiny \textit{a charcoal drawing of} \par\nobreak fox}
    \end{subfigure}
    \begin{subfigure}{0.106\textwidth}
        \centering
        \includegraphics[width=\textwidth]{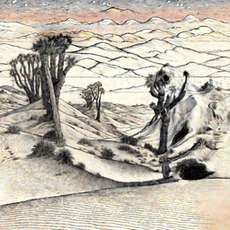}
        \vspace*{-6mm}
        \caption*{\centering \tiny \textit{a line drawing of} \par\nobreak desert dunes at sunset}
    \end{subfigure}
    \begin{subfigure}{0.106\textwidth}
        \centering
        \includegraphics[width=\textwidth]{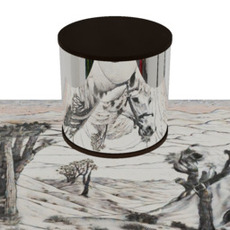}
        \vspace*{-6mm}
        \caption*{\centering \tiny \textit{\quad} \par\nobreak \quad}
    \end{subfigure}
    \begin{subfigure}{0.106\textwidth}
        \centering
        \includegraphics[width=\textwidth]{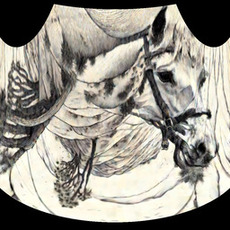}
        \vspace*{-6mm}
        \caption*{\centering \tiny \textit{a line drawing of} \par\nobreak horse}
    \end{subfigure}    \hfill
    \begin{subfigure}{0.106\textwidth}
        \centering
        \includegraphics[width=\textwidth]{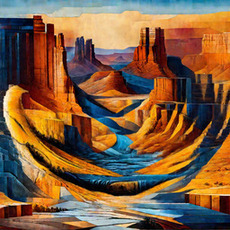}
        \vspace*{-6mm}
        \caption*{\centering \tiny \textit{a cubist interpretation of} \par\nobreak desert canyon floor}
    \end{subfigure}
    \begin{subfigure}{0.106\textwidth}
        \centering
        \includegraphics[width=\textwidth]{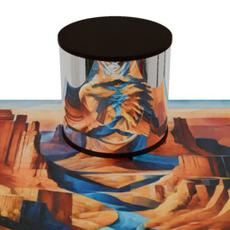}
        \vspace*{-6mm}
        \caption*{\centering \tiny \textit{\quad} \par\nobreak \quad}
    \end{subfigure}
    \begin{subfigure}{0.106\textwidth}
        \centering
        \includegraphics[width=\textwidth]{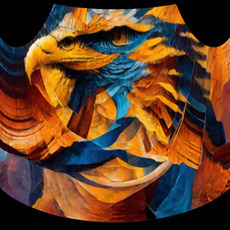}
        \vspace*{-6mm}
        \caption*{\centering \tiny \textit{a cubist interpretation of} \par\nobreak eagle}
    \end{subfigure}    \hfill
    \begin{subfigure}{0.106\textwidth}
        \centering
        \includegraphics[width=\textwidth]{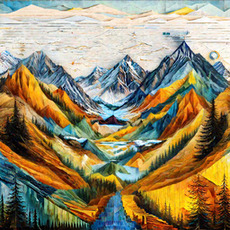}
        \vspace*{-6mm}
        \caption*{\centering \tiny \textit{a paper collage of} \par\nobreak mountain pass}
    \end{subfigure}
    \begin{subfigure}{0.106\textwidth}
        \centering
        \includegraphics[width=\textwidth]{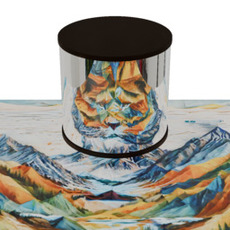}
        \vspace*{-6mm}
        \caption*{\centering \tiny \textit{\quad} \par\nobreak \quad}
    \end{subfigure}
    \begin{subfigure}{0.106\textwidth}
        \centering
        \includegraphics[width=\textwidth]{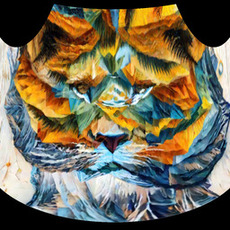}
        \vspace*{-6mm}
        \caption*{\centering \tiny \textit{a paper collage of} \par\nobreak lion}
    \end{subfigure}
    \begin{subfigure}{0.106\textwidth}
        \centering
        \includegraphics[width=\textwidth]{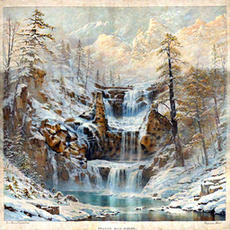}
        \vspace*{-6mm}
        \caption*{\centering \tiny \textit{a lithograph of} \par\nobreak frozen waterfall}
    \end{subfigure}
    \begin{subfigure}{0.106\textwidth}
        \centering
        \includegraphics[width=\textwidth]{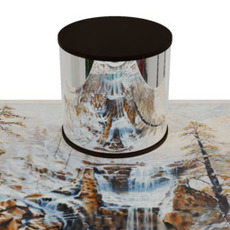}
        \vspace*{-6mm}
        \caption*{\centering \tiny \textit{\quad} \par\nobreak \quad}
    \end{subfigure}
    \begin{subfigure}{0.106\textwidth}
        \centering
        \includegraphics[width=\textwidth]{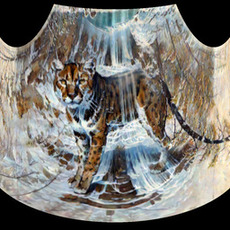}
        \vspace*{-6mm}
        \caption*{\centering \tiny \textit{a lithograph of} \par\nobreak cougar}
    \end{subfigure}    \hfill
    \begin{subfigure}{0.106\textwidth}
        \centering
        \includegraphics[width=\textwidth]{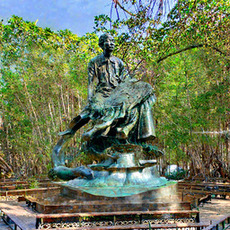}
        \vspace*{-6mm}
        \caption*{\centering \tiny \textit{a bronze statue of} \par\nobreak mangrove forest}
    \end{subfigure}
    \begin{subfigure}{0.106\textwidth}
        \centering
        \includegraphics[width=\textwidth]{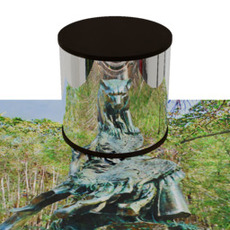}
        \vspace*{-6mm}
        \caption*{\centering \tiny \textit{\quad} \par\nobreak \quad}
    \end{subfigure}
    \begin{subfigure}{0.106\textwidth}
        \centering
        \includegraphics[width=\textwidth]{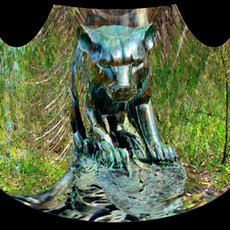}
        \vspace*{-6mm}
        \caption*{\centering \tiny \textit{a bronze statue of} \par\nobreak cougar}
    \end{subfigure}    \hfill
    \begin{subfigure}{0.106\textwidth}
        \centering
        \includegraphics[width=\textwidth]{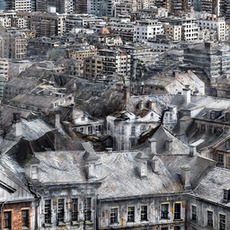}
        \vspace*{-6mm}
        \caption*{\centering \tiny \textit{a metal engraving of} \par\nobreak rooftops of a dense city}
    \end{subfigure}
    \begin{subfigure}{0.106\textwidth}
        \centering
        \includegraphics[width=\textwidth]{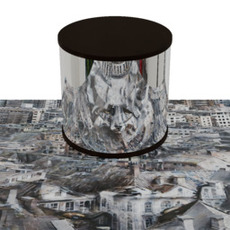}
        \vspace*{-6mm}
        \caption*{\centering \tiny \textit{\quad} \par\nobreak \quad}
    \end{subfigure}
    \begin{subfigure}{0.106\textwidth}
        \centering
        \includegraphics[width=\textwidth]{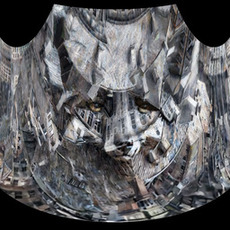}
        \vspace*{-6mm}
        \caption*{\centering \tiny \textit{a metal engraving of} \par\nobreak fox}
    \end{subfigure}
    \begin{subfigure}{0.106\textwidth}
        \centering
        \includegraphics[width=\textwidth]{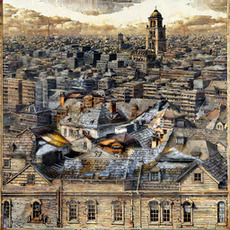}
        \vspace*{-6mm}
        \caption*{\centering \tiny \textit{a metal engraving of} \par\nobreak rooftops of a dense city}
    \end{subfigure}
    \begin{subfigure}{0.106\textwidth}
        \centering
        \includegraphics[width=\textwidth]{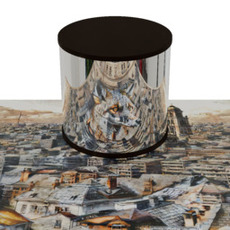}
        \vspace*{-6mm}
        \caption*{\centering \tiny \textit{\quad} \par\nobreak \quad}
    \end{subfigure}
    \begin{subfigure}{0.106\textwidth}
        \centering
        \includegraphics[width=\textwidth]{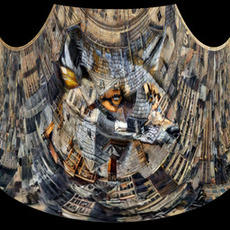}
        \vspace*{-6mm}
        \caption*{\centering \tiny \textit{a metal engraving of} \par\nobreak fox}
    \end{subfigure}    \hfill
    \begin{subfigure}{0.106\textwidth}
        \centering
        \includegraphics[width=\textwidth]{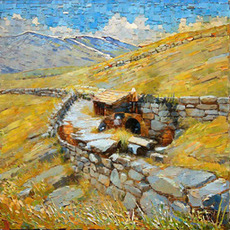}
        \vspace*{-6mm}
        \caption*{\centering \tiny \textit{oil painting of} highland moor with stone walls}
    \end{subfigure}
    \begin{subfigure}{0.106\textwidth}
        \centering
        \includegraphics[width=\textwidth]{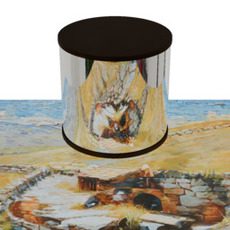}
        \vspace*{-6mm}
        \caption*{\centering \tiny \textit{\quad} \par\nobreak \quad}
    \end{subfigure}
    \begin{subfigure}{0.106\textwidth}
        \centering
        \includegraphics[width=\textwidth]{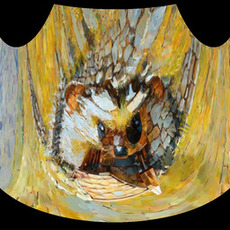}
        \vspace*{-6mm}
        \caption*{\centering \tiny \textit{oil painting of} \par\nobreak hedgehog}
    \end{subfigure}    \hfill
    \begin{subfigure}{0.106\textwidth}
        \centering
        \includegraphics[width=\textwidth]{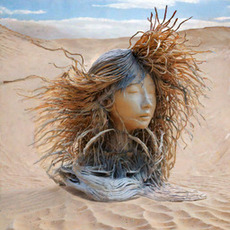}
        \vspace*{-6mm}
        \caption*{\centering \tiny \textit{a hyperrealistic sculpture of} windy desert}
    \end{subfigure}
    \begin{subfigure}{0.106\textwidth}
        \centering
        \includegraphics[width=\textwidth]{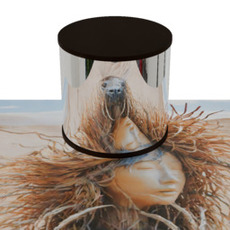}
        \vspace*{-6mm}
        \caption*{\centering \tiny \textit{\quad} \par\nobreak \quad}
    \end{subfigure}
    \begin{subfigure}{0.106\textwidth}
        \centering
        \includegraphics[width=\textwidth]{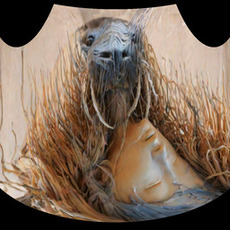}
        \vspace*{-6mm}
        \caption*{\centering \tiny \textit{a hyperrealistic sculpture of} walrus}
    \end{subfigure}
\end{figure*}

\begin{figure*}[h!]
    \centering
    \begin{subfigure}{0.490\textwidth}
        \centering
        \includegraphics[width=\textwidth]{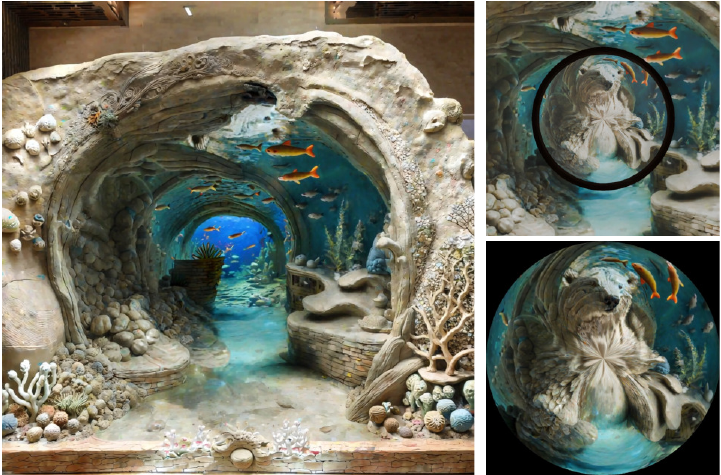}
        \caption*{\normalsize \centering \textit{a clay sculpture of} \par\nobreak aquarium tunnel / polar bear}
    \end{subfigure}    \hfill
    \begin{subfigure}{0.490\textwidth}
        \centering
        \includegraphics[width=\textwidth]{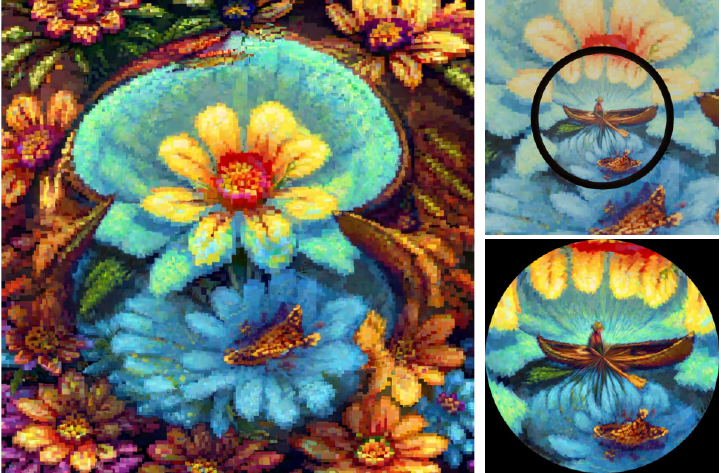}
        \caption*{\normalsize \centering \textit{a pixel art version of} \par\nobreak flower petals close-up / canoe}
    \end{subfigure}
    
    \vspace{1.0cm} %
    \begin{subfigure}{0.490\textwidth}
        \centering
        \includegraphics[width=\textwidth]{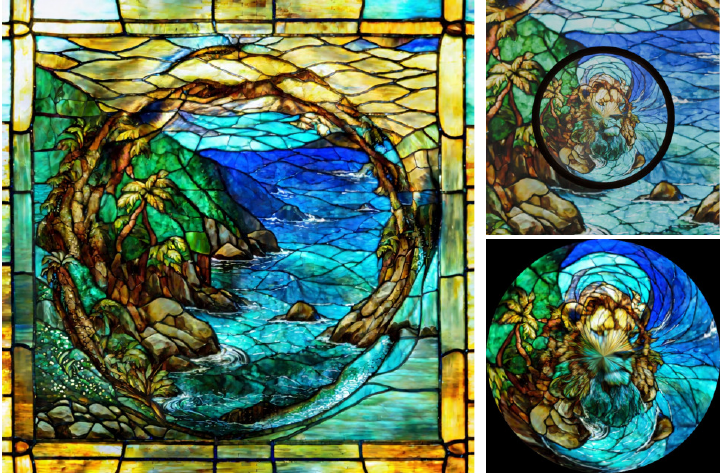}
        \caption*{\normalsize \centering \textit{a stained glass depiction of} \par\nobreak straight coastline / lion}
    \end{subfigure}    \hfill
    \begin{subfigure}{0.490\textwidth}
        \centering
        \includegraphics[width=\textwidth]{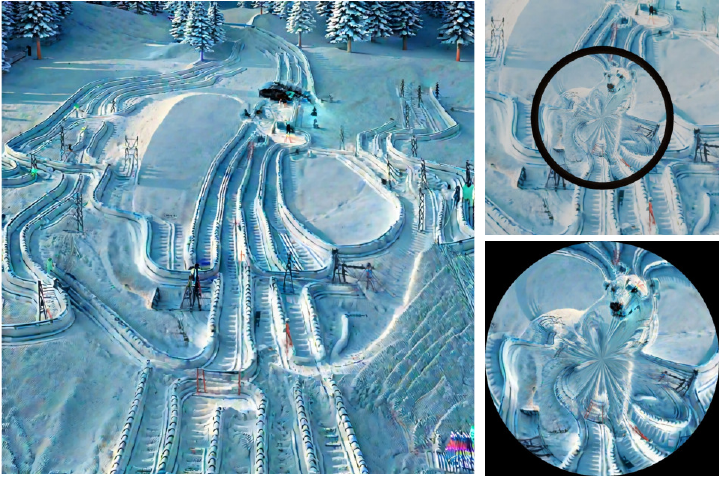}
        \caption*{\normalsize \centering \textit{a 3D rendering of} \par\nobreak straight ski tracks / polar bear}
    \end{subfigure}
    
    \vspace{1.0cm} %
    \begin{subfigure}{0.490\textwidth}
        \centering
        \includegraphics[width=\textwidth]{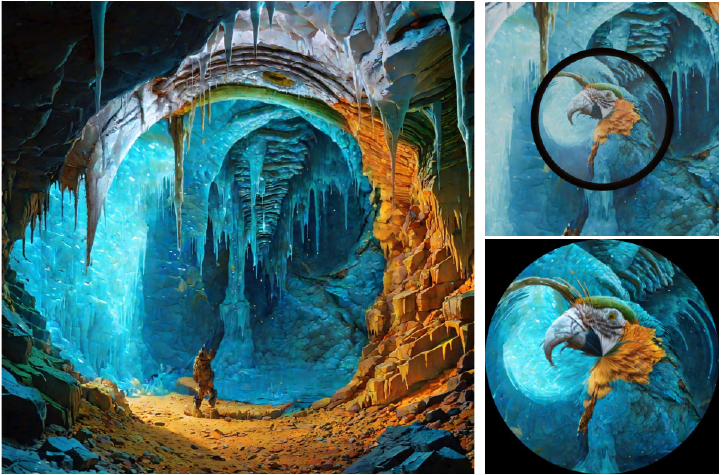}
        \caption*{\normalsize \centering \textit{a cinematic rendering of} \par\nobreak icy cave with stalactites / parrot}
    \end{subfigure}    \hfill
    \begin{subfigure}{0.490\textwidth}
        \centering
        \includegraphics[width=\textwidth]{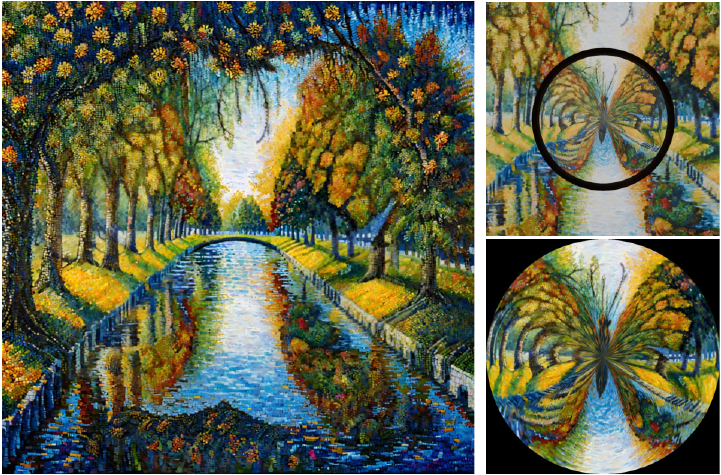}
        \caption*{\normalsize \centering \textit{a pointillism painting of} \par\nobreak straight canal lined with trees / butterfly}
    \end{subfigure}
        \caption{\textbf{Conic mirror anamorphosis.} In this figure and the two following ones, we show additional results for the conic mirror example. Each example contains the identity view, the mirror view as predicted by the flow model, and a rendering of the actual physical setting from the top to validate our examples. Kindly refer to the supplementary videos to see these results in action.}
    \label{fig:suppl_cone_b}
\end{figure*}

\begin{figure*}[h!]
    \centering
    \begin{subfigure}{0.158\textwidth}
        \centering
        \includegraphics[width=\textwidth]{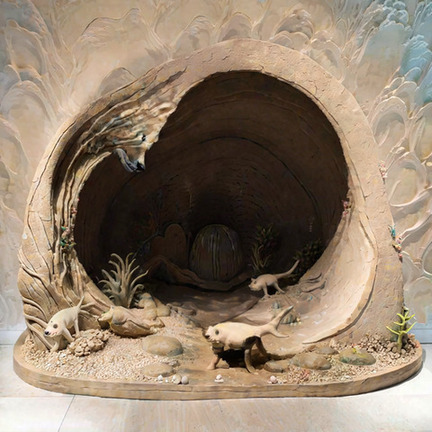}
        \vspace*{-6mm}
        \caption*{\centering \tiny \textit{a clay sculpture of} \par\nobreak aquarium tunnel}
    \end{subfigure}
    \begin{subfigure}{0.158\textwidth}
        \centering
        \includegraphics[width=\textwidth]{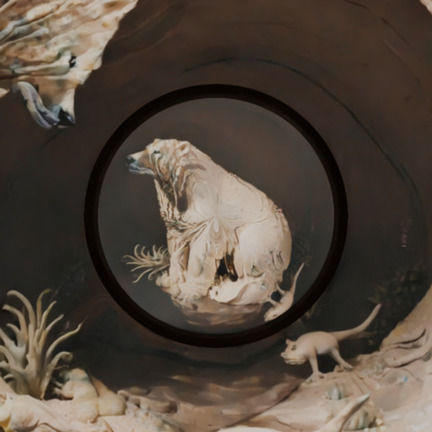}
        \vspace*{-6mm}
        \caption*{\centering \tiny \textit{\quad} \par\nobreak \quad}
    \end{subfigure}
    \begin{subfigure}{0.158\textwidth}
        \centering
        \includegraphics[width=\textwidth]{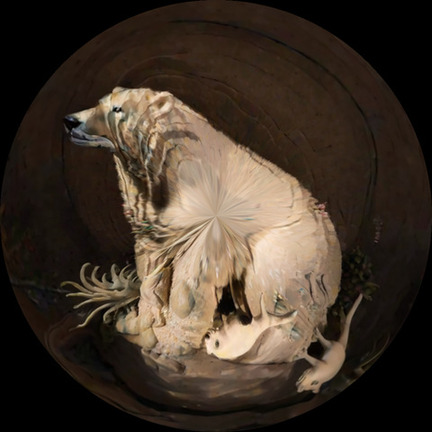}
        \vspace*{-6mm}
        \caption*{\centering \tiny \textit{a clay sculpture of} \par\nobreak polar bear}
    \end{subfigure}    \hfill
    \begin{subfigure}{0.158\textwidth}
        \centering
        \includegraphics[width=\textwidth]{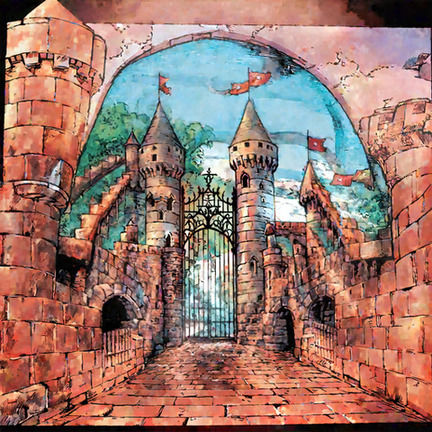}
        \vspace*{-6mm}
        \caption*{\centering \tiny \textit{a comic book panel of} \par\nobreak medieval castle gate}
    \end{subfigure}
    \begin{subfigure}{0.158\textwidth}
        \centering
        \includegraphics[width=\textwidth]{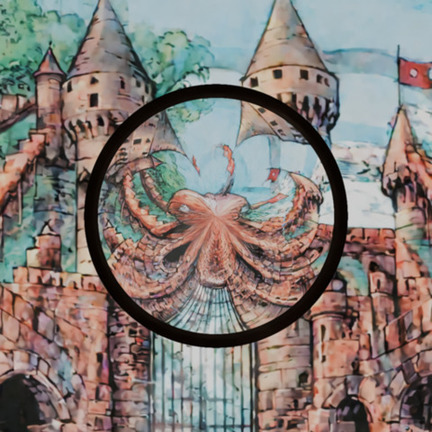}
        \vspace*{-6mm}
        \caption*{\centering \tiny \textit{\quad} \par\nobreak \quad}
    \end{subfigure}
    \begin{subfigure}{0.158\textwidth}
        \centering
        \includegraphics[width=\textwidth]{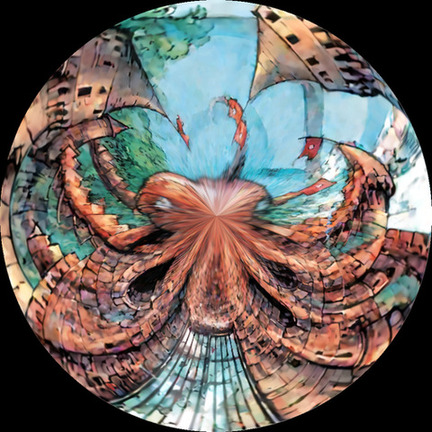}
        \vspace*{-6mm}
        \caption*{\centering \tiny \textit{a comic book panel of} \par\nobreak octopus}
    \end{subfigure}
    \begin{subfigure}{0.158\textwidth}
        \centering
        \includegraphics[width=\textwidth]{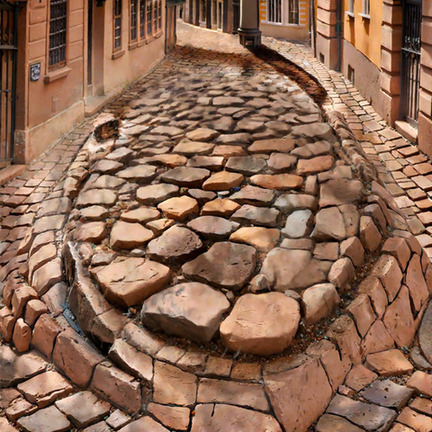}
        \vspace*{-6mm}
        \caption*{\centering \tiny \textit{a clay sculpture of} \par\nobreak cobblestone street}
    \end{subfigure}
    \begin{subfigure}{0.158\textwidth}
        \centering
        \includegraphics[width=\textwidth]{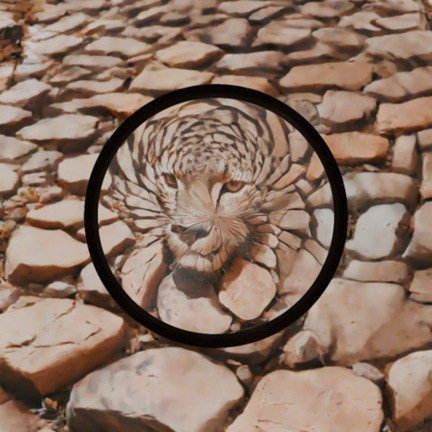}
        \vspace*{-6mm}
        \caption*{\centering \tiny \textit{\quad} \par\nobreak \quad}
    \end{subfigure}
    \begin{subfigure}{0.158\textwidth}
        \centering
        \includegraphics[width=\textwidth]{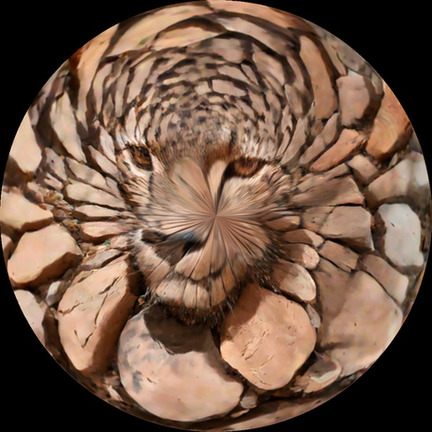}
        \vspace*{-6mm}
        \caption*{\centering \tiny \textit{a clay sculpture of} \par\nobreak cheetah}
    \end{subfigure}    \hfill
    \begin{subfigure}{0.158\textwidth}
        \centering
        \includegraphics[width=\textwidth]{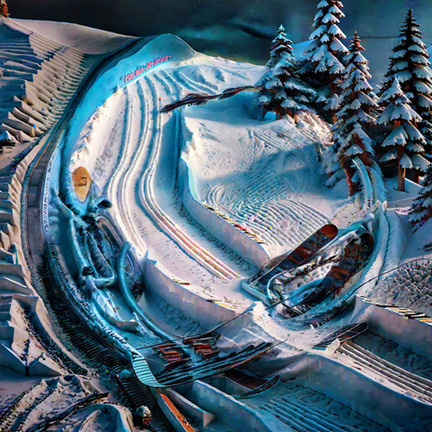}
        \vspace*{-6mm}
        \caption*{\centering \tiny \textit{a hyperrealistic sculpture of} \par\nobreak straight ski tracks}
    \end{subfigure}
    \begin{subfigure}{0.158\textwidth}
        \centering
        \includegraphics[width=\textwidth]{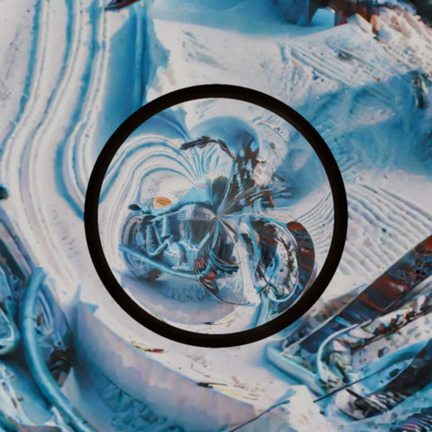}
        \vspace*{-6mm}
        \caption*{\centering \tiny \textit{\quad} \par\nobreak \quad}
    \end{subfigure}
    \begin{subfigure}{0.158\textwidth}
        \centering
        \includegraphics[width=\textwidth]{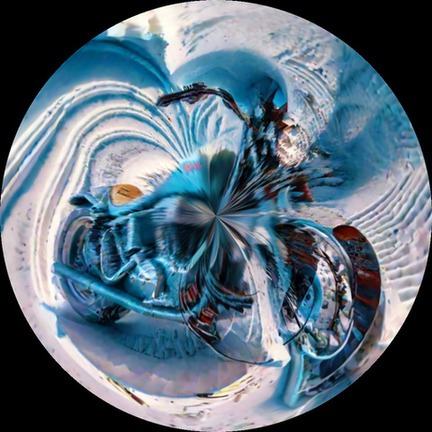}
        \vspace*{-6mm}
        \caption*{\centering \tiny \textit{a hyperrealistic sculpture of} \par\nobreak motorcycle}
    \end{subfigure}
    \begin{subfigure}{0.158\textwidth}
        \centering
        \includegraphics[width=\textwidth]{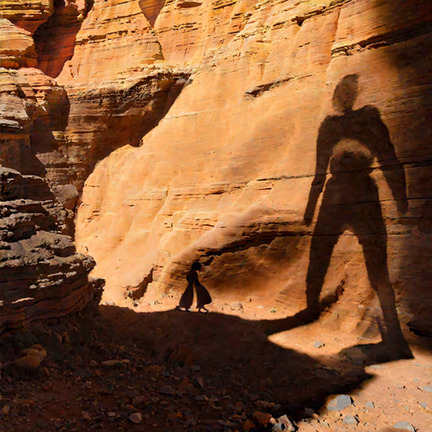}
        \vspace*{-6mm}
        \caption*{\centering \tiny \textit{a shadow puppet silhouette of} \par\nobreak desert canyon floor}
    \end{subfigure}
    \begin{subfigure}{0.158\textwidth}
        \centering
        \includegraphics[width=\textwidth]{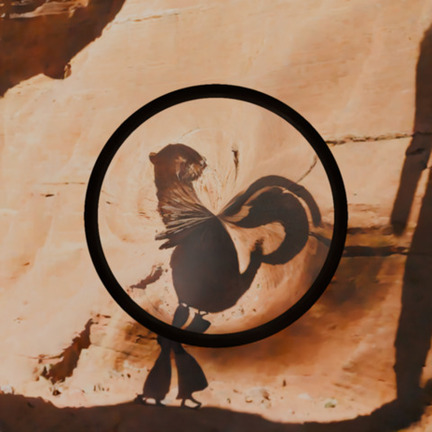}
        \vspace*{-6mm}
        \caption*{\centering \tiny \textit{\quad} \par\nobreak \quad}
    \end{subfigure}
    \begin{subfigure}{0.158\textwidth}
        \centering
        \includegraphics[width=\textwidth]{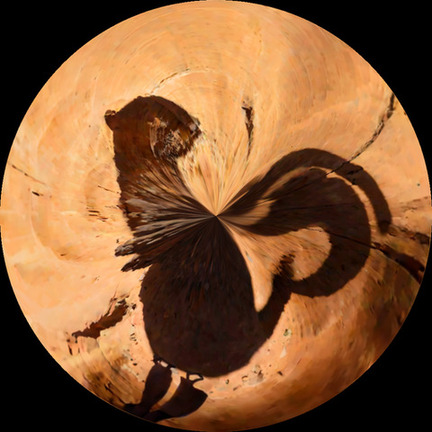}
        \vspace*{-6mm}
        \caption*{\centering \tiny \textit{a shadow puppet silhouette of} \par\nobreak otter}
    \end{subfigure}    \hfill
    \begin{subfigure}{0.158\textwidth}
        \centering
        \includegraphics[width=\textwidth]{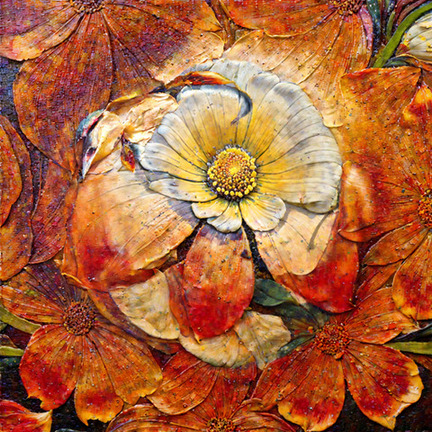}
        \vspace*{-6mm}
        \caption*{\centering \tiny \textit{a fresco painting of} \par\nobreak flower petals close-up}
    \end{subfigure}
    \begin{subfigure}{0.158\textwidth}
        \centering
        \includegraphics[width=\textwidth]{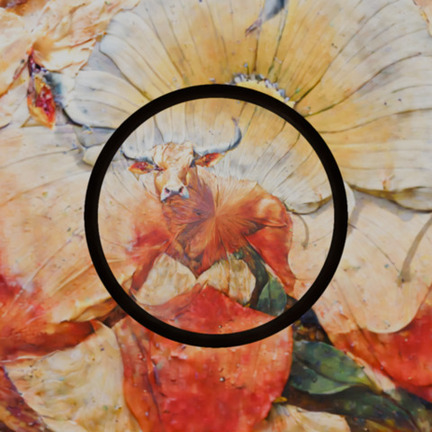}
        \vspace*{-6mm}
        \caption*{\centering \tiny \textit{\quad} \par\nobreak \quad}
    \end{subfigure}
    \begin{subfigure}{0.158\textwidth}
        \centering
        \includegraphics[width=\textwidth]{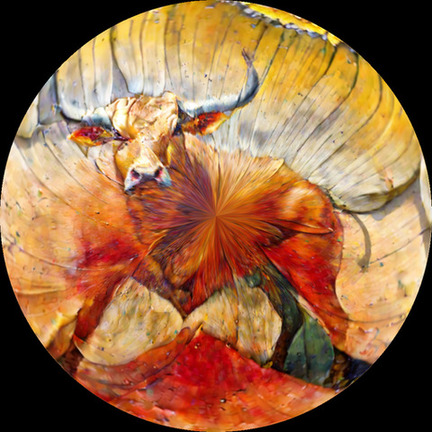}
        \vspace*{-6mm}
        \caption*{\centering \tiny \textit{a fresco painting of} \par\nobreak bull}
    \end{subfigure}
    \begin{subfigure}{0.158\textwidth}
        \centering
        \includegraphics[width=\textwidth]{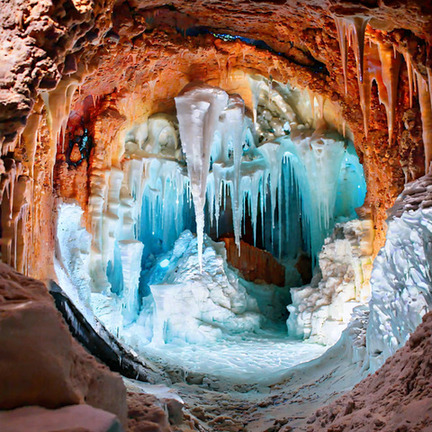}
        \vspace*{-6mm}
        \caption*{\centering \tiny \textit{a hyperrealistic sculpture of} \par\nobreak icy cave with stalactites}
    \end{subfigure}
    \begin{subfigure}{0.158\textwidth}
        \centering
        \includegraphics[width=\textwidth]{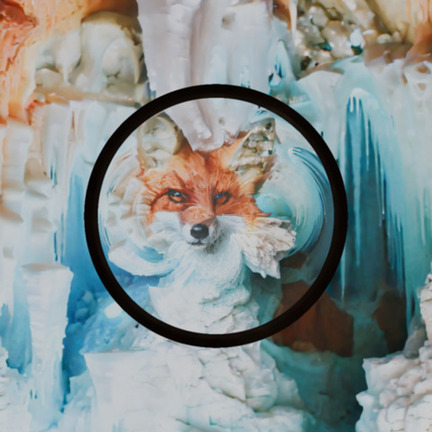}
        \vspace*{-6mm}
        \caption*{\centering \tiny \textit{\quad} \par\nobreak \quad}
    \end{subfigure}
    \begin{subfigure}{0.158\textwidth}
        \centering
        \includegraphics[width=\textwidth]{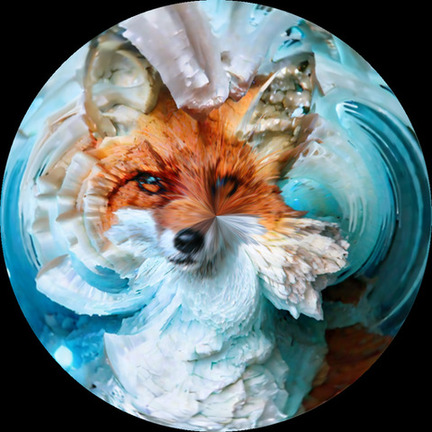}
        \vspace*{-6mm}
        \caption*{\centering \tiny \textit{a hyperrealistic sculpture of} \par\nobreak fox}
    \end{subfigure}    \hfill
    \begin{subfigure}{0.158\textwidth}
        \centering
        \includegraphics[width=\textwidth]{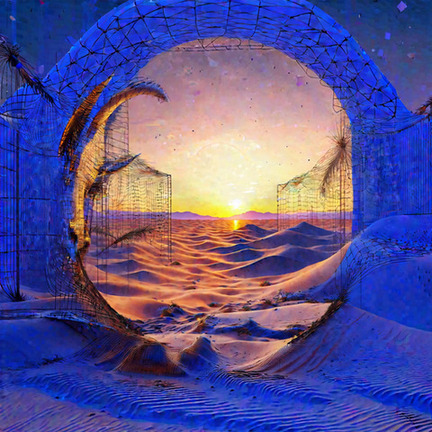}
        \vspace*{-6mm}
        \caption*{\centering \tiny \textit{a wireframe rendering of} \par\nobreak desert dunes at sunset}
    \end{subfigure}
    \begin{subfigure}{0.158\textwidth}
        \centering
        \includegraphics[width=\textwidth]{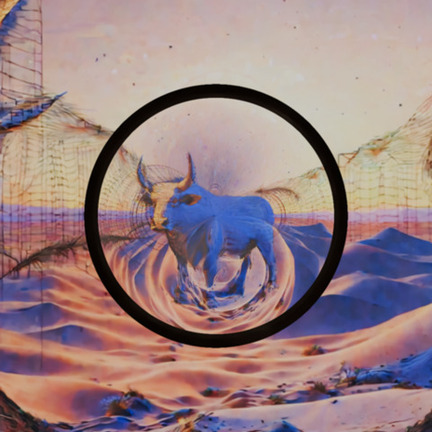}
        \vspace*{-6mm}
        \caption*{\centering \tiny \textit{\quad} \par\nobreak \quad}
    \end{subfigure}
    \begin{subfigure}{0.158\textwidth}
        \centering
        \includegraphics[width=\textwidth]{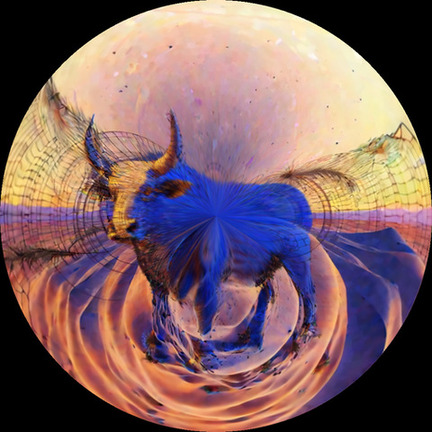}
        \vspace*{-6mm}
        \caption*{\centering \tiny \textit{a wireframe rendering of} \par\nobreak bull}
    \end{subfigure}
    \begin{subfigure}{0.158\textwidth}
        \centering
        \includegraphics[width=\textwidth]{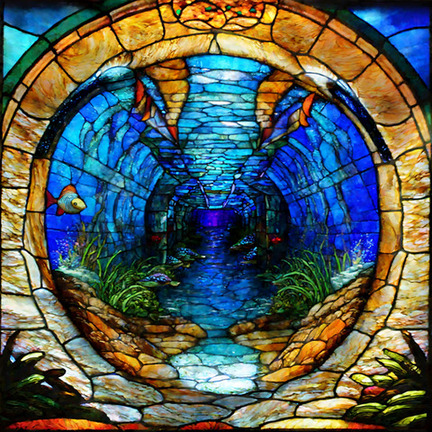}
        \vspace*{-6mm}
        \caption*{\centering \tiny \textit{a stained glass depiction of} \par\nobreak aquarium tunnel}
    \end{subfigure}
    \begin{subfigure}{0.158\textwidth}
        \centering
        \includegraphics[width=\textwidth]{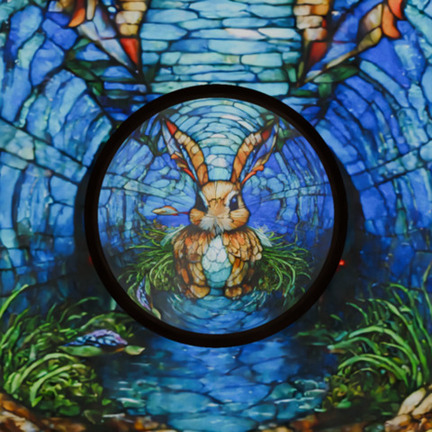}
        \vspace*{-6mm}
        \caption*{\centering \tiny \textit{\quad} \par\nobreak \quad}
    \end{subfigure}
    \begin{subfigure}{0.158\textwidth}
        \centering
        \includegraphics[width=\textwidth]{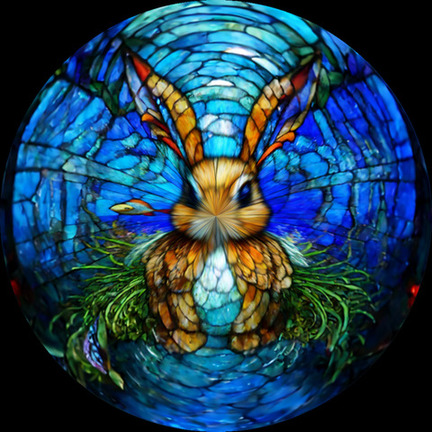}
        \vspace*{-6mm}
        \caption*{\centering \tiny \textit{a stained glass depiction of} \par\nobreak rabbit}
    \end{subfigure}    \hfill
    \begin{subfigure}{0.158\textwidth}
        \centering
        \includegraphics[width=\textwidth]{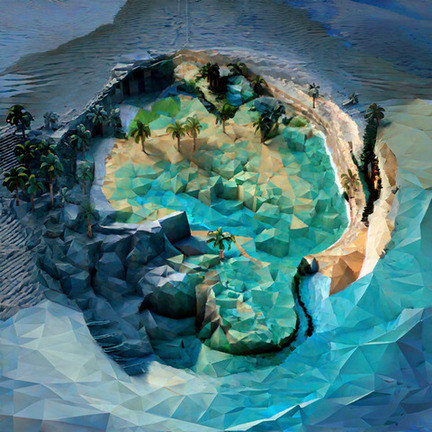}
        \vspace*{-6mm}
        \caption*{\centering \tiny \textit{a low-poly model of} \par\nobreak straight coastline}
    \end{subfigure}
    \begin{subfigure}{0.158\textwidth}
        \centering
        \includegraphics[width=\textwidth]{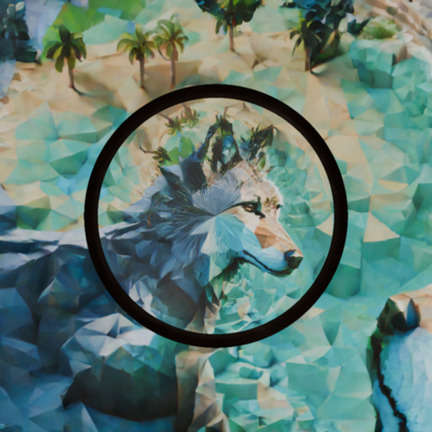}
        \vspace*{-6mm}
        \caption*{\centering \tiny \textit{\quad} \par\nobreak \quad}
    \end{subfigure}
    \begin{subfigure}{0.158\textwidth}
        \centering
        \includegraphics[width=\textwidth]{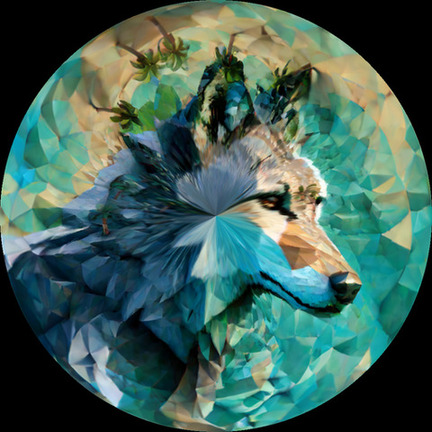}
        \vspace*{-6mm}
        \caption*{\centering \tiny \textit{a low-poly model of} \par\nobreak wolf}
    \end{subfigure}
    \begin{subfigure}{0.158\textwidth}
        \centering
        \includegraphics[width=\textwidth]{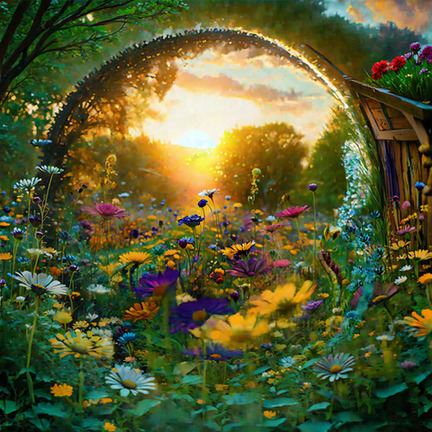}
        \vspace*{-6mm}
        \caption*{\centering \tiny \textit{a cinematic rendering of} \par\nobreak flower meadow at sunrise}
    \end{subfigure}
    \begin{subfigure}{0.158\textwidth}
        \centering
        \includegraphics[width=\textwidth]{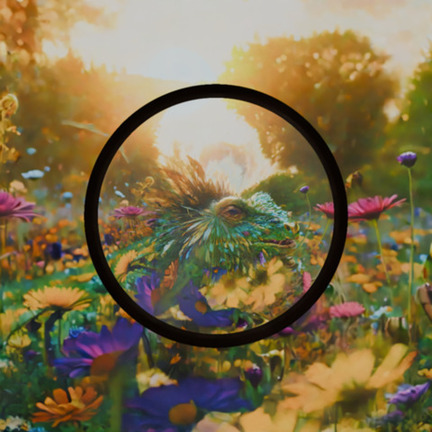}
        \vspace*{-6mm}
        \caption*{\centering \tiny \textit{\quad} \par\nobreak \quad}
    \end{subfigure}
    \begin{subfigure}{0.158\textwidth}
        \centering
        \includegraphics[width=\textwidth]{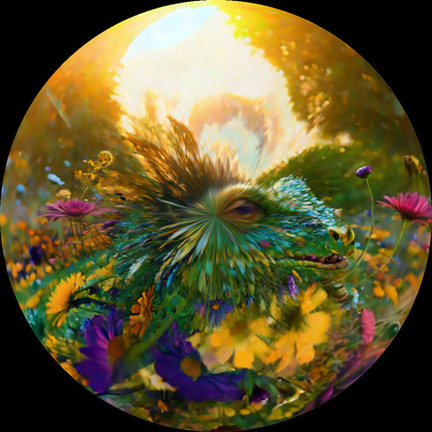}
        \vspace*{-6mm}
        \caption*{\centering \tiny \textit{a cinematic rendering of} \par\nobreak iguana}
    \end{subfigure}    \hfill
    \begin{subfigure}{0.158\textwidth}
        \centering
        \includegraphics[width=\textwidth]{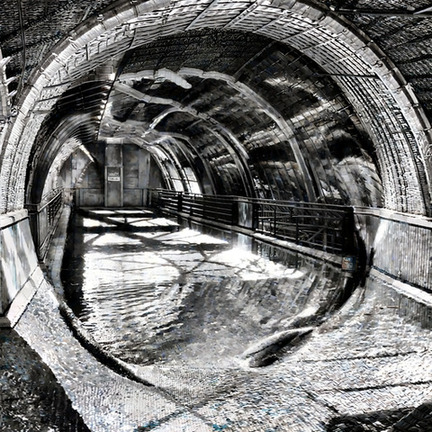}
        \vspace*{-6mm}
        \caption*{\centering \tiny \textit{a black-and-white photo of} \par\nobreak aquarium tunnel}
    \end{subfigure}
    \begin{subfigure}{0.158\textwidth}
        \centering
        \includegraphics[width=\textwidth]{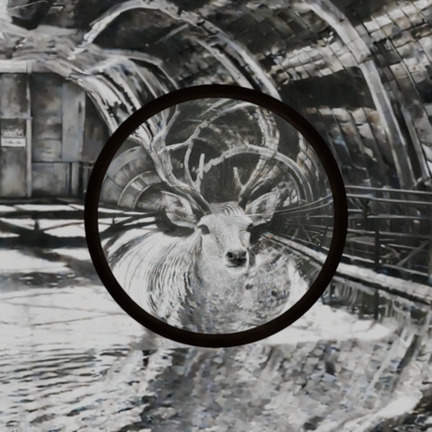}
        \vspace*{-6mm}
        \caption*{\centering \tiny \textit{\quad} \par\nobreak \quad}
    \end{subfigure}
    \begin{subfigure}{0.158\textwidth}
        \centering
        \includegraphics[width=\textwidth]{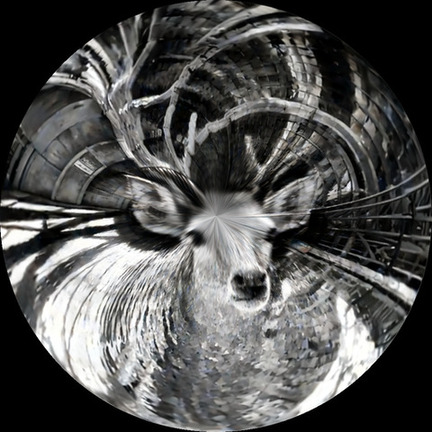}
        \vspace*{-6mm}
        \caption*{\centering \tiny \textit{a black-and-white photo of} \par\nobreak reindeer}
    \end{subfigure}
    \begin{subfigure}{0.158\textwidth}
        \centering
        \includegraphics[width=\textwidth]{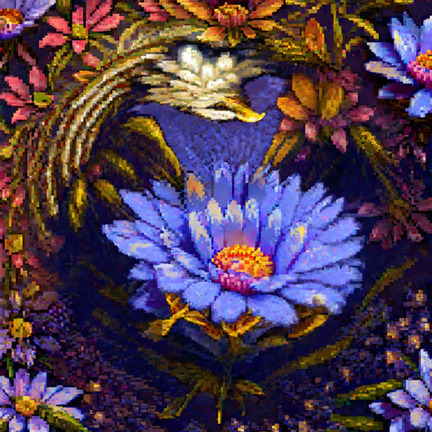}
        \vspace*{-6mm}
        \caption*{\centering \tiny \textit{a 16-bit sprite of} \par\nobreak flower petals close-up}
    \end{subfigure}
    \begin{subfigure}{0.158\textwidth}
        \centering
        \includegraphics[width=\textwidth]{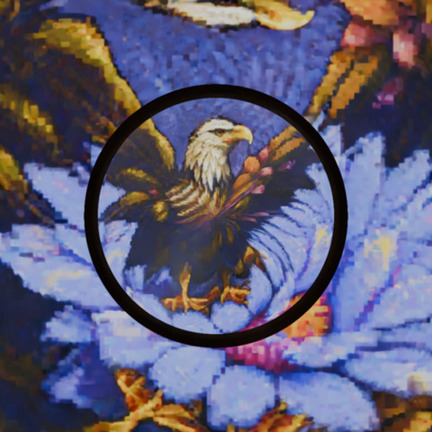}
        \vspace*{-6mm}
        \caption*{\centering \tiny \textit{\quad} \par\nobreak \quad}
    \end{subfigure}
    \begin{subfigure}{0.158\textwidth}
        \centering
        \includegraphics[width=\textwidth]{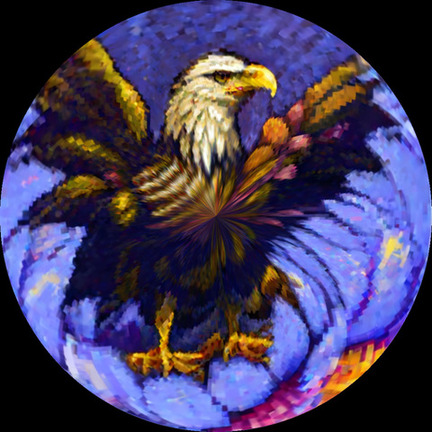}
        \vspace*{-6mm}
        \caption*{\centering \tiny \textit{a 16-bit sprite of} \par\nobreak eagle}
    \end{subfigure}    \hfill
    \begin{subfigure}{0.158\textwidth}
        \centering
        \includegraphics[width=\textwidth]{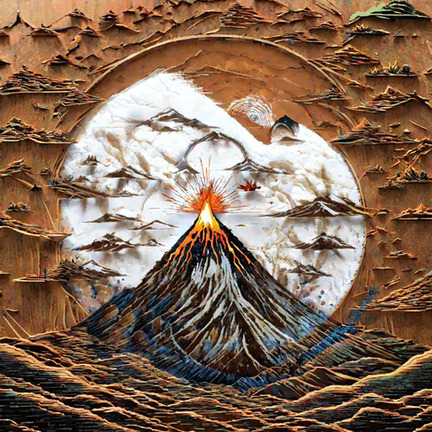}
        \vspace*{-6mm}
        \caption*{\centering \tiny \textit{a cut-paper silhouette of} \par\nobreak volcanic crater}
    \end{subfigure}
    \begin{subfigure}{0.158\textwidth}
        \centering
        \includegraphics[width=\textwidth]{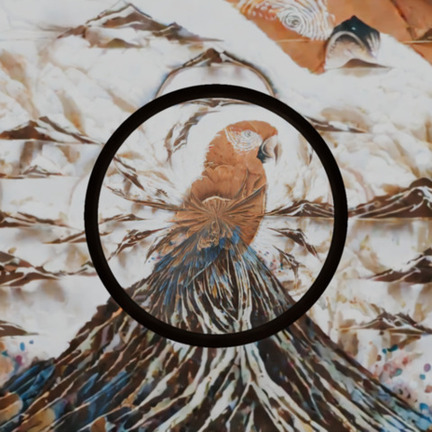}
        \vspace*{-6mm}
        \caption*{\centering \tiny \textit{\quad} \par\nobreak \quad}
    \end{subfigure}
    \begin{subfigure}{0.158\textwidth}
        \centering
        \includegraphics[width=\textwidth]{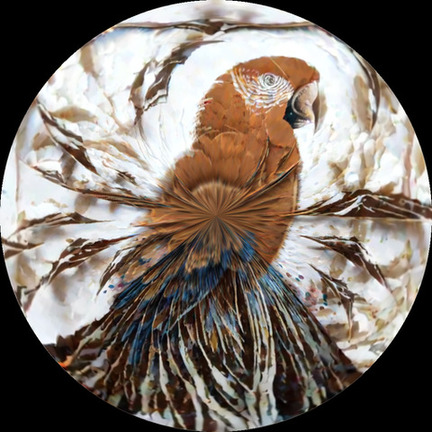}
        \vspace*{-6mm}
        \caption*{\centering \tiny \textit{a cut-paper silhouette of} \par\nobreak macaw}
    \end{subfigure}
\end{figure*}

\begin{figure*}[h!]
    \centering
    \begin{subfigure}{0.106\textwidth}
        \centering
        \includegraphics[width=\textwidth]{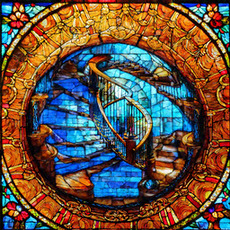}
        \vspace*{-6mm}
        \caption*{\centering \tiny \textit{a stained glass depiction} \par\nobreak \textit{of} spiral staircase}
    \end{subfigure}
    \begin{subfigure}{0.106\textwidth}
        \centering
        \includegraphics[width=\textwidth]{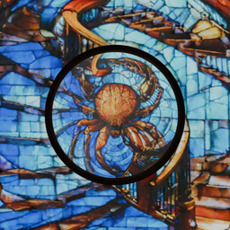}
        \vspace*{-6mm}
        \caption*{\centering \tiny \textit{\quad} \par\nobreak \quad}
    \end{subfigure}
    \begin{subfigure}{0.106\textwidth}
        \centering
        \includegraphics[width=\textwidth]{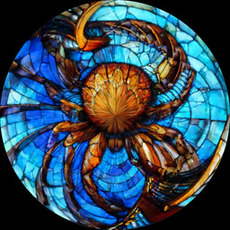}
        \vspace*{-6mm}
        \caption*{\centering \tiny \textit{a stained glass depiction} \par\nobreak \textit{of} hermit crab}
    \end{subfigure}    \hfill
    \begin{subfigure}{0.106\textwidth}
        \centering
        \includegraphics[width=\textwidth]{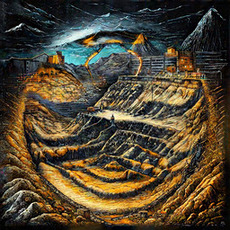}
        \vspace*{-6mm}
        \caption*{\centering \tiny \textit{a chalkboard drawing of} \par\nobreak open-pit mine}
    \end{subfigure}
    \begin{subfigure}{0.106\textwidth}
        \centering
        \includegraphics[width=\textwidth]{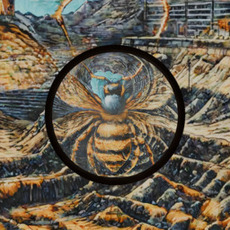}
        \vspace*{-6mm}
        \caption*{\centering \tiny \textit{\quad} \par\nobreak \quad}
    \end{subfigure}
    \begin{subfigure}{0.106\textwidth}
        \centering
        \includegraphics[width=\textwidth]{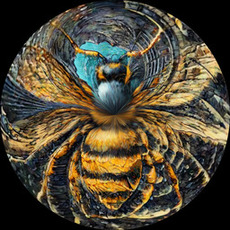}
        \vspace*{-6mm}
        \caption*{\centering \tiny \textit{a chalkboard drawing of} \par\nobreak bee}
    \end{subfigure}    \hfill
    \begin{subfigure}{0.106\textwidth}
        \centering
        \includegraphics[width=\textwidth]{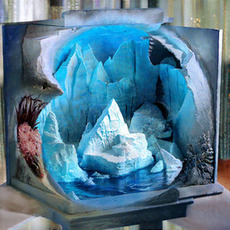}
        \vspace*{-6mm}
        \caption*{\centering \tiny \textit{a diorama of} \par\nobreak icebergs in the ocean}
    \end{subfigure}
    \begin{subfigure}{0.106\textwidth}
        \centering
        \includegraphics[width=\textwidth]{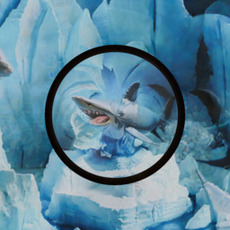}
        \vspace*{-6mm}
        \caption*{\centering \tiny \textit{\quad} \par\nobreak \quad}
    \end{subfigure}
    \begin{subfigure}{0.106\textwidth}
        \centering
        \includegraphics[width=\textwidth]{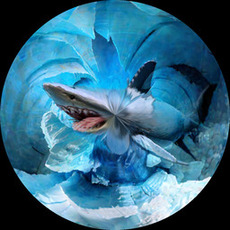}
        \vspace*{-6mm}
        \caption*{\centering \tiny \textit{a diorama of} \par\nobreak shark}
    \end{subfigure}
    \begin{subfigure}{0.106\textwidth}
        \centering
        \includegraphics[width=\textwidth]{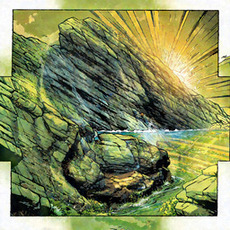}
        \vspace*{-6mm}
        \caption*{\centering \tiny \textit{a comic book panel of} \par\nobreak sunlit rocky outcrop}
    \end{subfigure}
    \begin{subfigure}{0.106\textwidth}
        \centering
        \includegraphics[width=\textwidth]{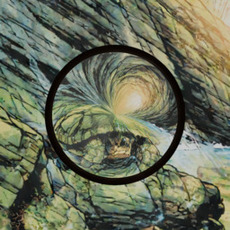}
        \vspace*{-6mm}
        \caption*{\centering \tiny \textit{\quad} \par\nobreak \quad}
    \end{subfigure}
    \begin{subfigure}{0.106\textwidth}
        \centering
        \includegraphics[width=\textwidth]{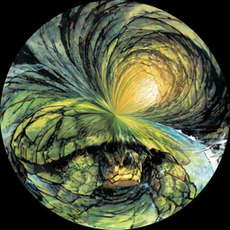}
        \vspace*{-6mm}
        \caption*{\centering \tiny \textit{a comic book panel of} \par\nobreak turtle}
    \end{subfigure}    \hfill
    \begin{subfigure}{0.106\textwidth}
        \centering
        \includegraphics[width=\textwidth]{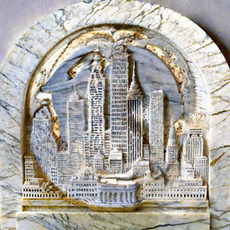}
        \vspace*{-6mm}
        \caption*{\centering \tiny \textit{a marble carving of} \par\nobreak city skyline at night}
    \end{subfigure}
    \begin{subfigure}{0.106\textwidth}
        \centering
        \includegraphics[width=\textwidth]{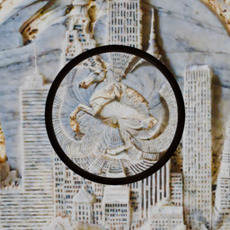}
        \vspace*{-6mm}
        \caption*{\centering \tiny \textit{\quad} \par\nobreak \quad}
    \end{subfigure}
    \begin{subfigure}{0.106\textwidth}
        \centering
        \includegraphics[width=\textwidth]{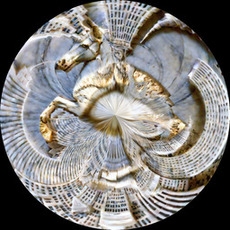}
        \vspace*{-6mm}
        \caption*{\centering \tiny \textit{a marble carving of} \par\nobreak horse}
    \end{subfigure}    \hfill
    \begin{subfigure}{0.106\textwidth}
        \centering
        \includegraphics[width=\textwidth]{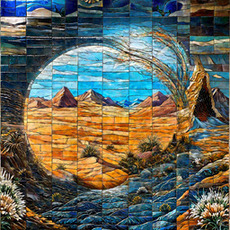}
        \vspace*{-6mm}
        \caption*{\centering \tiny \textit{a ceramic tile mural of} \par\nobreak windy desert}
    \end{subfigure}
    \begin{subfigure}{0.106\textwidth}
        \centering
        \includegraphics[width=\textwidth]{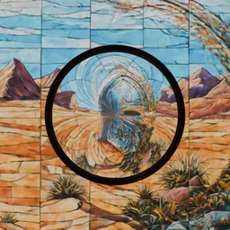}
        \vspace*{-6mm}
        \caption*{\centering \tiny \textit{\quad} \par\nobreak \quad}
    \end{subfigure}
    \begin{subfigure}{0.106\textwidth}
        \centering
        \includegraphics[width=\textwidth]{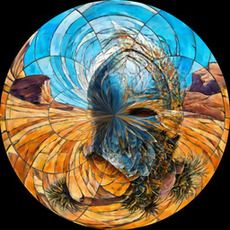}
        \vspace*{-6mm}
        \caption*{\centering \tiny \textit{a ceramic tile mural of} \par\nobreak knight's helmet}
    \end{subfigure}
    \begin{subfigure}{0.106\textwidth}
        \centering
        \includegraphics[width=\textwidth]{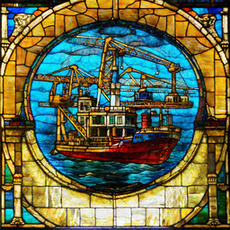}
        \vspace*{-6mm}
        \caption*{\centering \tiny \textit{a stained glass depiction} \par\nobreak \textit{of} shipyard with cranes}
    \end{subfigure}
    \begin{subfigure}{0.106\textwidth}
        \centering
        \includegraphics[width=\textwidth]{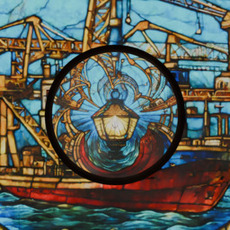}
        \vspace*{-6mm}
        \caption*{\centering \tiny \textit{\quad} \par\nobreak \quad}
    \end{subfigure}
    \begin{subfigure}{0.106\textwidth}
        \centering
        \includegraphics[width=\textwidth]{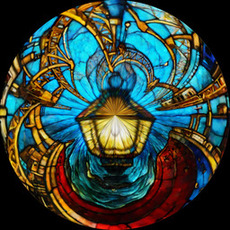}
        \vspace*{-6mm}
        \caption*{\centering \tiny \textit{a stained glass depiction} \par\nobreak \textit{of} lantern}
    \end{subfigure}    \hfill
    \begin{subfigure}{0.106\textwidth}
        \centering
        \includegraphics[width=\textwidth]{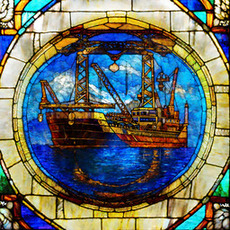}
        \vspace*{-6mm}
        \caption*{\centering \tiny \textit{a stained glass depiction} \par\nobreak \textit{of} shipyard with cranes}
    \end{subfigure}
    \begin{subfigure}{0.106\textwidth}
        \centering
        \includegraphics[width=\textwidth]{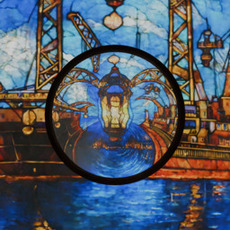}
        \vspace*{-6mm}
        \caption*{\centering \tiny \textit{\quad} \par\nobreak \quad}
    \end{subfigure}
    \begin{subfigure}{0.106\textwidth}
        \centering
        \includegraphics[width=\textwidth]{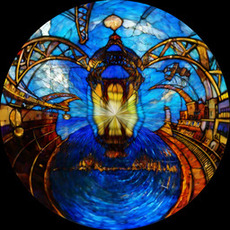}
        \vspace*{-6mm}
        \caption*{\centering \tiny \textit{a stained glass depiction} \par\nobreak \textit{of} lantern}
    \end{subfigure}    \hfill
    \begin{subfigure}{0.106\textwidth}
        \centering
        \includegraphics[width=\textwidth]{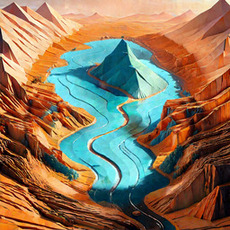}
        \vspace*{-6mm}
        \caption*{\centering \tiny \textit{a low-poly model of} \par\nobreak meandering river in valley}
    \end{subfigure}
    \begin{subfigure}{0.106\textwidth}
        \centering
        \includegraphics[width=\textwidth]{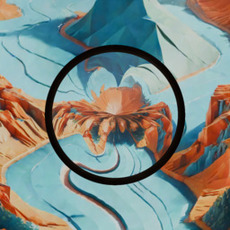}
        \vspace*{-6mm}
        \caption*{\centering \tiny \textit{\quad} \par\nobreak \quad}
    \end{subfigure}
    \begin{subfigure}{0.106\textwidth}
        \centering
        \includegraphics[width=\textwidth]{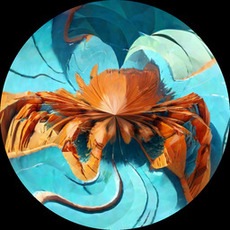}
        \vspace*{-6mm}
        \caption*{\centering \tiny \textit{a low-poly model of} \par\nobreak hermit crab}
    \end{subfigure}
    \begin{subfigure}{0.106\textwidth}
        \centering
        \includegraphics[width=\textwidth]{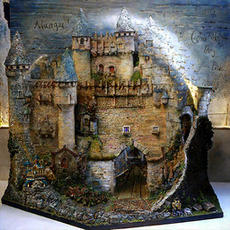}
        \vspace*{-6mm}
        \caption*{\centering \tiny \textit{a diorama of} \par\nobreak medieval castle gate}
    \end{subfigure}
    \begin{subfigure}{0.106\textwidth}
        \centering
        \includegraphics[width=\textwidth]{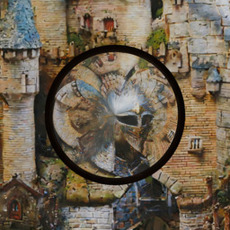}
        \vspace*{-6mm}
        \caption*{\centering \tiny \textit{\quad} \par\nobreak \quad}
    \end{subfigure}
    \begin{subfigure}{0.106\textwidth}
        \centering
        \includegraphics[width=\textwidth]{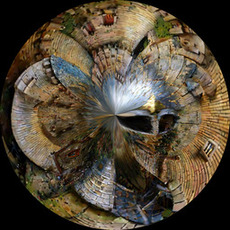}
        \vspace*{-6mm}
        \caption*{\centering \tiny \textit{a diorama of} \par\nobreak knight's helmet}
    \end{subfigure}    \hfill
    \begin{subfigure}{0.106\textwidth}
        \centering
        \includegraphics[width=\textwidth]{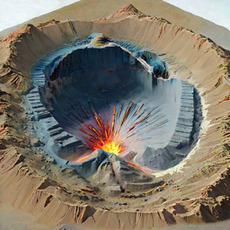}
        \vspace*{-6mm}
        \caption*{\centering \tiny \textit{a 3D rendering of} \par\nobreak volcanic crater}
    \end{subfigure}
    \begin{subfigure}{0.106\textwidth}
        \centering
        \includegraphics[width=\textwidth]{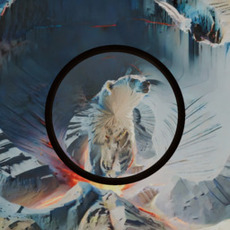}
        \vspace*{-6mm}
        \caption*{\centering \tiny \textit{\quad} \par\nobreak \quad}
    \end{subfigure}
    \begin{subfigure}{0.106\textwidth}
        \centering
        \includegraphics[width=\textwidth]{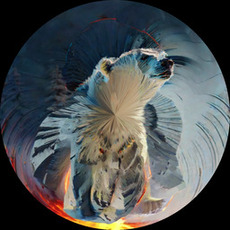}
        \vspace*{-6mm}
        \caption*{\centering \tiny \textit{a 3D rendering of} \par\nobreak polar bear}
    \end{subfigure}    \hfill
    \begin{subfigure}{0.106\textwidth}
        \centering
        \includegraphics[width=\textwidth]{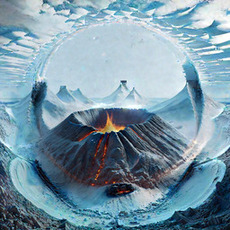}
        \vspace*{-6mm}
        \caption*{\centering \tiny \textit{a 3D rendering of} \par\nobreak volcanic crater}
    \end{subfigure}
    \begin{subfigure}{0.106\textwidth}
        \centering
        \includegraphics[width=\textwidth]{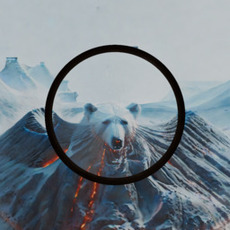}
        \vspace*{-6mm}
        \caption*{\centering \tiny \textit{\quad} \par\nobreak \quad}
    \end{subfigure}
    \begin{subfigure}{0.106\textwidth}
        \centering
        \includegraphics[width=\textwidth]{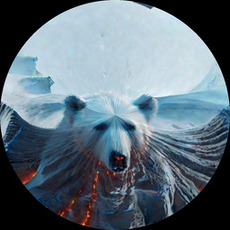}
        \vspace*{-6mm}
        \caption*{\centering \tiny \textit{a 3D rendering of} \par\nobreak polar bear}
    \end{subfigure}
    \begin{subfigure}{0.106\textwidth}
        \centering
        \includegraphics[width=\textwidth]{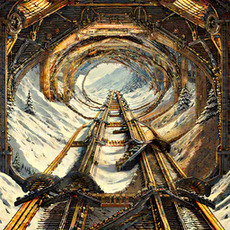}
        \vspace*{-6mm}
        \caption*{\centering \tiny \textit{a steampunk illustration} \par\nobreak \textit{of} straight ski tracks}
    \end{subfigure}
    \begin{subfigure}{0.106\textwidth}
        \centering
        \includegraphics[width=\textwidth]{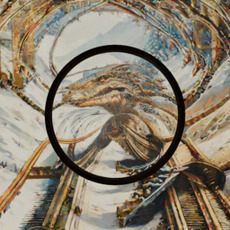}
        \vspace*{-6mm}
        \caption*{\centering \tiny \textit{\quad} \par\nobreak \quad}
    \end{subfigure}
    \begin{subfigure}{0.106\textwidth}
        \centering
        \includegraphics[width=\textwidth]{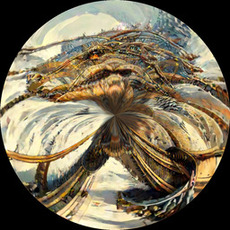}
        \vspace*{-6mm}
        \caption*{\centering \tiny \textit{a steampunk illustration} \par\nobreak \textit{of} komodo dragon}
    \end{subfigure}    \hfill
    \begin{subfigure}{0.106\textwidth}
        \centering
        \includegraphics[width=\textwidth]{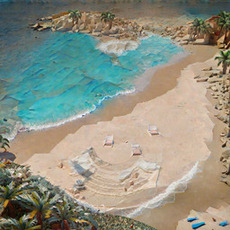}
        \vspace*{-6mm}
        \caption*{\centering \tiny \textit{low-poly model of} beach, shoreline \& waves}
    \end{subfigure}
    \begin{subfigure}{0.106\textwidth}
        \centering
        \includegraphics[width=\textwidth]{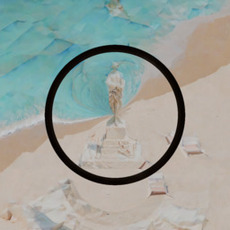}
        \vspace*{-6mm}
        \caption*{\centering \tiny \textit{\quad} \par\nobreak \quad}
    \end{subfigure}
    \begin{subfigure}{0.106\textwidth}
        \centering
        \includegraphics[width=\textwidth]{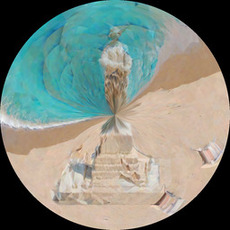}
        \vspace*{-6mm}
        \caption*{\centering \tiny \textit{low-poly model of} \par\nobreak fountain statue}
    \end{subfigure}    \hfill
    \begin{subfigure}{0.106\textwidth}
        \centering
        \includegraphics[width=\textwidth]{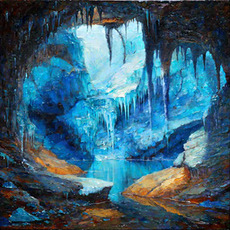}
        \vspace*{-6mm}
        \caption*{\centering \tiny \textit{an oil painting of} \par\nobreak icy cave with stalactites}
    \end{subfigure}
    \begin{subfigure}{0.106\textwidth}
        \centering
        \includegraphics[width=\textwidth]{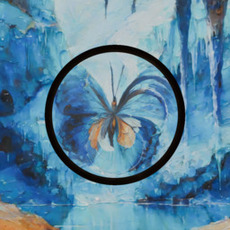}
        \vspace*{-6mm}
        \caption*{\centering \tiny \textit{\quad} \par\nobreak \quad}
    \end{subfigure}
    \begin{subfigure}{0.106\textwidth}
        \centering
        \includegraphics[width=\textwidth]{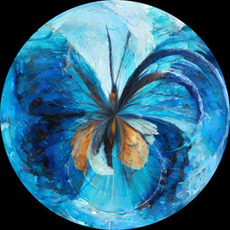}
        \vspace*{-6mm}
        \caption*{\centering \tiny \textit{an oil painting of} \par\nobreak butterfly}
    \end{subfigure}
    \begin{subfigure}{0.106\textwidth}
        \centering
        \includegraphics[width=\textwidth]{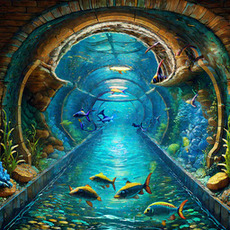}
        \vspace*{-6mm}
        \caption*{\centering \tiny \textit{a digital illustration of} \par\nobreak aquarium tunnel}
    \end{subfigure}
    \begin{subfigure}{0.106\textwidth}
        \centering
        \includegraphics[width=\textwidth]{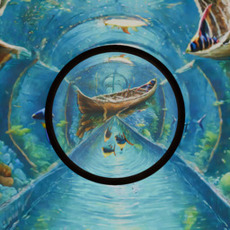}
        \vspace*{-6mm}
        \caption*{\centering \tiny \textit{\quad} \par\nobreak \quad}
    \end{subfigure}
    \begin{subfigure}{0.106\textwidth}
        \centering
        \includegraphics[width=\textwidth]{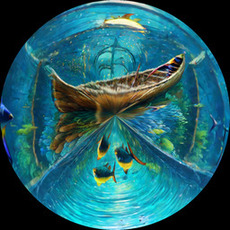}
        \vspace*{-6mm}
        \caption*{\centering \tiny \textit{a digital illustration of} \par\nobreak canoe}
    \end{subfigure}    \hfill
    \begin{subfigure}{0.106\textwidth}
        \centering
        \includegraphics[width=\textwidth]{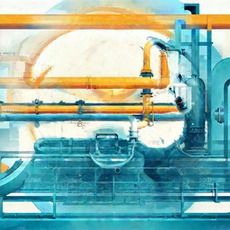}
        \vspace*{-6mm}
        \caption*{\centering \tiny \textit{a minimalist illustration} \par\nobreak \textit{of} industrial pipeline}
    \end{subfigure}
    \begin{subfigure}{0.106\textwidth}
        \centering
        \includegraphics[width=\textwidth]{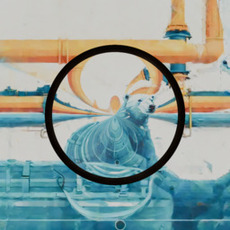}
        \vspace*{-6mm}
        \caption*{\centering \tiny \textit{\quad} \par\nobreak \quad}
    \end{subfigure}
    \begin{subfigure}{0.106\textwidth}
        \centering
        \includegraphics[width=\textwidth]{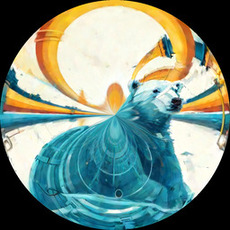}
        \vspace*{-6mm}
        \caption*{\centering \tiny \textit{a minimalist illustration} \par\nobreak \textit{of} polar bear}
    \end{subfigure}    \hfill
    \begin{subfigure}{0.106\textwidth}
        \centering
        \includegraphics[width=\textwidth]{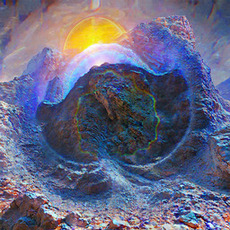}
        \vspace*{-6mm}
        \caption*{\centering \tiny \textit{a holographic image of} \par\nobreak sunlit rocky outcrop}
    \end{subfigure}
    \begin{subfigure}{0.106\textwidth}
        \centering
        \includegraphics[width=\textwidth]{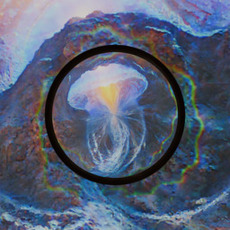}
        \vspace*{-6mm}
        \caption*{\centering \tiny \textit{\quad} \par\nobreak \quad}
    \end{subfigure}
    \begin{subfigure}{0.106\textwidth}
        \centering
        \includegraphics[width=\textwidth]{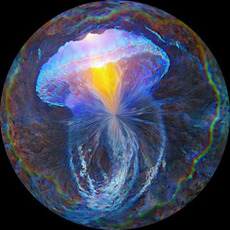}
        \vspace*{-6mm}
        \caption*{\centering \tiny \textit{a holographic image of} \par\nobreak jellyfish}
    \end{subfigure}
    \begin{subfigure}{0.106\textwidth}
        \centering
        \includegraphics[width=\textwidth]{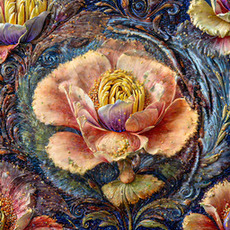}
        \vspace*{-6mm}
        \caption*{\centering \tiny \textit{a fresco painting of} \par\nobreak flower petals close-up}
    \end{subfigure}
    \begin{subfigure}{0.106\textwidth}
        \centering
        \includegraphics[width=\textwidth]{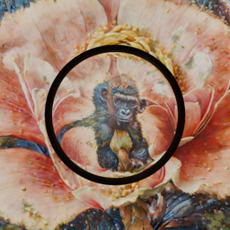}
        \vspace*{-6mm}
        \caption*{\centering \tiny \textit{\quad} \par\nobreak \quad}
    \end{subfigure}
    \begin{subfigure}{0.106\textwidth}
        \centering
        \includegraphics[width=\textwidth]{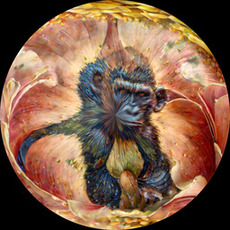}
        \vspace*{-6mm}
        \caption*{\centering \tiny \textit{a fresco painting of} \par\nobreak gorilla}
    \end{subfigure}    \hfill
    \begin{subfigure}{0.106\textwidth}
        \centering
        \includegraphics[width=\textwidth]{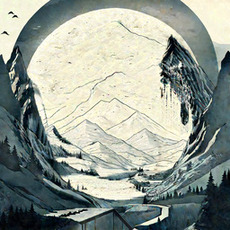}
        \vspace*{-6mm}
        \caption*{\centering \tiny \textit{a minimalist illustration} \par\nobreak \textit{of} mountain pass}
    \end{subfigure}
    \begin{subfigure}{0.106\textwidth}
        \centering
        \includegraphics[width=\textwidth]{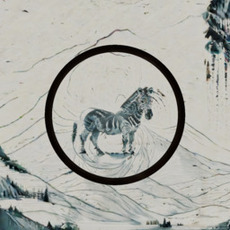}
        \vspace*{-6mm}
        \caption*{\centering \tiny \textit{\quad} \par\nobreak \quad}
    \end{subfigure}
    \begin{subfigure}{0.106\textwidth}
        \centering
        \includegraphics[width=\textwidth]{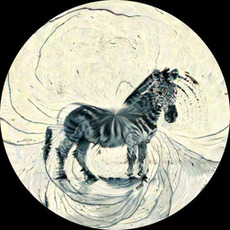}
        \vspace*{-6mm}
        \caption*{\centering \tiny \textit{a minimalist illustration} \par\nobreak \textit{of} zebra}
    \end{subfigure}    \hfill
    \begin{subfigure}{0.106\textwidth}
        \centering
        \includegraphics[width=\textwidth]{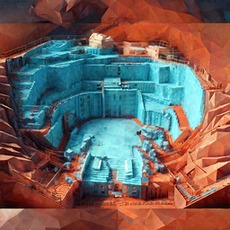}
        \vspace*{-6mm}
        \caption*{\centering \tiny \textit{a low-poly model of} \par\nobreak open-pit mine}
    \end{subfigure}
    \begin{subfigure}{0.106\textwidth}
        \centering
        \includegraphics[width=\textwidth]{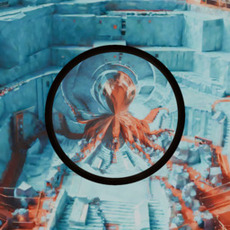}
        \vspace*{-6mm}
        \caption*{\centering \tiny \textit{\quad} \par\nobreak \quad}
    \end{subfigure}
    \begin{subfigure}{0.106\textwidth}
        \centering
        \includegraphics[width=\textwidth]{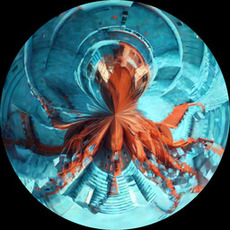}
        \vspace*{-6mm}
        \caption*{\centering \tiny \textit{a low-poly model of} \par\nobreak octopus}
    \end{subfigure}
    \begin{subfigure}{0.106\textwidth}
        \centering
        \includegraphics[width=\textwidth]{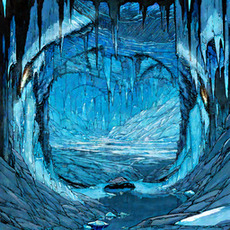}
        \vspace*{-6mm}
        \caption*{\centering \tiny \textit{a comic book panel of} \par\nobreak icy cave with stalactites}
    \end{subfigure}
    \begin{subfigure}{0.106\textwidth}
        \centering
        \includegraphics[width=\textwidth]{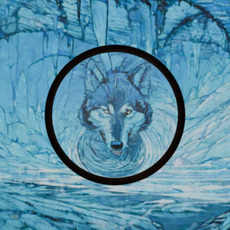}
        \vspace*{-6mm}
        \caption*{\centering \tiny \textit{\quad} \par\nobreak \quad}
    \end{subfigure}
    \begin{subfigure}{0.106\textwidth}
        \centering
        \includegraphics[width=\textwidth]{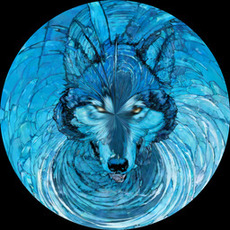}
        \vspace*{-6mm}
        \caption*{\centering \tiny \textit{a comic book panel of} \par\nobreak wolf}
    \end{subfigure}    \hfill
    \begin{subfigure}{0.106\textwidth}
        \centering
        \includegraphics[width=\textwidth]{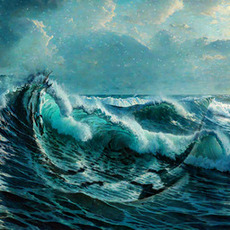}
        \vspace*{-6mm}
        \caption*{\centering \tiny \textit{a cinematic rendering of} \par\nobreak ocean waves}
    \end{subfigure}
    \begin{subfigure}{0.106\textwidth}
        \centering
        \includegraphics[width=\textwidth]{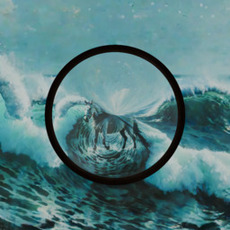}
        \vspace*{-6mm}
        \caption*{\centering \tiny \textit{\quad} \par\nobreak \quad}
    \end{subfigure}
    \begin{subfigure}{0.106\textwidth}
        \centering
        \includegraphics[width=\textwidth]{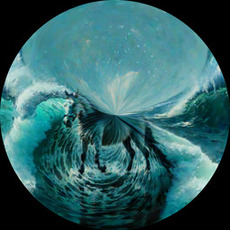}
        \vspace*{-6mm}
        \caption*{\centering \tiny \textit{a cinematic rendering of} \par\nobreak horse}
    \end{subfigure}    \hfill
    \begin{subfigure}{0.106\textwidth}
        \centering
        \includegraphics[width=\textwidth]{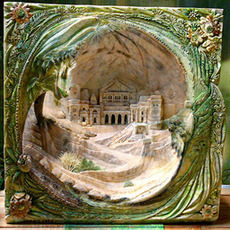}
        \vspace*{-6mm}
        \caption*{\centering \tiny \textit{a marble carving of} \par\nobreak desert oasis}
    \end{subfigure}
    \begin{subfigure}{0.106\textwidth}
        \centering
        \includegraphics[width=\textwidth]{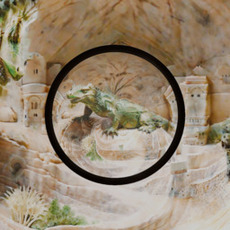}
        \vspace*{-6mm}
        \caption*{\centering \tiny \textit{\quad} \par\nobreak \quad}
    \end{subfigure}
    \begin{subfigure}{0.106\textwidth}
        \centering
        \includegraphics[width=\textwidth]{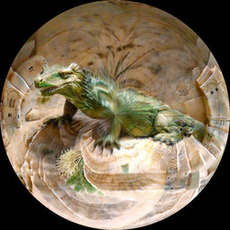}
        \vspace*{-6mm}
        \caption*{\centering \tiny \textit{a marble carving of} \par\nobreak komodo dragon}
    \end{subfigure}
    \begin{subfigure}{0.106\textwidth}
        \centering
        \includegraphics[width=\textwidth]{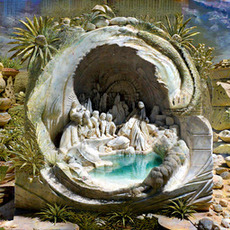}
        \vspace*{-6mm}
        \caption*{\centering \tiny \textit{a marble carving of} \par\nobreak desert oasis}
    \end{subfigure}
    \begin{subfigure}{0.106\textwidth}
        \centering
        \includegraphics[width=\textwidth]{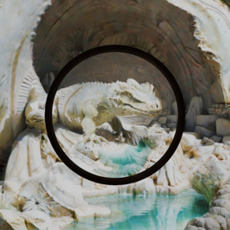}
        \vspace*{-6mm}
        \caption*{\centering \tiny \textit{\quad} \par\nobreak \quad}
    \end{subfigure}
    \begin{subfigure}{0.106\textwidth}
        \centering
        \includegraphics[width=\textwidth]{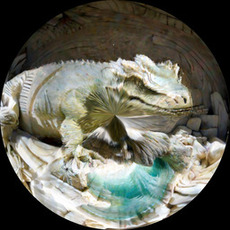}
        \vspace*{-6mm}
        \caption*{\centering \tiny \textit{a marble carving of} \par\nobreak komodo dragon}
    \end{subfigure}    \hfill
    \begin{subfigure}{0.106\textwidth}
        \centering
        \includegraphics[width=\textwidth]{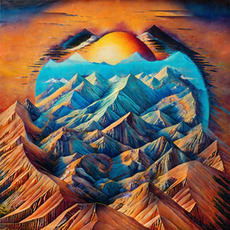}
        \vspace*{-6mm}
        \caption*{\centering \tiny \textit{a surrealist painting of} \par\nobreak mountain ridge}
    \end{subfigure}
    \begin{subfigure}{0.106\textwidth}
        \centering
        \includegraphics[width=\textwidth]{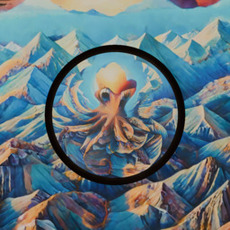}
        \vspace*{-6mm}
        \caption*{\centering \tiny \textit{\quad} \par\nobreak \quad}
    \end{subfigure}
    \begin{subfigure}{0.106\textwidth}
        \centering
        \includegraphics[width=\textwidth]{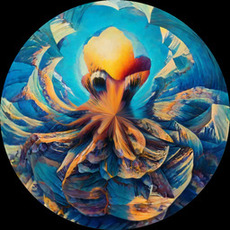}
        \vspace*{-6mm}
        \caption*{\centering \tiny \textit{a surrealist painting of} \par\nobreak octopus}
    \end{subfigure}    \hfill
    \begin{subfigure}{0.106\textwidth}
        \centering
        \includegraphics[width=\textwidth]{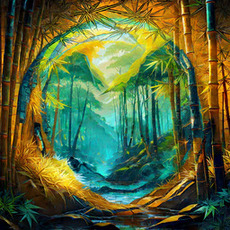}
        \vspace*{-6mm}
        \caption*{\centering \tiny \textit{a futuristic concept art of} \par\nobreak bamboo forest}
    \end{subfigure}
    \begin{subfigure}{0.106\textwidth}
        \centering
        \includegraphics[width=\textwidth]{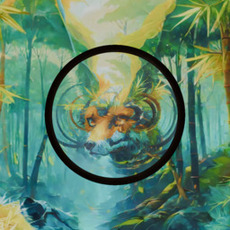}
        \vspace*{-6mm}
        \caption*{\centering \tiny \textit{\quad} \par\nobreak \quad}
    \end{subfigure}
    \begin{subfigure}{0.106\textwidth}
        \centering
        \includegraphics[width=\textwidth]{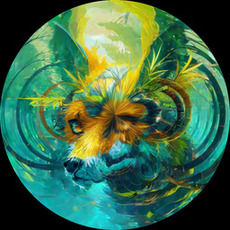}
        \vspace*{-6mm}
        \caption*{\centering \tiny \textit{a futuristic concept art of} \par\nobreak fox}
    \end{subfigure}
    \begin{subfigure}{0.106\textwidth}
        \centering
        \includegraphics[width=\textwidth]{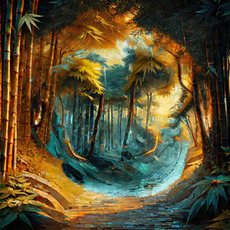}
        \vspace*{-6mm}
        \caption*{\centering \tiny \textit{a futuristic concept art of} \par\nobreak bamboo forest}
    \end{subfigure}
    \begin{subfigure}{0.106\textwidth}
        \centering
        \includegraphics[width=\textwidth]{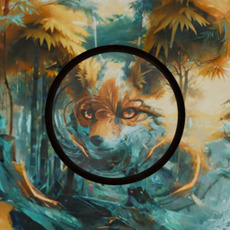}
        \vspace*{-6mm}
        \caption*{\centering \tiny \textit{\quad} \par\nobreak \quad}
    \end{subfigure}
    \begin{subfigure}{0.106\textwidth}
        \centering
        \includegraphics[width=\textwidth]{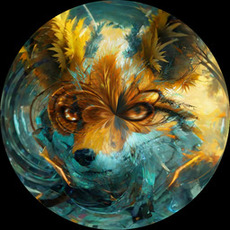}
        \vspace*{-6mm}
        \caption*{\centering \tiny \textit{a futuristic concept art of} \par\nobreak fox}
    \end{subfigure}    \hfill
    \begin{subfigure}{0.106\textwidth}
        \centering
        \includegraphics[width=\textwidth]{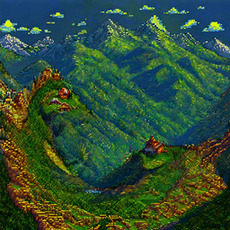}
        \vspace*{-6mm}
        \caption*{\centering \tiny \textit{a 16-bit sprite of} \par\nobreak mountain pass}
    \end{subfigure}
    \begin{subfigure}{0.106\textwidth}
        \centering
        \includegraphics[width=\textwidth]{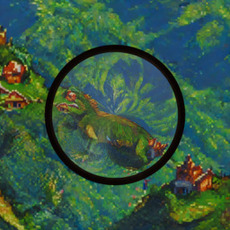}
        \vspace*{-6mm}
        \caption*{\centering \tiny \textit{\quad} \par\nobreak \quad}
    \end{subfigure}
    \begin{subfigure}{0.106\textwidth}
        \centering
        \includegraphics[width=\textwidth]{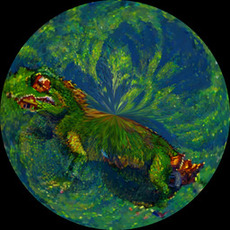}
        \vspace*{-6mm}
        \caption*{\centering \tiny \textit{a 16-bit sprite of} \par\nobreak gecko}
    \end{subfigure}    \hfill
    \begin{subfigure}{0.106\textwidth}
        \centering
        \includegraphics[width=\textwidth]{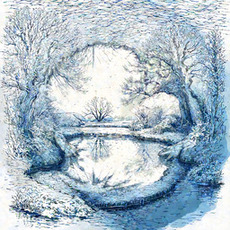}
        \vspace*{-6mm}
        \caption*{\centering \tiny \textit{a line drawing of} \par\nobreak mirror-like frozen pond}
    \end{subfigure}
    \begin{subfigure}{0.106\textwidth}
        \centering
        \includegraphics[width=\textwidth]{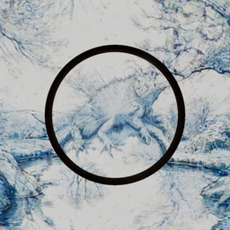}
        \vspace*{-6mm}
        \caption*{\centering \tiny \textit{\quad} \par\nobreak \quad}
    \end{subfigure}
    \begin{subfigure}{0.106\textwidth}
        \centering
        \includegraphics[width=\textwidth]{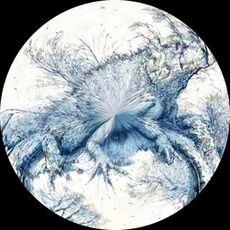}
        \vspace*{-6mm}
        \caption*{\centering \tiny \textit{a line drawing of} \par\nobreak lizard}
    \end{subfigure}
\end{figure*}

\begin{figure*}[h!]
    \centering
    \begin{subfigure}{0.490\textwidth}
        \centering
        \includegraphics[width=\textwidth]{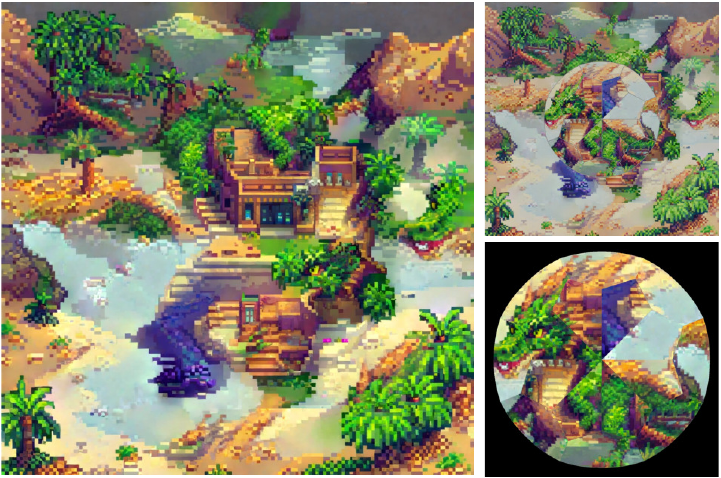}
        \caption*{\normalsize \centering \textit{a 16-bit sprite of} \par\nobreak desert oasis / dragon}
    \end{subfigure}    \hfill
    \begin{subfigure}{0.490\textwidth}
        \centering
        \includegraphics[width=\textwidth]{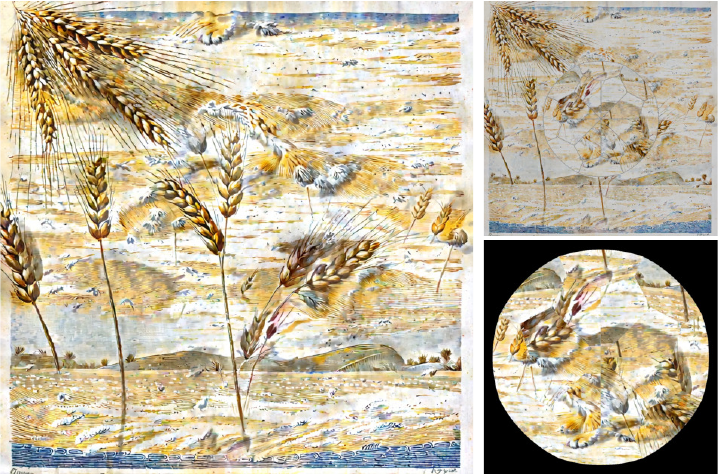}
        \caption*{\normalsize \centering \textit{a lithograph of} \par\nobreak horizon of a wheat field / rabbit}
    \end{subfigure}
    
    \vspace{1.0cm} %
    \begin{subfigure}{0.490\textwidth}
        \centering
        \includegraphics[width=\textwidth]{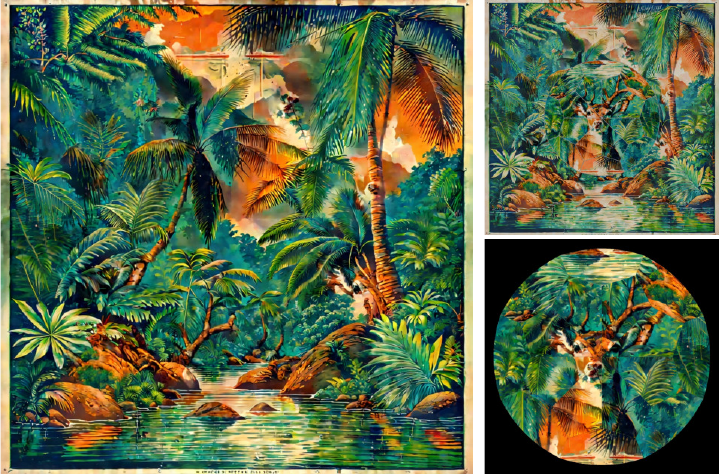}
        \caption*{\normalsize \centering \textit{a vintage poster of} \par\nobreak dense tropical rainforest / deer}
    \end{subfigure}    \hfill
    \begin{subfigure}{0.490\textwidth}
        \centering
        \includegraphics[width=\textwidth]{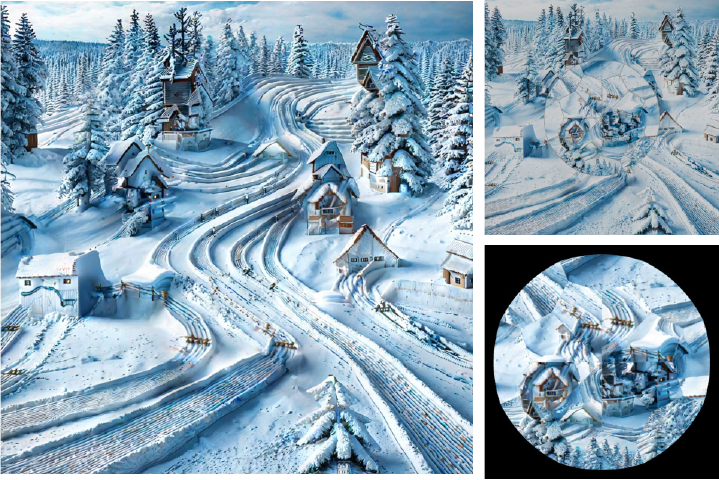}
        \caption*{\normalsize \centering \textit{a hyperrealistic sculpture of} \par\nobreak straight ski tracks / motorcycle}
    \end{subfigure}
    
    \vspace{1.0cm} %
    \begin{subfigure}{0.490\textwidth}
        \centering
        \includegraphics[width=\textwidth]{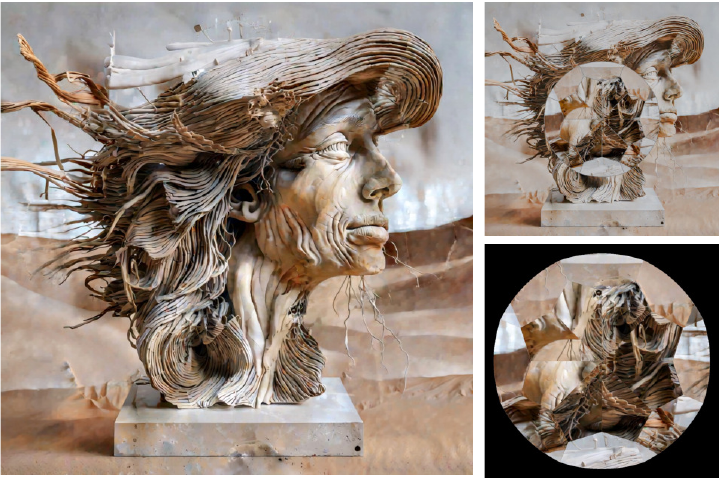}
        \caption*{\normalsize \centering \textit{a hyperrealistic sculpture of} \par\nobreak windy desert / walrus}
    \end{subfigure}    \hfill
    \begin{subfigure}{0.490\textwidth}
        \centering
        \includegraphics[width=\textwidth]{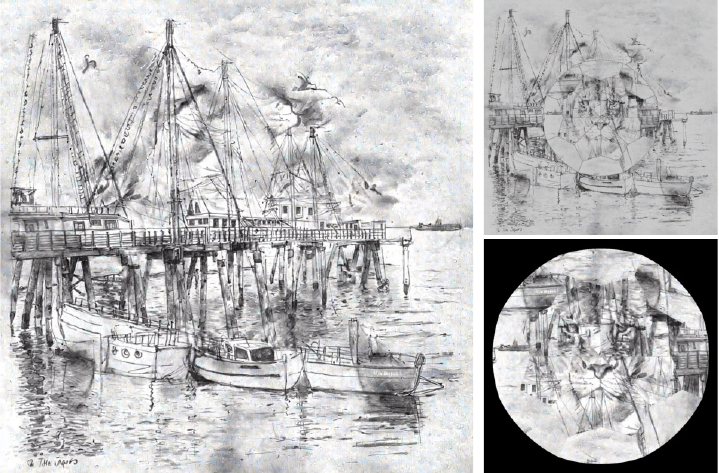}
        \caption*{\normalsize \centering \textit{a pencil sketch of} \par\nobreak harbor pier / lion}
    \end{subfigure}
        \caption{\textbf{Nicéron's lens anamorphosis.} In this figure and the two following ones, we show additional results for the lens example. Each example contains the identity view, the lens view as predicted by the flow model, and a rendering of the actual image through the lens to validate our examples. Kindly refer to the supplementary videos to see these results in action.}
    \label{fig:suppl_lens_b}
\end{figure*}

\begin{figure*}[h!]
    \centering
    \begin{subfigure}{0.158\textwidth}
        \centering
        \includegraphics[width=\textwidth]{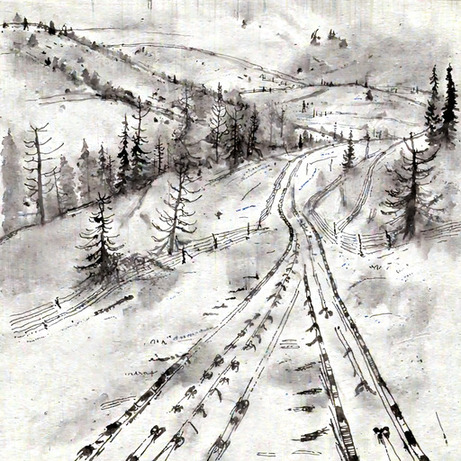}
        \vspace*{-6mm}
        \caption*{\centering \tiny \textit{an ink wash drawing of} \par\nobreak straight ski tracks}
    \end{subfigure}
    \begin{subfigure}{0.158\textwidth}
        \centering
        \includegraphics[width=\textwidth]{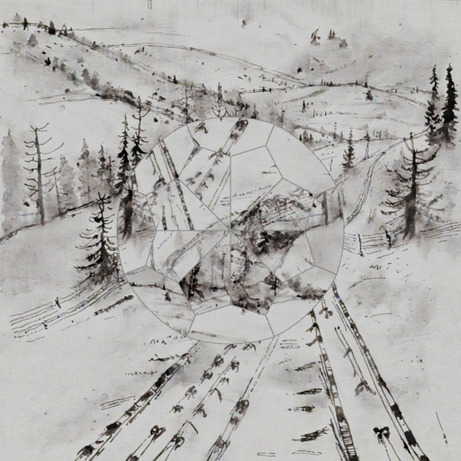}
        \vspace*{-6mm}
        \caption*{\centering \tiny \textit{\quad} \par\nobreak \quad}
    \end{subfigure}
    \begin{subfigure}{0.158\textwidth}
        \centering
        \includegraphics[width=\textwidth]{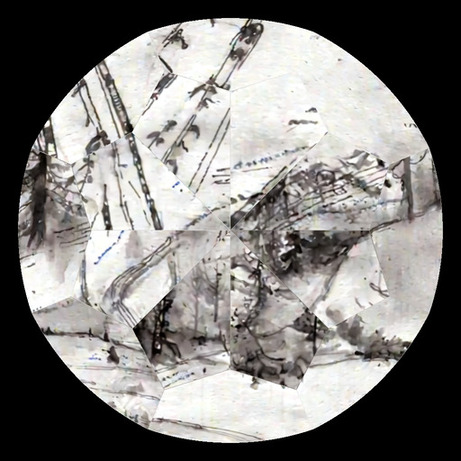}
        \vspace*{-6mm}
        \caption*{\centering \tiny \textit{an ink wash drawing of} \par\nobreak iguana}
    \end{subfigure}    \hfill
    \begin{subfigure}{0.158\textwidth}
        \centering
        \includegraphics[width=\textwidth]{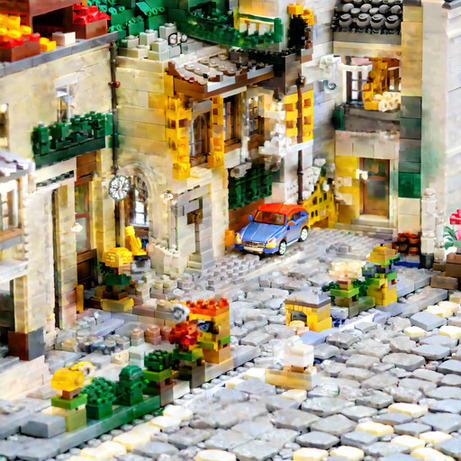}
        \vspace*{-6mm}
        \caption*{\centering \tiny \textit{a LEGO model of} \par\nobreak cobblestone street}
    \end{subfigure}
    \begin{subfigure}{0.158\textwidth}
        \centering
        \includegraphics[width=\textwidth]{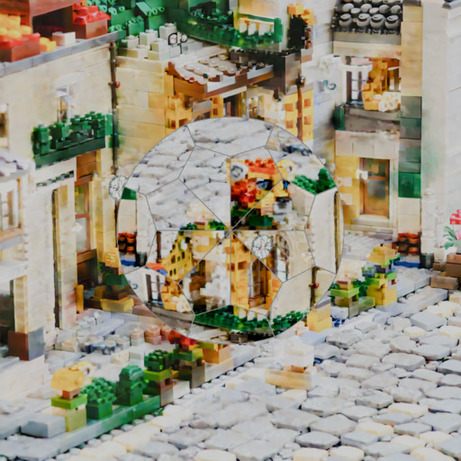}
        \vspace*{-6mm}
        \caption*{\centering \tiny \textit{\quad} \par\nobreak \quad}
    \end{subfigure}
    \begin{subfigure}{0.158\textwidth}
        \centering
        \includegraphics[width=\textwidth]{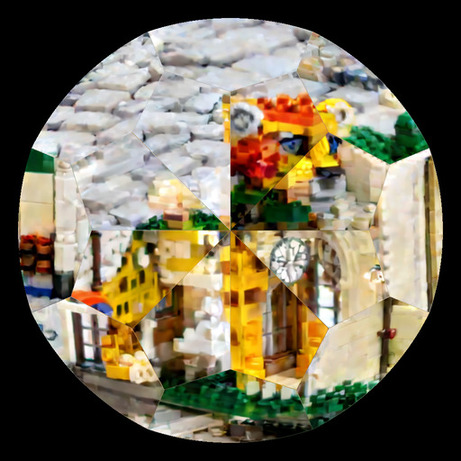}
        \vspace*{-6mm}
        \caption*{\centering \tiny \textit{a LEGO model of} \par\nobreak cheetah}
    \end{subfigure}
    \begin{subfigure}{0.158\textwidth}
        \centering
        \includegraphics[width=\textwidth]{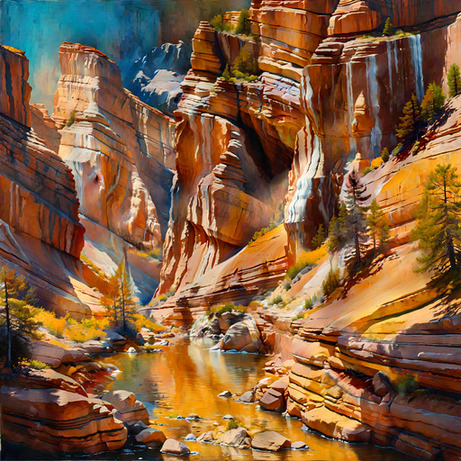}
        \vspace*{-6mm}
        \caption*{\centering \tiny \textit{an oil painting of} \par\nobreak sunlit canyon with straight cliffs}
    \end{subfigure}
    \begin{subfigure}{0.158\textwidth}
        \centering
        \includegraphics[width=\textwidth]{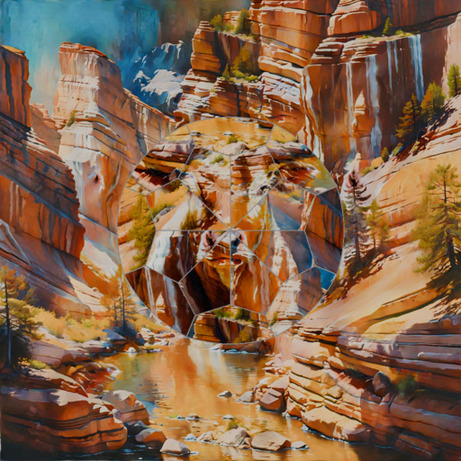}
        \vspace*{-6mm}
        \caption*{\centering \tiny \textit{\quad} \par\nobreak \quad}
    \end{subfigure}
    \begin{subfigure}{0.158\textwidth}
        \centering
        \includegraphics[width=\textwidth]{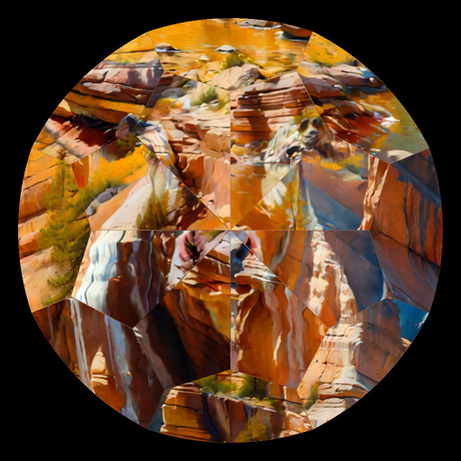}
        \vspace*{-6mm}
        \caption*{\centering \tiny \textit{an oil painting of} \par\nobreak bull}
    \end{subfigure}    \hfill
    \begin{subfigure}{0.158\textwidth}
        \centering
        \includegraphics[width=\textwidth]{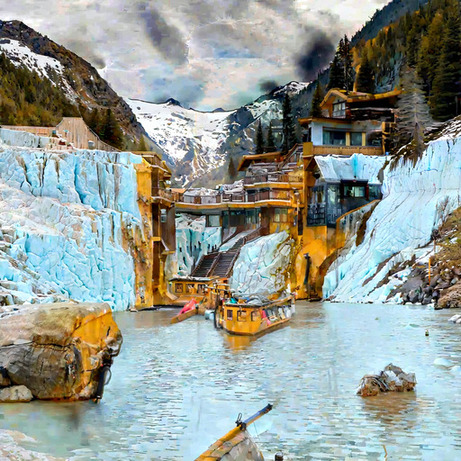}
        \vspace*{-6mm}
        \caption*{\centering \tiny \textit{a photo of} \par\nobreak glacier valley}
    \end{subfigure}
    \begin{subfigure}{0.158\textwidth}
        \centering
        \includegraphics[width=\textwidth]{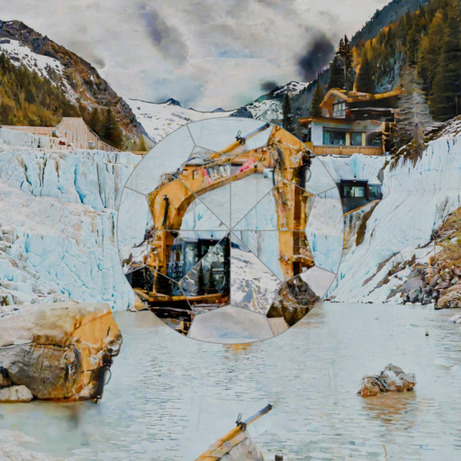}
        \vspace*{-6mm}
        \caption*{\centering \tiny \textit{\quad} \par\nobreak \quad}
    \end{subfigure}
    \begin{subfigure}{0.158\textwidth}
        \centering
        \includegraphics[width=\textwidth]{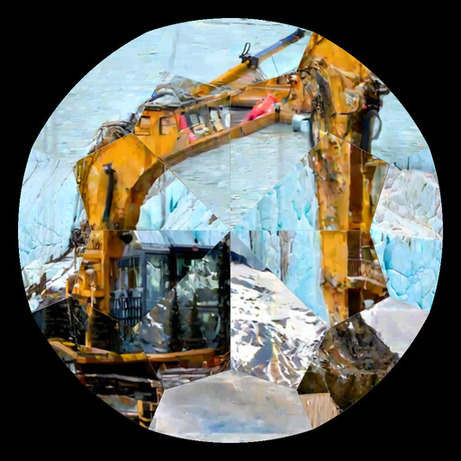}
        \vspace*{-6mm}
        \caption*{\centering \tiny \textit{a photo of} \par\nobreak excavator}
    \end{subfigure}
    \begin{subfigure}{0.158\textwidth}
        \centering
        \includegraphics[width=\textwidth]{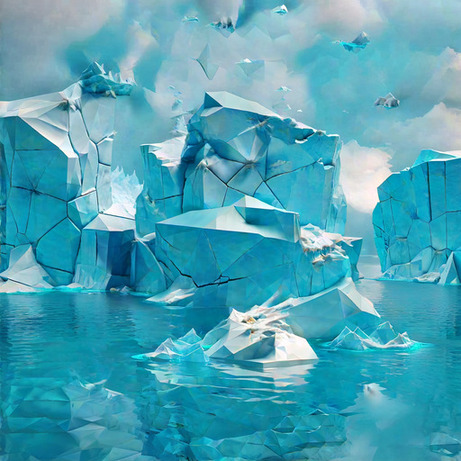}
        \vspace*{-6mm}
        \caption*{\centering \tiny \textit{a low-poly model of} \par\nobreak icebergs in the ocean}
    \end{subfigure}
    \begin{subfigure}{0.158\textwidth}
        \centering
        \includegraphics[width=\textwidth]{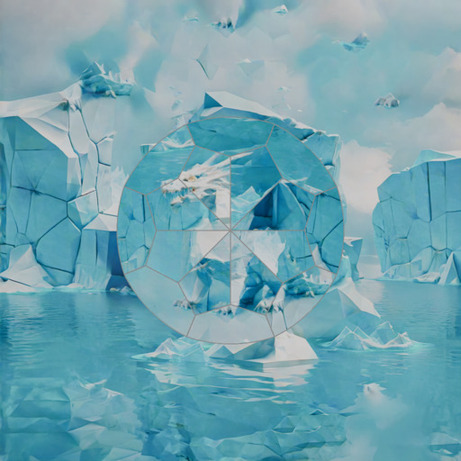}
        \vspace*{-6mm}
        \caption*{\centering \tiny \textit{\quad} \par\nobreak \quad}
    \end{subfigure}
    \begin{subfigure}{0.158\textwidth}
        \centering
        \includegraphics[width=\textwidth]{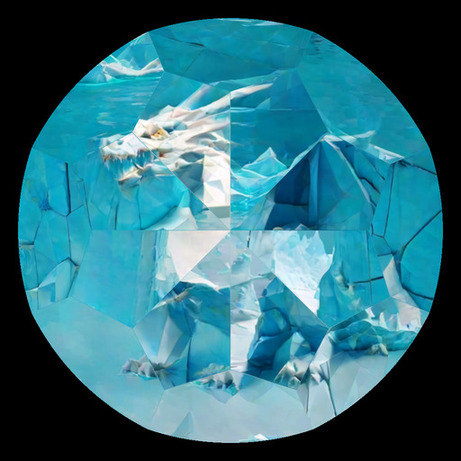}
        \vspace*{-6mm}
        \caption*{\centering \tiny \textit{a low-poly model of} \par\nobreak dragon}
    \end{subfigure}    \hfill
    \begin{subfigure}{0.158\textwidth}
        \centering
        \includegraphics[width=\textwidth]{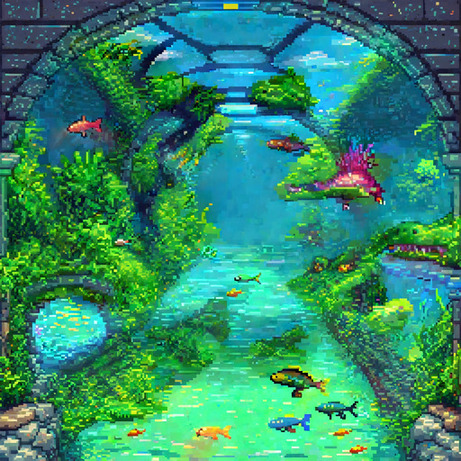}
        \vspace*{-6mm}
        \caption*{\centering \tiny \textit{a pixel art version of} \par\nobreak aquarium tunnel}
    \end{subfigure}
    \begin{subfigure}{0.158\textwidth}
        \centering
        \includegraphics[width=\textwidth]{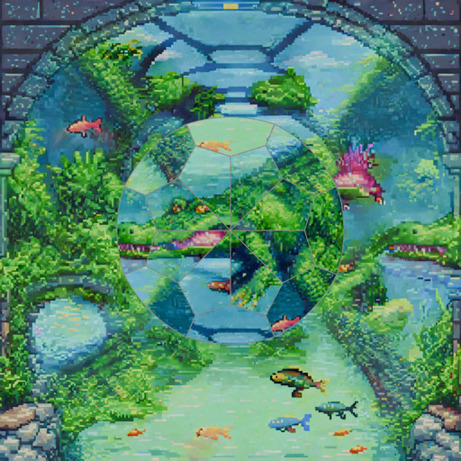}
        \vspace*{-6mm}
        \caption*{\centering \tiny \textit{\quad} \par\nobreak \quad}
    \end{subfigure}
    \begin{subfigure}{0.158\textwidth}
        \centering
        \includegraphics[width=\textwidth]{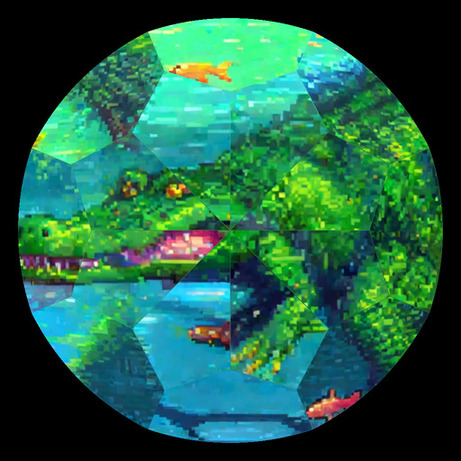}
        \vspace*{-6mm}
        \caption*{\centering \tiny \textit{a pixel art version of} \par\nobreak alligator}
    \end{subfigure}
    \begin{subfigure}{0.158\textwidth}
        \centering
        \includegraphics[width=\textwidth]{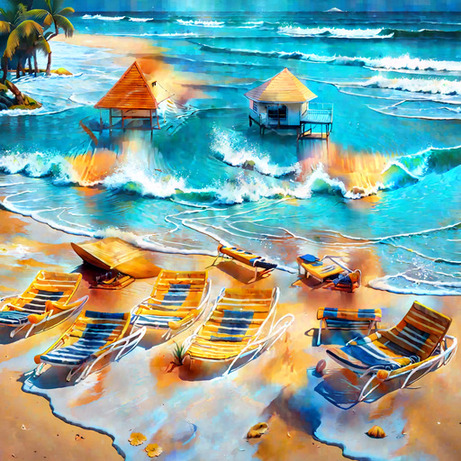}
        \vspace*{-6mm}
        \caption*{\centering \tiny \textit{a digital illustration of} \par\nobreak beach with straight shoreline \& waves}
    \end{subfigure}
    \begin{subfigure}{0.158\textwidth}
        \centering
        \includegraphics[width=\textwidth]{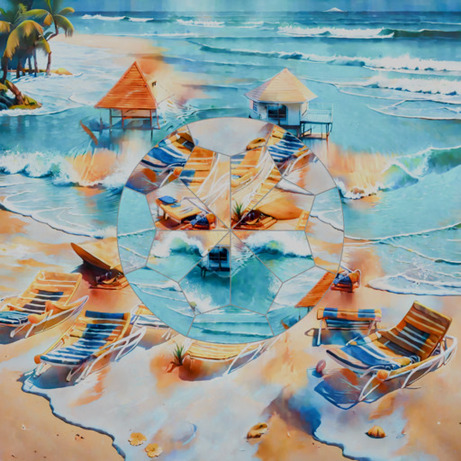}
        \vspace*{-6mm}
        \caption*{\centering \tiny \textit{\quad} \par\nobreak \quad}
    \end{subfigure}
    \begin{subfigure}{0.158\textwidth}
        \centering
        \includegraphics[width=\textwidth]{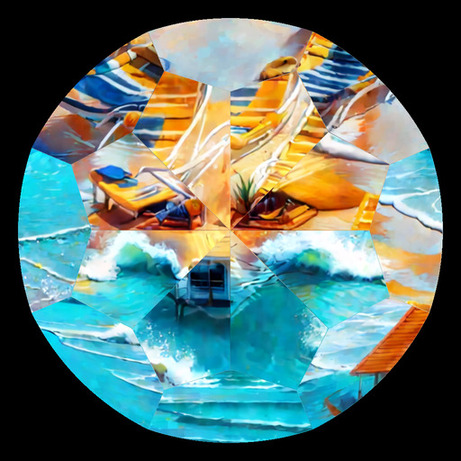}
        \vspace*{-6mm}
        \caption*{\centering \tiny \textit{a digital illustration of} \par\nobreak fox}
    \end{subfigure}    \hfill
    \begin{subfigure}{0.158\textwidth}
        \centering
        \includegraphics[width=\textwidth]{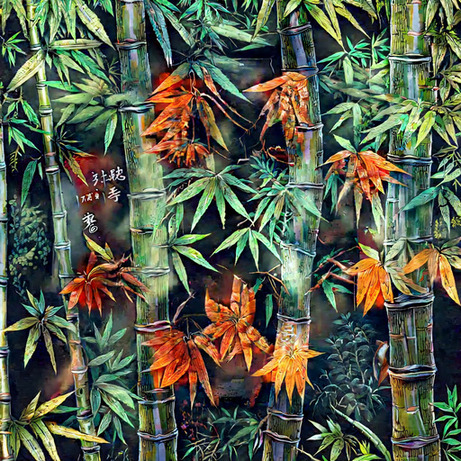}
        \vspace*{-6mm}
        \caption*{\centering \tiny \textit{a chalkboard drawing of} \par\nobreak bamboo forest}
    \end{subfigure}
    \begin{subfigure}{0.158\textwidth}
        \centering
        \includegraphics[width=\textwidth]{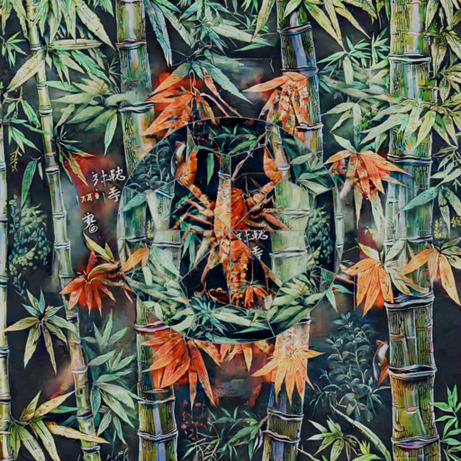}
        \vspace*{-6mm}
        \caption*{\centering \tiny \textit{\quad} \par\nobreak \quad}
    \end{subfigure}
    \begin{subfigure}{0.158\textwidth}
        \centering
        \includegraphics[width=\textwidth]{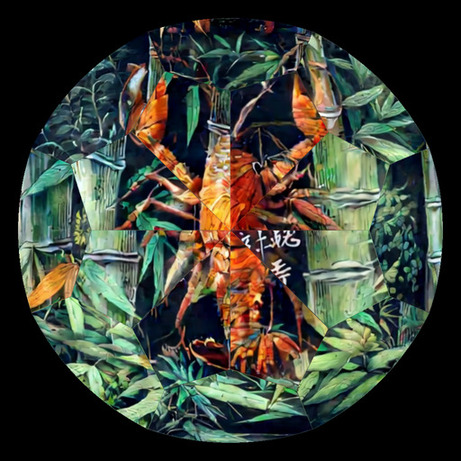}
        \vspace*{-6mm}
        \caption*{\centering \tiny \textit{a chalkboard drawing of} \par\nobreak lobster}
    \end{subfigure}
    \begin{subfigure}{0.158\textwidth}
        \centering
        \includegraphics[width=\textwidth]{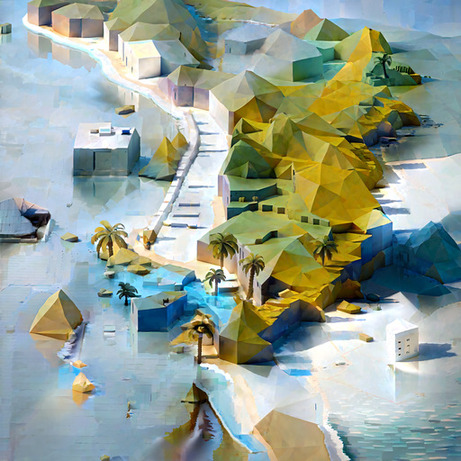}
        \vspace*{-6mm}
        \caption*{\centering \tiny \textit{a low-poly model of} \par\nobreak straight coastline}
    \end{subfigure}
    \begin{subfigure}{0.158\textwidth}
        \centering
        \includegraphics[width=\textwidth]{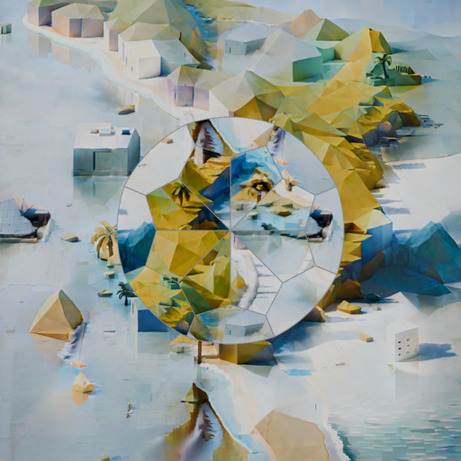}
        \vspace*{-6mm}
        \caption*{\centering \tiny \textit{\quad} \par\nobreak \quad}
    \end{subfigure}
    \begin{subfigure}{0.158\textwidth}
        \centering
        \includegraphics[width=\textwidth]{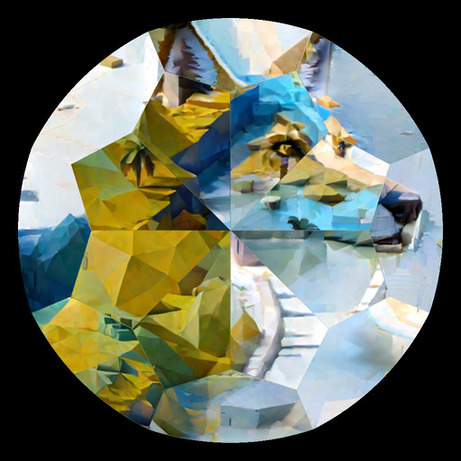}
        \vspace*{-6mm}
        \caption*{\centering \tiny \textit{a low-poly model of} \par\nobreak wolf}
    \end{subfigure}    \hfill
    \begin{subfigure}{0.158\textwidth}
        \centering
        \includegraphics[width=\textwidth]{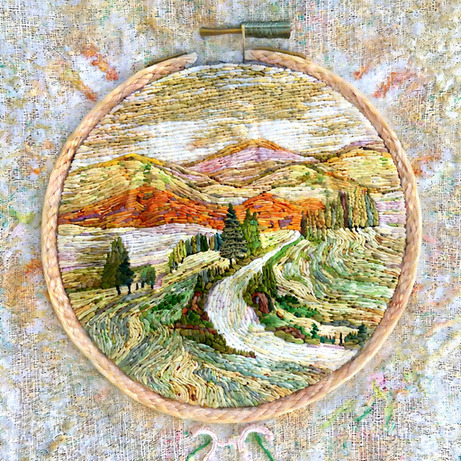}
        \vspace*{-6mm}
        \caption*{\centering \tiny \textit{an embroidered version of} \par\nobreak rolling hills in golden light}
    \end{subfigure}
    \begin{subfigure}{0.158\textwidth}
        \centering
        \includegraphics[width=\textwidth]{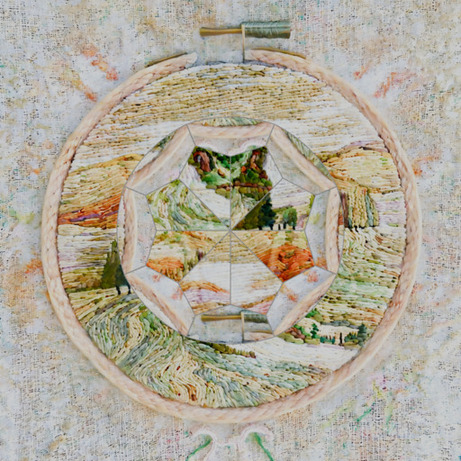}
        \vspace*{-6mm}
        \caption*{\centering \tiny \textit{\quad} \par\nobreak \quad}
    \end{subfigure}
    \begin{subfigure}{0.158\textwidth}
        \centering
        \includegraphics[width=\textwidth]{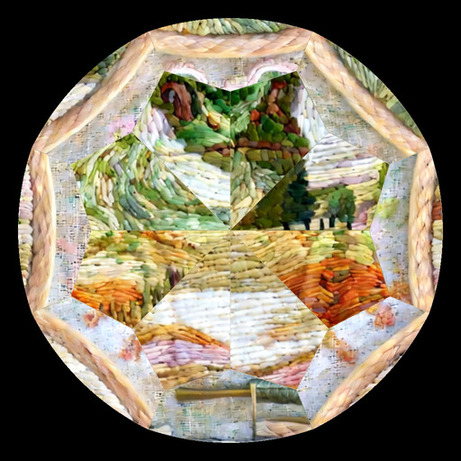}
        \vspace*{-6mm}
        \caption*{\centering \tiny \textit{an embroidered version of} \par\nobreak frog}
    \end{subfigure}
    \begin{subfigure}{0.158\textwidth}
        \centering
        \includegraphics[width=\textwidth]{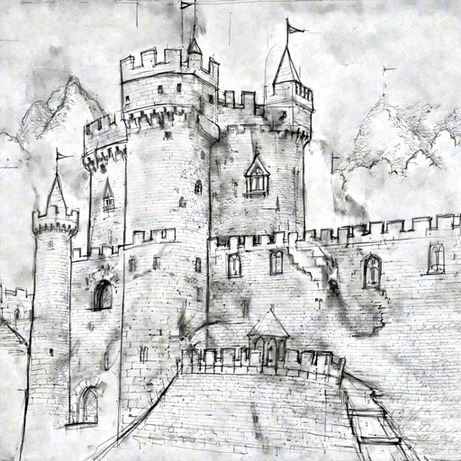}
        \vspace*{-6mm}
        \caption*{\centering \tiny \textit{a pencil sketch of} \par\nobreak castle walls with battlements}
    \end{subfigure}
    \begin{subfigure}{0.158\textwidth}
        \centering
        \includegraphics[width=\textwidth]{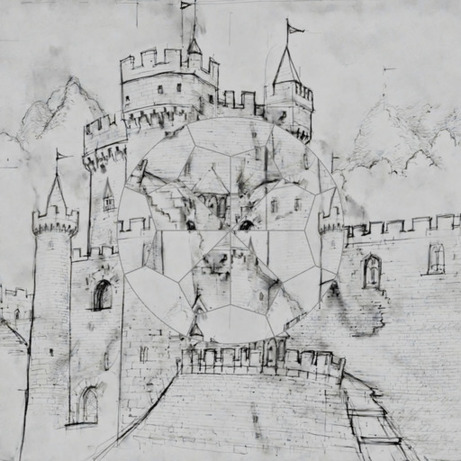}
        \vspace*{-6mm}
        \caption*{\centering \tiny \textit{\quad} \par\nobreak \quad}
    \end{subfigure}
    \begin{subfigure}{0.158\textwidth}
        \centering
        \includegraphics[width=\textwidth]{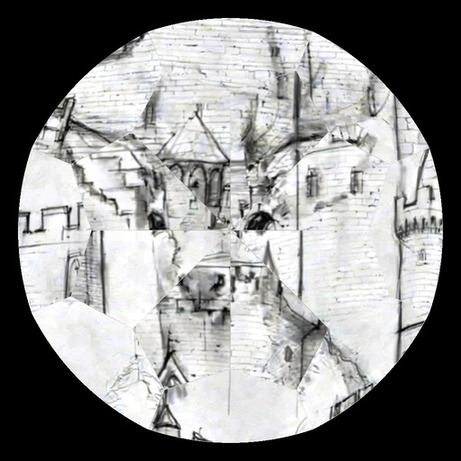}
        \vspace*{-6mm}
        \caption*{\centering \tiny \textit{a pencil sketch of} \par\nobreak deer}
    \end{subfigure}    \hfill
    \begin{subfigure}{0.158\textwidth}
        \centering
        \includegraphics[width=\textwidth]{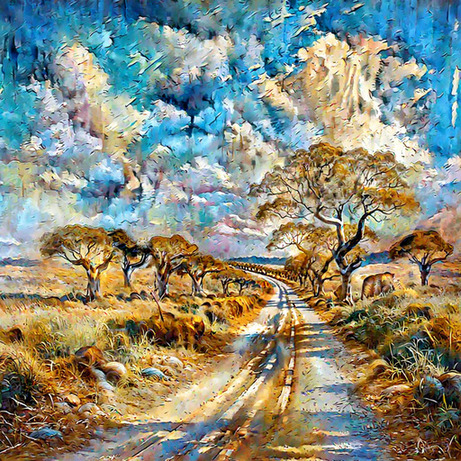}
        \vspace*{-6mm}
        \caption*{\centering \tiny \textit{a pastel artwork of} \par\nobreak straight dirt road through a savannah}
    \end{subfigure}
    \begin{subfigure}{0.158\textwidth}
        \centering
        \includegraphics[width=\textwidth]{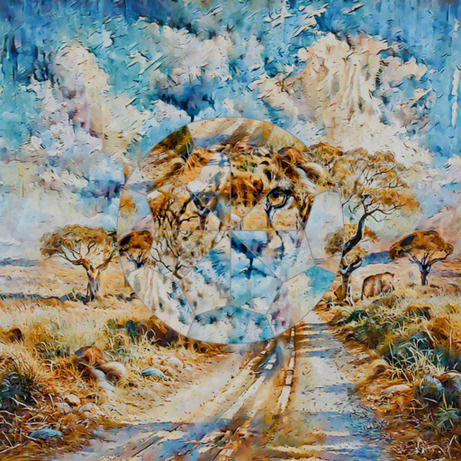}
        \vspace*{-6mm}
        \caption*{\centering \tiny \textit{\quad} \par\nobreak \quad}
    \end{subfigure}
    \begin{subfigure}{0.158\textwidth}
        \centering
        \includegraphics[width=\textwidth]{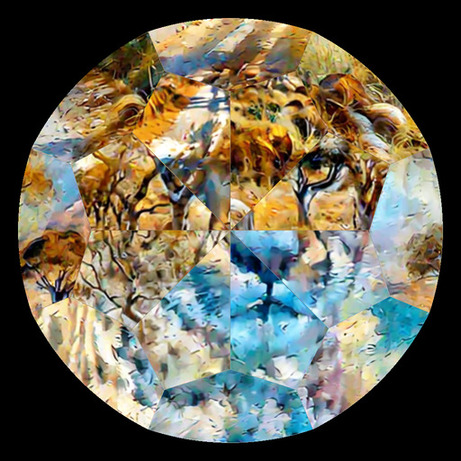}
        \vspace*{-6mm}
        \caption*{\centering \tiny \textit{a pastel artwork of} \par\nobreak cheetah}
    \end{subfigure}
    \begin{subfigure}{0.158\textwidth}
        \centering
        \includegraphics[width=\textwidth]{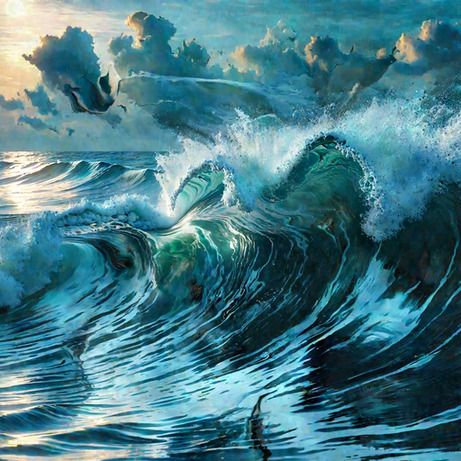}
        \vspace*{-6mm}
        \caption*{\centering \tiny \textit{a cinematic rendering of} \par\nobreak ocean waves}
    \end{subfigure}
    \begin{subfigure}{0.158\textwidth}
        \centering
        \includegraphics[width=\textwidth]{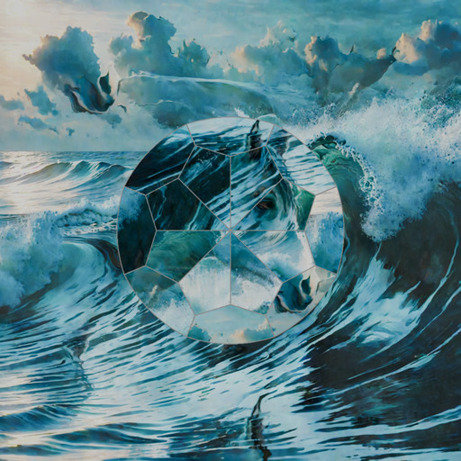}
        \vspace*{-6mm}
        \caption*{\centering \tiny \textit{\quad} \par\nobreak \quad}
    \end{subfigure}
    \begin{subfigure}{0.158\textwidth}
        \centering
        \includegraphics[width=\textwidth]{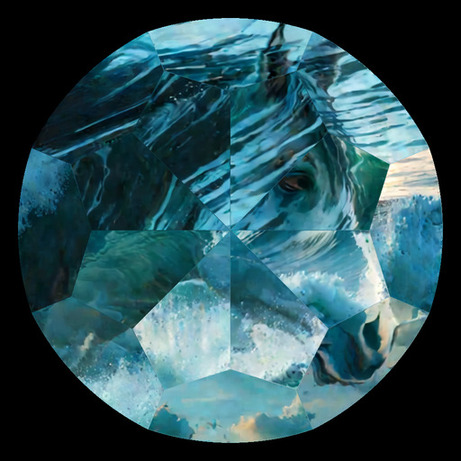}
        \vspace*{-6mm}
        \caption*{\centering \tiny \textit{a cinematic rendering of} \par\nobreak horse}
    \end{subfigure}    \hfill
    \begin{subfigure}{0.158\textwidth}
        \centering
        \includegraphics[width=\textwidth]{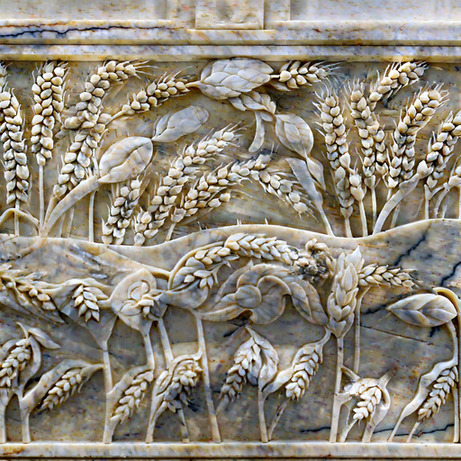}
        \vspace*{-6mm}
        \caption*{\centering \tiny \textit{a marble carving of} \par\nobreak horizon of a wheat field}
    \end{subfigure}
    \begin{subfigure}{0.158\textwidth}
        \centering
        \includegraphics[width=\textwidth]{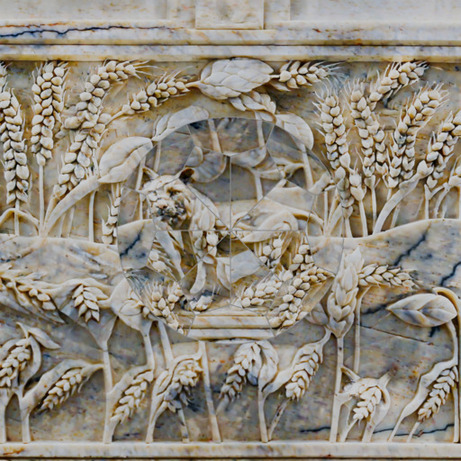}
        \vspace*{-6mm}
        \caption*{\centering \tiny \textit{\quad} \par\nobreak \quad}
    \end{subfigure}
    \begin{subfigure}{0.158\textwidth}
        \centering
        \includegraphics[width=\textwidth]{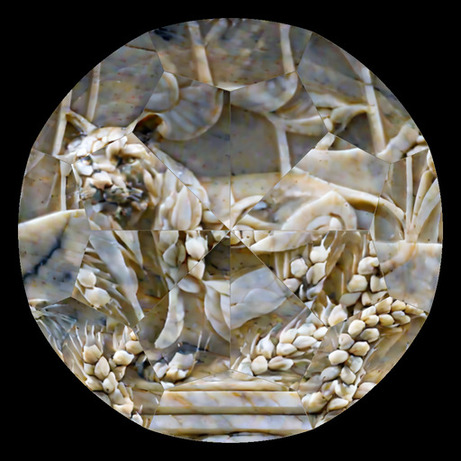}
        \vspace*{-6mm}
        \caption*{\centering \tiny \textit{a marble carving of} \par\nobreak cougar}
    \end{subfigure}
\end{figure*}

\begin{figure*}[h!]
    \centering
    \begin{subfigure}{0.106\textwidth}
        \centering
        \includegraphics[width=\textwidth]{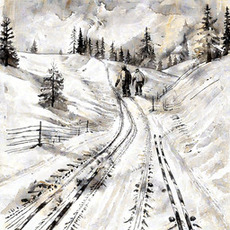}
        \vspace*{-6mm}
        \caption*{\centering \tiny \textit{an ink wash drawing of} \par\nobreak straight ski tracks}
    \end{subfigure}
    \begin{subfigure}{0.106\textwidth}
        \centering
        \includegraphics[width=\textwidth]{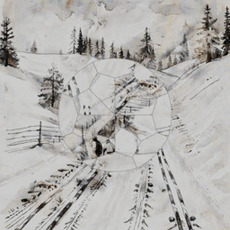}
        \vspace*{-6mm}
        \caption*{\centering \tiny \textit{\quad} \par\nobreak \quad}
    \end{subfigure}
    \begin{subfigure}{0.106\textwidth}
        \centering
        \includegraphics[width=\textwidth]{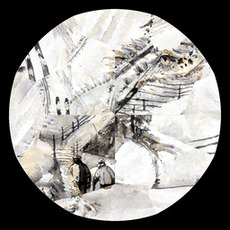}
        \vspace*{-6mm}
        \caption*{\centering \tiny \textit{an ink wash drawing of} \par\nobreak iguana}
    \end{subfigure}    \hfill
    \begin{subfigure}{0.106\textwidth}
        \centering
        \includegraphics[width=\textwidth]{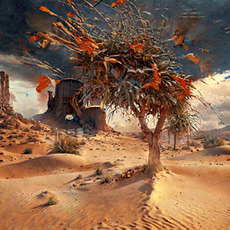}
        \vspace*{-6mm}
        \caption*{\centering \tiny \textit{a cinematic rendering of} \par\nobreak windy desert}
    \end{subfigure}
    \begin{subfigure}{0.106\textwidth}
        \centering
        \includegraphics[width=\textwidth]{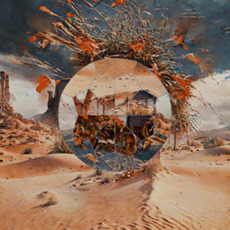}
        \vspace*{-6mm}
        \caption*{\centering \tiny \textit{\quad} \par\nobreak \quad}
    \end{subfigure}
    \begin{subfigure}{0.106\textwidth}
        \centering
        \includegraphics[width=\textwidth]{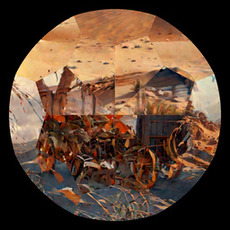}
        \vspace*{-6mm}
        \caption*{\centering \tiny \textit{a cinematic rendering of} \par\nobreak wagon}
    \end{subfigure}    \hfill
    \begin{subfigure}{0.106\textwidth}
        \centering
        \includegraphics[width=\textwidth]{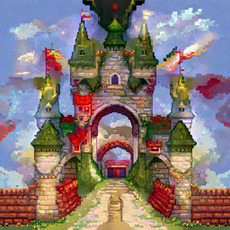}
        \vspace*{-6mm}
        \caption*{\centering \tiny \textit{a pixel art version of} \par\nobreak medieval castle gate}
    \end{subfigure}
    \begin{subfigure}{0.106\textwidth}
        \centering
        \includegraphics[width=\textwidth]{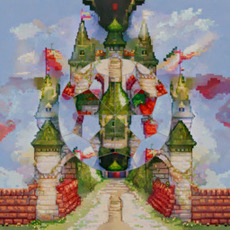}
        \vspace*{-6mm}
        \caption*{\centering \tiny \textit{\quad} \par\nobreak \quad}
    \end{subfigure}
    \begin{subfigure}{0.106\textwidth}
        \centering
        \includegraphics[width=\textwidth]{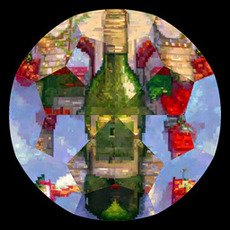}
        \vspace*{-6mm}
        \caption*{\centering \tiny \textit{a pixel art version of} \par\nobreak wine bottle}
    \end{subfigure}
    \begin{subfigure}{0.106\textwidth}
        \centering
        \includegraphics[width=\textwidth]{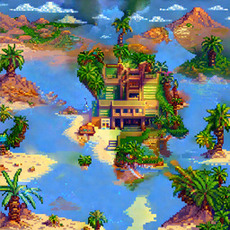}
        \vspace*{-6mm}
        \caption*{\centering \tiny \textit{a 16-bit sprite of} \par\nobreak desert oasis}
    \end{subfigure}
    \begin{subfigure}{0.106\textwidth}
        \centering
        \includegraphics[width=\textwidth]{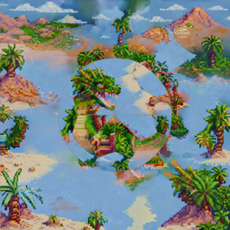}
        \vspace*{-6mm}
        \caption*{\centering \tiny \textit{\quad} \par\nobreak \quad}
    \end{subfigure}
    \begin{subfigure}{0.106\textwidth}
        \centering
        \includegraphics[width=\textwidth]{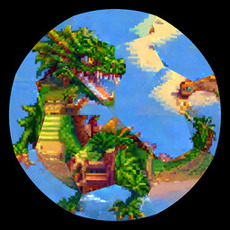}
        \vspace*{-6mm}
        \caption*{\centering \tiny \textit{a 16-bit sprite of} \par\nobreak dragon}
    \end{subfigure}    \hfill
    \begin{subfigure}{0.106\textwidth}
        \centering
        \includegraphics[width=\textwidth]{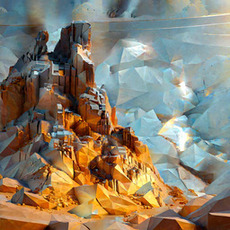}
        \vspace*{-6mm}
        \caption*{\centering \tiny \textit{a low-poly model of} \par\nobreak sunlit rocky outcrop}
    \end{subfigure}
    \begin{subfigure}{0.106\textwidth}
        \centering
        \includegraphics[width=\textwidth]{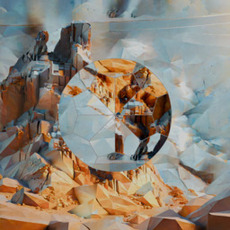}
        \vspace*{-6mm}
        \caption*{\centering \tiny \textit{\quad} \par\nobreak \quad}
    \end{subfigure}
    \begin{subfigure}{0.106\textwidth}
        \centering
        \includegraphics[width=\textwidth]{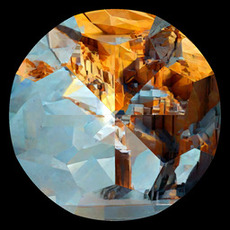}
        \vspace*{-6mm}
        \caption*{\centering \tiny \textit{a low-poly model of} \par\nobreak fox}
    \end{subfigure}    \hfill
    \begin{subfigure}{0.106\textwidth}
        \centering
        \includegraphics[width=\textwidth]{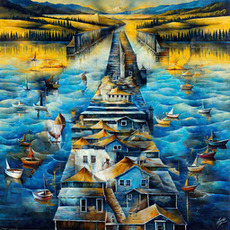}
        \vspace*{-6mm}
        \caption*{\centering \tiny \textit{a surrealist painting of} levee dividing water}
    \end{subfigure}
    \begin{subfigure}{0.106\textwidth}
        \centering
        \includegraphics[width=\textwidth]{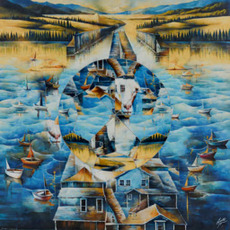}
        \vspace*{-6mm}
        \caption*{\centering \tiny \textit{\quad} \par\nobreak \quad}
    \end{subfigure}
    \begin{subfigure}{0.106\textwidth}
        \centering
        \includegraphics[width=\textwidth]{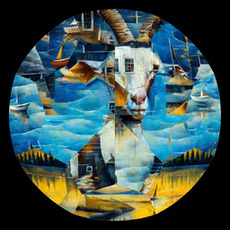}
        \vspace*{-6mm}
        \caption*{\centering \tiny \textit{a surrealist painting of} \par\nobreak goat}
    \end{subfigure}
    \begin{subfigure}{0.106\textwidth}
        \centering
        \includegraphics[width=\textwidth]{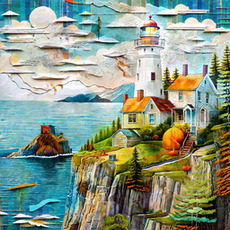}
        \vspace*{-6mm}
        \caption*{\centering \tiny \textit{a paper collage of} \par\nobreak cliffside lighthouse}
    \end{subfigure}
    \begin{subfigure}{0.106\textwidth}
        \centering
        \includegraphics[width=\textwidth]{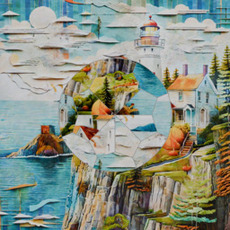}
        \vspace*{-6mm}
        \caption*{\centering \tiny \textit{\quad} \par\nobreak \quad}
    \end{subfigure}
    \begin{subfigure}{0.106\textwidth}
        \centering
        \includegraphics[width=\textwidth]{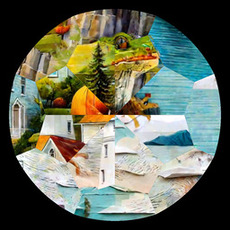}
        \vspace*{-6mm}
        \caption*{\centering \tiny \textit{a paper collage of} \par\nobreak lizard}
    \end{subfigure}    \hfill
    \begin{subfigure}{0.106\textwidth}
        \centering
        \includegraphics[width=\textwidth]{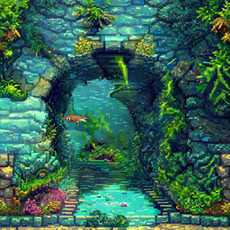}
        \vspace*{-6mm}
        \caption*{\centering \tiny \textit{a pixel art version of} \par\nobreak aquarium tunnel}
    \end{subfigure}
    \begin{subfigure}{0.106\textwidth}
        \centering
        \includegraphics[width=\textwidth]{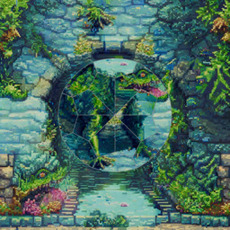}
        \vspace*{-6mm}
        \caption*{\centering \tiny \textit{\quad} \par\nobreak \quad}
    \end{subfigure}
    \begin{subfigure}{0.106\textwidth}
        \centering
        \includegraphics[width=\textwidth]{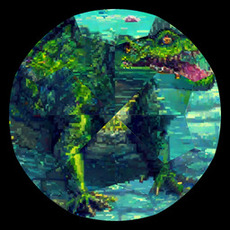}
        \vspace*{-6mm}
        \caption*{\centering \tiny \textit{a pixel art version of} \par\nobreak alligator}
    \end{subfigure}    \hfill
    \begin{subfigure}{0.106\textwidth}
        \centering
        \includegraphics[width=\textwidth]{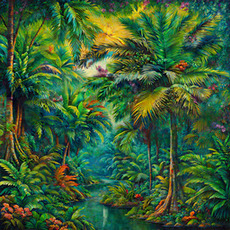}
        \vspace*{-6mm}
        \caption*{\centering \tiny \textit{a surrealist painting of} \par\nobreak dense tropical rainforest}
    \end{subfigure}
    \begin{subfigure}{0.106\textwidth}
        \centering
        \includegraphics[width=\textwidth]{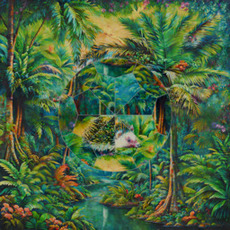}
        \vspace*{-6mm}
        \caption*{\centering \tiny \textit{\quad} \par\nobreak \quad}
    \end{subfigure}
    \begin{subfigure}{0.106\textwidth}
        \centering
        \includegraphics[width=\textwidth]{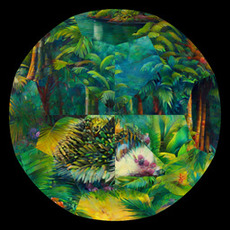}
        \vspace*{-6mm}
        \caption*{\centering \tiny \textit{a surrealist painting of} \par\nobreak hedgehog}
    \end{subfigure}
    \begin{subfigure}{0.106\textwidth}
        \centering
        \includegraphics[width=\textwidth]{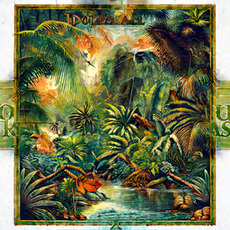}
        \vspace*{-6mm}
        \caption*{\centering \tiny \textit{a vintage poster of} \par\nobreak dense tropical rainforest}
    \end{subfigure}
    \begin{subfigure}{0.106\textwidth}
        \centering
        \includegraphics[width=\textwidth]{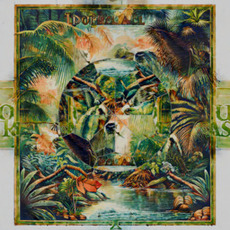}
        \vspace*{-6mm}
        \caption*{\centering \tiny \textit{\quad} \par\nobreak \quad}
    \end{subfigure}
    \begin{subfigure}{0.106\textwidth}
        \centering
        \includegraphics[width=\textwidth]{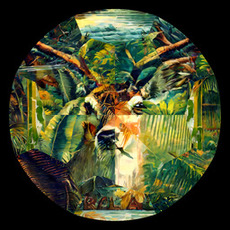}
        \vspace*{-6mm}
        \caption*{\centering \tiny \textit{a vintage poster of} \par\nobreak deer}
    \end{subfigure}    \hfill
    \begin{subfigure}{0.106\textwidth}
        \centering
        \includegraphics[width=\textwidth]{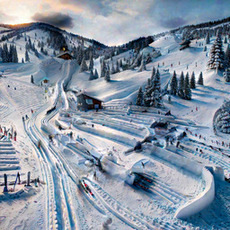}
        \vspace*{-6mm}
        \caption*{\centering \tiny \textit{a hyperrealistic sculpture} \par\nobreak \textit{of} straight ski tracks}
    \end{subfigure}
    \begin{subfigure}{0.106\textwidth}
        \centering
        \includegraphics[width=\textwidth]{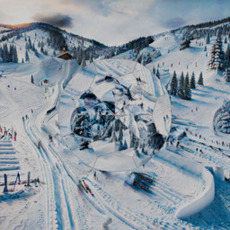}
        \vspace*{-6mm}
        \caption*{\centering \tiny \textit{\quad} \par\nobreak \quad}
    \end{subfigure}
    \begin{subfigure}{0.106\textwidth}
        \centering
        \includegraphics[width=\textwidth]{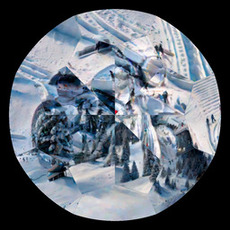}
        \vspace*{-6mm}
        \caption*{\centering \tiny \textit{a hyperrealistic sculpture} \par\nobreak \textit{of} motorcycle}
    \end{subfigure}    \hfill
    \begin{subfigure}{0.106\textwidth}
        \centering
        \includegraphics[width=\textwidth]{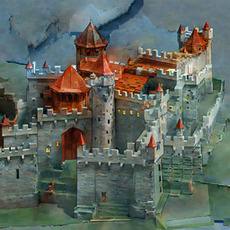}
        \vspace*{-6mm}
        \caption*{\centering \tiny \textit{a low-poly model of} castle walls}
    \end{subfigure}
    \begin{subfigure}{0.106\textwidth}
        \centering
        \includegraphics[width=\textwidth]{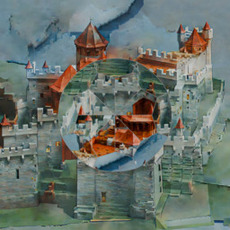}
        \vspace*{-6mm}
        \caption*{\centering \tiny \textit{\quad} \par\nobreak \quad}
    \end{subfigure}
    \begin{subfigure}{0.106\textwidth}
        \centering
        \includegraphics[width=\textwidth]{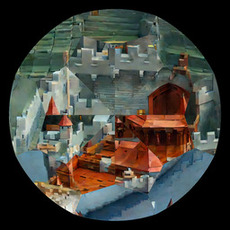}
        \vspace*{-6mm}
        \caption*{\centering \tiny \textit{a low-poly model of} \par\nobreak canoe}
    \end{subfigure}
    \begin{subfigure}{0.106\textwidth}
        \centering
        \includegraphics[width=\textwidth]{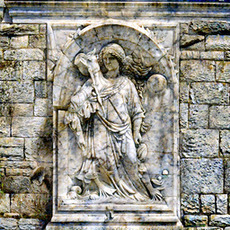}
        \vspace*{-6mm}
        \caption*{\centering \tiny \textit{a marble carving of} highland moor}
    \end{subfigure}
    \begin{subfigure}{0.106\textwidth}
        \centering
        \includegraphics[width=\textwidth]{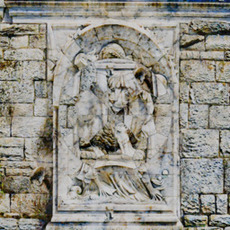}
        \vspace*{-6mm}
        \caption*{\centering \tiny \textit{\quad} \par\nobreak \quad}
    \end{subfigure}
    \begin{subfigure}{0.106\textwidth}
        \centering
        \includegraphics[width=\textwidth]{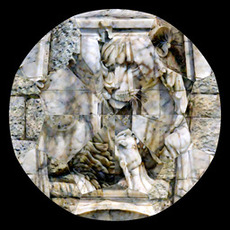}
        \vspace*{-6mm}
        \caption*{\centering \tiny \textit{a marble carving of} \par\nobreak cougar}
    \end{subfigure}    \hfill
    \begin{subfigure}{0.106\textwidth}
        \centering
        \includegraphics[width=\textwidth]{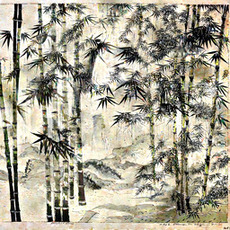}
        \vspace*{-6mm}
        \caption*{\centering \tiny \textit{a lithograph of} \par\nobreak bamboo forest}
    \end{subfigure}
    \begin{subfigure}{0.106\textwidth}
        \centering
        \includegraphics[width=\textwidth]{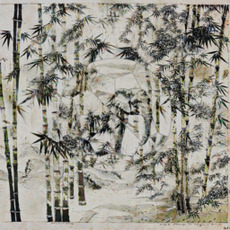}
        \vspace*{-6mm}
        \caption*{\centering \tiny \textit{\quad} \par\nobreak \quad}
    \end{subfigure}
    \begin{subfigure}{0.106\textwidth}
        \centering
        \includegraphics[width=\textwidth]{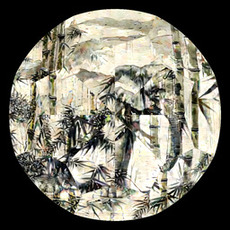}
        \vspace*{-6mm}
        \caption*{\centering \tiny \textit{a lithograph of} \par\nobreak elephant}
    \end{subfigure}    \hfill
    \begin{subfigure}{0.106\textwidth}
        \centering
        \includegraphics[width=\textwidth]{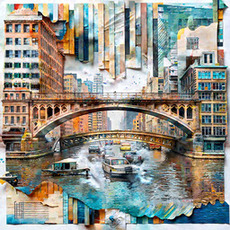}
        \vspace*{-6mm}
        \caption*{\centering \tiny \textit{a paper collage of} \par\nobreak urban bridge over a river}
    \end{subfigure}
    \begin{subfigure}{0.106\textwidth}
        \centering
        \includegraphics[width=\textwidth]{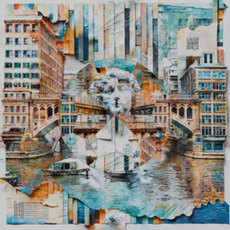}
        \vspace*{-6mm}
        \caption*{\centering \tiny \textit{\quad} \par\nobreak \quad}
    \end{subfigure}
    \begin{subfigure}{0.106\textwidth}
        \centering
        \includegraphics[width=\textwidth]{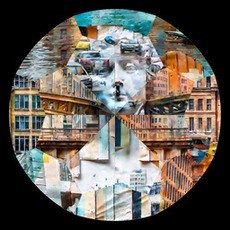}
        \vspace*{-6mm}
        \caption*{\centering \tiny \textit{a paper collage of} \par\nobreak statue}
    \end{subfigure}
    \begin{subfigure}{0.106\textwidth}
        \centering
        \includegraphics[width=\textwidth]{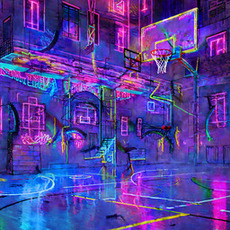}
        \vspace*{-6mm}
        \caption*{\centering \tiny \textit{a neon light artwork of} \par\nobreak urban basketball court}
    \end{subfigure}
    \begin{subfigure}{0.106\textwidth}
        \centering
        \includegraphics[width=\textwidth]{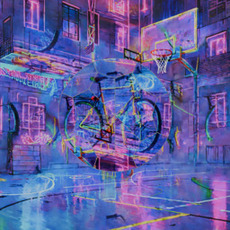}
        \vspace*{-6mm}
        \caption*{\centering \tiny \textit{\quad} \par\nobreak \quad}
    \end{subfigure}
    \begin{subfigure}{0.106\textwidth}
        \centering
        \includegraphics[width=\textwidth]{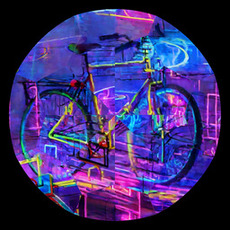}
        \vspace*{-6mm}
        \caption*{\centering \tiny \textit{a neon light artwork of} \par\nobreak bicycle}
    \end{subfigure}    \hfill
    \begin{subfigure}{0.106\textwidth}
        \centering
        \includegraphics[width=\textwidth]{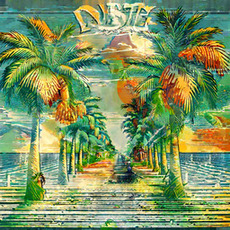}
        \vspace*{-6mm}
        \caption*{\centering \tiny \textit{a vintage poster of} rows of palm trees}
    \end{subfigure}
    \begin{subfigure}{0.106\textwidth}
        \centering
        \includegraphics[width=\textwidth]{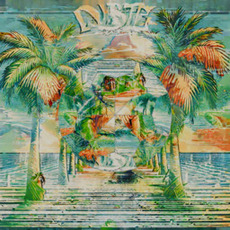}
        \vspace*{-6mm}
        \caption*{\centering \tiny \textit{\quad} \par\nobreak \quad}
    \end{subfigure}
    \begin{subfigure}{0.106\textwidth}
        \centering
        \includegraphics[width=\textwidth]{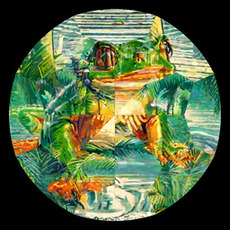}
        \vspace*{-6mm}
        \caption*{\centering \tiny \textit{a vintage poster of} \par\nobreak frog}
    \end{subfigure}    \hfill
    \begin{subfigure}{0.106\textwidth}
        \centering
        \includegraphics[width=\textwidth]{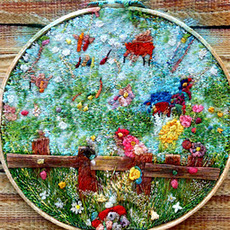}
        \vspace*{-6mm}
        \caption*{\centering \tiny \textit{an embroidered version} \par\nobreak \textit{of} grassy meadow}
    \end{subfigure}
    \begin{subfigure}{0.106\textwidth}
        \centering
        \includegraphics[width=\textwidth]{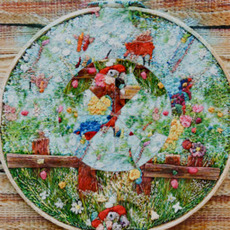}
        \vspace*{-6mm}
        \caption*{\centering \tiny \textit{\quad} \par\nobreak \quad}
    \end{subfigure}
    \begin{subfigure}{0.106\textwidth}
        \centering
        \includegraphics[width=\textwidth]{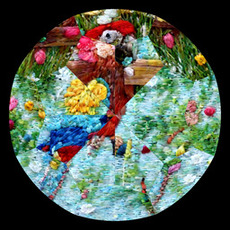}
        \vspace*{-6mm}
        \caption*{\centering \tiny \textit{an embroidered version} \par\nobreak \textit{of} parrot}
    \end{subfigure}
    \begin{subfigure}{0.106\textwidth}
        \centering
        \includegraphics[width=\textwidth]{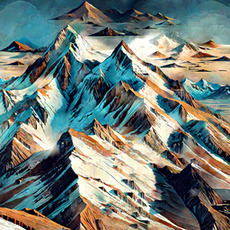}
        \vspace*{-6mm}
        \caption*{\centering \tiny \textit{a digital illustration of} \par\nobreak mountain ridge}
    \end{subfigure}
    \begin{subfigure}{0.106\textwidth}
        \centering
        \includegraphics[width=\textwidth]{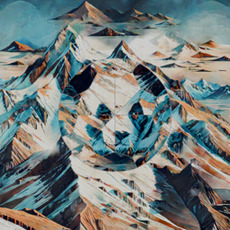}
        \vspace*{-6mm}
        \caption*{\centering \tiny \textit{\quad} \par\nobreak \quad}
    \end{subfigure}
    \begin{subfigure}{0.106\textwidth}
        \centering
        \includegraphics[width=\textwidth]{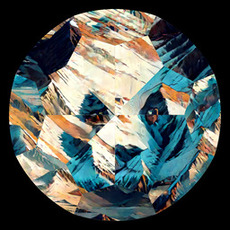}
        \vspace*{-6mm}
        \caption*{\centering \tiny \textit{a digital illustration of} \par\nobreak panda}
    \end{subfigure}    \hfill
    \begin{subfigure}{0.106\textwidth}
        \centering
        \includegraphics[width=\textwidth]{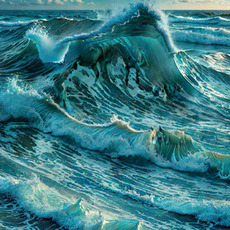}
        \vspace*{-6mm}
        \caption*{\centering \tiny \textit{a cinematic rendering of} \par\nobreak ocean waves}
    \end{subfigure}
    \begin{subfigure}{0.106\textwidth}
        \centering
        \includegraphics[width=\textwidth]{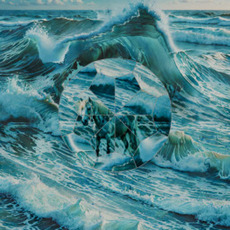}
        \vspace*{-6mm}
        \caption*{\centering \tiny \textit{\quad} \par\nobreak \quad}
    \end{subfigure}
    \begin{subfigure}{0.106\textwidth}
        \centering
        \includegraphics[width=\textwidth]{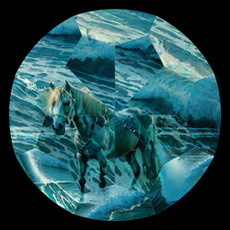}
        \vspace*{-6mm}
        \caption*{\centering \tiny \textit{a cinematic rendering of} \par\nobreak horse}
    \end{subfigure}    \hfill
    \begin{subfigure}{0.106\textwidth}
        \centering
        \includegraphics[width=\textwidth]{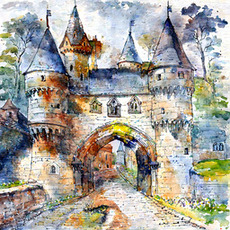}
        \vspace*{-6mm}
        \caption*{\centering \tiny \textit{a watercolor painting of} \par\nobreak medieval castle gate}
    \end{subfigure}
    \begin{subfigure}{0.106\textwidth}
        \centering
        \includegraphics[width=\textwidth]{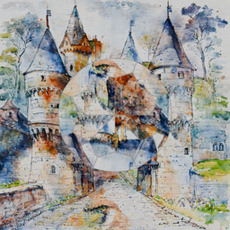}
        \vspace*{-6mm}
        \caption*{\centering \tiny \textit{\quad} \par\nobreak \quad}
    \end{subfigure}
    \begin{subfigure}{0.106\textwidth}
        \centering
        \includegraphics[width=\textwidth]{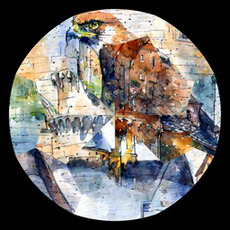}
        \vspace*{-6mm}
        \caption*{\centering \tiny \textit{a watercolor painting of} \par\nobreak hawk}
    \end{subfigure}
    \begin{subfigure}{0.106\textwidth}
        \centering
        \includegraphics[width=\textwidth]{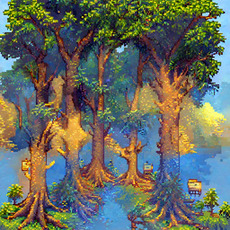}
        \vspace*{-6mm}
        \caption*{\centering \tiny \textit{a pixel art version of} \par\nobreak forest, tall straight trees}
    \end{subfigure}
    \begin{subfigure}{0.106\textwidth}
        \centering
        \includegraphics[width=\textwidth]{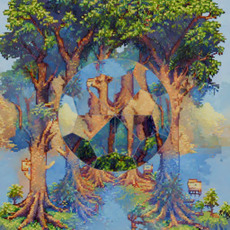}
        \vspace*{-6mm}
        \caption*{\centering \tiny \textit{\quad} \par\nobreak \quad}
    \end{subfigure}
    \begin{subfigure}{0.106\textwidth}
        \centering
        \includegraphics[width=\textwidth]{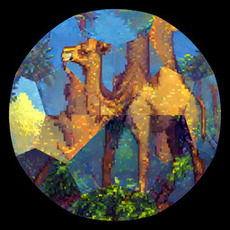}
        \vspace*{-6mm}
        \caption*{\centering \tiny \textit{a pixel art version of} \par\nobreak camel}
    \end{subfigure}    \hfill
    \begin{subfigure}{0.106\textwidth}
        \centering
        \includegraphics[width=\textwidth]{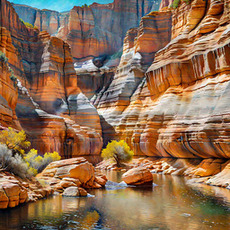}
        \vspace*{-6mm}
        \caption*{\centering \tiny \textit{a photorealistic painting} \par\nobreak \textit{of} sunlit canyon, cliffs}
    \end{subfigure}
    \begin{subfigure}{0.106\textwidth}
        \centering
        \includegraphics[width=\textwidth]{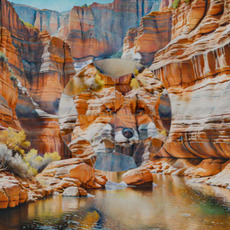}
        \vspace*{-6mm}
        \caption*{\centering \tiny \textit{\quad} \par\nobreak \quad}
    \end{subfigure}
    \begin{subfigure}{0.106\textwidth}
        \centering
        \includegraphics[width=\textwidth]{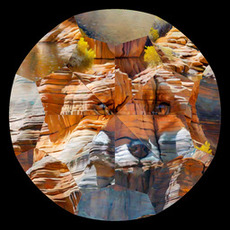}
        \vspace*{-6mm}
        \caption*{\centering \tiny \textit{a photorealistic painting} \par\nobreak \textit{of} fox}
    \end{subfigure}    \hfill
    \begin{subfigure}{0.106\textwidth}
        \centering
        \includegraphics[width=\textwidth]{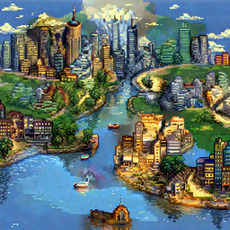}
        \vspace*{-6mm}
        \caption*{\centering \tiny \textit{a 16-bit sprite of} \par\nobreak city river with reflections}
    \end{subfigure}
    \begin{subfigure}{0.106\textwidth}
        \centering
        \includegraphics[width=\textwidth]{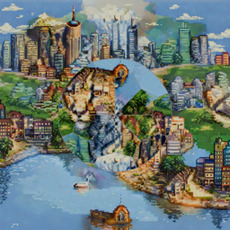}
        \vspace*{-6mm}
        \caption*{\centering \tiny \textit{\quad} \par\nobreak \quad}
    \end{subfigure}
    \begin{subfigure}{0.106\textwidth}
        \centering
        \includegraphics[width=\textwidth]{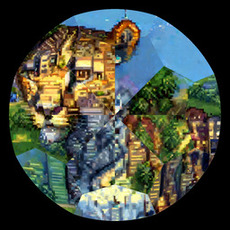}
        \vspace*{-6mm}
        \caption*{\centering \tiny \textit{a 16-bit sprite of} \par\nobreak cheetah}
    \end{subfigure}
    \begin{subfigure}{0.106\textwidth}
        \centering
        \includegraphics[width=\textwidth]{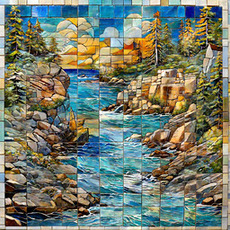}
        \vspace*{-6mm}
        \caption*{\centering \tiny \textit{a ceramic tile mural of} \par\nobreak rocky coastline}
    \end{subfigure}
    \begin{subfigure}{0.106\textwidth}
        \centering
        \includegraphics[width=\textwidth]{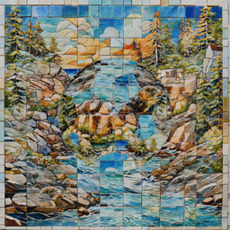}
        \vspace*{-6mm}
        \caption*{\centering \tiny \textit{\quad} \par\nobreak \quad}
    \end{subfigure}
    \begin{subfigure}{0.106\textwidth}
        \centering
        \includegraphics[width=\textwidth]{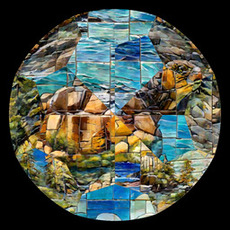}
        \vspace*{-6mm}
        \caption*{\centering \tiny \textit{a ceramic tile mural of} \par\nobreak turtle}
    \end{subfigure}    \hfill
    \begin{subfigure}{0.106\textwidth}
        \centering
        \includegraphics[width=\textwidth]{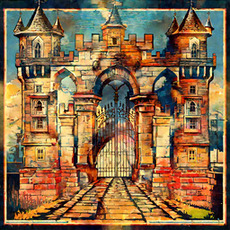}
        \vspace*{-6mm}
        \caption*{\centering \tiny \textit{a vintage poster of} \par\nobreak medieval castle gate}
    \end{subfigure}
    \begin{subfigure}{0.106\textwidth}
        \centering
        \includegraphics[width=\textwidth]{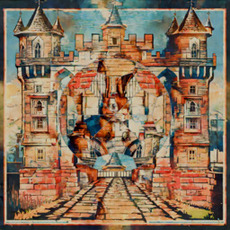}
        \vspace*{-6mm}
        \caption*{\centering \tiny \textit{\quad} \par\nobreak \quad}
    \end{subfigure}
    \begin{subfigure}{0.106\textwidth}
        \centering
        \includegraphics[width=\textwidth]{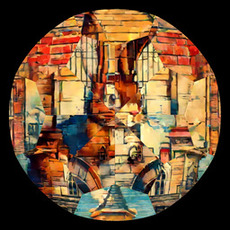}
        \vspace*{-6mm}
        \caption*{\centering \tiny \textit{a vintage poster of} \par\nobreak rabbit}
    \end{subfigure}    \hfill
    \begin{subfigure}{0.106\textwidth}
        \centering
        \includegraphics[width=\textwidth]{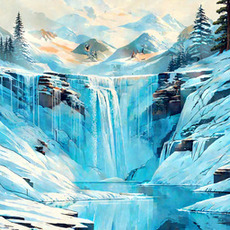}
        \vspace*{-6mm}
        \caption*{\centering \tiny \textit{a minimalist illustration} \par\nobreak \textit{of} frozen waterfall}
    \end{subfigure}
    \begin{subfigure}{0.106\textwidth}
        \centering
        \includegraphics[width=\textwidth]{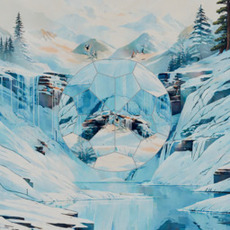}
        \vspace*{-6mm}
        \caption*{\centering \tiny \textit{\quad} \par\nobreak \quad}
    \end{subfigure}
    \begin{subfigure}{0.106\textwidth}
        \centering
        \includegraphics[width=\textwidth]{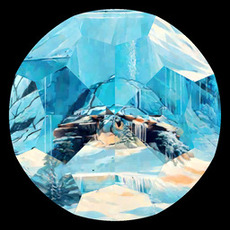}
        \vspace*{-6mm}
        \caption*{\centering \tiny \textit{a minimalist illustration} \par\nobreak \textit{of} turtle}
    \end{subfigure}
    \begin{subfigure}{0.106\textwidth}
        \centering
        \includegraphics[width=\textwidth]{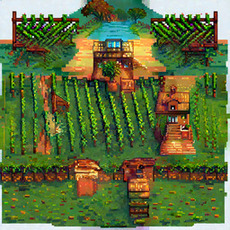}
        \vspace*{-6mm}
        \caption*{\centering \tiny \textit{a pixel art version of} \par\nobreak vineyard, straight trellises}
    \end{subfigure}
    \begin{subfigure}{0.106\textwidth}
        \centering
        \includegraphics[width=\textwidth]{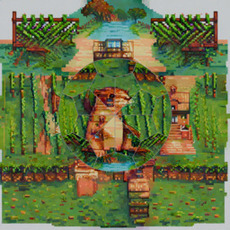}
        \vspace*{-6mm}
        \caption*{\centering \tiny \textit{\quad} \par\nobreak \quad}
    \end{subfigure}
    \begin{subfigure}{0.106\textwidth}
        \centering
        \includegraphics[width=\textwidth]{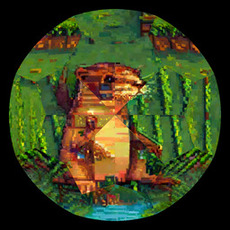}
        \vspace*{-6mm}
        \caption*{\centering \tiny \textit{a pixel art version of} \par\nobreak otter}
    \end{subfigure}    \hfill
    \begin{subfigure}{0.106\textwidth}
        \centering
        \includegraphics[width=\textwidth]{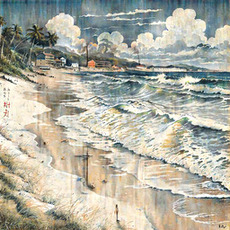}
        \vspace*{-6mm}
        \caption*{\centering \tiny \textit{a lithograph of} beach, shoreline \& waves}
    \end{subfigure}
    \begin{subfigure}{0.106\textwidth}
        \centering
        \includegraphics[width=\textwidth]{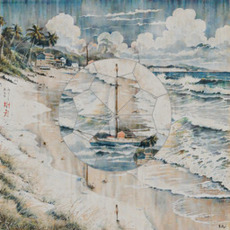}
        \vspace*{-6mm}
        \caption*{\centering \tiny \textit{\quad} \par\nobreak \quad}
    \end{subfigure}
    \begin{subfigure}{0.106\textwidth}
        \centering
        \includegraphics[width=\textwidth]{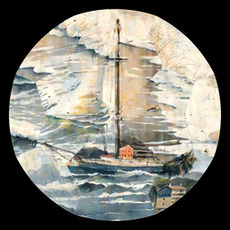}
        \vspace*{-6mm}
        \caption*{\centering \tiny \textit{a lithograph of} \par\nobreak sailboat}
    \end{subfigure}    \hfill
    \begin{subfigure}{0.106\textwidth}
        \centering
        \includegraphics[width=\textwidth]{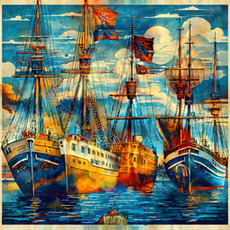}
        \vspace*{-6mm}
        \caption*{\centering \tiny \textit{a vintage poster of} \par\nobreak harbor with docked ships}
    \end{subfigure}
    \begin{subfigure}{0.106\textwidth}
        \centering
        \includegraphics[width=\textwidth]{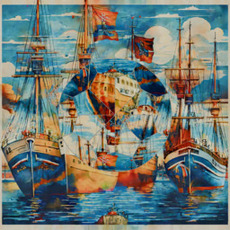}
        \vspace*{-6mm}
        \caption*{\centering \tiny \textit{\quad} \par\nobreak \quad}
    \end{subfigure}
    \begin{subfigure}{0.106\textwidth}
        \centering
        \includegraphics[width=\textwidth]{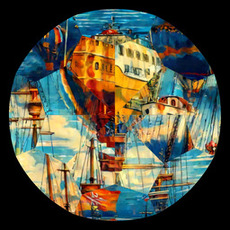}
        \vspace*{-6mm}
        \caption*{\centering \tiny \textit{a vintage poster of} \par\nobreak hot air balloon}
    \end{subfigure}
\end{figure*}

\end{document}